\documentclass{article}

\usepackage{microtype}
\usepackage{graphicx}
\usepackage{subcaption}
\usepackage{booktabs} 
\usepackage{comment}
\usepackage{amsmath}
\usepackage{physics}
\usepackage{placeins}
\usepackage{bbm}
\usepackage{float}
\usepackage{amssymb}
\usepackage{booktabs,chemformula}

\newcommand\bs[1]{\boldsymbol{#1}}

\usepackage{hyperref}

\usepackage[accepted]{icml2021}

\icmltitlerunning{Catapult Dynamics and Phase Transitions in Quadratic Nets}

\begin{document}

\twocolumn[
\icmltitle{Catapult Dynamics and Phase Transitions in Quadratic Nets}



\icmlsetsymbol{equal}{*}

\begin{icmlauthorlist}
\icmlauthor{David Meltzer}{Cornell}
\icmlauthor{Min Chen}{Pittsburgh}
\icmlauthor{Junyu Liu}{Pittsburgh,Pritzker}
\end{icmlauthorlist}

\icmlaffiliation{Cornell}{Department of Physics, Cornell University, Ithaca, NY 14850, USA}
\icmlaffiliation{Pittsburgh}{Department of Computer Science, The University of Pittsburgh, Pittsburgh, PA 15260, USA}
\icmlaffiliation{Pritzker}{Pritzker School of Molecular Engineering, The University of Chicago, Chicago, IL 60637, USA}

\icmlcorrespondingauthor{David Meltzer, Junyu Liu}{dm694@cornell.edu,junyuliu@uchicago.edu}


\vskip 0.3in
]


\printAffiliationsAndNotice{}  

\begin{abstract}
Neural networks trained with gradient descent can undergo non-trivial phase transitions as a function of the learning rate. 
In \cite{lewkowycz2020large} it was discovered that wide neural nets can exhibit a catapult phase for super-critical learning rates, where the training loss grows exponentially quickly at early times before rapidly decreasing to a small value. 
During this phase the top eigenvalue of the neural tangent kernel (NTK) also undergoes significant evolution. 
In this work, we will prove that the catapult phase exists in a large class of models, including quadratic models and two-layer, homogenous neural nets. 
To do this, we show that for a certain range of learning rates the weight norm decreases whenever the loss becomes large.
We also empirically study learning rates beyond this theoretically derived range and show that the activation map of ReLU nets trained with super-critical learning rates becomes increasingly sparse as we increase the learning rate.
\end{abstract}

\section{Introduction}
\label{intro}

In recent years, the field of deep learning has seen a marked revival and lead to remarkable progress in machine learning \cite{goodfellow2016deep}. However, there remain many open problems concerning its theoretical foundations. 
Recent progress has been driven in large part by the development of neural tangent kernel (NTK) theory \cite{lee2017deep,jacot2018neural,lee2019wide,https://doi.org/10.48550/arxiv.1910.00019,arora2019exact,sohl2020infinite,Roberts:2021fes}.
In particular, it has been proven, under certain conditions on the learning rate and weight initialization, that infinitely wide neural nets reduce to linear models.
To study more realistic models, which can learn new representations of the data, one can go beyond the strict infinite width and study perturbative, $1/\text{width}$ corrections \cite{DBLP:journals/corr/abs-1909-08156,DBLP:journals/corr/abs-1909-11304,yaida2020non}.

In this work we are interested in understanding models that display non-trivial, non-perturbative dynamics when trained with full-batch gradient descent.
Specifically, we will study the \emph{catapult mechanism} of \cite{lewkowycz2020large}, which occurs when we train the model with a learning rate $\eta$ that is larger than the maximal learning rate of the corresponding linearized, or infinite-width, model.
We will refer to learning rates above the na\"ive, linear stability threshold as ``super-critical".

\textcolor{black}{The catapult phase refers to an intermediate dynamical regime of gradient descent that lies between the so-called ``lazy phase" and ``divergent phase". In this regime, the training loss initially increases before turning around and decreasing to a small value. The phase is characterized by a transient growth of the NTK or related metrics, and is enabled by large learning rates that exceed the linearized stability threshold termed as super-critical learning rate above.} While in the lazy phase the learning rate is sufficiently small such that the model weights do not deviate significantly from their values at initialization.
In this regime one can use large width perturbation theory to make predictions.
In the ``divergent phase" the learning rate is taken to be large enough such that the loss diverges .
In \cite{lewkowycz2020large} it was demonstrated in a variety of examples that the catapult phase leads to a smaller generalization loss than the lazy phase.

Overall, our understanding of the catapult phase, e.g. why it exists and how the neural net evolves in this phase, pales in comparison to our knowledge about neural nets in the lazy phase.
To gain a better understanding of the catapult phase, we will study two classes of toy models: quadratic models and homogenous MLPs with one hidden layer.
Quadratic models are minimal models for representation learning \cite{Roberts:2021fes}, as they are the simplest deviation from linear models whose NTK evolves under gradient descent.
We will derive sufficient, but not necessary, conditions for quadratic models to converge at large learning rates in terms of the maximal eigenvalue of the meta-feature function, which we will define later.
We will also derive sufficient conditions for homogenous MLPs to have a catapult phase, which will depend on the slope of the activation function in the positive/negative regions.
For both classes of models, the existence of the catapult phase is proven by showing that there exists a range of super-critical learning rates such that the weight norm $\bs{\theta}^2$ decreases whenever the loss becomes large.
We will provide evidence for all analytical results by numerically studying various quadratic models and two-layer MLPs. 

More generally, the weight norm $\bs{\theta}^2$ is an interesting quantity to study because its behavior changes qualitatively as we change the learning rate \textit{inside} the catapult phase.
We observe that for sufficiently large learning rates, the weight norm in ReLU nets can increase significantly over the course of training.
We empirically observe that this increase in the weight norm does not hurt the generalization performance of ReLU MLPs.
We conjecture that these models still generalize well because the activation map of ReLU nets becomes increasingly sparse as we increase the learning rate.
We believe this result points to interesting future directions for the study of ReLU nets at large learning rates.

\textbf{Notations: }We use bold letters, e.g. $\bs{\theta}$, $\bs{\psi}$ to refer to vectors, matrices, etc. in weight space. We use the corresponding un-bolded letter when referring to individual indices, e.g. $\theta_{\mu}$ and $\psi_{\mu\nu}$ refer to specific components. 
We also use un-bolded letters for one-dimensional quantities in weight space.
We use Greek letters from the middle of the alphabet to refer to indices in weight space, as above, and Greek letters from the beginning of the alphabet to refer to points in sample space, e.g. $(\bs{x}_\alpha,\bs{y}_\alpha)$ refers to the $\alpha^{\text{th}}$ data-point and its corresponding label.
We will keep the indices in sample space explicit, except when we consider toy examples with only one data-point, in which case we drop this index for compactness.
The symbols $\mathcal{N}$ and $\mathcal{U}$ refer to the normal and uniform distribution, respectively.

For vector-norms we use the notation $\bs{\theta}^2\equiv|\!|\bs{\theta}|\!|_{2}^{2}$,
where $|\!|\!\cdot\!|\!|_2$ is the $L_2$ norm. For matrices we use the norm induced by the vector $L_2$ norm, $|\!|\bs{\psi}|\!|\equiv |\!|\bs{\psi}|\!|_2$.
In addition, we use $\lambda_i(\bs{\psi})$ to refer to the $i^{\text{th}}$ eigenvalue. We use the $\lambda_i$ notation for matrices in both weight and sample space.
For the NTK we use $\lambda_{\text{max}}(H_{\alpha\beta})$ and $|\!|H_{\alpha\beta}|\!|$ interchangeably.\footnote{Since we reserve index-free notation for objects in weight space, $\lambda_{\text{max}}(H_{\alpha\beta})=|\!|H_{\alpha\beta}|\!|$ is the top eigenvalue of the full NTK matrix, and does not refer to the norm of individual components.}

Finally, we use the $t$-subscript to refer to quantities at step $t$ of gradient descent.
We will exclusively use the $0$ subscript to denote quantities at initialization, e.g. $\bs{\theta}_{0}$ is the vector $\bs{\theta}$ at initialization and not its $0^{\text{th}}$ component. 

\section{Related works}
\textcolor{black}{The catapult phase was first analyzed by \cite{lewkowycz2020large}, who studied gradient descent dynamics in a variety of architectures, including linear models, deep fully-connected networks, convolutional nets, and residual networks. They showed that in the catapult phase, models can access large learning rates stably, and often achieve better generalization performance compared to those trained in the lazy phase. Theoretically, they analyzed a finite-width linear MLP trained with mean-squared error loss, and identified phase boundaries based on the learning rate and Hessian curvature. Empirically, they demonstrated that models trained in the catapult regime not only converge, but may also exhibit more favorable generalization ability.} Besides, the catapult phase in models trained with logistic loss was studied in \cite{DBLP:journals/corr/abs-2011-12547}. The effect of large learning rates on matrix factorization was studied in \cite{https://doi.org/10.48550/arxiv.2110.03677}.

The work we present here is similar to the interesting analysis of \cite{Belkin_quadratic}.
There they proved that the catapult phase exists in the quadratic model that approximates a two-layer ReLU MLP. 
They also demonstrated that the generalization loss of neural nets and their quadratic approximations are qualitatively similar in the catapult phase.
Our work differs from \cite{Belkin_quadratic} in a few key areas.
The first is, we will show that the catapult phase exists in a large class of quadratic models, while in \cite{Belkin_quadratic} they focused on the quadratic approximation of a two-layer ReLU MLP.
We also show that the catapult phase exists in general two-layer, homogenous MLPs, without making a quadratic approximation.
We believe it would be interesting to understand how their analysis for the quadratic approximation to the ReLU net could be used to study the class of quadratic models considered in this work.

Quadratic models have also been studied more generally in the literature.
In \cite{Roberts:2021fes} the quadratic model was proposed as a minimal model to study feature learning.
In \cite{Pennington_second_order} they proved that in quadratic models the maximal eigenvalue of the NTK displays the edge of stability (EOS) phenomena \cite{https://doi.org/10.48550/arxiv.2103.00065}.\footnote{In \cite{https://doi.org/10.48550/arxiv.2103.00065} the EOS was studied for the top eigenvalue of the Hessian. As explained in Appendix A of \cite{Pennington_second_order}, if the model is trained with MSE then the spectrum of the NTK approaches that of the Hessian as the model converges.}
Taylor expansions of neural nets were also studied in \cite{DBLP:journals/corr/abs-2002-04010}. \textcolor{black}{The pure quadratic model with \(\psi(x)\) defined in Section \ref{sec:model_definition} with specific form being \( xx^\top\) is equivalent to the classical phase retrieval problem, which has been extensively studied in both signal processing and machine learning \cite{candes2015phase,sun2018geometric,sarao2020complex,davis2020nonsmooth,sarao2020optimization,mignacco2021stochasticity,arnaboldi2023escaping,martin2024impact}. These works analyze the optimization landscape and the dynamics of GD/SGD under small and moderate learning rates with various regimes. In contrast, we aim to analyze the dynamics and implicit regularization in the large learning rate regime, particularly the catapult phase and associated NTK evolution.}

Finally, we also empirically study the sparsity of ReLU MLPs in the catapult phase as a function of the learning rate. 
For a review on sparsity see \cite{https://doi.org/10.48550/arxiv.2102.00554}.
Two recent works \cite{https://doi.org/10.48550/arxiv.2210.05337} and \cite{https://doi.org/10.48550/arxiv.2210.06313} show that increasing the learning rate promotes sparsity, but to our knowledge they do not consider training the model in the catapult phase.
Large initial learning rates in ReLU nets have also been studied in \cite{https://doi.org/10.48550/arxiv.2212.07295}.

\section{Critical points in the learning rate}



\subsection{Model definitions}
\label{sec:model_definition}
\textbf{Quadratic models}. We start by defining the general quadratic model. 
We take the data-points to be $d$-dimensional, $\bs{x}_\alpha\in \mathbb{R}^{d}$, where $\alpha=1,\ldots,D$, and the labels to be one-dimensional $y_\alpha\in\mathbb{R}$.

The general quadratic model (for a $1d$ output) is defined by:
\begin{align}
z(\bs{x}_{\alpha})=\boldsymbol{\theta}^T \boldsymbol{\phi}(\mathbf{x}_{\alpha})+\frac{\zeta}{2}\boldsymbol{\theta}^T\boldsymbol{\psi}(\mathbf{x}_{\alpha})\boldsymbol{\theta},
\label{eq:quad_vector}
 \end{align}
where $z(\bs{x})\in\mathbb{R}$, $\boldsymbol{\theta}\in\mathbb{R}^{n}$, $\bs{\phi}(\mathbf{x})\in\mathbb{R}^{n}$, $\boldsymbol{\psi}(\mathbf{x})\in\mathbb{R}^{n \times n}$, and $\zeta\in\mathbb{R}$.
The weights are drawn from a normal distribution with unit variance, $\bs{\theta}\sim \mathcal{N}(0,\mathbb{I}_{n\times n})$.
The functions $\bs{\phi}$ and $\bs{\psi}$ are part of the definition of the model and are not trainable.
We will refer to $\bs{\phi}$ and $\bs{\psi}$ as the feature and meta-feature functions, respectively.

The parameter $\zeta$ is in principle arbitrary and measures the deviation of \eqref{eq:quad_vector} from linearity.
When $\zeta=0$ the quadratic model \eqref{eq:quad_vector} reduces to a linear model, which has static features.
When $\zeta\neq0$ the above model can exhibit non-trivial representation learning.
In this work we take $0<\zeta\ll1$, but we will still see large deviations from linearity when the learning rate $\eta$ is taken to be sufficiently large.

We train the model to minimize the mean-squared error (MSE):
\begin{align}
L=\frac{1}{2D}\sum\limits_{\alpha=1}^{D}\epsilon_{\alpha}^2=\frac{1}{2D}\sum\limits_{\alpha=1}^{D}(z_{\alpha}-y_{\alpha})^2,
\end{align}
where the error term is $\epsilon_{\alpha}=z_{\alpha}-y_{\alpha}$.
Throughout this work we will assume $y_\alpha=O(1)$.
Under gradient descent the parameters evolve as:
\begin{align}
\bs{\theta}_{t+1}&=\bs{\theta}_{t}-\frac{\eta}{D}\frac{\partial L_t}{\partial \bs{\theta}_t}
\nonumber
\\
& = \bs{\theta}_t-\frac{\eta}{D}\sum\limits_{\alpha=1}^D\epsilon_{\alpha,t}\bigg(\bs{\phi}(\bs{x}_\alpha)+\zeta\bs{\psi}(\bs{x}_\alpha)\bs{\theta}_t\bigg),
\end{align}
where $\eta$ is the learning rate and we explicitly introduced the time-dependence.

One quantity of particular interest is the Neural Tangent Kernel (NTK), $H_{\alpha\beta}$, which is defined by:
\begin{align}
H_{\alpha\beta}\equiv H(\mathbf{x}_{\alpha},\mathbf{x}_{\beta})&=\frac{1}{D}\sum\limits_{\mu=1}^{n}\frac{\partial z(\mathbf{x}_{\alpha})}{\partial \theta_{\mu}}\frac{\partial z(\mathbf{x}_{\beta})}{\partial \theta_{\mu}}.
\end{align}
In vector notation $H_{\alpha\beta}=D^{-1}(\partial_{\bs{\theta}}z(x_{\alpha}))^T\partial_{\bs{\theta}}z(x_{\beta})$.
In the quadratic model the NTK is given by:
\begin{align}
H_{\alpha\beta}&=\frac{1}{D}\sum\limits_{\mu=1}^n\phi^{E}_{\mu}(\mathbf{x}_\alpha,\bs{\theta})\phi^E_{\mu}(\mathbf{x}_\beta,\bs{\theta}),
\\
\phi^E_{\mu}(\mathbf{x}_\alpha,\bs{\theta})&=\phi_{\mu}(\mathbf{x}_\alpha)+\zeta\sum\limits_{\nu=1}^n\psi_{\mu\nu}(\mathbf{x}_{\alpha})\theta_{\nu},
\end{align}
where $\bs{\phi}^E(\mathbf{x}_\alpha,\bs{\theta})$ are the effective feature functions of the quadratic model.
When $\zeta\neq0$ they are $\theta$-dependent and evolve under gradient descent, while when $\zeta=0$ they reduce to the static feature functions $\bs{\phi}$.
Correspondingly, the NTK only evolves when $\zeta\neq0$.

We study two classes of quadratic models, the \textit{pure quadratic model} which is defined to have $\bs{\phi}=0$ and the \textit{quadratic model with bias} which is defined by the condition that $\bs{\psi}\bs{\phi}=0$, i.e. that $\bs{\phi}$ is an eigenvector of $\bs{\psi}$ with eigenvalue 0.

\textbf{Homogenous Nets}. The quadratic models capture some aspects of representation learning in finite-width neural nets. Unfortunately, the quadratic approximation of neural nets, i.e. simply taking the output function $z(\bs{x}_{\alpha},\bs{\theta})$ and Taylor expanding $\bs{\theta}$ around its value at initialization, breaks down when the weights $\bs{\theta}$ evolve significantly. When this happens we need to keep higher-order terms in the Taylor expansion to get a reasonable approximation to the full neural net.\footnote{The breakdown of the quadratic approximation is discussed in \cite{Roberts:2021fes,Pennington_second_order,Belkin_quadratic}.}

For this reason, we will also consider a more realistic set of toy models: neural nets with a single hidden layer and a scale-invariant activation function.
These also scale quadratically in the weights, but will not generically be differentiable.
They can therefore be thought of as \textit{generalized quadratic models}.
We will see how the methods which proved essential in studying quadratic models can be carried over to homogenous nets with one hidden layer.

We study the following MLP:
\begin{align}
z(x)=\frac{1}{\sqrt{n}}\bs{v}^T\sigma\left(\bs{u}x\right)=\frac{1}{\sqrt{n}}\sum\limits_{\mu=1}^{n}v_\mu\sigma(u_\mu x), \label{eq:generic_scale_invariant_z}
\end{align}
where $x\in \mathbb{R}$, $\bs{u}\in \mathbb{R}^{n}$, $\bs{v}\in \mathbb{R}^{n}$.
The activation function is:
\begin{align}
\sigma(x)=a_+x\hspace{.05cm}\mathbbm{1}_{x\geq0}+a_{-}x\hspace{.05cm}\mathbbm{1}_{x\leq0},\label{eq:sigma_act_definition}
\end{align}
where $\mathbbm{1}$ is the indicator function.
We assume that $0\leq a_-\leq a_+$. 
The ReLU function corresponds to $(a_{-},a_+)=(0,1)$.


The NTK in this model is:\footnote{The derivative of the activation function is not defined at the origin, but for definiteness we can define:
\begin{align}
\sigma'(0)\equiv\frac{1}{2}(a_++a_-).
\end{align}}
\begin{align}
H&=\frac{1}{n}\left(\sigma(\bs{u}x)^2+\left(\bs{v}\circ \sigma'(\bs{u}x)\right)^2\right)\nonumber\\
&=\frac{1}{n}\sum\limits_{\mu=1}^{n}(\sigma^2(u_\mu x)+(v_\mu\sigma'(u_\mu x))^{2}),\label{eq:NTK_generic_scale}
\end{align}
where $\circ$ is the Hadamard product.
Finally, we need that the weight norm is simply given by $\bs{\theta}^2=\bs{u}^2+\bs{v}^2$. 

\begin{table*}[!htbp]
\caption{Rigorous guarantees for the existence of the catapult phase in different models trained on the toy dataset $(x,y)=(1,0)$. 
 }
  \centering
  \begin{tabular}{@{}ccccc@{}}
    \toprule
    Model  & $z$  & $H$    & Derived catapult phase range for $\eta$\\
    \midrule
    Pure quadratic & $\frac{\zeta}{2}\bs{\theta}^T\bs{\psi}\bs{\theta}$ & $\zeta^2\bs{\theta}^T\bs{\psi}^2\bs{\theta}$ & $(\frac{2}{H_{0}},\frac{4}{\zeta^2\bs{\theta}^2_{0}\lambda_{\text{max}}(\bs{\psi}^2)})$ 
    \\[13pt]
    Quadratic with biases & $\bs{\phi}^T\bs{\theta}+\frac{\zeta}{2}\bs{\theta}^T\bs{\psi}\bs{\theta}, \hspace{.2cm} \bs{\psi}\bs{\phi}=0$ & $\bs{\phi}^2+\zeta^2\bs{\theta}^T\bs{\psi}^2\bs{\theta}$ & $(\frac{2}{H_{0}},\frac{4}{2\bs{\phi}^{2}+\zeta^2\lambda_{\text{max}}(\bs{\psi}^2)\left(\bs{\theta}^{2}_{0}+\frac{(\bs{\phi}^T\bs{\theta}_{0})^2}{\bs{\phi}^2}\right)})$ 
    \\[13pt]
    Homogenous nets & $ \begin{matrix}
   \frac{1}{\sqrt{n}}\bs{v}^T\sigma\left(\bs{u}x\right)   
      \\
   \sigma=a_+x\hspace{.05cm}\mathbbm{1}_{x\geq0}+a_{-}x\hspace{.05cm}\mathbbm{1}_{x\leq0}\\
   a_->0  \\
\end{matrix}$ & $\frac{1}{n}\left(\sigma(\bs{u}x)^2+\left(\bs{v}\circ \sigma'(\bs{u}x)\right)^2\right)$ & $(\frac{2}{H_0},\frac{4n}{a_+^2(\bs{u}_0^2+\bs{v}_0^2)})$
\\[20pt]
    ReLu nets & $ \begin{matrix}
   \frac{1}{\sqrt{n}}\bs{v}^T\sigma\left(\bs{u}x\right)   \\
   \sigma=a_+x\hspace{.05cm}\mathbbm{1}_{x\geq0}+a_{-}x\hspace{.05cm}\mathbbm{1}_{x\leq0}   \\
   
   (a_{-},a_+)=(0,1)   \\
\end{matrix}$ & $\frac{1}{n}\left(\sigma(\bs{u}x)^2+\left(\bs{v}\circ \sigma'(\bs{u}x)\right)^2\right)$ & $(\frac{2}{{{H}_{0}}},\frac{4}{{{H}_{0}}})$
\\
    \bottomrule
  \end{tabular}
\label{catares}
\end{table*}

\subsection{Summary of Results}
Here we will summarize the main results.
Table \ref{catares}\footnote{\textcolor{black}{In particular, the derived learning rate bounds ensure that \( \theta_t^2 \) decays during training, which can be viewed as an implicit \( L_2 \) regularization induced by the dynamics of gradient descent at large learning rates.}} gives the primary analytic results of our paper, where we summarize our sufficiency conditions for the existence of the catapult phase using a toy dataset. 
The toy dataset is defined to have a single, one-dimensional datapoint, $x=1$, with label $y=0$.
The proofs are given in Appendix \ref{app:single_datapoint}.
Although this dataset is too simple to model realistic datasets, it does give qualitatively good insight into the dynamics of these models at large learning rates.
The generalization of these analytic results to more generic datasets is given in Appendix \ref{app:multiple_datapoints}.
Below, we will briefly comment on the results derived.

\begin{itemize}
\item Our results in Table \ref{catares} are rigorously derived and significantly extend the results of \cite{lewkowycz2020large,Belkin_quadratic}. 
These conditions are in general sufficient, but not necessary, for the existence of the catapult phase.
Our conditions are sufficient because they are derived by imposing that the weight norm $\bs{\theta}_t^2$ decays monotonically during training.
In quadratic models and homogenous MLPs, the loss can only diverge if some of the weights $\theta_{\mu}$ diverge. Therefore, if $\bs{\theta}^2_t$ is bounded the loss must be finite.
On the other hand, our conditions are not necessary because imposing that $\bs{\theta}_t^2$ monotonically decreases is stronger than only requiring that $\bs{\theta}_{t}^2$ is finite.
We find numerically that the catapult phase generically does exist beyond the range we have derived.

\item Given that our conditions were derived by requiring that the weight norm decreases monotonically under gradient descent, we can restate our results as follows: large learning rates which satisfy the conditions listed in Table \ref{catares} lead to an implicit $L_2$ regularization of the weight norm $\bs{\theta}_t^2$, \textcolor{black}{i.e., these conditions imply that the weight norm \( \theta_t^2 \) decreases monotonically during training, acting as an implicit \( L_2 \)-type regularization, even though no explicit regularization term is included in the loss function.}
This result, along with the fact that the catapult phase lands on flatter minima \cite{DBLP:journals/corr/KeskarMNST16,lewkowycz2020large},\footnote{In the two-layer linear net with $1d$ input, the NTK is proportional to the weight norm \cite{lewkowycz2020large}, so a decrease in the weight norm implies that the minima is flatter.} may help explain why models in the catapult phase generalize better than models in the lazy phase.\footnote{In general however, flatness is not necessarily correlated with better generalization behavior \cite{https://doi.org/10.48550/arxiv.1703.04933,https://doi.org/10.48550/arxiv.2206.10654}.}

\item The analysis of ReLU nets is more subtle than the analysis of general homogenous nets.
If we take our bounds for homogenous nets and set $a_-=0$ and $a_+=1$, the allowed range for $\eta$ shrinks to zero size.\footnote{To see this from the third row of Table \ref{catares}, we use that in the ReLU net with one datapoint on average only half of the weights at $t=0$ contribute to the NTK. Therefore, at $t=0$ the weight norm is twice the NTK on average and the allowed range disappears.} 
However, by using special properties of the ReLU function, we can use a different argument to show that a two-layer ReLU net with one, $1d$ datapoint converges for $\eta H_0\leq 4$. This agrees with the bound found in \cite{Belkin_quadratic} for the quadratic approximation of the same ReLU net with one datapoint of generic dimension.

On the other hand, in practice we know that the catapult phase for ReLU nets exists for larger learning rates, up to $\eta H_0 \lesssim 12$ \cite{lewkowycz2020large}.
This presents an interesting tension and suggests that the two-layer ReLU net may have characteristically different behaviors depending on the sign of $4-\eta H_0$.
We observe empirically that there is a qualitative difference: for $\eta H_0\lesssim 4$ the weight norm $\bs{\theta}^2$ decreases during training, while for $\eta H_0\gtrsim 4$ we find that $\bs{\theta}^2$ receives positive updates during training. In particular, if we take $\eta H_0$ large enough, then the final weight norm can be greater than its value at initialization, $\bs{\theta}^2_{\infty}>\bs{\theta}^2_{0}$.
Despite this increase in $\bs{\theta}^2$, models trained with $\eta H_0\gtrsim 4$ still generalize well \cite{lewkowycz2020large}. We conjecture that this is the case because the activation map of ReLU nets becomes sparse in the catapult phase.
\end{itemize}

\begin{figure*}[!ht]
\centering

\begin{subfigure}[t]{0.3\textwidth}
\centering
\includegraphics[scale=.25]{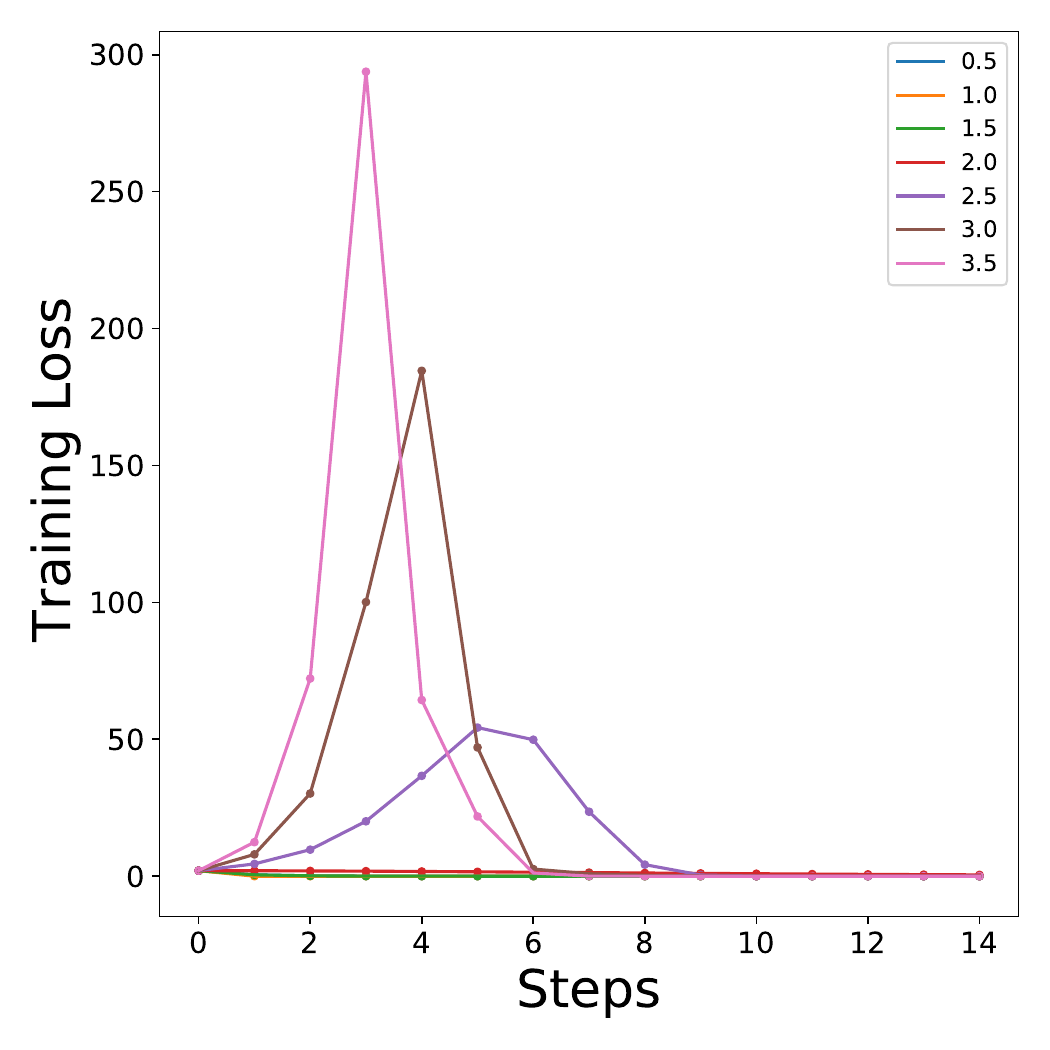}
\caption{}
\end{subfigure}
\hfill
\begin{subfigure}[t]{0.3\textwidth}
\centering
\includegraphics[scale=.25]{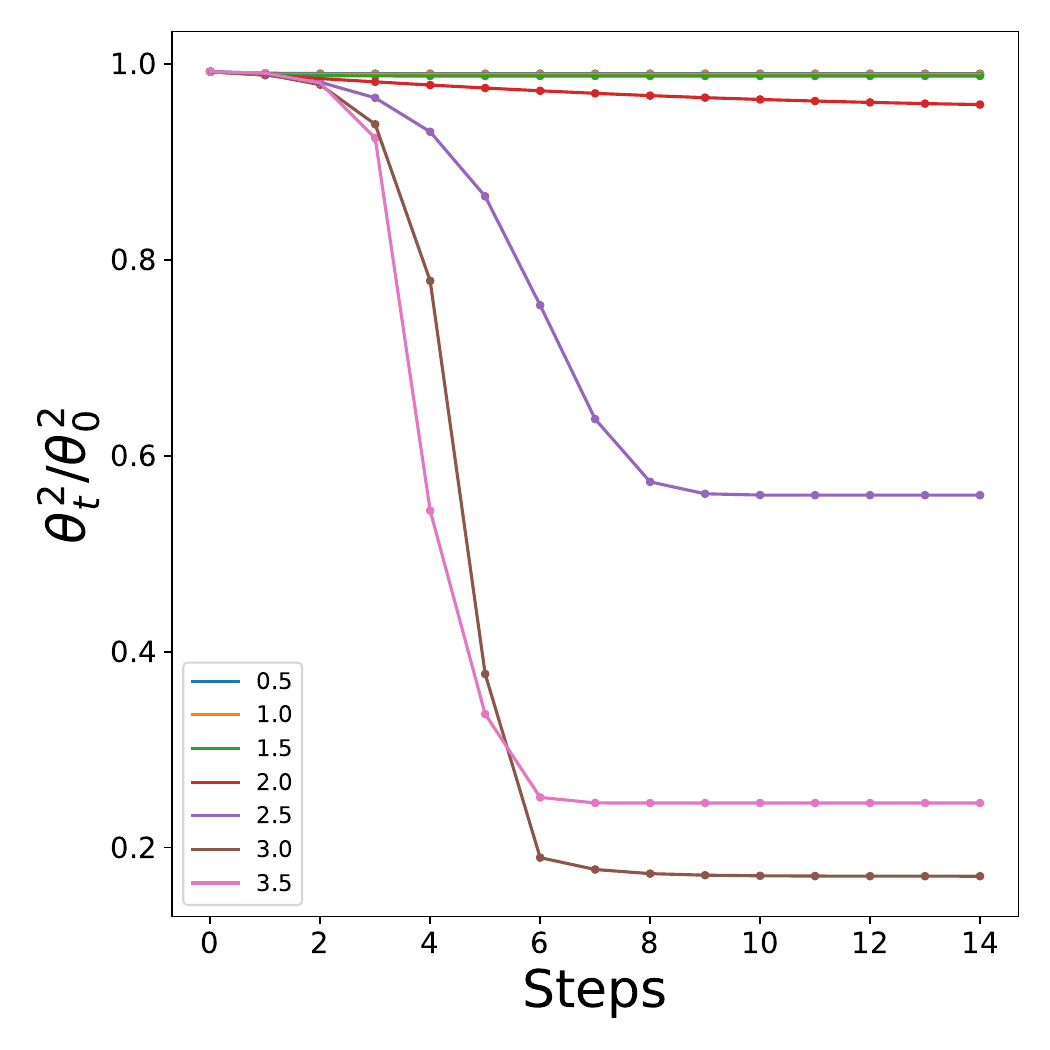}
\caption{}
\end{subfigure}
\hfill
\begin{subfigure}[t]{0.3\textwidth}
\centering
\includegraphics[scale=.25]{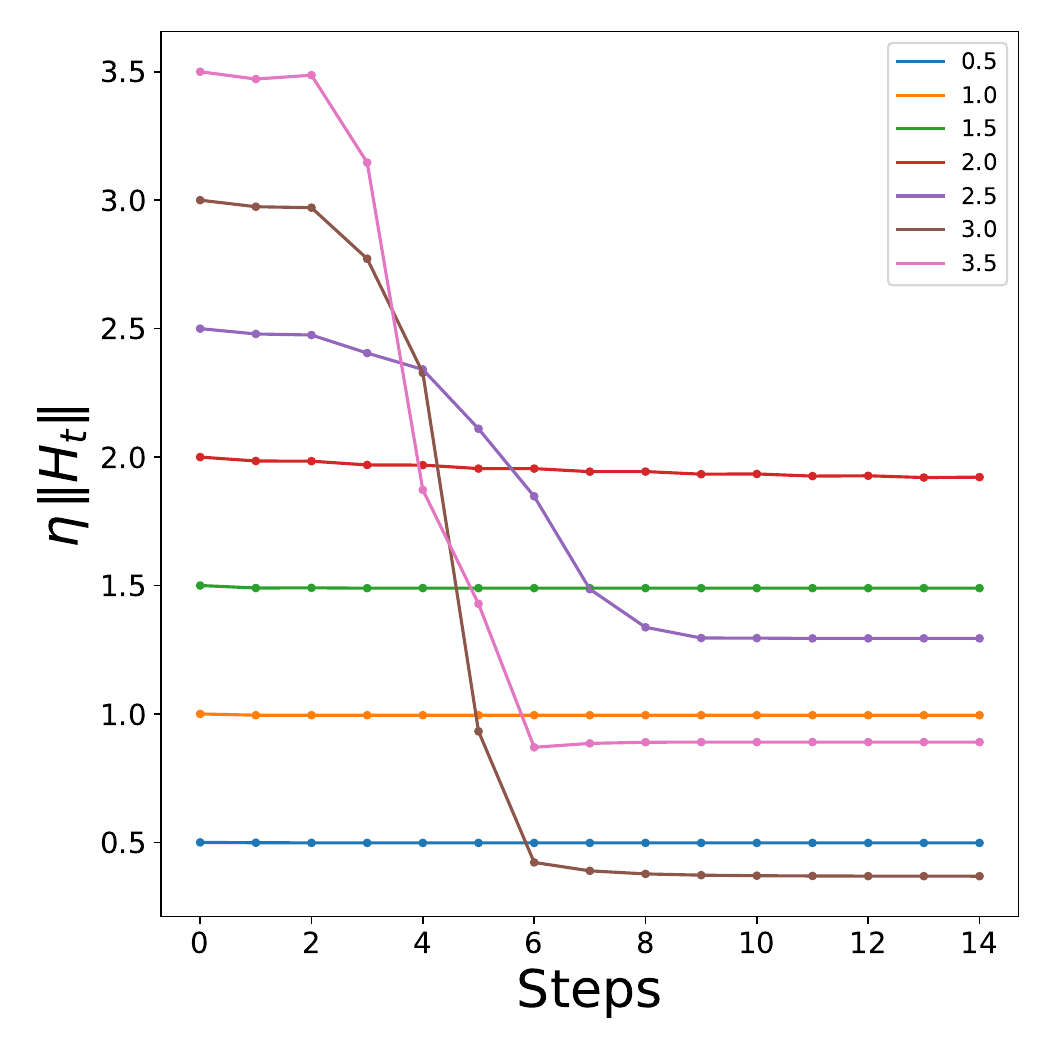}
\caption{}
\end{subfigure}
\centering
\begin{subfigure}[b]{0.4\textwidth}
\centering
\includegraphics[scale=.25]{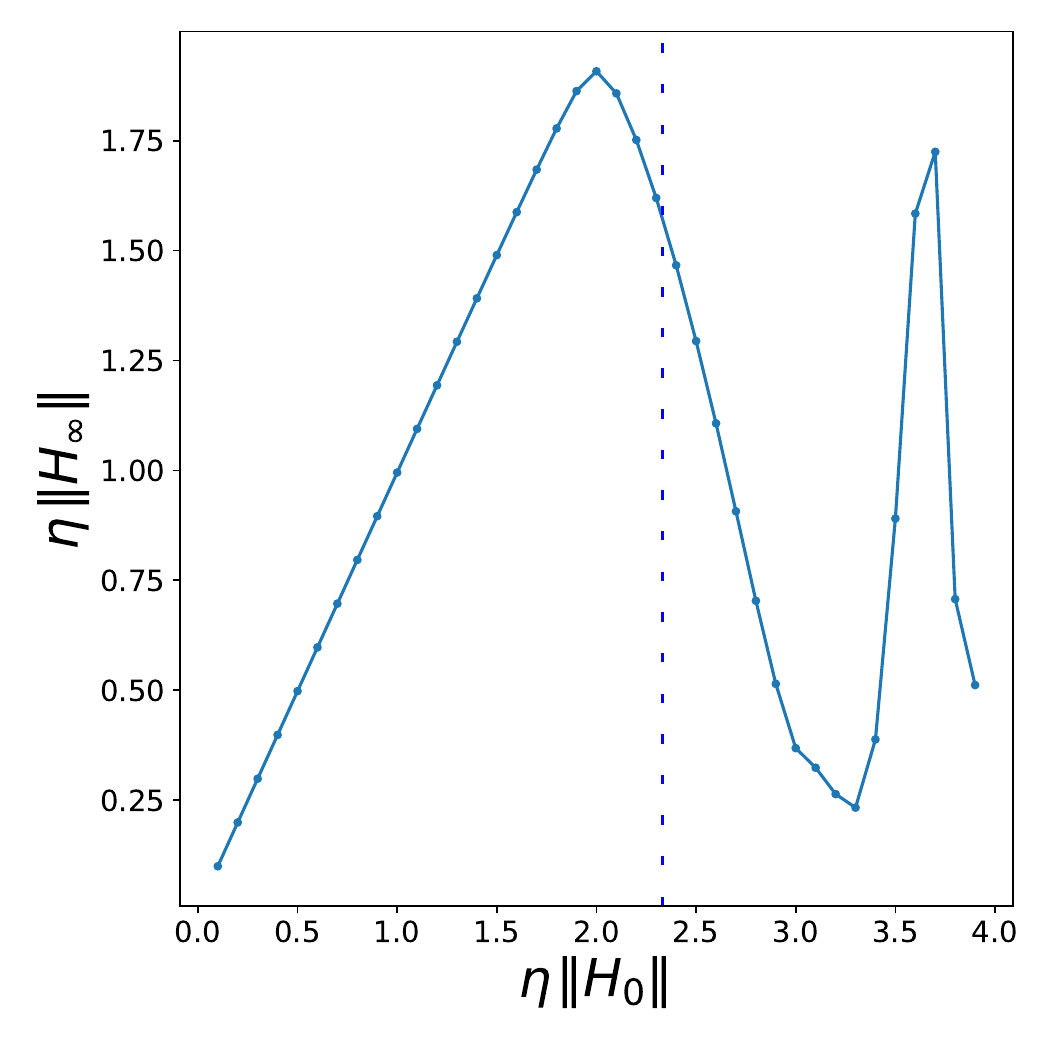}
\caption{}
\end{subfigure}
\hspace{.25in}
\begin{subfigure}[b]{0.4\textwidth}
\centering
\includegraphics[scale=.25]{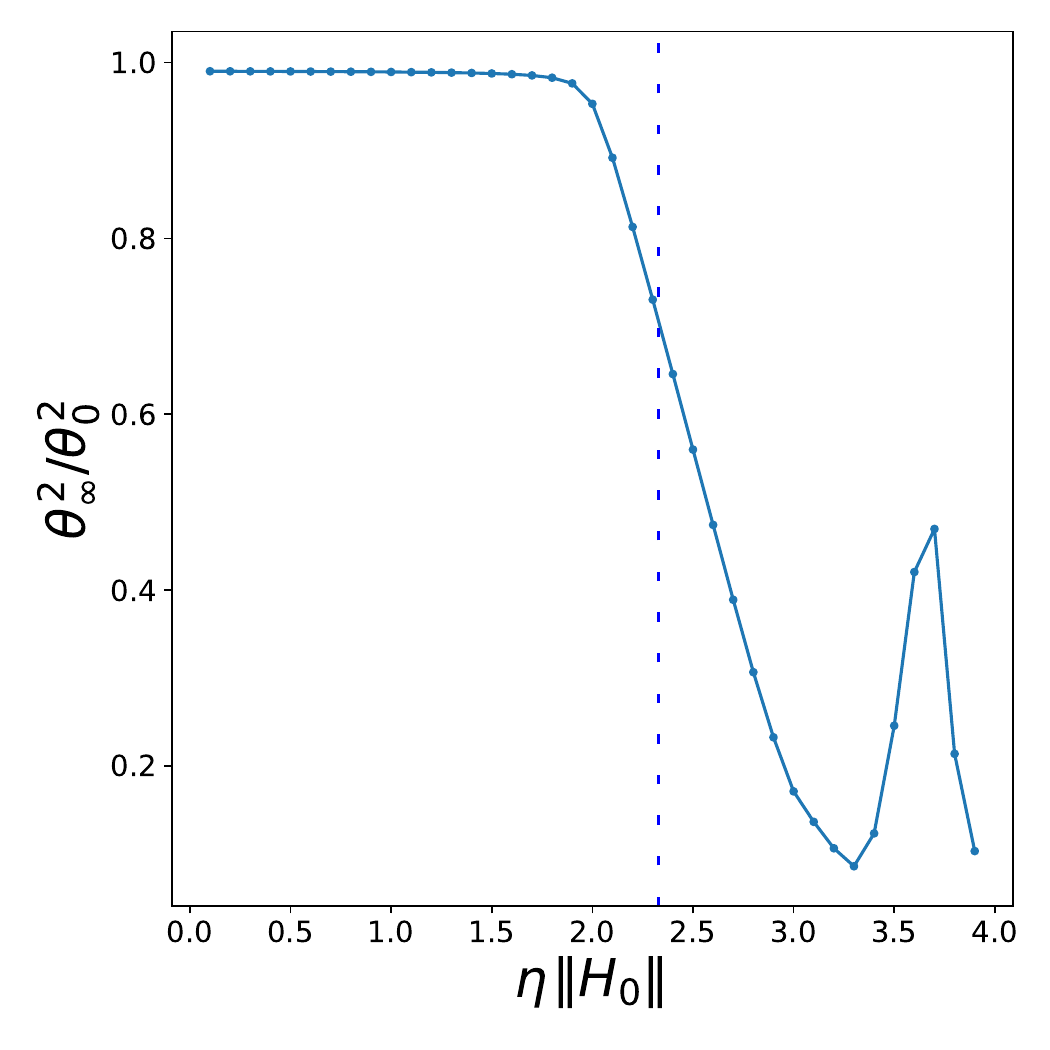}
\caption{}
\end{subfigure}
\hfill
\caption{Results for the pure quadratic model with a linear meta-feature function trained on the toy dataset $(x,y)=(1,0)$. (a)-(c) give the evolution of the loss $L_t$, the weight norm $\bs{\theta}^2_t$, and $\eta H_t$ as a function of time, respectively. The different colors in (a)-(c) correspond to different choices of $\eta H_0$. In (a) we see that the loss experiences a transient phase of exponential growth before decreasing down to 0. In (b)-(c) we see that in the catapult phase both $\bs{\theta}_t^2$ and $\eta H_t$ decrease significantly. In (d)-(e) we show the final value of $\eta H_{t}$ and $\bs{\theta}_t^2$ as a function of $\eta H_0$. 
The dashed vertical lines correspond to the upper bound in the first row of Table \ref{catares}.
We correctly predict that the model does have a catapult phase and empirically observe the model converges up to $\eta H_0=4$.}
\label{fig:linear_meta_toy_dataset_ev1_2}
\end{figure*}

\begin{figure*}[!ht]
\centering

\begin{subfigure}[t]{0.3\textwidth}
\centering
\includegraphics[scale=.25]{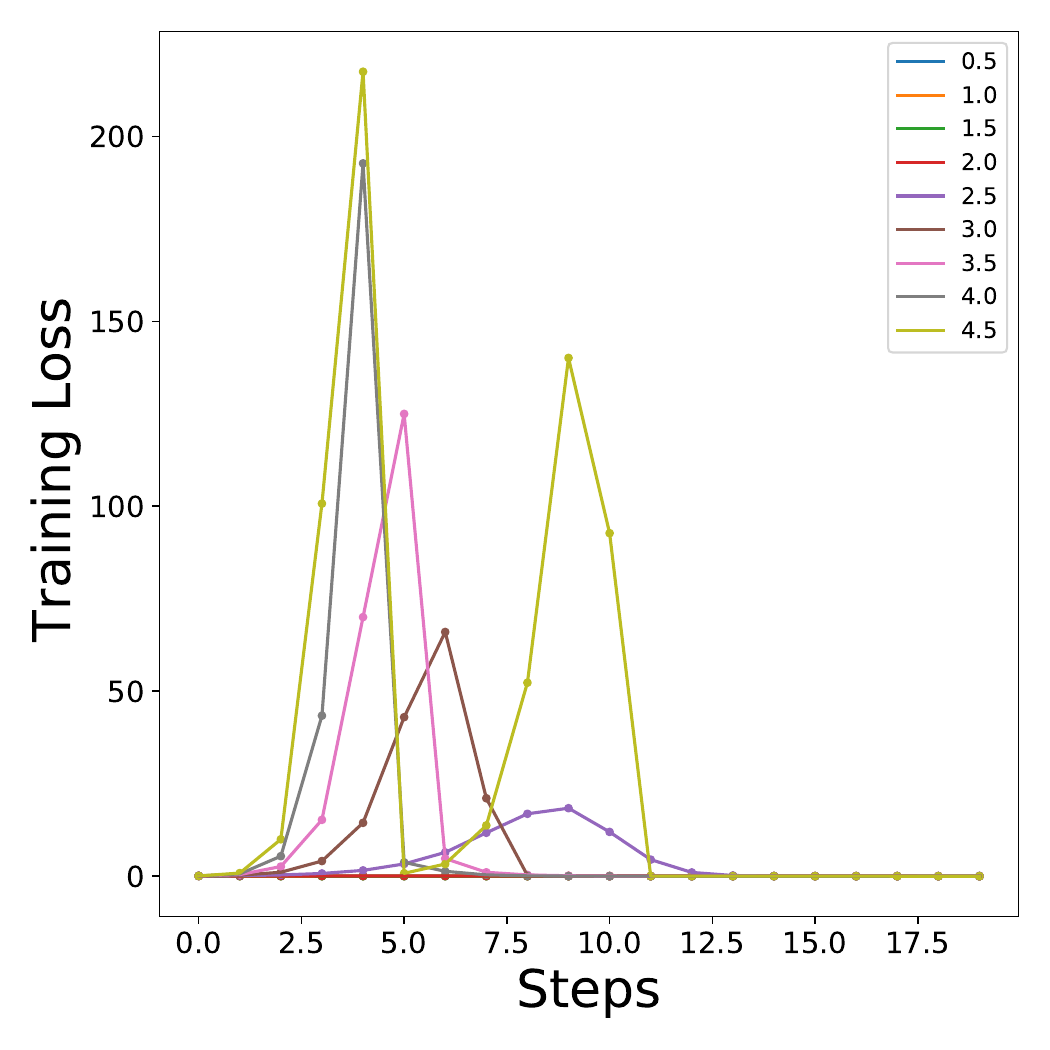}
\caption{}
\end{subfigure}
\hfill
\begin{subfigure}[t]{0.3\textwidth}
\centering
\includegraphics[scale=.25]{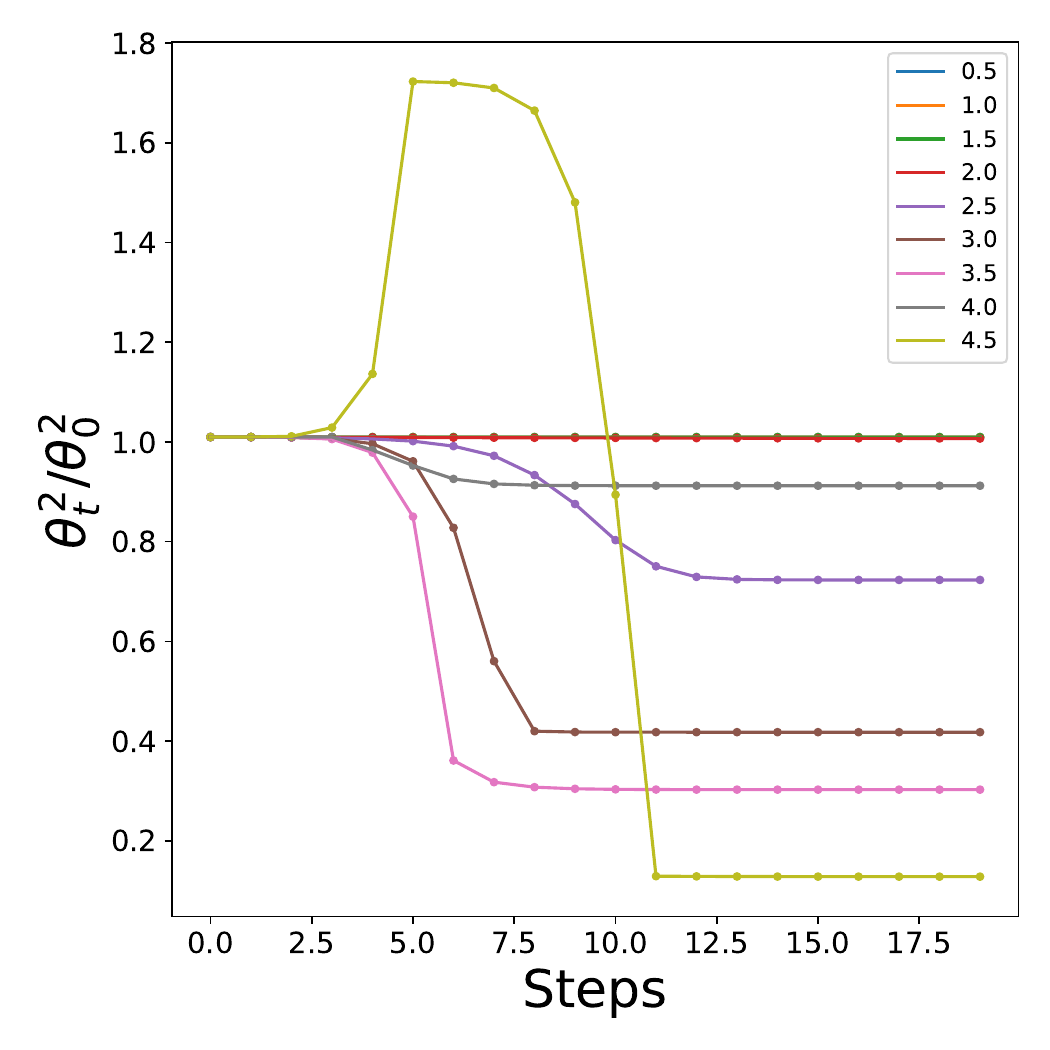}
\caption{}
\end{subfigure}
\hfill
\begin{subfigure}[t]{0.3\textwidth}
\centering
\includegraphics[scale=.25]{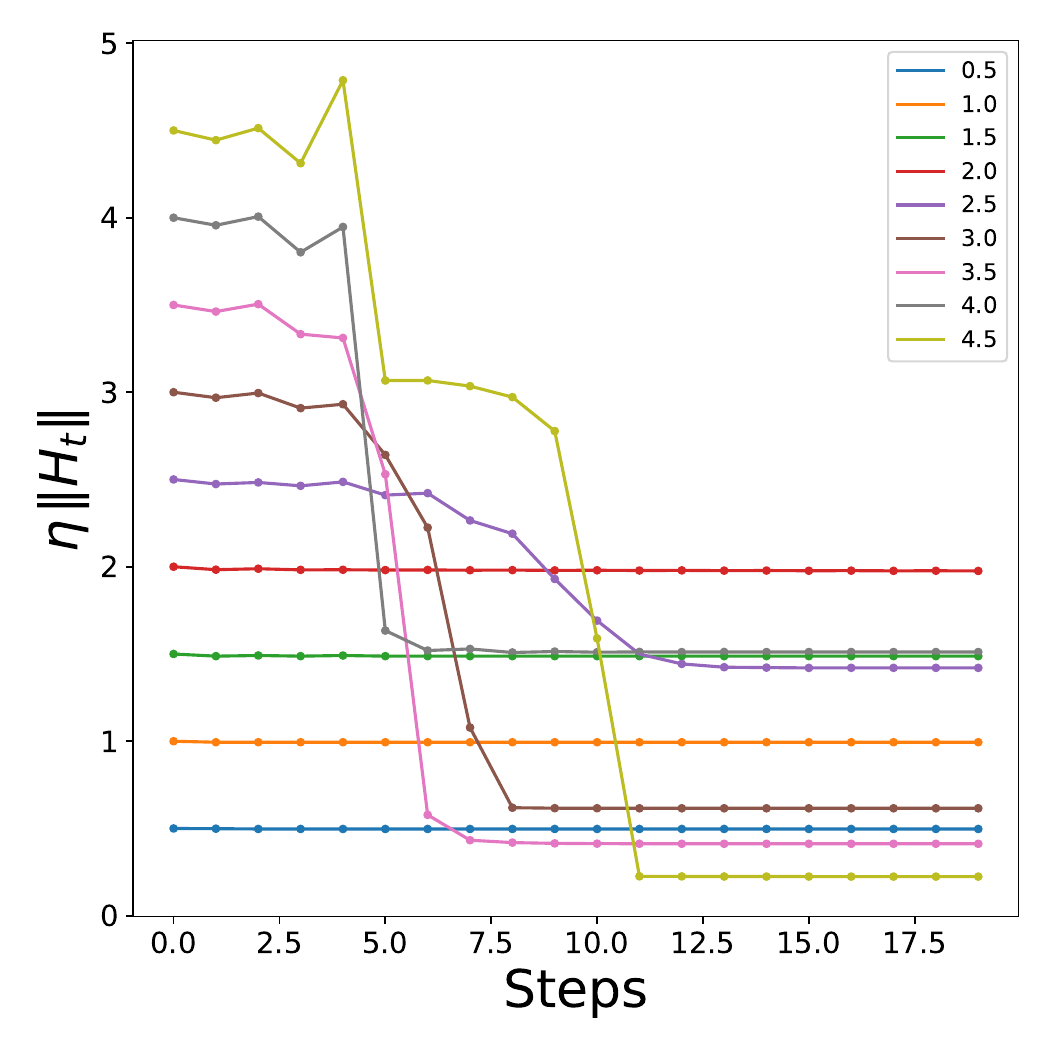}
\caption{}
\end{subfigure}
\centering
\begin{subfigure}[b]{0.4\textwidth}
\centering
\includegraphics[scale=.25]{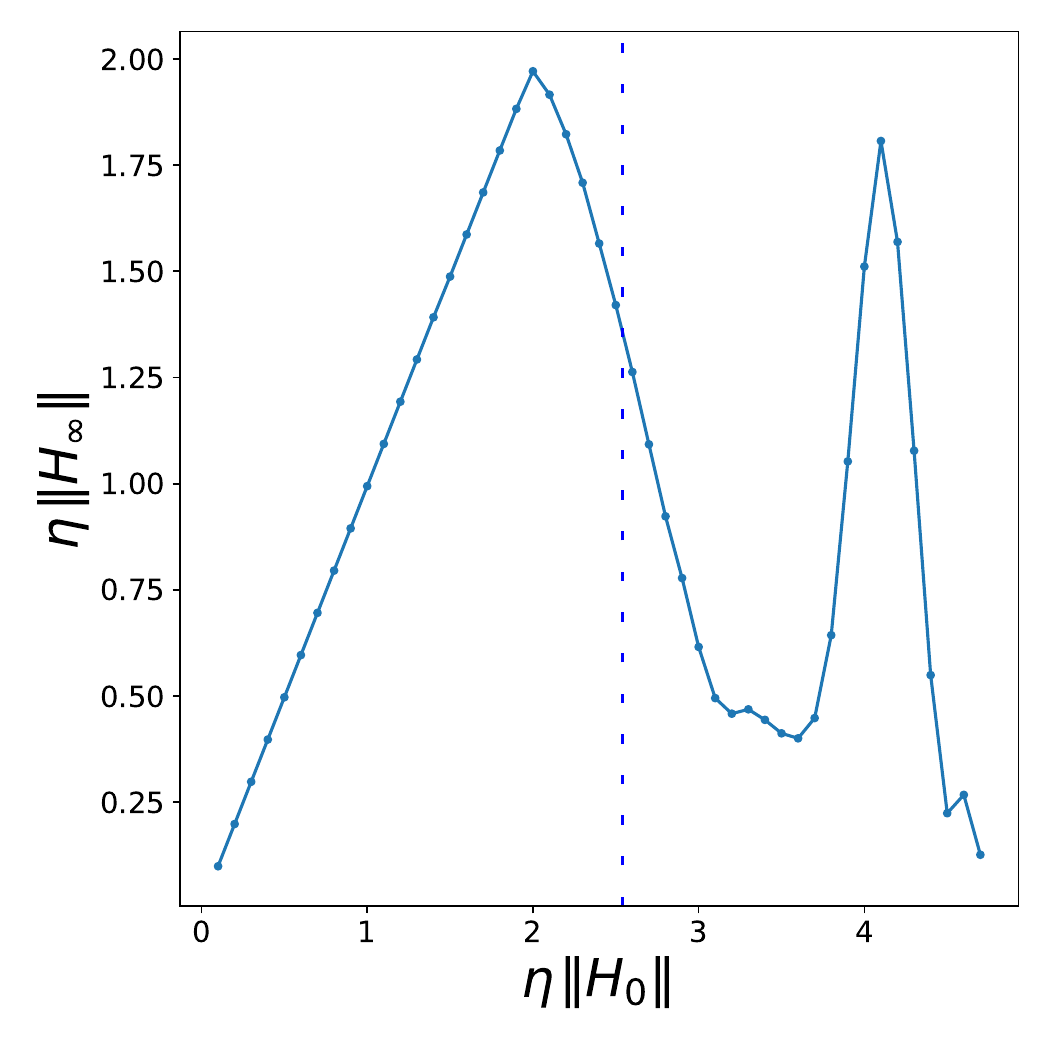}
\caption{}
\end{subfigure}
\hspace{.25in}
\begin{subfigure}[b]{0.4\textwidth}
\centering
\includegraphics[scale=.25]{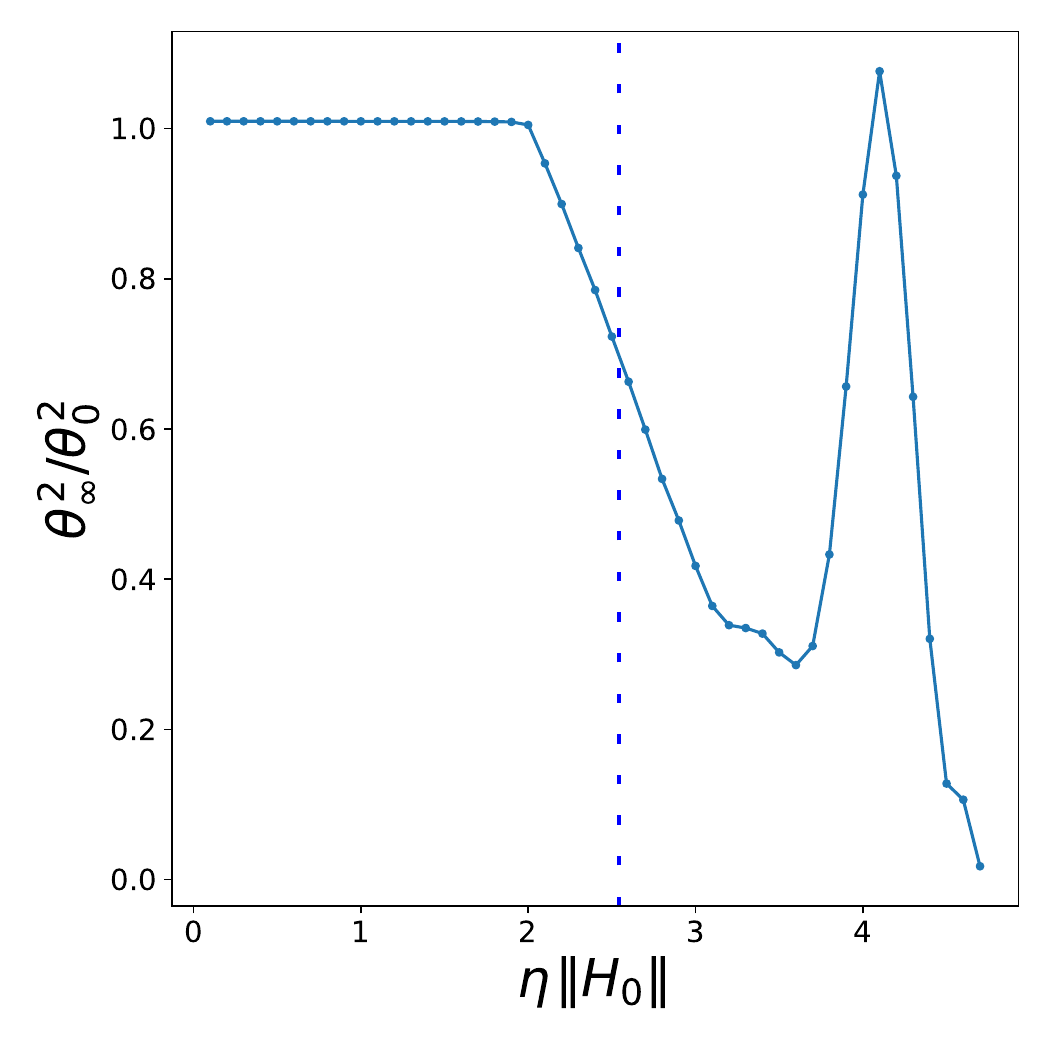}
\caption{}
\label{fig:linear_scale_invariant_toy_dataset_half_final_weight}
\end{subfigure}
\hfill
\caption{Results for the two-layer, homogenous net with $a_+=1$ and $a_-=1/2$. 
(a)-(c) give the evolution of the loss $L_t$, the weight norm $\bs{\theta}^2_t$, and $\eta H_t$ as a function of time, respectively.
The different colors in (a)-(c) correspond to different choices of $\eta H_0$.
(d) and (e) give the final values for $\eta H_t$ and $\bs{\theta}^2_t$ after the model has converged as a function of $\eta H_0$.
The hidden layer has width 1024. The vertical dashed line corresponds to the upper bound in the third row of Table \ref{catares}.
We again correctly predict that the model converges for super-critical learning rates to the left of this line.}
\label{fig:linear_scale_invariant_toy_dataset_half}
\end{figure*}

\subsection{Sketch of a Proof}
\label{ssec:sketch_proof}
Here we will sketch the proof for the results in Table \ref{catares} and leave the details for Appendix \ref{app:single_datapoint}.
In both two-layer, homogenous nets and the pure quadratic model, the output $z$ is a homogenous function of $\bs{\theta}$ with scaling weight two.\footnote{A slightly modified argument is needed for the quadratic model with bias, see Appendix \ref{app:single_datapoint}.}
Restoring the explicit weight dependence we have:
\begin{align}
z(\bs{x}_{\alpha},\lambda \bs{\theta})=\lambda^2z(\bs{x}_{\alpha},\bs{\theta}).\label{eq:homogeneity_weight_two}
\end{align}
This homogeneity property simplifies the update equation for $\bs{\theta}_{t}^2$ and implies that the NTK is also a homogenous function of $\bs{\theta}_t$ with weight two.
Both of these properties are important in deriving the results of Table \ref{catares}.

If we train a net with the property \eqref{eq:homogeneity_weight_two} on the toy dataset $(x,y)=(1,0)$ we find:
\begin{align}
\bs{\theta}_{t+1}^2=\bs{\theta}_{t}^2+\eta z_{t}^2(\eta H_{t}-4).\label{eq:weight_norm_update_simple}
\end{align}
From \eqref{eq:weight_norm_update_simple} we observe that the value $\eta H_t=4$ is special: if $\eta H_t<4$ then the weight norm decreases, $\bs{\theta}_{t+1}^2<\bs{\theta}_{t}^2$, while for $\eta H_t>4$ it increases. 
Therefore, if we can guarantee that $\eta H_t<4$ for all $t$, then \eqref{eq:weight_norm_update_simple} implies that the weight norm monotonically decreases.
This condition is sufficient, but not necessary, to ensure the loss does not diverge.
It is also clear that imposing $\eta H_0<4$ does not guarantee convergence: it ensures that the first step of gradient descent decreases $\bs{\theta}^2_t$, but $H_t$ can increase during training and cause the model to diverge.

Therefore, to ensure convergence, we want to bound the NTK $H_t$ throughout training.
Specifically, we will bound $H_t$ in terms of the weight norm $\bs{\theta}_t^2$:
\begin{align}
H_{t}\leq C \bs{\theta}_{t}^2,\label{eq:bound_H_theta}
\end{align}
where $C$ is some constant, positive number.\footnote{In principle, one can also consider bounding the NTK in terms of a positive, monotonic function of $\bs{\theta}_{t}^2$, but we will find the above bound sufficient for the models we study.}
We will show that the bound \eqref{eq:bound_H_theta} holds in the pure quadratic model and for two-layer, homogenous MLPs in Appendix \ref{app:single_datapoint}.
The bound \eqref{eq:bound_H_theta} allows us to ensure that $H_t$ cannot become too large during training and cause the weight norm to get positive updates.
Here it is important that the homogeneity property \eqref{eq:homogeneity_weight_two} implies $H_t$ also has scaling weight two in $\bs{\theta}_t$ in order that both sides of \eqref{eq:bound_H_theta} have the same scaling in $\bs{\theta}_t$.

Finally, if the inequality \eqref{eq:bound_H_theta} holds, then $\bs{\theta}_{t}^2$ is a monotonically decreasing quantity when:
\begin{align}
\eta < \frac{4}{C \bs{\theta}_{t=0}^2}.\label{eq:general_bound_eta}
\end{align}
The argument is simple, if \eqref{eq:general_bound_eta} holds then the first step of gradient descent decreases the weight norm:
\begin{align}
\bs{\theta}_{t=1}^2-\bs{\theta}_{t=0}^2&=\eta z_{t=0}^2(\eta H_{t=0}-4)
\nonumber
\\
& \leq \eta z_{t=0}^2(\eta C \bs{\theta}_{t=0}^2-4)<0.
\label{eq:first_step}
\end{align}
To get the first inequality we plugged in \eqref{eq:bound_H_theta} and to get the second inequality we used \eqref{eq:general_bound_eta}. \textcolor{black}{\eqref{eq:general_bound_eta} guarantees that the weight norm decreases monotonically during training, which acts as an effective or implicit \( L_2 \)-type regularization. We emphasize that although no explicit regularization term is present, the dynamical effect of large learning rates in this regime suppresses weight growth similarly to \( L_2 \) regularization.} 

Now after a second step of gradient descent we have:
\begin{align}
\bs{\theta}_{t=2}^2-\bs{\theta}_{t=1}^2&=\eta z_{t=1}^2(\eta H_{t=1}-4)
\nonumber
\\
& \leq \eta z_{t=1}^2(\eta C \bs{\theta}_{t=1}^2-4)
\nonumber
\\
& < \eta z_{t=1}^2(\eta C \bs{\theta}_{t=0}^2-4)<0.
\end{align}
Once again, to get the first inequality we used \eqref{eq:bound_H_theta}. To get the second inequality we used \eqref{eq:first_step}, or that the first step decreased $\bs{\theta}^2_t$.
The final inequality follows from condition \eqref{eq:general_bound_eta}.
It \textcolor{black}{is} now clear that we can extend this argument to all $t$ using an inductive proof, which is given in Appendix \ref{app:single_datapoint}.

\section{Experiments}\label{exp}
Here we will present experimental results for quadratic models and two-layer, homogenous nets. \textcolor{black}{Notably, the experimental choices are for clarity and numerical stability. They are not essential to observe catapult dynamics. In the following we find qualitatively similar behavior under a wide range of settings.}
Further details and experiments, including extensions to more generic quadratic models and datasets, can be found in Appendix \ref{app:more_experiments}.
\subsection{Linear Meta-Feature Function}
In general, the meta-feature function $\bs{\psi}$ is an arbitrary function of $\bs{x}$. In this section we study the pure quadratic model with a linear, meta-feature function:\begin{eqnarray}\psi^{\text{lin}}_{\mu\nu}(\bs{x}_\alpha)=\sum\limits_{i=1}^{d}x_{\alpha,i}W^{i}_{\mu\nu}.
\end{eqnarray}
This is arguably the simplest class of meta-feature functions to consider.
The tensor $W^{i}_{\mu\nu}$ is symmetric under $\mu\leftrightarrow \nu$ and can be diagonalized for each $i$:
\begin{align}
W^{i}_{\mu\nu}=\sum\limits_{\sigma=1}^n\lambda^{i}_{\sigma}q_{\sigma \mu}^{i}q_{\sigma \nu}^{i}.
\label{eq:W_definition}
\end{align}
Here for each, fixed $i$ the matrix $q^i_{\sigma\mu}$ is an orthogonal matrix.

We will train the pure quadratic model on the toy dataset $(x,y)=(1,0)$.\footnote{\textcolor{black}{The toy setting $(x, y) = (1, 0)$ is a simplification that offers qualitatively meaningful insights into the dynamics of the model under large learning rates. This allows us to analytically demonstrate the existence of the catapult phase in a clean and transparent manner. Though not conducting traditional representation learning, it reveals that the model can still exhibit nontrivial training dynamics.} In Appendix \ref{app:exp_quad} we will consider experiments for both the pure quadratic model and the quadratic model with bias trained on random datasets and teacher-student set-ups.}
We take the number of weights to be $n=1000$ and set $\zeta^2=2/n$. \textcolor{black}{This setting serves as a form of NTK normalization. It ensures that the effective feature functions and NTK remain well-behaved in the large-width limit, similar to the \(1/\sqrt{d}\) scaling used in Transformer attention mechanisms~\cite{10.5555/3295222.3295349}. 
See also the discussions in~\cite{roberts2022principles} for related justifications.} For each $i$ we take $q^i_{\mu\nu}$ to be a random orthogonal matrix. In addition, for each $i$ we split the eigenvalues $\lambda^i_\sigma$ into two sets of identical size, corresponding to the positive and negative eigenvalues. We draw the positive eigenvalues from $\mathcal{U}([1,2])$ and take the negative eigenvalues to be exactly $-1$ times the positive eigenvalues.
We impose that the eigenvalues come in positive/negative pairs so that $\mathbb{E}[z_{0}]=0$.
We also choose the range $[1,2]$ because the two-layer, linear MLP with one datapoint, $x=1$, corresponds to a quadratic model with eigenvalues $\pm 1$ \cite{Belkin_quadratic,Pennington_second_order}. We can then think of this quadratic model as a simple deformation of the linear MLP.
The results for this set-up are shown in Figure \ref{fig:linear_meta_toy_dataset_ev1_2}.
We see that this quadratic model undergoes catapult dynamics and our bounds correctly predict the model does converge for a finite window above $\eta H_0=2$.

\begin{figure*}[!ht]
\centering
\begin{subfigure}[b]{0.3\textwidth}
\centering
\includegraphics[scale=.25]{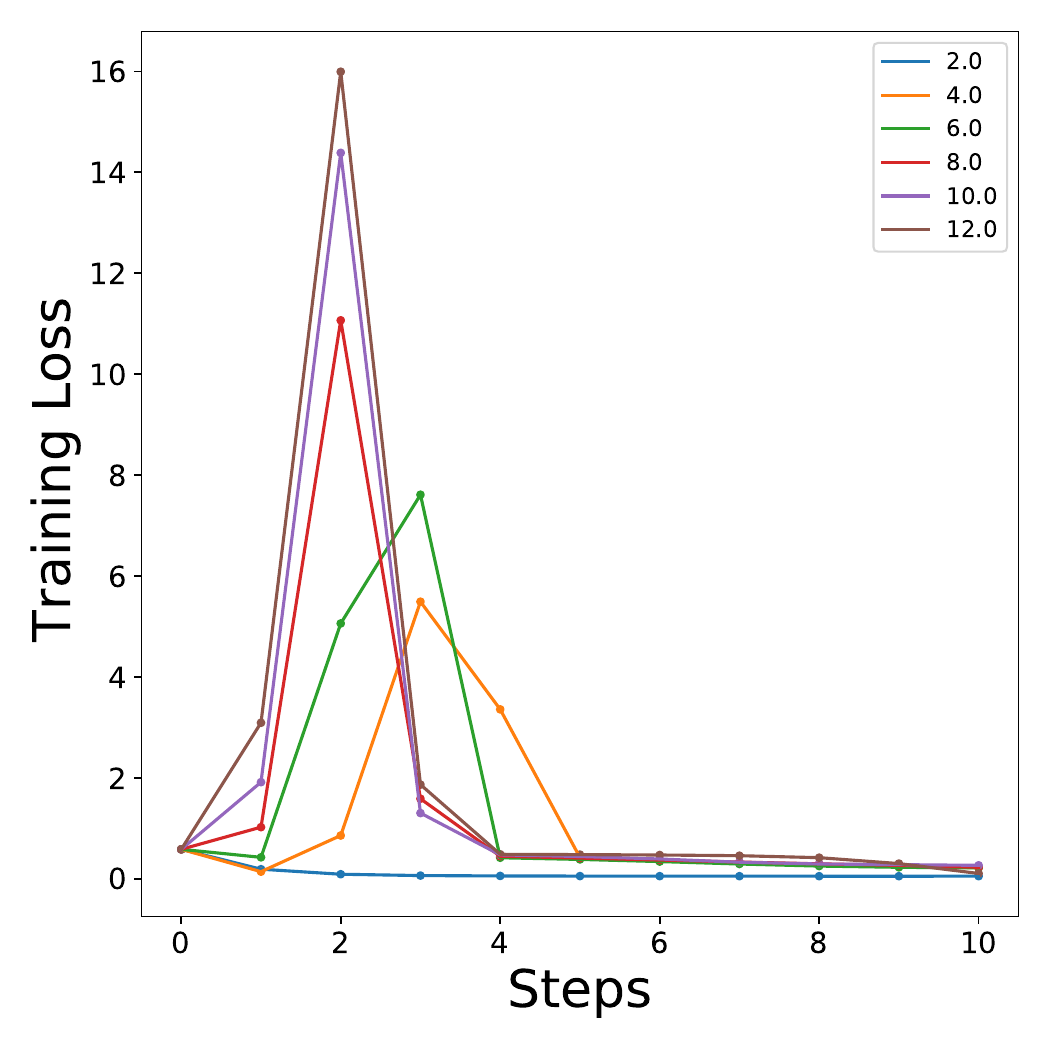}
\caption{}
\end{subfigure}
\hfill
\begin{subfigure}[b]{0.3\textwidth}
\centering
\includegraphics[scale=.25]{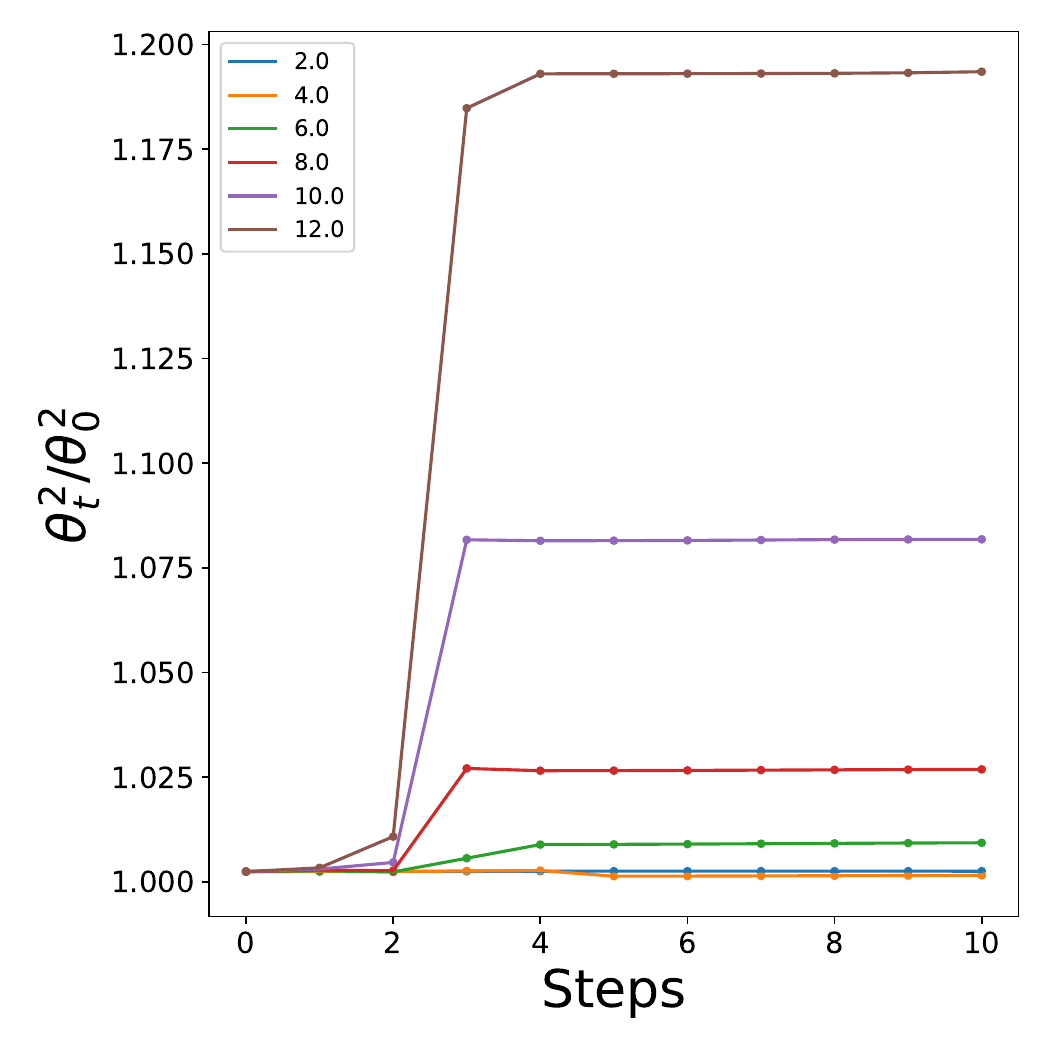}
\caption{}
\label{fig:MNIST_011_ReLU_one_hidden_layer_weight_norm}

\end{subfigure}
\hfill
\begin{subfigure}[b]{0.3\textwidth}
\centering
\includegraphics[scale=.25]{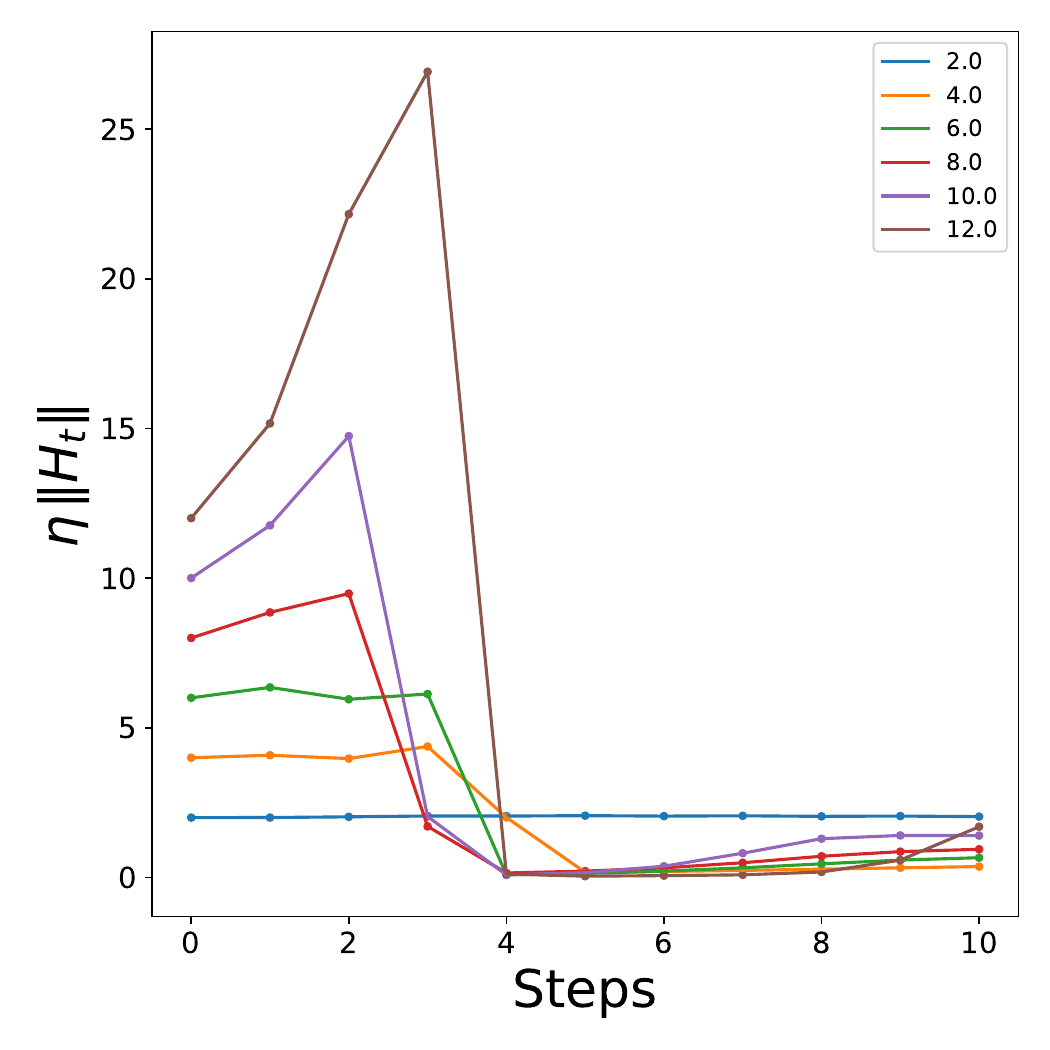}
\caption{}
\label{fig:MNIST_011_ReLU_one_hidden_layer_NTK}

\end{subfigure}
\centering
\begin{subfigure}[t]{0.3\textwidth}
\centering
\includegraphics[scale=.25]{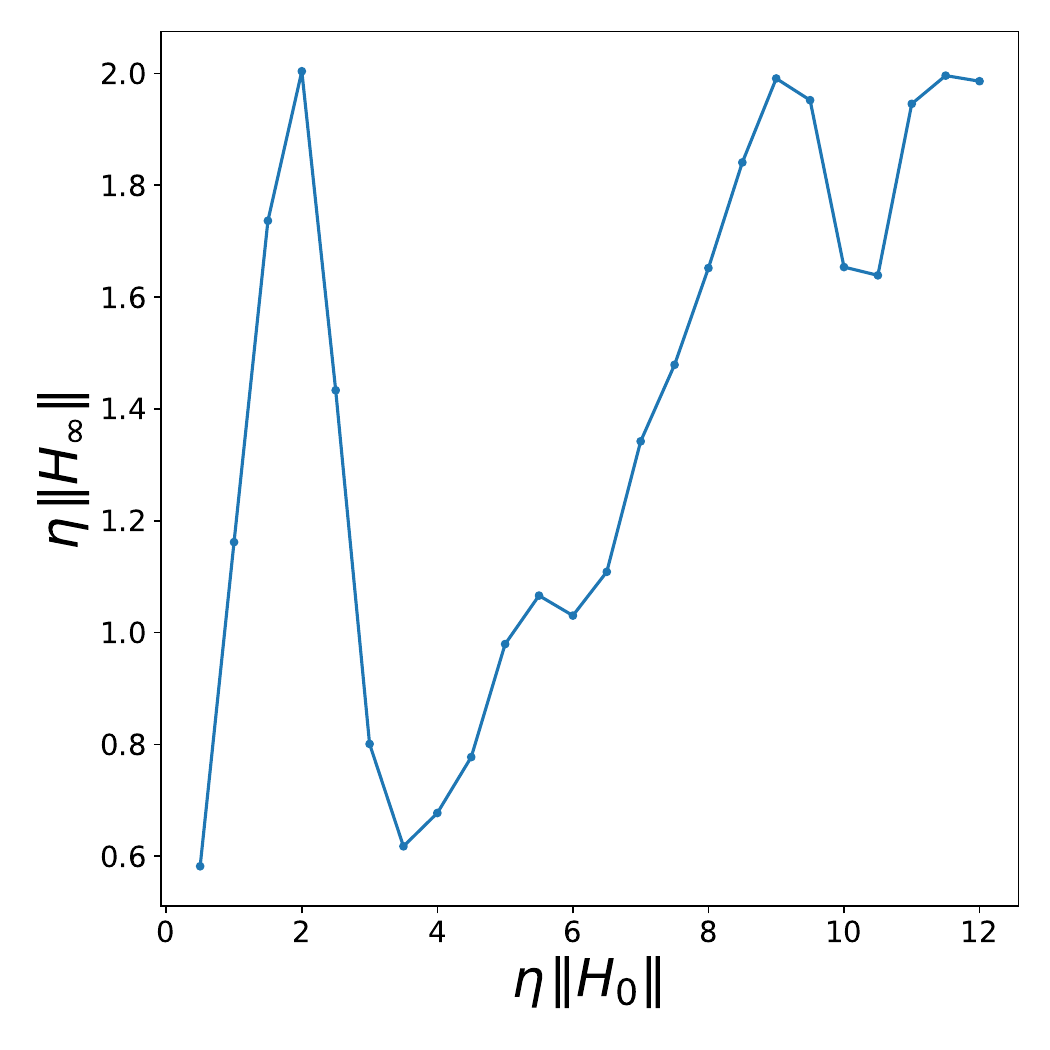}
\caption{}
\label{fig:MNIST_011_ReLU_one_hidden_layer_final_NTK}
\end{subfigure}
\hfill
\begin{subfigure}[t]{0.3\textwidth}
\centering
\includegraphics[scale=.25]{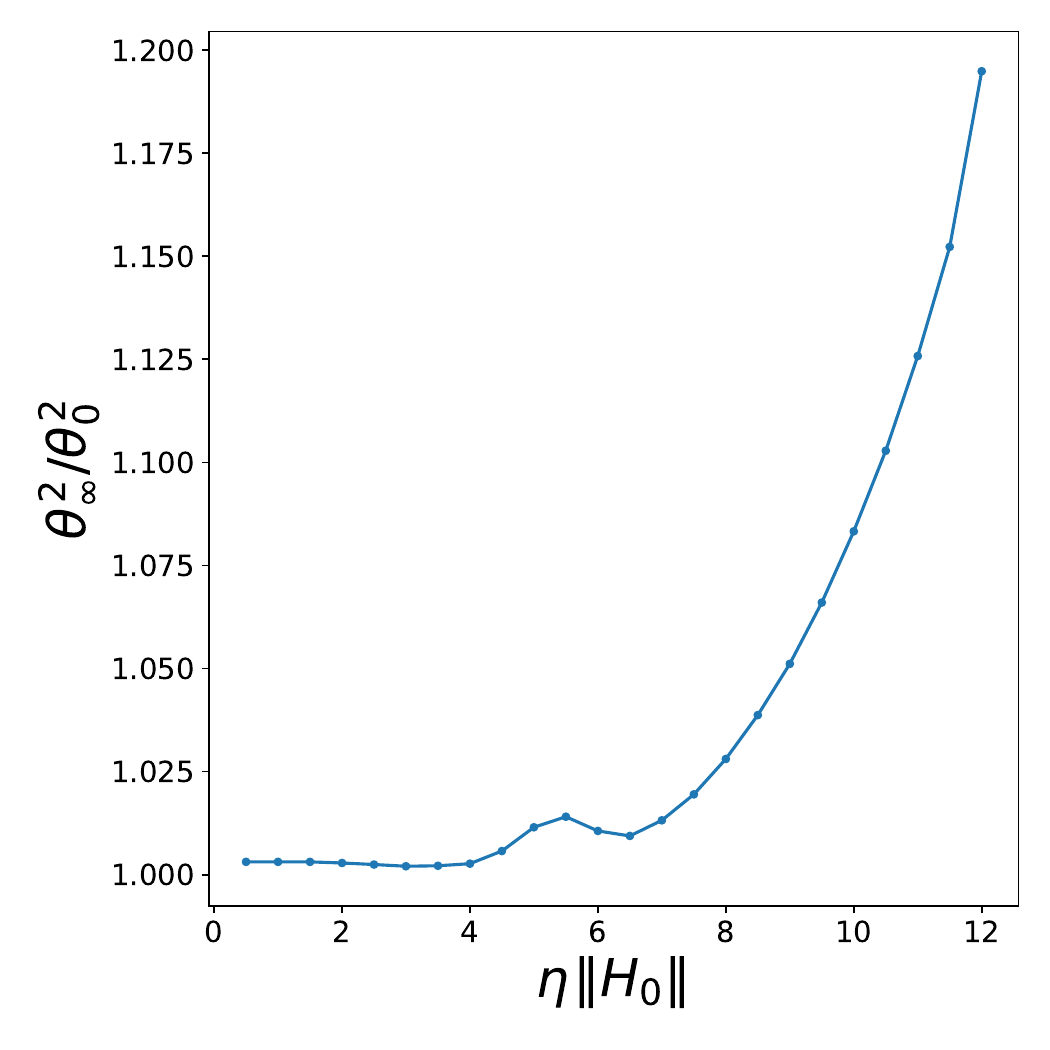}
\caption{}
\label{fig:MNIST_011_ReLU_one_hidden_layer_final_weight}

\end{subfigure}
\hfill
\begin{subfigure}[t]{0.3\textwidth}
\centering
\includegraphics[scale=.25]{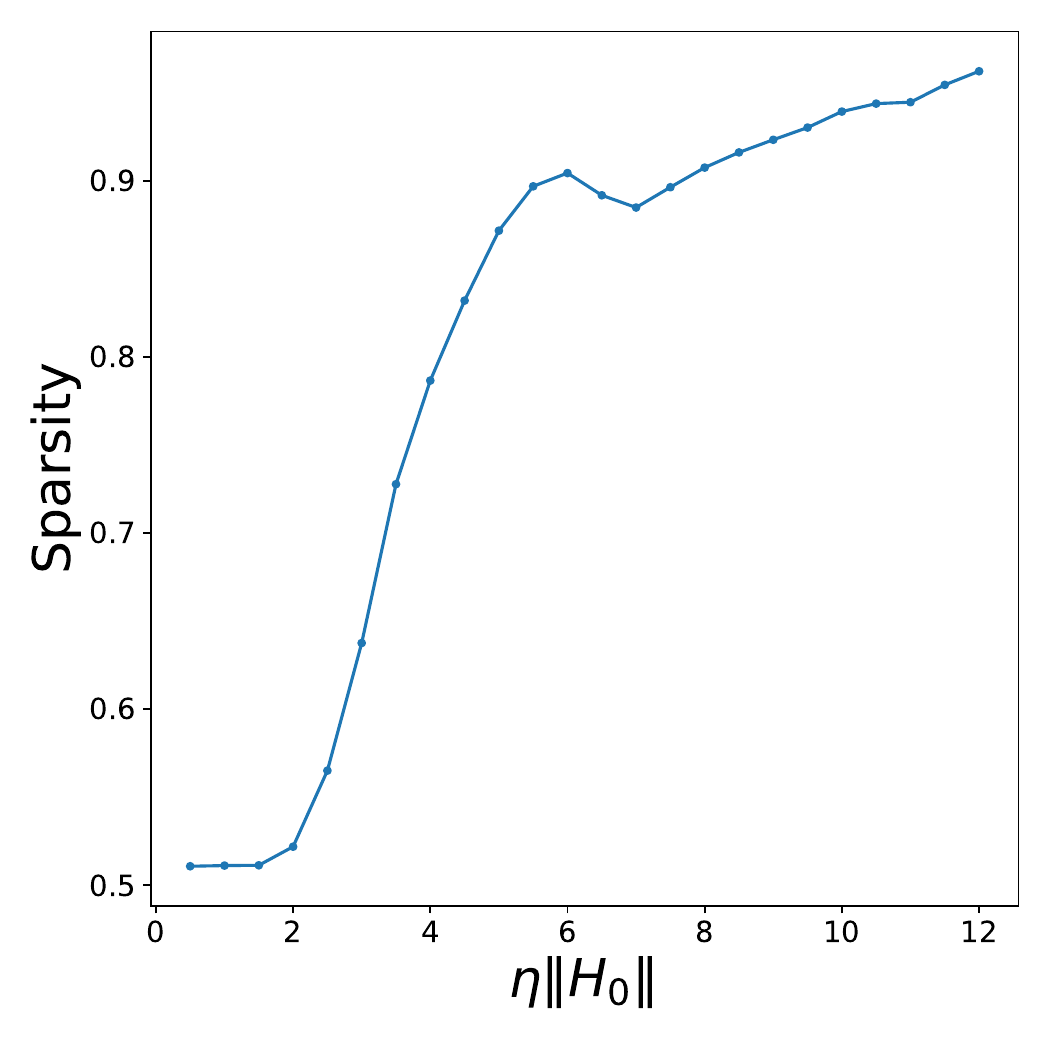}
\caption{}
\label{fig:MNIST_011_ReLU_one_hidden_layer_sparsity}
\end{subfigure}
\\
\hfill
\begin{subfigure}[t]{\textwidth}
\centering
\includegraphics[scale=.25]{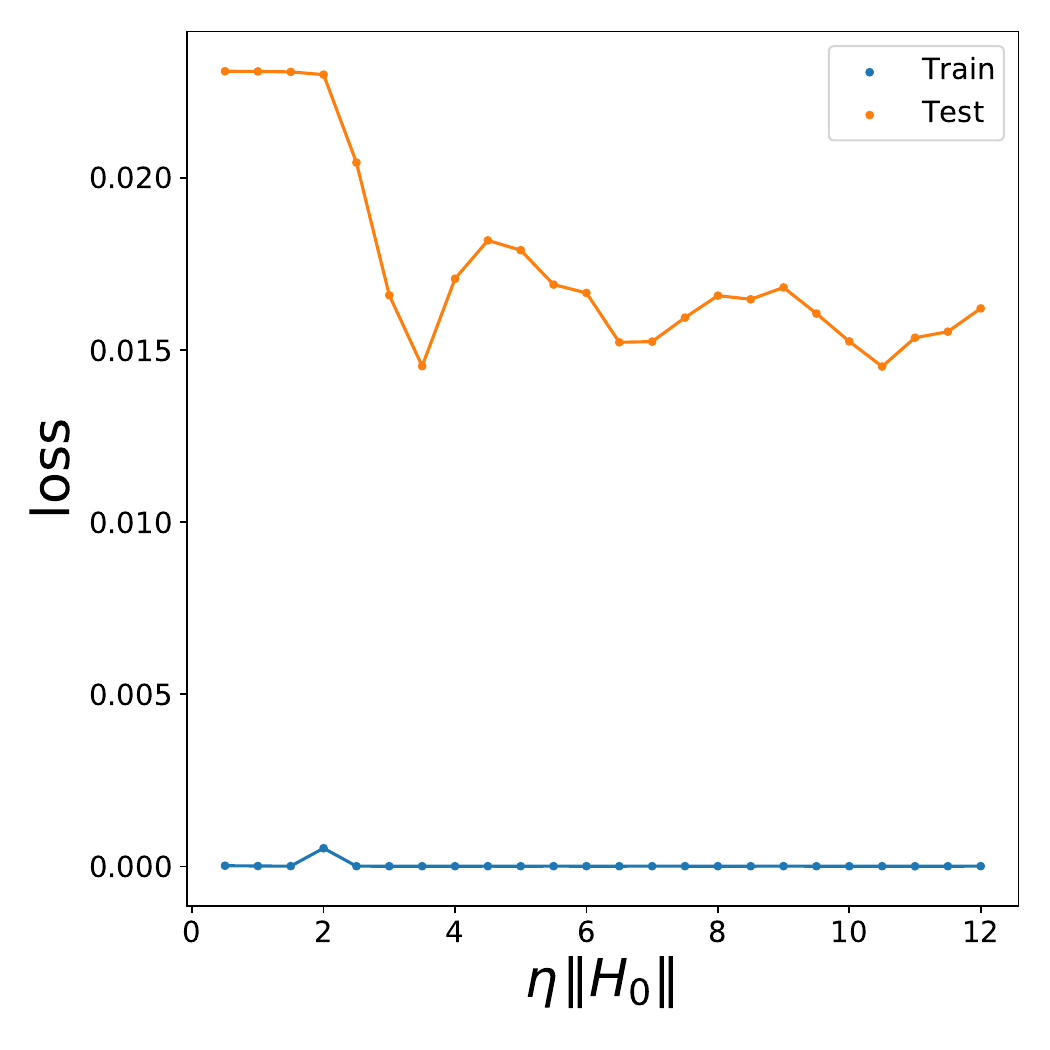}
\caption{}
\label{fig:MNIST_011_ReLU_one_hidden_layer_gen_loss}
\end{subfigure}
\caption{Results for the two-layer ReLU net trained on a two-class version of MNIST. The hidden layer has width 1024.
(a)-(c) give the early-time evolution of the loss, the weight norm, and $\eta\lambda_{\text{max}}(H_{\alpha\beta,t})$, respectively. The different lines in (a)-(c) correspond to different choices of $\eta\lambda_{\text{max}}(H_{\alpha\beta,0})$. (d)-(g) give the final values for $\eta \lambda_{\text{max}}(H_{\alpha\beta,t})$, $\bs{\theta}^2_t$, the sparsity of the activation map, and the generalization gap, respectively, for the converged model as a function of $\eta\lambda_{\text{max}}(H_{\alpha\beta,0})$. From (f) we observe that increasing the learning rate in the catapult phase promotes sparsity.
}
\label{fig:MNIST_011_ReLU_one_hidden_layer}
\end{figure*}

\subsection{Generic Homogenous Net}
We also perform a similar experiment for homogenous nets with one hidden layer of width 1024. To be concrete, we consider a net with $a_+=1$ and $a_-=1/2$. 
We train this model on the toy dataset $(x,y)=(1,0)$.
The results are shown in Figure \ref{fig:linear_scale_invariant_toy_dataset_half}. When $\eta H_0\lesssim 4$, the results are qualitatively the same as for the pure quadratic model. 
However, one new feature is this model converges for larger learning rates. 
In particular, for $\eta H_0=4.5$ we see that $\bs{\theta}_{t}^2$ initially grows during the catapult phase, before decreasing to a small value. 
This growth in $\bs{\theta}^2_t$ occurs because the product $\eta H_t>4$ when $z_{t}=O( \sqrt{n})$. 
From the update equation for $\bs{\theta}^2_t$ \eqref{eq:weight_norm_update_simple}, this means that $\bs{\theta}^2_t$ receives a large positive correction during training. However, eventually $\eta H_t$ is pushed below 4 and the model converges.

This result is consistent with our analytic results summarized in Table \ref{catares}. 
When $\eta$ obeys the bounds summarized there, the weight norm $\bs{\theta}_{t}^2$ must decrease monotonically. 
Since $\eta H_0=4.5$ exceeds our bounds, there is no guarantee $\bs{\theta}_{t}^2$ must decrease.
We can also observe in Figure \ref{fig:linear_scale_invariant_toy_dataset_half_final_weight} that there is a transition in $\bs{\theta}^2_{\infty}/\bs{\theta}^2_{0}$ when $\eta H_0\approx 4$. When $2\leq \eta H_0 \lesssim4$ the final weight norm is smaller than its value at initialization. As we approach $\eta H_0\rightarrow 4$ this trend reverses and there is a peak at $\eta H_0=4$, before $\bs{\theta}^2_t$ decreases again.
To make this change in the evolution of $\bs{\theta}_t^2$ as a function of $\eta$ more pronounced, we will study ReLU nets.

\subsection{ReLU Net}
In this section we will study the behavior of fully-connected ReLU nets trained on a two-class version of MNIST.
We will train the model to distinguish images of ``0" and ``1", which are assigned labels $-1$ and $1$, respectively. The training set has size 128 and the test set has size 2115.
Our analytic bounds for ReLU nets is restricted to $1d$ input data, so here the results will be purely empirical.\footnote{A comparison between analytics and numerics for one-dimensional data and ReLU nets will be given in Appendix \ref{app:scale_inv_MLP_exp}.}

We train a two-layer net with width 1024 using full-batch gradient descent.
The model is trained for each learning rate until the change in the training loss is $<10^{-8}$. The results are shown in Figure \ref{fig:MNIST_011_ReLU_one_hidden_layer}. The new feature, in comparison to the previous experiments, is that the final weight norm $\bs{\theta}_{\infty}^2$ can be greater than the value at initialization, see figures \ref{fig:MNIST_011_ReLU_one_hidden_layer_weight_norm} and \ref{fig:MNIST_011_ReLU_one_hidden_layer_final_weight}.
Despite the increase in $\bs{\theta}^2_t$ the model still generalizes well, see Figure \ref{fig:MNIST_011_ReLU_one_hidden_layer_gen_loss}.
We conjecture that the model is able to generalize well, despite the increase in $\bs{\theta}^{2}_t$, because the activation map becomes sparse in the catapult phase, see Figure \ref{fig:MNIST_011_ReLU_one_hidden_layer_sparsity}.
The sparsity of a given layer is the fraction of nodes in that layer which are zero after acting with the ReLU function for a single input.
To get Figure \ref{fig:MNIST_011_ReLU_one_hidden_layer_sparsity} we averaged the sparsity of the activation map over all inputs.

Finally, we can note that the final value of $\eta \lambda_{\text{max}}(H_{\alpha\beta,t})$, which measures the sharpness of the final minima, first decreases in the catapult phase before increasing again to the edge of stability, $\eta \lambda_{\text{max}}(H_t)=2$, see figure \ref{fig:MNIST_011_ReLU_one_hidden_layer_final_NTK}.
This is consistent with existing work on the edge of stability phenomena for the top eigenvalue of the NTK \cite{https://doi.org/10.48550/arxiv.2103.00065,Pennington_second_order}.

\section{Conclusions}
In this work we have studied the properties of simple machine learning models, the quadratic model and two-layer, homogenous nets, for super-critical learning rates. We have proven that these models can converge beyond the stability threshold of the corresponding linearized model. 
In the process, we have shown that training these models for super-critical learning rates can lead to an implicit $L_2$ regularization where the weight norm $\bs{\theta}^2$ decreases significantly during training. \textcolor{black}{While our theoretical analysis mainly rely on weight norm decay, we emphasize that the catapult phase exhibits richer behavior beyond this. For instance,}
we have also empirically studied ReLU nets for even larger learning rates, $\eta H_0\gtrsim 4$, and showed that training models in this range leads to an increase in both the sparsity and the weight norm. \textcolor{black}{The ReLU nets can generalize well even when the weight norm increases during training, likely due to emergent sparsity in the activation map.} These results, both analytical and empirical, shed new light on the implicit biases of gradient descent at large learning rates.

There are many interesting open questions to consider. One is to understand how to weaken our bounds, which in general can miss large portions of the catapult regime. The reason our bounds can be too strong is because they are designed to hold for all possible configurations of $\bs{\theta}$.
In particular, they hold when inequalities of the form \eqref{eq:bound_H_theta} are saturated. 
To derive weaker bounds we need a better understanding on how the weights evolve on average for super-critical learning rates. 
In addition, proving that ReLU nets can converge for $4\leq \eta H_0 \lesssim12$ remains an open question.

Another question is: why do models trained in the catapult phase generalize well?
Our empirical and analytic results align with the expectation that large learning rates can lead to implicit regularization \cite{DBLP:journals/corr/abs-1907-04595} which improves model performance.
It would be interesting to understand what features of the catapult mechanism lead to a small generalization error for more generic, wide neural nets.

\section*{Acknowledgements}
We thank Yasaman Bahri, Daniel Roberts, Yi-Hsien Du, and Han Zheng for discussions.
We additionally thank Daniel Roberts for comments on the draft.
DM is supported by the NSF grant PHY-2014071. JL is supported in part by International Business Machines (IBM) Quantum through the Chicago Quantum Exchange, and the Pritzker School of Molecular Engineering at the University of Chicago through AFOSR MURI (FA9550-21-1-0209). MC and JL are supported in part by the University of Pittsburgh, School of Computing and Information, Department of Computer Science, Pitt Cyber, PQI Community Collaboration Awards, John C. Mascaro Faculty Scholar in Sustainability, NASA under award number 80NSSC25M7057, and Fluor Marine Propulsion LLC (U.S. Naval Nuclear Laboratory) under award number 140449-R08. This research used resources of the Oak Ridge Leadership Computing Facility, which is a DOE Office of Science User Facility supported under Contract DE-AC05-00OR22725.

\bibliography{main}
\bibliographystyle{icml2021}

\newpage
\onecolumn
\appendix

\section{Early Time Dynamics}
\label{app:early_time}
In this appendix, we study the early time dynamics of quadratic models and two-layer homogenous MLPs.
The goal of this appendix is to show that when $\eta \lambda_{\text{max}}(H_{\alpha\beta,0})>2$, the loss grows exponentially quickly at early times.
We will then estimate the scale at which perturbation theory breaks down.
This appendix builds off of the theoretical analysis done in \cite{lewkowycz2020large}.
We will also study the evolution of the weight norm and show that it only receives large corrections near the peak of the catapult phase.

The analysis in this appendix will be important in proving that models trained on generic datasets can converge for super-critical learning rates but will not be necessary for proving convergence on the toy dataset $(x,y)=(1,0)$.

\subsection{Catapult Phase in the Quadratic Model}
Here we will study the pure quadratic model and the quadratic model with bias. 
The general quadratic model is:
\begin{align}
z(\bs{x}_{\alpha})=\bs{\theta}^T\bs{\phi}_{\alpha}+\frac{\zeta}{2}\bs{\theta}^{T}\bs{\psi}_{\alpha}\bs{\theta},
\end{align}
where $\bs{\theta}\in\mathbb{R}^{n}$, $\zeta^2=O(1/n)$ and $n\gg1$.
We also use the notation, $\bs{\phi}_{\alpha}\equiv\bs{\phi}(\bs{x}_{\alpha})$ and $\bs{\psi}_{\alpha}\equiv\bs{\psi}(\bs{x}_{\alpha})$.
We train the model with MSE:
\begin{align}
L=\frac{1}{2D}\sum\limits_{\alpha=1}^D \epsilon_{\alpha}^2=\frac{1}{2D}\sum\limits_{\alpha=1}^D(z_{\alpha}-y_{\alpha})^2.
\end{align}
Here $y_{\alpha}\in\mathbb{R}$ and $\bs{x}_{\alpha}\in\mathbb{R}^{d}$.
The weights evolve as:
\begin{align}
\bs{\theta}_{t+1}=\bs{\theta}_{t}-\frac{\eta}{D}\sum\limits_{\alpha=1}^{D}\epsilon_{\alpha,t}(\bs{\phi}_{\alpha}+\zeta \bs{\psi}_{\alpha}\bs{\theta}_{t}).
\end{align}

To study the quadratic model with bias we impose the condition,
\begin{align}
\bs{\psi}_{\alpha}\bs{\phi}_{\beta}=0, \quad \forall \ \alpha,\beta.
\end{align}
To study the pure quadratic model we can set $\bs{\phi}=0$.
The NTK in this model is
\begin{align}
H_{\alpha\beta}=\frac{1}{D}(\bs{\phi}_{\alpha}^T\bs{\phi}_{\beta}+\zeta^2\bs{\theta}^T\bs{\psi}_{\alpha}\bs{\psi}_{\beta}\bs{\theta}).
\end{align}

We draw the initial weights from a normal distribution with zero mean and unit variance, $\bs{\theta}\sim \mathcal{N}(0,\mathbb{I}_{n\times n})$.
Then at initialization we have:
\begin{align}
\mathbb{E}[z_{\alpha,0}]&=\frac{\zeta}{2}\Tr(\bs{\psi}_{\alpha}), 
\\ 
\mathbb{E}[z_{\alpha,0}z_{\beta,0}]&=\bs{\phi}_{\alpha}^{T}\bs{\phi}_{\beta}+\frac{\zeta^2}{2}\Tr(\bs{\psi}_{\alpha}\bs{\psi}_{\beta})+\mathbb{E}[z_{\alpha}]\mathbb{E}[z_{\beta}],
\\
\mathbb{E}[H_{\alpha\beta,0}]&=\frac{1}{D}\left(\bs{\phi}^{T}_{\alpha}\bs{\phi}_{\beta}+\zeta^2\Tr(\bs{\psi}_{\alpha}\bs{\psi}_{\beta})\right),
\\
\mathbb{E}[\bs{\theta}_0^2]&=n.
\end{align}
We will choose the (meta-)feature functions such that $z_{\alpha,0}, H_{\alpha\beta,0} = O(\zeta^0)$. To impose this, we will assume that the eigenvalues of the meta-feature functions are order-one, $\lambda_{i}(\bs{\psi}_{\alpha})= O(\zeta^0)$ and come in approximately positive/negative pairs.
In this case:
\begin{align}
\zeta \Tr(\bs{\psi}_{\alpha})&=\zeta\sum\limits_{a=1}^{n}\lambda_{a}(\bs{\psi}_{\alpha})\ll1,
\\
\zeta^2\Tr(\bs{\psi}^2_{\alpha})&=\zeta^2\sum\limits_{a=1}^{n}\lambda_{a}^2(\bs{\psi}_{\alpha})=O(\zeta^0).
\end{align}
Finally, we impose $\bs{\phi}_{\alpha}^2=O(\zeta^0)$.

The update equations for $\epsilon_{\alpha}$ and $H_{\alpha\beta}$ are:
\begin{align}
\epsilon_{\alpha,t+1}&=\sum\limits_{\beta=1}^{D}(\delta_{\alpha\beta}-\eta H_{\alpha\beta,t})\epsilon_{\beta,t}+\frac{\eta^2\zeta^3}{2D^2}\sum\limits_{\beta,\gamma=1}^{D}\epsilon_{\beta,t}\epsilon_{\gamma,t}\bs{\theta}^{T}_t\bs{\psi}_{\beta}\bs{\psi}_{\alpha}\bs{\psi}_{\gamma}\bs{\theta}_t, \label{eq:pure_quad_z_mult}
\\
H_{\alpha\beta,t+1}&=H_{\alpha\beta,t}-\frac{\eta\zeta^3}{D}\sum\limits_{\gamma=1}^{D}\left[\epsilon_{\gamma,t}\bs{\theta}^{T}_{t}\bs{\psi}_{\gamma}\bs{\psi}_{\alpha}\bs{\psi}_{\beta}\bs{\theta}_t+(\alpha\leftrightarrow \beta)\right]+\frac{\eta^2\zeta^4}{D^{2}}\sum\limits_{\gamma,\rho=1}^{D}\epsilon_{\gamma,t}\epsilon_{\rho,t}\bs{\theta}^{T}_t\bs{\psi}_{\gamma}\bs{\psi}_{\alpha}\bs{\psi}_{\beta}\bs{\psi}_{\rho}\bs{\theta}_t ,\label{eq:pure_quad_H_mult}
\end{align}
where $\delta_{\alpha\beta}$ is the Kronecker delta function.
Next we will show that when $\eta \lambda_{\text{max}}(H_{\alpha\beta,0})>2$ and $\zeta \ll1$ the loss increases exponentially quickly at early times and that small $\zeta$-perturbation theory breaks down when $t=O(\log(\zeta^{-1}))$.
First we will show that when $t\ll \log(\zeta^{-1})$ we can approximate the update equations by:
\begin{align}
\epsilon_{\alpha,t+1}&\approx \sum\limits_{\beta=1}^{D}(\delta_{\alpha\beta}-\eta H_{\alpha\beta,t})\epsilon_{\beta,t},\label{eq:approx_z_update}
\\
H_{\alpha\beta,t+1}&\approx H_{\alpha\beta,t}. \label{eq:approx_H_update}
\end{align}
That is, we are dropping terms in \eqref{eq:pure_quad_z_mult} and \eqref{eq:pure_quad_H_mult} which are explicitly suppressed in $\zeta$.
When this approximation holds the NTK $H_{\alpha\beta,t}$ is constant and the error $\epsilon_{\alpha,t}$ increases fastest in the direction parallel to the top eigenvector of the NTK.
To be more explicit, we write
\begin{align}
\epsilon_{\alpha,t}=\sum\limits_{i=1}^{D}c^{i}_{t}e^{i}_{\alpha,t},
\end{align}
where $e^{i}_{\alpha,t}$ is the $i^{\text{th}}$ eigenvector of the NTK at time $t$. 
Assuming \eqref{eq:approx_z_update} and \eqref{eq:approx_H_update} hold, the eigenvectors are approximately constant and the coefficients $c^i_{t}$ are given by:
\begin{align}
c^i_{t}=\left(1-\eta \lambda_{i}(H_{\alpha\beta,0})\right)^t c^{i}_{t=0},\label{eq:0th_order_solution_coefficients}
\end{align}
so the error grows fastest in the direction of the top eigenvector of the NTK.

To show that \eqref{eq:approx_z_update} and \eqref{eq:approx_H_update} are valid at early times we will bound the sub-leading terms in \eqref{eq:pure_quad_z_mult} and \eqref{eq:pure_quad_H_mult}.
We will first bound the size of the $\zeta^3$ term in \eqref{eq:pure_quad_z_mult} using Cauchy-Schwarz inequalities:
\begin{align}
\bigg|\frac{\eta^2\zeta^3}{2D^2}\sum\limits_{\beta,\gamma=1}^{D}\epsilon_{\beta,t}\epsilon_{\gamma,t}\bs{\theta}^{T}_t\bs{\psi}_{\beta}\bs{\psi}_{\alpha}\bs{\psi}_{\gamma}\bs{\theta}_t\bigg| 
&\leq \frac{\eta^2\zeta^3}{2D^2}\sum\limits_{\beta,\gamma=1}^{D}|\epsilon_{\beta,t}\epsilon_{\gamma,t}\bs{\theta}^{T}_t\bs{\psi}_{\beta}\bs{\psi}_{\alpha}\bs{\psi}_{\gamma}\bs{\theta}_t|
\nonumber
\\
&\leq \frac{\eta^2\zeta^3}{2D^2}\bs{\theta}^2_{t}\sqrt{\lambda_{\text{max}}(\bs{\psi}^2_{\alpha})}\sum\limits_{\beta,\gamma=1}^{D}|\epsilon_{\beta,t}\epsilon_{\gamma,t}|\sqrt{\lambda_{\text{max}}(\bs{\psi}^2_{\beta})\lambda_{\text{max}}(\bs{\psi}^2_{\gamma})}
\nonumber
\\
& \leq \frac{\eta^2\zeta^3}{2D^2}\bs{\theta}^2_{t}\sqrt{\lambda_{\text{max}}(\bs{\psi}_{\alpha}^{2})}\sum\limits_{\beta=1}^{D}\epsilon_{\beta,t}^{2}\sum\limits_{\gamma=1}^{D}\lambda_{\text{max}}(\bs{\psi}_{\gamma}^{2})
\nonumber
\\
&= O\left(\zeta^3\bs{\theta}_{t}^2 D^{-1} \sum\limits_{\beta=1}^D\epsilon^{2}_{\beta,t}\right). \label{eq:z_last_term}
\end{align}
To obtain the last line we used the assumption that the eigenvalues of the meta-feature function are all $O(1)$.

We can similarly bound the size of the $\zeta^3$ term in \eqref{eq:pure_quad_H_mult} using Cauchy-Schwarz inequalities:
\begin{align}
\left|\frac{\eta \zeta^3}{D}\sum\limits_{\gamma=1}^{D}\epsilon_{\gamma,t}\bs{\theta}^{T}_{t}\bs{\psi}_{\gamma}\bs{\psi}_{\alpha}\bs{\psi}_{\beta}\bs{\theta}_{t}\right|
&\leq \frac{\eta \zeta^3}{D} \bs{\theta}^{2}_{t} \sqrt{\lambda_{\text{max}}(\bs{\psi}_{\alpha}^{2}) \lambda_{\text{max}}(\bs{\psi}_{\beta}^{2})}\sum\limits_{\gamma=1}^{D}|\epsilon_{\gamma,t}| \sqrt{\lambda_{\text{max}}(\bs{\psi}_{\gamma}^{2})}
\nonumber
\\
&\leq  \frac{\eta \zeta^3}{D} \bs{\theta}^{2}_{t} \sqrt{ \lambda_{\text{max}}(\bs{\psi}_{\alpha}^{2}) \lambda_{\text{max}}(\bs{\psi}_{\beta}^{2})}\sqrt{\sum\limits_{\gamma=1}^D\epsilon_{\gamma,t}^2}\sqrt{\sum\limits_{\rho=1}^{D} \lambda_{\text{max}}(\bs{\psi}^{2}_{\rho})} 
\nonumber
\\
& = O\left(\zeta^3 \bs{\theta}_{t}^2 D^{-1/2}\sqrt{\sum\limits_{\gamma=1}^D\epsilon_{\gamma,t}^{2}}\right). \label{eq:H_second_term}
\end{align}
Finally, we can bound the $\zeta^4$ term of \eqref{eq:pure_quad_H_mult}:
\begin{align}
\left|\frac{\eta^2 \zeta^4}{D^2}\sum\limits_{\gamma,\rho=1}^{D}\epsilon_{\gamma,t}\epsilon_{\rho,t}\bs{\theta}^{T}_t\bs{\psi}_{\gamma}\bs{\psi}_{\alpha}\bs{\psi}_{\beta}\bs{\psi}_{\rho}\bs{\theta}_t\right| &
\leq \frac{\eta^2\zeta^4}{D^2}\bs{\theta}_{t}^2 \sqrt{\lambda_{\text{max}}(\bs{\psi}_{\alpha}^{2})\lambda_{\text{max}}(\bs{\psi}_{\beta}^{2})}\sum\limits_{\gamma,\rho=1}^{D}|\epsilon_{\gamma,t}\epsilon_{\rho,t}|\sqrt{\lambda_{\text{max}}(\bs{\psi}_{\gamma}^{2})\lambda_{\text{max}}(\bs{\psi}_{\rho}^{2})}
\nonumber
\\
&\leq \frac{\eta^2\zeta^4}{D^2}\bs{\theta}_{t}^2\sqrt{\lambda_{\text{max}}(\bs{\psi}_{\alpha}^{2})\lambda_{\text{max}}(\bs{\psi}_{\beta}^{2})}\sum\limits_{\gamma=1}^{D}\epsilon_{\gamma,t}^{2} \sum\limits_{\rho=1}^{D}\lambda_{\text{max}}(\bs{\psi}_{\rho}^{2})
\nonumber
\\
& = O\left(\zeta^{4}\bs{\theta}_{t}^{2}D^{-1}\sum\limits_{\gamma}\epsilon_{\gamma,t}^{2}\right).\label{eq:H_third_term}
\end{align}

Next, we can argue that \eqref{eq:z_last_term}-\eqref{eq:H_third_term} are all small at initialization when $\zeta \ll1$. We will use that at $t=0$:
\begin{align}
\bs{\theta}_{0}^2&= O(\zeta^{-2} D^0),
\\
|\epsilon_{\beta,0}^{2}|&= O(1),
\\
H_{\alpha\beta,0}&= O(\zeta^0 D^{-1}).
\end{align}
If we use these estimates in \eqref{eq:z_last_term}-\eqref{eq:H_third_term}, we find that at $t=0$ these terms scale like:
\begin{align}
\bigg|\frac{\eta^2\zeta^3}{2D^2}\sum\limits_{\beta,\gamma=1}^{D}\epsilon_{\beta,0}\epsilon_{\gamma,0}\bs{\theta}^{T}_0\bs{\psi}_{\beta}\bs{\psi}_{\alpha}\bs{\psi}_{\gamma}\bs{\theta}_0\bigg| &= O(\zeta), \label{eq:z_sub-leading_init}
\\
\left|\frac{\eta \zeta^3}{D}\sum\limits_{\gamma=1}^{D}\epsilon_{\gamma,0}\bs{\theta}^{T}_{0}\bs{\psi}_{\gamma}\bs{\psi}_{\alpha}\bs{\psi}_{\beta}\bs{\theta}_{0}\right|&=O(\zeta)\label{eq:H_sub-leading_init_second},
\\
\left|\frac{\eta^2\zeta^4}{D^2}\sum\limits_{\gamma,\rho=1}^{D}\epsilon_{\gamma,0}\epsilon_{\rho,0}\bs{\theta}^{T}_0\bs{\psi}_{\gamma}\bs{\psi}_{\alpha}\bs{\psi}_{\beta}\bs{\psi}_{\rho}\bs{\theta}_0\right|&=O(\zeta^2).\label{eq:H_sub-leading_init_third}
\end{align}
At initialization $\epsilon_{\alpha,0}=O(1)$, so the sub-leading term \eqref{eq:z_sub-leading_init} is negligible if $\zeta \ll1$.
Similarly, at $t=0$ we have $H_{\alpha\beta,0}=O(D^{-1})$, so we can drop the sub-leading terms \eqref{eq:H_sub-leading_init_second}-\eqref{eq:H_sub-leading_init_third} if $\zeta D\ll1$.
Both conditions are satisfied if assume $\zeta\ll1$ and $D$ is generic.

This proves that, at $t=0$, we can ignore the sub-leading terms in the small $\zeta$ expansion. 
We can now determine at what scale perturbation theory breaks down by finding at what scale the sub-leading terms \eqref{eq:z_last_term}-\eqref{eq:H_third_term} are of the same order as the leading order terms.
If we plug in the $0^{\text{th}}$-order solutions, \eqref{eq:approx_H_update}-\eqref{eq:0th_order_solution_coefficients} into the sub-leading terms \eqref{eq:z_last_term}-\eqref{eq:H_third_term}, we find that perturbation theory breaks down when $c^{\text{max}}_{t}=O(\zeta^{-1})$, which happens when $t=O(\log(\zeta^{-1}))$. 

\subsection{Weight Norm in the Quadratic Model}
In this section we will study the evolution of $\bs{\theta}^2$ and $\bs{\phi}_{\alpha}^T\bs{\theta}$ and show that small $\zeta$-perturbation theory is valid when $t\ll \log(\zeta^{-1})$. 
Specifically, we will show:
\begin{align}
\bs{\theta}_{t}^2=\bs{\theta}_{0}^2+O(1) \quad \text{for } \ \ t \ll \log(\zeta^{-1}).
\end{align}

The update equation for $\bs{\theta}_{t}^2$ is:
\begin{align}
\bs{\theta}_{t+1}^2&=\bs{\theta}^2_t-\frac{2\eta}{D}\sum\limits_{\alpha=1}^D\left(\epsilon_{\alpha,t}\bs{\phi}_{\alpha}^T\bs{\theta}_t+\zeta\epsilon_{\alpha,t}\bs{\theta}^T_t\bs{\psi}_{\alpha}\bs{\theta}_t\right)+\frac{\eta^2}{D}\sum\limits_{\alpha,\beta=1}^D\epsilon_{\alpha,t}\epsilon_{\beta,t}H_{\alpha\beta,t}
\nonumber
\\
&=\bs{\theta}^2_t-\frac{4\eta}{D}\sum\limits_{\alpha=1}\epsilon_{\alpha,t}z_{\alpha,t}+\frac{\eta^2}{D}\sum\limits_{\alpha,\beta=1}^D\epsilon_{\alpha,t}\epsilon_{\beta,t}H_{\alpha\beta,t}+\frac{2\eta}{D}\sum\limits_{\alpha=1}^{D}\epsilon_{\alpha,t}\bs{\phi}_{\alpha}^T\bs{\theta}_{t}.
\label{eq:weight_norm_early_time}
\end{align}
With the exception of the last term, we already know the time evolution of the right-hand side of \eqref{eq:weight_norm_early_time}.
For early times, $t\ll \log(\zeta^{-1})$, the terms $\epsilon_{\alpha,t}$ and $z_{\alpha,t}$ grow exponentially quickly and the NTK $H_{\alpha\beta,t}$ is approximately constant, see \eqref{eq:approx_z_update}-\eqref{eq:0th_order_solution_coefficients}. 
In addition, at initialization we have $\bs{\theta}^2_0=O(\zeta^{-2})$.
Therefore, these sub-leading terms in $\zeta$ become comparable to $\bs{\theta}^2_0$ when $t=O(\log(\zeta^{-1}))$.
This is the same scale at which perturbation theory breaks down for the $z_{\alpha,t}$ and $H_{\alpha\beta,t}$ update equations, see the previous section.

In the pure quadratic model $\bs{\phi}_{\alpha}=0$ and we could stop here.
The above analysis proves that, in the pure quadratic model, we have:
\begin{align}
\bs{\theta}_{t}^2= \bs{\theta}_{0}^2 + O(1), \quad \text{for} \ \ t\ll\log(\zeta^{-1}). 
\end{align}
To complete the proof for the quadratic model with bias we need to study the last term in \eqref{eq:weight_norm_early_time}.
Using our assumption that $\bs{\psi}_{\alpha}\bs{\phi}_{\beta}=0$, we find its update equation is:
\begin{align}
\bs{\phi}_{\alpha}^T\bs{\theta}_{t+1}=\bs{\phi}_{\alpha}^T\bs{\theta}_{t}-\frac{\eta}{D}\sum\limits_{\beta=1}^D \bs{\phi}_{\alpha}^T\bs{\phi}_{\beta}\epsilon_{\beta,t},
\label{eq:update_phitheta_bias}
\end{align}
whose solution we can find in closed form:
\begin{align}
\bs{\phi}_{\alpha}^T\bs{\theta}_{t}=\bs{\phi}_{\alpha}^T\bs{\theta}_{0}-\frac{\eta}{D}\sum\limits_{\beta=1}^{D}\sum\limits_{i=1}^{t-1}\bs{\phi}_{\alpha}^T\bs{\phi}_{\beta}\epsilon_{\beta,i}.
\label{eq:exact_solution_phitheta}
\end{align}

The time evolution of the last term in \eqref{eq:weight_norm_early_time} is then:
\begin{align}
\frac{2\eta}{D}\sum\limits_{\alpha=1}^{D}\epsilon_{\alpha,t}\bs{\phi}_{\alpha}^T\bs{\theta}_{t}=\frac{2\eta}{D}\sum\limits_{\alpha=1}^{D}\epsilon_{\alpha,t}\bs{\phi}_{\alpha}^{T}\left(\bs{\theta}_{0}-\frac{\eta}{D}\sum\limits_{\beta=1}^{D}\sum\limits_{i=1}^{t-1}\bs{\phi}_{\beta}\epsilon_{\beta,i}\right).
\end{align}
Note that this is an exact solution and does not require any assumptions on $t$.

We can now bound the size of the last term in \eqref{eq:weight_norm_early_time}:
\begin{align}
\left|\frac{2\eta}{D}\sum\limits_{\alpha=1}^{D}\epsilon_{\alpha,t}\bs{\phi}_{\alpha}^T\bs{\theta}_{t} \right|&= \frac{2\eta}{D}\left|\sum\limits_{\alpha=1}^{D}\epsilon_{\alpha,t}\bs{\phi}_{\alpha}^{T}\left(\bs{\theta}_{0}-\frac{\eta}{D}\sum\limits_{\beta=1}^{D}\sum\limits_{i=1}^{t-1}\bs{\phi}_{\beta}\epsilon_{\beta,i}\right) \right|
\nonumber
\\
&\leq \frac{2\eta}{D}
\bigg(
\left|\sum\limits_{\alpha=1}^{D}\epsilon_{\alpha,t}\bs{\phi}_{\alpha}^{T}\bs{\theta}_{0}\right|
+
\frac{\eta}{D}\sum\limits_{\alpha,\beta=1}^{D}\sum\limits_{i=1}^{t-1}|\epsilon_{\alpha,t}\bs{\phi}_{\alpha}^{T}\bs{\phi}_{\beta}\epsilon_{\beta,i}
|
\bigg)
\nonumber
\\
&
\leq\frac{2\eta}{D}
\bigg(
\sqrt{\bs{\theta}_{0}^{2}\lambda_{\text{max}}(\bs{\phi}^T_{\alpha}\bs{\phi}^T_{\beta})
\sum\limits_{\alpha=1}^D\epsilon_{\alpha}^2}
+
\frac{\eta}{D}(t-1)\lambda_{\text{max}}(\bs{\phi}^T_{\alpha}\bs{\phi}_{\beta})\sum\limits_{\alpha=1}^{D}\epsilon_{\alpha,t}^2
\bigg).
\label{eq:bound_last_term_thetasq}
\end{align}
To go from the second to the third line we used Cauchy-Schwarz inequalities and also replaced $\epsilon_{\beta,i}\rightarrow \epsilon_{\beta,t}$ for each $i$. This gives a very weak, but valid, upper bound because the errors $\epsilon_{\beta,t}$ grow exponentially quickly when $\eta \lambda_{\text{max}}(H_{\alpha\beta,0})>2$.

If we define our model such that $D^{-1}\lambda_{\text{max}}(\bs{\phi}^T_{\alpha}\bs{\phi}_{\beta})=O(1)$, then the first term in \eqref{eq:bound_last_term_thetasq} becomes order $O(\zeta^{-2})=O(n)$
when $t=O(\log(\zeta^{-2}))$. The second term is more non-trivial to analyze due to the explicit factor of $t$ and we find it becomes order $O(\zeta^{-2})$ when $t=O(\log(\zeta^{-2})-\log\log(\zeta^{-2}))$.
If in \eqref{eq:bound_last_term_thetasq} we had instead assumed:
\begin{align}
\frac{\eta}{D}\sum\limits_{\alpha,\beta=1}^{D}\sum\limits_{i=1}^{t-1}|\epsilon_{\alpha,t}\bs{\phi}_{\alpha}^{T}\bs{\phi}_{\beta}\epsilon_{\beta,i}
|\approx \frac{\eta}{D}\sum\limits_{\alpha,\beta=1}^{D}|\epsilon_{\alpha,t}\bs{\phi}_{\alpha}^{T}\bs{\phi}_{\beta}\epsilon_{\beta,t}
|,\label{eq:second_approx_bias}
\end{align}
then we would have found this term becomes of order $O(\zeta^{-2})$ for $t=O(\log(\zeta^{-1}))$. 
We think \eqref{eq:second_approx_bias} is a more reasonable assumption because of the exponential growth of $\epsilon_{\beta,t}$.
With either assumption, we find that the final term in \eqref{eq:weight_norm_early_time} becomes $O(\zeta^{-2})$ in a time-scale that scales logarithmically in $\zeta^{-1}$ and that we can use perturbation theory if $t\ll \log(\zeta^{-1})$.

\subsection{Homogenous, Two-Layer MLP}
Here we study the early-time dynamics of the homogenous, two-layer MLP:
\begin{align}
z_{\alpha}&=\frac{1}{\sqrt{n}}\bs{v}^T\sigma(\bs{U}\bs{x}_{\alpha}),
\label{eq:z_homogenous_two_layer_MLP}
\\
H_{\alpha\beta}&=\frac{1}{nD}\left(\sigma(\bs{U}\bs{x}_{\alpha})^T\sigma(\bs{U}\bs{x}_{\beta})
+
\bs{x}_{\alpha}^T\bs{x}_{\beta}(\bs{v}\circ \sigma'(\bs{U}\bs{x}_{\alpha}))^T (\bs{v}\circ \sigma'(\bs{U}\bs{x}_{\beta}))
\right),
\label{eq:H_homogenous_two_layer_MLP}
\end{align}
where $\bs{v}\in\mathbb{R}^n$, $\bs{U}\in\mathbb{R}^{n\times d}$, $\bs{x}_{\alpha}\in\mathbb{R}^{d}$, $\alpha,\beta\in\{1,\ldots,D\}$, and $\circ$ is the Hadamard product.
The scale-invariant activation function $\sigma$ is defined in \eqref{eq:sigma_act_definition}.
We will choose an activation function with $0\leq a_-\leq a_+$.
The weights $\bs{U}$ and $\bs{v}$ evolve as:
\begin{align}
\bs{v}_{t+1}&=\bs{v}_t-\frac{\eta}{\sqrt{n}D}\sum\limits_{\alpha=1}^{D}\epsilon_{\alpha,t}\sigma(\bs{U}_t\bs{x}_{\alpha}),
\\
\bs{U}_{t+1}&=\bs{U}_{t}-\frac{\eta}{\sqrt{n}D}\sum\limits_{\alpha=1}^{D}\epsilon_{\alpha,t}
\left(\bs{v}_t\circ\sigma'(\bs{U}_t\bs{x}_{\alpha})\right)
\bs{x}_{\alpha}^T.
\end{align}
Then the update equation for $z_{\alpha,t}$ is:
\begin{align}
z_{\alpha,t+1}=\frac{1}{\sqrt{n}}\left(\bs{v}_t-\frac{\eta}{\sqrt{n}D}\sum\limits_{\beta=1}^{D}\epsilon_{\beta,t}\sigma(\bs{U}_t\bs{x}_{\beta})\right)^T\sigma\left(\bs{U}_{t}\bs{x}_{\alpha}-\frac{\eta}{\sqrt{n}D}\sum\limits_{\gamma=1}^{D}\epsilon_{\gamma,t}
\left(\bs{v}_t\circ\sigma'(\bs{U}_t\bs{x}_{\gamma})\right)
\bs{x}_{\gamma}^T\bs{x}_{\alpha}\right).\label{eq:z_update_generic}
\end{align}
In the infinite width limit the update equation for the error $\epsilon$ is:
\begin{align}
\epsilon_{\alpha,t+1}=\sum\limits_{\beta=1}^{D}(\delta_{\alpha\beta}-\eta H_{\alpha\beta,0})\epsilon_{\beta,t}+O(1/n),
\label{eq:update_largen_MLP}
\end{align}
where the NTK is frozen at its initial value, see e.g. chapter 10 of \cite{Roberts:2021fes} for a review.
When $\eta \lambda_{\text{max}}(H_{\alpha\beta,0})<2$ the errors converge to zero exponentially fast while when $\eta \lambda_{\text{max}}(H_{\alpha\beta,0})>2$ the errors grow exponentially quickly.

Here we run into an important subtlety, the activation function $\sigma$ is not differentiable around $x=0$. 
More generally, the series expansion of $\sigma(x)$ around a point $x=x_0$ has a radius of convergence $r=|x_0|$.
This means that we cannot necessarily expand \eqref{eq:z_update_generic} around $n=\infty$ if $\epsilon_{\alpha}$ is large enough to flip the sign of $\bs{U}\bs{x}_{\alpha}$.\footnote{To be more precise, the large $n$ expansion breaks down at finite width because Taylor expanding the activation function produces increasingly singular $\delta$-functions. These distributions are not integrable when we compute expectation values over the initial weights.
Therefore, we cannot use the large $n$ expansion to study the evolution of the NTK.
For more details see the discussion around equation $(\infty.39)$ of \cite{Roberts:2021fes}.}

This breakdown of perturbation theory is of a different nature than what we saw for the quadratic model in the catapult phase.
In the quadratic model the Taylor expansion of the \textit{update equations}, \eqref{eq:pure_quad_z_mult}-\eqref{eq:pure_quad_H_mult}, truncates at order $\zeta^2$ for all $t$. Perturbation theory breaks down for super-critical learning rates because we cannot similarly truncate $z_{t}$ and $H_{\alpha\beta,t}$ at a low order in $\zeta$ once $z_{t}=O(\zeta^{-1})$. Instead, we are forced to work to all orders in the small $\zeta$ expansion.
On the other hand, for the homogenous MLPs studied here, the large-$n$ expansion of the update equations is simply not valid when $z_{\alpha,t}$ is large because $\sigma$ is not differentiable at the origin.
This breakdown of perturbation theory also occurs when $\eta$ is small and the model is in the lazy phase.

For this reason, to estimate at what time-scale the large-$n$ expansion breaks down we will look at the update equation for $\bs{U}_t$ directly.
At initialization we have $v_{i,0},U_{ij,0}\sim \mathcal{N}(0,1)$ for all $i$ and $j$.
Assuming $\bs{x}_{\alpha}$ and $y_{\alpha}$ do not scale with $n$, we expect an order-one fraction of the components of $\bs{U}_t\bs{x}_{\alpha}$ to flip signs when $z_{t}=O(\sqrt{n})$, which occurs after $O(\log(\sqrt{n}))$ time-steps.
This argument is less rigorous than the one given above for the quadratic model, but it agrees with the numerical results of \cite{lewkowycz2020large} and what we find numerically.

\section{Derivations for a Single Datapoint}
\label{app:single_datapoint}
In this section we will prove that the catapult phase exists for super-critical learning rates when training on a single datapoint. With the exception of the ReLU MLP, we will work with the toy dataset $(x,y)=(1,0)$. For the ReLU MLP we will take the label $y>0$ to avoid the trivial solution where all the first layer weights are negative.
We will drop the sample index everywhere since we only have one data-point.

\subsection{Pure Quadratic Model}
\label{app:proof_single_data_quad}
We will start with the pure quadratic model:
\begin{align}
z=\frac{\zeta}{2}\bs{\theta}^T\bs{\psi}\bs{\theta},\qquad H=\zeta^2\bs{\theta}^T\bs{\psi}^2\bs{\theta}.
\end{align}
We will prove that the weight norm $\bs{\theta}_{t}^2$ decreases monotonically if:
\begin{align}
\eta <\frac{4}{\zeta^2\bs{\theta}_{0}^2\lambda_{\text{max}}(\bs{\psi}^2)}\label{eq:bound_eta_init_pure_quadratic}
\end{align}
and therefore that the loss will remain finite.
The proof was sketched in Section \ref{ssec:sketch_proof} and here we will fill in the details.

First, we use that under gradient descent the weight norm evolves as:
\begin{align}
\bs{\theta}^2_{t+1}=\bs{\theta}^2_t+\eta z_{t}^2(\eta H_{t}-4).
\label{eq:weight_sq_evolution}
\end{align}
Next, we use that the meta-feature function $\bs{\psi}$ is a real symmetric matrix.
Therefore $\bs{\psi}^2$ is a positive semi-definite matrix and the NTK in this model obeys the bound:
\begin{align}
H_{t}\leq \zeta^{2}\lambda_{\text{max}}(\bs{\psi}^2)\bs{\theta}_{t}^2.\label{eq:H_bound_pure_quad}
\end{align}
This has the same form as the bound \eqref{eq:bound_H_theta}.
We will now assume that at some time-step $t_{*}$ we have:
\begin{align}
\eta < \frac{4}{\zeta^2\bs{\theta}_{t_{*}}^2 \lambda_{\text{max}}(\bs{\psi}^2)}.\label{eq:bound_eta_t0}
\end{align}
If \eqref{eq:bound_eta_t0} holds at some time $t_{*}$, then we can prove that the weight norm $\bs{\theta}^2_t$ is a monotonically decreasing function for all subsequent steps:
\begin{align}
\bs{\theta}^{2}_{t+1}-\bs{\theta}^{2}_{t}<0 \quad \forall \ t\geq t_{*}.
\end{align}
This statement follows directly from the update equation \eqref{eq:weight_sq_evolution}:
\begin{align}
\bs{\theta}_{t_{*}+1}^2-\bs{\theta}_{t_{*}}^2&=\eta z_{t_{*}}^2(\eta H_{t_{*}}-4)
\nonumber \\
& \leq \eta z_{t_{*}}^2(\eta \zeta^2 \bs{\theta}_{t_{*}}^2 \lambda_{\text{max}}(\bs{\psi}^2) -4)
\nonumber \\
&<0.
\end{align}
To get the second line we used \eqref{eq:H_bound_pure_quad} and to get the third line we used \eqref{eq:bound_eta_t0}.
This shows that the weight norm decreases under one step of gradient descent.
Moreover, since $\bs{\theta}^2_{t_{*}+1}<\bs{\theta}^2_{t_{*}}$ we also have that \eqref{eq:bound_eta_t0} continues to hold at time $t=t_{*}+1$:
\begin{align}
\eta < \frac{4}{\zeta^2\bs{\theta}_{t_{*}}^2 \lambda_{\text{max}}(\bs{\psi}^2)}  < \frac{4}{\zeta^2\bs{\theta}_{t_{*}+1}^2 \lambda_{\text{max}}(\bs{\psi}^2)}.
\end{align}
Therefore, once \eqref{eq:bound_eta_t0} holds at some time $t_{*}$, it holds for all subsequent times. Then from \eqref{eq:weight_sq_evolution} this implies that the weight norm $\bs{\theta}^2_t$ also decreases for all subsequent times.
To complete the proof we then just need to assume the bound on $\eta$ \eqref{eq:bound_eta_t0} holds at $t=0$ to guarantee that the weight norm decreases monotonically for all time.
This gives the original condition \eqref{eq:bound_eta_init_pure_quadratic} and completes the proof.

Note that for this argument to work we did not need to use that the model undergoes any catapult dynamics.
We then have two separate cases:
\begin{enumerate}
\item If $2H_0>\zeta^2\bs{\theta}_{0}^2\lambda_{\text{max}}(\bs{\psi}^2)$, then there is a finite window above the linear stability threshold $\eta H_0=2$  where the model converges and the loss exhibits the catapult mechanism.
\item If $2H_0<\zeta^2\bs{\theta}_{0}^2\lambda_{\text{max}}(\bs{\psi}^2)$, then this argument does not guarantee that the model has a catapult phase, but it does imply there exists a region in the lazy phase where the weight norm decays monotonically.
\end{enumerate}
These conditions are dependent on the values of the weights at initialization.
We can also ask when these conditions hold in expectation by averaging over the weights. 
For example, the inequality $2H_0>\zeta^2\bs{\theta}_{0}^2\lambda_{\text{max}}(\bs{\psi}^2)$ holds in expectation if:
\begin{align}
\mathbb{E}_{\bs{\theta}_0}[2H_0-\zeta^2\bs{\theta}_{0}^2\lambda_{\text{max}}(\bs{\psi}^2)]>0
\nonumber
\\
\Longrightarrow  \lambda_{\text{max}}(\bs{\psi}^2)<\frac{2}{n}\Tr(\bs{\psi}^2).
\end{align}
That is, the maximum eigenvalue of $\bs{\psi}^2$ cannot differ too significantly from the mean. This inequality is closest to being saturated when all the eigenvalues of $\bs{\psi}^2$ are the same, which is what happens in a two-layer MLP with a linear activation function \cite{Belkin_quadratic,Pennington_second_order}.

We can use the same type of argument to prove that if:
\begin{align}
\eta > \frac{4}{\zeta^2 \bs{\theta}_{0}^2\lambda_{\text{min}}(\bs{\psi}^2)},\label{eq:eta_lower_bound_pure_init}
\end{align}
then the loss diverges.
The proof is almost the same as before, except we need to use the following lower bound on the NTK in terms of the minimal eigenvalue of $\bs{\psi}^2$:
\begin{align}
H_t\geq \zeta^2\lambda_{\text{min}}(\bs{\psi}^2)\bs{\theta}^2_t.\label{eq:lower_bound_H_pure}
\end{align}
Following the same ideas as before, we start by assuming that at some time-step $t_{*}$ we have:
\begin{align}
\eta > \frac{4}{\zeta^2 \bs{\theta}_{t_{*}}^2\lambda_{\text{min}}(\bs{\psi}^2)}.\label{eq:eta_lower_bound_pure_t0}
\end{align}
Then from the update equation for $\bs{\theta}^2$ \eqref{eq:weight_sq_evolution} we have:
\begin{align}
\bs{\theta}_{t_{*}+1}^{2}-\bs{\theta}_{t_{*}}^{2}&=\eta z_{t_{*}}^2(\eta H_{t_{*}}-4)
\nonumber
\\
&\geq \eta z_{t_{*}}^2(\eta \zeta^2\lambda_{\text{min}}(\bs{\psi}^2)\bs{\theta}^2_{t_{*}}-4)
\nonumber
\\
& >0.
\end{align}
To get the second line we used \eqref{eq:lower_bound_H_pure} and to get the third line we used our assumption \eqref{eq:eta_lower_bound_pure_t0}.
Since here the weight norm has increased after one step of gradient descent, this implies \eqref{eq:eta_lower_bound_pure_t0} continues to hold at time-step $t=t_{*}+1$:
\begin{align}
\eta > \frac{4}{\zeta^2 \bs{\theta}_{t_{*}}^2\lambda_{\text{min}}(\bs{\psi}^2)}>\frac{4}{\zeta^2 \bs{\theta}_{t_{*}+1}^2\lambda_{\text{min}}(\bs{\psi}^2)}.
\end{align}
Therefore, by induction, the weight norm monotonically increases for all future time-steps.
Finally, this means if the lower bound holds at initialization, see \eqref{eq:eta_lower_bound_pure_init}, then the weight norm increases monotonically for all time.

The fact the weight norm monotonically increases for all times does not, by itself, imply the loss diverges.
However, we expect the loss will diverge since if the lower bound \eqref{eq:eta_lower_bound_pure_init} holds, then the learning rate $\eta$ is necessarily super-critical:
\begin{align}
\eta H_0>\eta \lambda_{\text{min}}(\bs{\psi}^2)\zeta^2\bs{\theta}_{0}^2>4,\label{eq:lowe_bound_H}
\end{align}
where to find the first inequality we used \eqref{eq:lower_bound_H_pure} and to find the second inequality we used \eqref{eq:eta_lower_bound_pure_init}.
Moreover, since $\bs{\theta}_t^2$ is a monotonically increasing function, this implies $\eta H_t>4$ for all time $t$ and the model can never re-enter the lazy phase.
Altogether, this implies the point $z_{t}=0$ is an unstable fixed point of the update equations when the learning rate satisfies the lower bound \eqref{eq:eta_lower_bound_pure_init}.
If $z_{t}$ vanishes exactly, then the model stays at $z_t=0$ for all time.
However, if $z_t$ is small, but non-zero, then we can use the small $\zeta$-perturbation theory arguments of Appendix \ref{app:early_time} to show that the loss will increase exponentially quickly.
Therefore, barring fine-tuning where the update equations set $z_t=0$ exactly, the output will be exponentially large and $\bs{\theta}_{t}^2$ receives, large, positive updates for all $t$ and will diverge.\footnote{The weight norm $\bs{\theta}^2$ can receive small updates if $\eta H_t\rightarrow 4$ as $t\rightarrow\infty$, but given our assumption \eqref{eq:eta_lower_bound_pure_init} we have that $\eta H_t-4$ is strictly bounded away from 0 for all $t$.}

To summarize, we have shown that if $\eta$ satisfies the bound \eqref{eq:bound_eta_init_pure_quadratic} then the weights decay monotonically and if $\eta$ satisfies the bound \eqref{eq:eta_lower_bound_pure_init} then the weights increase monotonically.
This leaves the in-between region:
\begin{align}
\frac{4}{\zeta^2 \bs{\theta}^2_{0}\lambda_{\text{max}}(\bs{\psi}^2)}<\eta<\frac{4}{\zeta^2 \bs{\theta}^2_{0}\lambda_{\text{min}}(\bs{\psi}^2)},
\end{align}
where we cannot say anything definite with our methods.
To say something definite about this region would require understanding how the details of the meta-feature eigensystem affects the evolution of the quadratic model.
Here we can note that this region shrinks to zero size when:
\begin{align}
\lambda_{\text{min}}(\bs{\psi}^2)=\lambda_{\text{max}}(\bs{\psi}^2).\label{eq:min_max_psi_evals}
\end{align}
In this case there is a very sharp delineation between the catapult and divergent phases.
The condition \eqref{eq:min_max_psi_evals} is satisfied in the two-layer MLP with linear activation functions. The phase boundary between the catapult and divergent phase for this model was first found in \cite{lewkowycz2020large}.

\subsection{Quadratic Model With Bias}
\label{app:proof_single_data_quad_bias}
In this appendix we study the quadratic model with bias,
\begin{align}
z=\bs{\theta}^T\bs{\phi}+\frac{\zeta}{2}\bs{\theta}^T\bs{\psi}\bs{\theta}, \qquad H=\bs{\phi}^2+\zeta^2\bs{\theta}^T\bs{\psi}^2\bs{\theta},
\end{align}
where $\bs{\psi}\bs{\phi}=0$. We train this model on the toy dataset $(x,y)=(1,0)$.
We will prove this model converges if:
\begin{align}
\eta<\frac{4}{2\bs{\phi}^{2}+\zeta^2\lambda_{\text{max}}(\bs{\psi}^2)\left(\bs{\theta}^{2}_{0}+\frac{(\bs{\phi}^T\bs{\theta}_{0})^2}{\bs{\phi}^2}\right)}.
\label{eq:inequality_eta_bias_V2_init}
\end{align}

In the quadratic model with bias, the update equation for $\bs{\theta}_t^2$ is:
\begin{align}
\bs{\theta}^{2}_{t+1}&=\bs{\theta}^{2}_{t}-2\eta  z_{t}\left[\bs{\phi}^T\bs{\theta}_t+\zeta(\bs{\theta}_t^T\bs{\psi}\bs{\theta}_t)\right]+\eta^2  z^2_tH_t. \label{eq:theta_sq_update_bias}
\end{align}
Unlike the pure quadratic model, here the update equation for $\bs{\theta}_t^2$ does not have a nice sign-definiteness property.
In both the pure quadratic model and the quadratic model with bias, the $\eta^2$ term in the $\bs{\theta}^2_t$ update equation is manifestly positive semi-definite since $z_t$ is real and the NTK is positive semi-definite.
On the other hand, for the pure quadratic model the order $\eta$ term was manifestly negative semi-definite, see \eqref{eq:weight_sq_evolution}, while in \eqref{eq:theta_sq_update_bias} it is not clear if the order $\eta$ term has a definite sign. The fact the order $\eta$ term in \eqref{eq:weight_sq_evolution} is negative semi-definite was important in proving that the pure quadratic model can have a catapult phase.
To remedy the lack of sign-definiteness in \eqref{eq:theta_sq_update_bias} we will study a slightly different update equation.

Specifically, we will study the evolution of:
\begin{align}
\bs{\theta}^{2}_t+\frac{(\bs{\phi}^T\bs{\theta}_t)^{2}}{\bs{\phi}^2}.\label{eq:new_quant_bias}
\end{align}
Under gradient descent the second term of \eqref{eq:new_quant_bias} evolves as:
\begin{align}
\frac{(\bs{\phi}^T\bs{\theta}_{t+1})^{2}}{\bs{\phi}^2}=
\frac{(\bs{\phi}^T\bs{\theta}_{t})^{2}}{\bs{\phi}^2}-2\eta z_t \bs{\phi}^T\bs{\theta}_t+\eta^2z^2_t \bs{\phi}^2,
\end{align}
where we used that $\bs{\psi}\bs{\phi}=0$.
Using the above update equation we find the quantity \eqref{eq:new_quant_bias} evolves as:
\begin{align}
\bs{\theta}^{2}_{t+1}+\frac{(\bs{\theta}^T_{t+1}\bs{\phi})^{2}}{\bs{\phi}^2}&=\bs{\theta}^{2}_{t}+\frac{(\bs{\theta}^T_{t}\bs{\phi})^{2}}{\bs{\phi}^2}-2\eta z_{t}\left(2(\bs{\theta}_{t}\bs{\phi})+\zeta(\bs{\theta}_{t}^T\bs{\psi}\bs{\theta}_{t})\right)+\eta^2z_{t}^{2}(H_{t}+\bs{\phi}^2)
\nonumber
\\
&=\bs{\theta}^{2}_{t}+\frac{(\bs{\theta}^T_{t}\bs{\phi})^{2}}{\bs{\phi}^2}+\eta z_{t}^{2}\left(\eta(H_t+\bs{\phi}^2)-4\right)\label{eq:update_combo_bias}.
\end{align}
Now the order $\eta$ and order $\eta^2$ terms in \eqref{eq:update_combo_bias} are sign-definite and we can prove that $\bs{\theta}^{2}_{t}+\frac{(\bs{\theta}^T_{t}\bs{\phi})^{2}}{\bs{\phi}^2}$ decays monotonically if \eqref{eq:inequality_eta_bias_V2_init} holds.

The proof will be the same as for the pure quadratic model. To start, we use that the NTK obeys the following upper bound:
\begin{align}
H_t=\bs{\phi}^2+\zeta^2\bs{\theta}^T_t\bs{\psi}^2\bs{\theta}_t&\leq \bs{\phi}^2+\zeta^2\lambda_{\text{max}}(\bs{\psi}^2)\bs{\theta}_t^2
\nonumber
\\
&\leq \bs{\phi}^2+\zeta^2\lambda_{\text{max}}(\bs{\psi}^2)\left(\bs{\theta}_t^2+\frac{(\bs{\phi}^T\bs{\theta}_t)^2}{\bs{\phi}^2}\right).\label{eq:H_chain_inequalitites}
\end{align}
The first inequality follows from the fact $\bs{\psi}^2$ is a positive semi-definite matrix.
The second inequality is trivial as we just added a manifestly positive quantity to the previous expression.
Therefore the $\eta(H_{t}+\bs{\phi}^2)-4$ term in \eqref{eq:update_combo_bias} will be negative at time-step $t=t_{*}$ if we impose the inequality:
\begin{align}
\eta<\frac{4}{2\bs{\phi}^{2}+\zeta^2\lambda_{\text{max}}(\bs{\psi}^2)\left(\bs{\theta}^{2}_{t_{*}}+\frac{(\bs{\phi}^T\bs{\theta}_{t_{*}})^2}{\bs{\phi}^2}\right)}.\label{eq:inequality_eta_bias_V2}
\end{align}
This condition is sufficient to guarantee $\eta(H_{t_{*}}+\bs{\phi}^2)-4<0$, but may not be necessary.
If the condition \eqref{eq:inequality_eta_bias_V2} holds at time $t_{*}$, then from \eqref{eq:update_combo_bias} we have that at time $t=t_{*}+1$:
\begin{align}
\bs{\theta}_{t_{*}+1}^{2}+\frac{(\bs{\theta}^{T}_{t_{*}+1}\bs{\phi})^2}{\bs{\phi}^2}< \bs{\theta}_{t_{*}}^{2}+\frac{(\bs{\theta}^{T}_{t_{*}}\bs{\phi})^2}{\bs{\phi}^2}.
\end{align}
The denominator in \eqref{eq:inequality_eta_bias_V2} is a sum of manifestly positive quantities, so if $\bs{\theta}^{2}+\frac{\bs{\theta}^{T}\bs{\phi}}{\bs{\phi}^2}$ is decreasing the right-hand side must increase:
\begin{align}
\eta<\frac{4}{2\bs{\phi}^{2}+\zeta^2\lambda_{\text{max}}(\bs{\psi}^2)\left(\bs{\theta}^{2}_{t_*}+\frac{(\bs{\phi}^T\bs{\theta}_{t_*})^2}{\bs{\phi}^2}\right)}<\frac{4}{2\bs{\phi}^{2}+\zeta^2\lambda_{\text{max}}(\bs{\psi}^2)\left(\bs{\theta}^{2}_{t_*+1}+\frac{(\bs{\phi}^T\bs{\theta}_{t_*+1})^2}{\bs{\phi}^2}\right)}.
\end{align}
This shows that if \eqref{eq:inequality_eta_bias_V2} holds at time $t_{*}$, then it also holds for all subsequent steps and therefore $\bs{\theta}^{2}_t+\frac{(\bs{\phi}^T\bs{\theta}_t)^{2}}{\bs{\phi}^2}$ will decrease monotonically for all subsequent steps as well.
Finally, we can guarantee that the weight norm $\bs{\theta}_t^2$ does not diverge, and therefore the loss remains finite, by imposing this condition at initialization. This gives the original inequality \eqref{eq:inequality_eta_bias_V2_init}.\footnote{Here we are using the trivial inequality $\bs{\theta}^2\leq \bs{\theta}^2+\frac{(\bs{\theta}^{T}\bs{\phi})^2}{\bs{\phi}^2}$, so if the right-hand side is finite, the left-hand side will be finite as well.}
\\
\\
\textbf{Example: Linear Net with Bias}
\\
\\
Here we will make the previous analysis more concrete by considering a simple example of a quadratic model with bias, the two-layer, linear net with bias on the final layer:
\begin{align}
z=\frac{1}{\sqrt{n}}\bs{v}^T\bs{u} +b,
\end{align}
where $\bs{u},\bs{v}\in\mathbb{R}^n$ and $b\in \mathbb{R}$.\footnote{We could also allow for bias in the hidden layer, but this will not change anything qualitatively.}
The NTK is given by:
\begin{align}
H=\frac{1}{n}(\bs{u}^2+\bs{v}^2)+1.
\end{align}
We train this model to minimize the MSE loss:
\begin{align}
L_t=\frac{1}{2}z_{t}^2.
\end{align}

To explain the connection between the linear net with bias and the quadratic model with bias, it is simplest to work in index notation. First, we define the abstract weights $\bs{\theta} \in \mathbb{R}^{2n+1}$ as the concatenation of $\bs{u}$, $\bs{v}$ and $b$:
\begin{align}
\bs{\theta}=(\bs{u},\bs{v},b)^T.
\end{align}
Here it is understood that the first $n$ indices of $\theta_{\mu}$ correspond to $\bs{u}$, the second $n$ indices to $\bs{v}$ and last index to $b$.
Then to make the map between the quadratic model with bias and the linear net with bias more explicit, we will label the indices of $\bs{\phi} $ and $\bs{\psi}$ by the corresponding weights, $\bs{u}$, $\bs{v}$ and $b$.
For example, we will write the output $z$ as:
\begin{align}
z&=\bs{\phi}^T\bs{\theta}+\frac{\zeta}{2}\bs{\theta}^T\bs{\psi}\bs{\theta}
\nonumber\\
&=\phi_{b}b+\zeta \sum\limits_{i,j=1}^n u_{i} \psi_{u_i v_j}v_j,
\end{align}
where $\zeta^2=1/n$. Here the only non-trivial components of the (meta-)feature functions are given by:
\begin{align}
\phi_{b}=1, \qquad
\psi_{u_iv_j}=\delta_{ij},
\end{align}
where $\delta_{ij}$ is the Kronecker delta function. The other components, e.g. $\phi_{u_i}$, $\phi_{v_i}$, $\psi_{u_iu_j}$, $\psi_{v_iv_j}$, etc. all vanish in the linear net with bias.

Then the sufficiency condition that this model converges \eqref{eq:inequality_eta_bias_V2_init} becomes:
\begin{align}
\eta < \frac{4}{2+1/n(\bs{u}_0^2+\bs{v}_0^2+2b^2_0)}.
\end{align}
If we assume that at initialization the bias is set to zero, $b_0=0$, then this reduces to:
\begin{align}
\eta < \frac{4}{2+1/n(\bs{u}_0^2+\bs{v}_0^2)}=\frac{4}{H_0+1}.
\end{align}

\subsection{Homogenous, Two-Layer, MLPs}
\label{app:homogenous_two_layer_one_datapoint_der}
Here we will study the homogenous, two-layer net,
\begin{align}
z=\frac{1}{\sqrt{n}}\sum\limits_{i=1}^{n}v_i\sigma(u_i), \qquad H=\frac{1}{n}\left(\sigma^2(u_i)+v_i^2\sigma'^2(u_i)\right),
\label{eq:z_ntk_single_datapoint_homogenous}
\end{align}
trained on the toy dataset $(x,y)=(1,0)$.
The activation function $\sigma$ is given in \eqref{eq:sigma_act_definition} and we assume $0<a_-\leq a_+\leq1$.
We will prove this model converges if:
\begin{align}
\eta < \frac{4n}{a_+^2(\bs{u}_{0}^2+\bs{v}_0^2)}=\frac{4n}{a_+^2\bs{\theta}_0^2}.\label{eq:upper_bound_generic_scale}
\end{align}
This analysis will be almost identical to the analysis of the pure quadratic model given in Appendix \ref{app:proof_single_data_quad}, so we will be brief.

First, the weight norm in this model is $\bs{\theta}^2=\bs{u}^2+\bs{v}^2$. 
Under gradient descent $\bs{\theta}^2$ evolves as:
\begin{align}
\bs{\theta}_{t+1}^2=\bs{\theta}_{t}^2+\eta z_{t}^2(\eta H_{t}-4). \label{eq:weight_sq_evolution_MLP}
\end{align}
Note that this is identical to the weight update equation for the pure quadratic model \eqref{eq:weight_sq_evolution}.
This is not a coincidence and instead comes from the fact that in both models $z_t$ is a homogenous, quadratic function of $\bs{\theta}$ and that we are training the models with MSE.

Next, we use the following upper bound for the NTK in terms of the weight norm:
\begin{align}
H_{t}\leq a_{+}^2\frac{\bs{u}_t^2+\bs{v}_{t}^2}{n}=\frac{a_+^2}{n}\bs{\theta}_{t}^2.
\label{eq:bound_NTK_single_data_scale_invariant}
\end{align}
This bound is identical in form to the bound used when studying the pure quadratic model, see \eqref{eq:H_bound_pure_quad}, except with the replacement:
\begin{align}
\zeta^2\lambda_{\text{max}}(\bs{\psi}^2) \ \ \Longrightarrow \ \ \frac{a_+^2}{n}. \label{eq:dictionaryMLP_quad}
\end{align}
Since the weight update equation \eqref{eq:weight_sq_evolution_MLP} and NTK bound \eqref{eq:bound_NTK_single_data_scale_invariant} have the same functional form as in the pure quadratic model, we can repeat the steps used in Appendix \ref{app:proof_single_data_quad} to derive the new bound \eqref{eq:upper_bound_generic_scale} for the homogenous MLP.
In particular, if we take the bound derived for the pure quadratic model \eqref{eq:bound_eta_init_pure_quadratic} and make the replacement \eqref{eq:dictionaryMLP_quad}, we land exactly on \eqref{eq:upper_bound_generic_scale}.

As with the pure quadratic model trained on $(x,y)=(1,0)$, we now have two separate cases to consider:
\begin{enumerate}
\item If $n^{-1}a_+^2\bs{\theta}_0^2<2H_0$, then there exists a finite window above the linear stability threshold, $\eta H_0=2$, where the model still converges. In this case the loss exhibits the catapult mechanism.
\item If $n^{-1}a_+^2\bs{\theta}_0^2>2H_0$, then our bound only guarantees there is a choice of learning rate in the lazy phase such that $\bs{\theta}^2_t$ decreases monotonically.
\end{enumerate}

To determine which case happens on average, we need the expectation values of the NTK and the weight norm at initialization:
\begin{align}
\mathbb{E}[H_0]&=(a_-^2+a_+^2),
\\
\mathbb{E}[\bs{\theta}_0^2]&=2n.
\end{align}
Therefore:
\begin{align}
\mathbb{E}\left[2H_0-\frac{a_+^2}{n}\bs{\theta}_0^2\right]>0 \ \ \Longrightarrow \ \ a_-\neq 0,
\end{align}
which includes all scale-invariant activation functions with the exception of the ReLU net.
Of course, this does not mean that the ReLU net cannot obey the bound \eqref{eq:upper_bound_generic_scale} for some choice of initial weights, but on average it obeys this condition only half of the time.

Finally, we can also prove that above a certain learning rate the model will diverge.
Specifically, if
\begin{align}
\eta >\frac{4n}{a_-^2\bs{\theta}_{0}^2},\label{eq:MLP_diverge}
\end{align}
then the weight norm grows monotonically and the loss diverges.
To prove this, we use that the NTK obeys the following lower bound:
\begin{align}
H_{t}\geq \frac{a_-^2}{n}\bs{\theta}^2_t.\label{eq:lower_bound_NTK_MLP_toy}
\end{align}
Note that here we are using our assumption that $0<a_- \leq a_+$.
If $a_{-}=0$ this lower bound is trivial.
The lower bound \eqref{eq:lower_bound_NTK_MLP_toy} also has the same form as the lower bound \eqref{eq:eta_lower_bound_pure_init} for the pure quadratic model.
To go from the pure quadratic model to the homogenous net we make the replacement:
\begin{align}
\zeta^2\lambda_{\text{min}}(\bs{\psi}^2) \ \ \Longrightarrow \ \ \frac{a_-^2}{n}. \label{eq:dictionaryMLP_quad_MIN}
\end{align}
Then, since the update equation for $\bs{\theta}_t^2$ and the lower bound on the NTK have the same form as in the pure quadratic model, we can again borrow from the analysis of Appendix \ref{app:proof_single_data_quad}.
We will not repeat the argument here, but can instead use the replacement \eqref{eq:dictionaryMLP_quad_MIN} in \eqref{eq:eta_lower_bound_pure_init} to give exactly the lower bound \eqref{eq:MLP_diverge} for the homogenous MLP.

\subsection{ReLU Net}
\label{ssec:relu_net}
In the previous section we derived sufficiency conditions for the catapult phase to exist in homogenous, two-layer nets.
However, these conditions were too strong for ReLU nets: the allowed range for the learning rate $\eta$ vanished on average.
In this section we will derive weaker bounds on $\eta$ by using some simplifying features of the ReLU function.

We again study a two-layer net:
\begin{align}
z=\frac{1}{\sqrt{n}}\sum\limits_{i=1}^{n}v_i\sigma_{\text{ReLU}}(u_i),
\qquad H=\frac{1}{n}\left(\sigma^2_{\text{ReLU}}(u_i)+v_i^2\sigma'^2_{\text{ReLU}}(u_i)\right).
\end{align}
Here we will train the model using MSE loss:
\begin{align}
L=\frac{1}{2}(z-y)^2,
\end{align}
where $y\geq 0$. We will consider two cases, $y=0$ and $y>0$ with $y=O(1)$.
The first case is trivial to solve, we can simply take $u_i<0$ for all $i$.
However, it is a useful case to consider to set up notation.
\\
\\
$\bs{y=0}:$
\\
\\
The reason our bounds for the generic homogenous net are too strong for the ReLU net is that, on average, only half of the weights will evolve in the ReLU MLP when we have one datapoint. 
It is then overkill to study the evolution of the full weight norm $\bs{\theta}^2_t=\bs{u}^2_t+\bs{v}^2_t$ and we should instead only focus on the weights which evolve for at least one step of gradient descent.
We can do this by decomposing the weights at initialization as follows:
\begin{align}
\bs{u}_0&=\bs{u}_{+,0}+\bs{u}_{-,0},
\\
\bs{u}_{+,0}&=\sigma_{\text{ReLU}}(\bs{u}_0),
\\
\bs{u}_{-,0}&=\sigma_{\text{ReLU}}(-\bs{u}_0).
\end{align}
That is, $\bs{u}_{\pm,0}$ are two $n$-dimensional vectors whose non-zero components correspond to the components of $\bs{u}_0$ which are positive/negative at $t=0$, respectively.
By definition they are orthogonal $\bs{u}_{+,0}^T\bs{u}_{-,0}=0$.
We can then define matrices which project onto the $\bs{u}_{\pm}$ directions:
\begin{align}
\bs{P}^+&=\mathbb{I}_{n\times n} \circ \text{diag}(\mathbbm{1}_{\sigma(\bs{u})\geq0}),
\\
\bs{P}^-&=\mathbb{I}_{n\times n} \circ \text{diag}(\mathbbm{1}_{\sigma(\bs{u})<0}).
\end{align}
Here $``\text{diag}"$ produces a diagonal matrix from the given vector.
These projectors may be clearer in index notation, where we can write:
\begin{align}
P^{+}_{ij}=\delta_{ij}\mathbbm{1}_{u_{i,0}\geq0},
\\
P^{-}_{ij}=\delta_{ij}\mathbbm{1}_{u_{i,0}<0},
\end{align}
where $\delta_{ij}$ is the Kronecker delta function.
We can then define $\bs{v}_{\pm,t}$ as:
\begin{align}
\bs{v}_{+,t}=\bs{P}^+\bs{v}_t,
\\
\bs{v}_{-,t}=\bs{P}^-\bs{v}_t.
\end{align}
That is, $\bs{v}_{\pm,t}$ is the projection of $\bs{v}_t$ onto the directions where $\bs{u}_{\pm,0}$ are non-zero.

Under gradient descent the weights evolve as:
\begin{align}
\bs{u}_{+,t+1}&=\bs{u}_{+,t}-\frac{1}{\sqrt{n}}\eta z_t \bs{v}_{+,t}\circ\sigma'_{\text{ReLU}}(\bs{u}_{+,t}),
\\
\bs{v}_{+,t+1}&=\bs{v}_{+,t}-\frac{1}{\sqrt{n}}\eta z_t \sigma_{\text{ReLU}}(\bs{u}_{+,t}),
\\
\bs{u}_{-,t+1}&=\bs{u}_{-,t},
\\
\bs{v}_{-,t+1}&=\bs{v}_{-,t}.
\end{align}
We can rewrite the output $z$ and NTK as:
\begin{align}
z_{t}&=\frac{1}{\sqrt{n}}\bs{v}_{+,t}^T\sigma_{\text{ReLU}}(\bs{u}_{+,t}),
\\
H_{t}&=\frac{1}{n}\left((\sigma_{\text{ReLU}}(\bs{u}_{+,t}))^2 + (\bs{v}_{+,t}\circ \sigma'_{\text{ReLU}}(\bs{u}_{+,t}))^2\right).
\end{align}

Then, instead of studying the full weight norm $\bs{u}^2_t+\bs{v}^2_t$, we can study the reduced weight norm $\bs{\theta}_{+,t}^2\equiv\bs{u}_{+,t}^2+\bs{v}_{+,t}^2$,
\begin{align}
\bs{\theta}^2_{+,t+1}=\bs{\theta}^2_{+,t}
+\eta z_{t}^2(\eta H_{t}-4)\label{eq:plus_weight_evolution}.
\end{align}
Note that the update equation for the reduced weight norm is identical in form to the update equation for the full weight norm in the pure quadratic model and two-layer, homogenous net, see \eqref{eq:weight_sq_evolution} and \eqref{eq:weight_sq_evolution_MLP}.
Therefore, we can now recycle our previous proofs for the existence of the catapult phase for this model.
As a first step, we use that $H_t$ is bounded from above by this reduced weight norm:
\begin{align}
H_{t}\leq \frac{1}{n}\left(\bs{u}_{+,t}^2+\bs{v}_{+,t}^2\right)=\frac{\bs{\theta}_{+,t}^2}{n}.\label{eq:H_inequality_ReLU}
\end{align}
At $t=0$ this inequality is saturated but when $t>0$ it is in general a strict inequality.
This is because at $t=0$ each component of $\bs{u}_{+,t}$ is positive and we have $\sigma_{\text{ReLU}}(\bs{u}_{+,0})=\bs{u}_{+,0}$ by definition.
However, under gradient descent some of the components of $\bs{u}_{+,t}$ can become negative, in which case they are set to zero by the ReLU activation function and only contribute to the right hand side of \eqref{eq:H_inequality_ReLU}.

Finally, we can use the inequality \eqref{eq:H_inequality_ReLU} to show that if
\begin{align}
\eta&<\frac{4n}{(\bs{u}_{+,0}^2+\bs{v}_{+,0}^2)}=\frac{4}{H_0},
\label{eq:eta_ineq_ReLU}
\end{align}
then the weight norm monotonically decreases under gradient descent.
We will not write out the proof here since it is identical in form to the proof used for the pure quadratic model, see Appendix \ref{app:proof_single_data_quad}.
The only differences are that here we are studying $\bs{\theta}^2_+$ instead of $\bs{\theta}^2$ and to go from the bound \eqref{eq:H_bound_pure_quad} to \eqref{eq:eta_ineq_ReLU} we make the replacement $\zeta^2\lambda_{\text{max}}(\bs{\psi}^2)\bs{\theta}^2\rightarrow \bs{\theta}_+^2/n$.
If we make this replacement in \eqref{eq:bound_eta_init_pure_quadratic} we find the bound \eqref{eq:eta_ineq_ReLU}.
\\
\\
$\bs{y>0}:$
\\
\\
We can now generalize the previous argument to the case where $y>0$, but is still $O(1)$.
The main difference is now the update equations for $\bs{u}_+$ and $\bs{v}_+$ are:
\begin{align}
\bs{u}_{+,t+1}&=\bs{u}_{+,t}-\frac{1}{\sqrt{n}}\eta \epsilon_t \bs{v}_{+,t}\circ\sigma'_{\text{ReLU}}(\bs{u}_{+,t}),
\\
\bs{v}_{+,t+1}&=\bs{v}_{+,t}-\frac{1}{\sqrt{n}}\eta \epsilon_t \sigma_{\text{ReLU}}(\bs{u}_{+,t}).
\end{align}
The reduced weight norm $\bs{\theta}_+^2$ then evolves as:
\begin{align}
\bs{\theta}_{+,t+1}^2=\bs{\theta}_{+,t}^2
-4\eta z_t\epsilon_t 
+\eta^2 \epsilon_t^2 H_t.
\label{eq:update_relu_mlp_generic}
\end{align}
Unlike the previous case, where $y=0$, now the order $\eta$ term does not generically have a definite sign.
However, this is a minor issue since when $\eta H_0>2$ the output $z_t$ grows exponentially quickly.
When $|z_{t}|\gg 1$ we have:
\begin{align}
\epsilon_t=z_t-y \approx z_t.
\end{align}
Therefore, if $\eta H_0>2$ and we study the model at a time $t_{*}$ such that $1\ll t_{*} \ll \log(n)$, then $|y|\ll |z_{t_{*}}|\ll 1/\sqrt{n}$ and the update equation simplifies to:\footnote{In practice, because $z_t$ increases exponentially fast and $|y|$ is an order-one number, we can take $t_*$ also to be an order-one number. The condition $t_*\gg1$ ensures that $\epsilon_{t_*}\approx z_{t_*}$ up to exponentially suppressed corrections.}
\begin{align}
\bs{\theta}_{+,t_{*}+1}^2\approx\bs{\theta}_{+,t_{*}}^2
+\eta z_{t_{*}}^2 (\eta H_{t_{*}}-4), \quad \text{for } \  1\ll t_{*} \ll \log(n).
\end{align}
In this case, the weight norm receives large, $O(n)$, negative corrections when $z_{t}=O(\sqrt{n})$ if we impose:
\begin{align}
\eta <\frac{4n}{a_+^2\bs{\theta}_{+,t_{*}}^2}= \frac{4n}{a_+^2\bs{\theta}_{+,0}^2}+O(1/n).
\label{eq:eta_bound_generic_label_relu}
\end{align}
Here we used that when $t_{*} \ll \log(n)$ the weight norm $\bs{\theta}_t^2$ changes at most by an $O(1)$ amount (recall $\bs{\theta}^2_0$ is $O(n)$), see Appendix \ref{app:early_time}. To conclude, if $\eta$ is smaller than $\frac{4n}{a_+^2 \bs{\theta}_{+,0}^2}$ by more than a parametrically order $1/n$ amount, then the reduced weight norm $\bs{\theta}_{+,t}^2$ receives $O(n)$ negative corrections in the catapult phase. Since the weight norm is a positive-definite quantity, it must eventually stop receiving large, additive, negative updates. This will happen when $|z_t|=O(1)$, since at this point the order $\eta$ term in \eqref{eq:update_relu_mlp_generic} does not have a definite sign. 
When this happens, the model has either re-entered the lazy phase ($\eta H_t<2$) and converges or the model still has a super-critical learning rate ($\eta H_t>2)$.

To complete our argument, we need to show that if the model has a super-critical learning rate after the first catapult mechanism has finished, then subsequent catapult mechanisms will eventually drive the model to the lazy phase.\footnote{Here by ``catapult mechanism" we mean that the loss curve exhibits the characteristic catapult shape where it rapidly increases and then decreases. This should be contrasted with the ``catapult phase" which is the choice of super-critical learning rates where the model converges. For a sufficiently large learning rate the model can in principle undergo many catapult mechanisms before the loss converges, i.e. the loss curve has multiple peaks.} 
To do this, we will use that during the first catapult mechanism $\bs{\theta}^2$ decreased by an $O(n)$ amount, but that between the end of the first catapult mechanism and the beginning of the second one, the weight norm $\bs{\theta}^2$ can change at most by an $O(1)$ amount.
It then follows that \eqref{eq:eta_bound_generic_label_relu} will hold during the second catapult mechanism and the weight norm will again receive $O(\zeta^{-2})$ negative updates.

To make this concrete, we will denote the time at which the first catapult mechanism ends by $t_1$.
At $t_1$ we have:
\begin{align}
|z_{t_1}|&=O(1),
\\
\eta H_{t_1}&>2,
\\
|\bs{\theta}_{+,t_{*}}^2-\bs{\theta}_{+,t_1}^2|&=O(n), \label{eq:change_norm_first_catapult_mag}
\\
\bs{\theta}_{+,t_1}^2&<\bs{\theta}_{+,t_{*}}^2.
\label{eq:change_norm_first_catapult}
\end{align}
Since the model still has a super-critical learning rate at time $t_1$, the output $z_t$ will again grow exponentially fast.
Therefore, at time $t_2$, such that $1\ll t_2-t_1\ll \log(n)$, we again have $\epsilon_{t_2}\approx z_{t_2}$.
We then want to show the analog of \eqref{eq:eta_bound_generic_label_relu} holds at the beginning of the second catapult mechanism:
\begin{align}
\eta <\frac{4n}{a_+^2\bs{\theta}^2_{+,t_2}} \quad \text{ for } \ \ 1\ll t_2-t_1\ll\log(n).
\label{eq:etaboundsecondcatapult}
\end{align}
If \eqref{eq:etaboundsecondcatapult} is true, then $\bs{\theta}^2_t$ also decreases by an $O(n)$ amount during the second catapult mechanism.
To prove \eqref{eq:etaboundsecondcatapult} holds, we use that under gradient descent $\bs{\theta}^2_{+,t}$ changes by:
\begin{align}
|\bs{\theta}_{+,t+1}^{2}-\bs{\theta}_{+,t}^{2}|=O(\eta z_t^2).
\end{align}
We can now track how $\bs{\theta}_{+,t}^2$ evolves from time $t_1$, when the first catapult mechanism ends and $|z_{t_1}|=O(1)$, to time $t_2$, when the output again becomes large, or $|z_t|\gg1$.
Since $y=O(1)$ and $z_t$ grows exponentially fast, we can take $t_2-t_1=O(1)$.
In this case, the weight norm cannot change significantly from time $t_1$ to time $t_2$:
\begin{align}
|\bs{\theta}_{+,t_2}^{2}-\bs{\theta}_{+,t_1}^{2}|\ll O(n).\label{eq:change_norm_first_catapult_small_change}
\end{align}
Here we used that $t_2-t_1=O(1)$ and that if $t_2-t_1\ll \log(n)$ then $|z_{t_2}|\ll \sqrt{n}$.
Combining \eqref{eq:change_norm_first_catapult_mag}, \eqref{eq:change_norm_first_catapult} and \eqref{eq:change_norm_first_catapult_small_change}, we have $\bs{\theta}_{+,t_2}^{2}<\bs{\theta}_{+,t_{*}}^{2}$ and the inequality \eqref{eq:etaboundsecondcatapult} follows from the inequality \eqref{eq:eta_bound_generic_label_relu}.
Finally, this means $\bs{\theta}_{+,t}^2$ receives large, negative updates during the second catapult mechanism.
One can extend this argument by induction to all possible subsequent catapult mechanisms.

We can note that the above subtlety, that once we include non-trivial labels $y_{\alpha}$ we can only argue that the updates to $\bs{\theta}_t^2$ are negative when the loss is large, is not unique to the ReLU MLP.
This is a generic feature of the $\bs{\theta}^2_t$ update equation with non-trivial labels and will arise repeatedly in Appendix \ref{app:multiple_datapoints}.
We should also note that the same type of argument was used in \cite{lewkowycz2020large} to argue that the NTK of the two-layer, linear MLP trained on a generic dataset receives large negative updates whenever the loss is large.
For the models studied in this work, we generally find it is the weight norm $\bs{\theta}^2_t$, and not the NTK, which decreases monotonically when the loss is large.

\section{Multiple Datapoints}
\label{app:multiple_datapoints}
In this section we will explain how to derive bounds for generic datasets $(\bs{x}_{\alpha},y_{\alpha})$ with $\alpha=1,\ldots, D$. 
We will always take the labels to be one-dimensional, although the generalization to multi-dimensional output is straightforward.
We will also assume that the labels $y_{\alpha}$ do not scale with $\zeta$ or $D$, $y_{\alpha}=O(\zeta^0D^0)$.
The main results for the pure quadratic model are given in \eqref{eq:bound_H_mult_data_claim1} and \eqref{eq:window_eta_claim2}, for the quadratic model with bias in \eqref{eq:inequality_eta_bias_mult_init}, and for the two-layer, homogenous MLP in \eqref{eq:final_bound_generic_scale_invariant}.

\subsection{Pure Quadratic Model}
\label{app_multi_data_pure}
We will consider two different methods to prove that pure quadratic models with generic data can converge at super-critical learning rates. The first method will be more rigorous but will give upper bounds that can be too strong.
The second method gives weaker upper bounds, but will involve approximations on how the NTK eigenvectors evolve in the catapult phase.
\subsubsection*{\textbf{Method 1}}
For the first method we claim that if the learning rate $\eta$ lies in the range:
\begin{align}
\frac{2}{\lambda_{\text{max}}(H_{\alpha\beta,0})}<\eta \lesssim \frac{4}{\lambda_{\text{max}}(\Omega_{\alpha\mu,\beta\nu})\bs{\theta}^{2}_0},\label{eq:bound_H_mult_data_claim1}
\end{align}
then the model is in the catapult phase.
We will define $\Omega$ momentarily and will also explain why we use the approximate inequality ``$\lesssim$" above.

The update equation for $\bs{\theta}_{t}^2$ is:
\begin{align}
\bs{\theta}_{t+1}^{2}=\bs{\theta}_{t}^2-\frac{4\eta}{D}\sum\limits_{\alpha=1}^{D}z_{\alpha,t}\epsilon_{\alpha,t}+\sum\limits_{\alpha,\beta=1}^{D}\frac{\eta^2}{D}\epsilon_{\alpha,t}\epsilon_{\beta,t}H_{\alpha\beta,t}.
\label{eq:update_equation_thetasq_pure}
\end{align}
To simplify this update equation we use that when $\eta \lambda_{\text{max}}(H_{\alpha\beta,0})>2$ the output $z_{\alpha,t}$ grows exponentially quickly at early times in the direction of the top eigenvector of $H_{\alpha\beta,0}$, see \eqref{eq:0th_order_solution_coefficients}.
Here we will only need to know that $z_{\alpha,t}$ is large and will not need to know that it is aligned with the top eigenvector.
We will write:\footnote{Here $|\!|z_{\alpha,t}|\!|^2=\sum\limits_{\alpha=1}^{D}z_{\alpha,t}^2$, i.e. it is the $L_2$ norm in sample-space.}
\begin{align}
z_{\alpha,t}=|\!|z_{\alpha,t}|\!| \hat{z}_{\alpha,t},
\end{align}
where $\hat{z}_{\alpha,t}$ is a unit vector in sample space, i.e. $\sum\limits_{\alpha=1}^{D} \hat{z}_{\alpha,t}^2=1$.

At time $t_*$ such that $1\ll t_* \ll \log(\zeta^{-1})$, the output becomes much larger than the labels, $|\!|z_{\alpha,t_*}|\!|\gg |\!|y_{\alpha}|\!|$, and the update equation for $\bs{\theta}_{t}^2$ simplifies to:
\begin{align}
\bs{\theta}_{t_*+1}^{2}\approx \bs{\theta}_{t_*}^2+\frac{\eta}{D}|\!|z_{\alpha,t_*}|\!|^2\left(\sum\limits_{\alpha,\beta=1}^{D}\eta \hat{z}_{\alpha,t_*}\hat{z}_{\beta,t_*}H_{\alpha\beta,t_*}-4\right).
\label{eq:approx_thetasq_update_claim1}
\end{align}
By taking $t_{*}\gg1$ we are ensuring that $z_{\alpha,t_{*}}$ has a large norm and the approximation $\epsilon_{\alpha,t_*}\approx z_{\alpha,t_*}$ is valid. 
Moreover, since $t_*\ll\log(n)$ we also have $|\!|z_{\alpha,t_*}|\!| \ll \zeta^{-1}$ and can still use small $\zeta$-perturbation theory.
This means that the weight norm at $t_*$ has not deviated significantly from its value at initialization, $\bs{\theta}^2_{t_{*}}-\bs{\theta}^2_{0}=O(1)$ where $\bs{\theta}_0^2=O(\zeta^{-2})$.

Next, we want to derive conditions on $\eta$ such that after $t=t_{*}$, the weight norm $\bs{\theta}_t^2$ decreases by an $O(\zeta^{-2})$ amount.
As in our previous analysis, we start by bounding the NTK $H_{\alpha\beta,t}$ in terms of the weight norm $\bs{\theta}^2_t$.
We will do this by expressing the NTK in terms of the meta-feature functions:
\begin{align}
\sum\limits_{\alpha,\beta=1}^{D}\hat{z}^{\alpha}\hat{z}^{\beta}H_{\alpha\beta}
=
\frac{\zeta^2}{D}\sum\limits_{\alpha,\beta=1}^{D}\sum\limits_{\mu,\nu,\rho=1}^{n}\hat{z}^{\alpha}\theta^{\mu}\psi_{\alpha,\mu\nu}\psi_{\beta,\nu\rho}\theta^{\rho}\hat{z}^{\beta}=\sum\limits_{\alpha,\beta=1}^{D}\sum\limits_{\mu,\rho=1}^{n}\hat{z}^{\alpha}\theta^{\mu}\Omega_{\alpha\mu,\beta\rho}\hat{z}^{\beta}\theta^{\rho},
\end{align}
where we defined the matrix:
\begin{align}
\Omega_{\alpha\mu,\beta\rho}=\frac{\zeta^2}{D}\sum\limits_{\nu=1}^{n}\psi_{\alpha,\mu\nu}\psi_{\beta,\nu\rho}.
\label{eq:omega_definition}
\end{align}
We can think of $\Omega$ as a matrix acting in $z\otimes \theta$ space, i.e. $\Omega\in\mathbb{R}^{(n\times d) \times (n\times d)}$.
The matrix $\Omega_{\alpha\mu,\beta\rho}$ is also positive semi-definite in $z\otimes \theta$ space, which implies:
\begin{align}
\sum\limits_{\alpha,\beta=1}^{D}\hat{z}^{\alpha}_tH_{\alpha\beta}\hat{z}^{\beta}_t\leq \lambda_{\text{max}}\left(\Omega_{\alpha\mu,\beta\rho}\right)\bs{\theta}^{2}_t,
\label{eq:H_bound_Theta}
\end{align}
where we are treating $\Omega$ as a $(n\times d) \times (n\times d)$-dimensional matrix and used that $\hat{z}_{\alpha,t}$ is a unit vector.

Using the approximate update equation \eqref{eq:approx_thetasq_update_claim1} and the bound \eqref{eq:H_bound_Theta} we claim that if:
\begin{align}
\eta < \frac{4}{\lambda_{\text{max}}\left(\Omega_{\alpha\mu,\beta\rho}\right)\bs{\theta}_{t_{*}}^2}, \quad \text{for} \ \ \ 1\ll t_{*}\ll \log(\zeta^{-1}),
\label{eq:eta_bound_upper_mult_data}
\end{align}
then the weight norm $\bs{\theta}_t^2$ decreases by an order $O(\zeta^{-2})$ during the catapult phase.
The argument follows the usual inductive proof we have used so far. If the inequality \eqref{eq:eta_bound_upper_mult_data} holds at some time $t_{*}$ such that $1\ll t_{*}\ll \log(\zeta^{-1})$, then from \eqref{eq:approx_thetasq_update_claim1} the weight norm decreases:
\begin{align}
\bs{\theta}_{t_{*}+1}^2<\bs{\theta}_{t_{*}}^2.
\end{align}
Then, since the matrix $\Omega$ is constant, this means the inequality \eqref{eq:eta_bound_upper_mult_data} continues to hold at time $t=t_{*}+1$:
\begin{align}
\eta < \frac{4}{\lambda_{\text{max}}\left(\Omega_{\alpha\mu,\beta\rho}\right)\bs{\theta}_{t_{*}}^2}< \frac{4}{\lambda_{\text{max}}\left(\Omega_{\alpha\mu,\beta\rho}\right)\bs{\theta}_{t_{*}+1}^2}, \quad \text{for} \ \ \ 1\ll t_{*}\ll \log(\zeta^{-1}).
\end{align}
Therefore, if the inequality \eqref{eq:eta_bound_upper_mult_data} holds, then while $|\!|z_{\alpha,t}|\!|\gg |\!|y_{\alpha,t}|\!|$ the weight norm $\bs{\theta}^2_t$ decreases.
When $t= O(\log(\zeta^{-1})$) we have $|\!|z_{\alpha,t}|\!|=O(\zeta^{-1})$ and the squared weights $\bs{\theta}^2_t$ receive negative $O(\zeta^{-2})$ corrections.
Since $\bs{\theta}^2_t$ is a manifestly positive semi-definite quantity, eventually this process will terminate.
This happens when $|\!|z_{\alpha,t}|\!|=O(1)$, so the updates to $\bs{\theta}_{t}^2$ become small and no longer have a definite sign.\footnote{The other possibility one can think of is that $\eta\sum_{\alpha\beta}\hat{z}_{\alpha}\hat{z}_{\beta}H_{\alpha\beta}-4\rightarrow 0$ so the updates to $\bs{\theta}_{t}^2$ become small while $|\!|z_{\alpha}|\!|$ remains large. However, this is not compatible with the bound \eqref{eq:H_bound_Theta} and the inequality \eqref{eq:eta_bound_upper_mult_data} which imply $\eta\sum_{\alpha\beta}\hat{z}_{\alpha}\hat{z}_{\beta}H_{\alpha\beta}-4$ is strictly negative and bounded away from $0$.}
In practice, the output decreases from $|\!|z_{\alpha,t}|\!|=O(\zeta^{-1})$ to $|\!|z_{\alpha,t}|\!|=O(1)$ in an order-one number of steps.

When $|\!|z_{\alpha,t}|\!|= O(1)$ we have two possibilities.
The first is that $\eta \lambda_{\text{max}}(H_{\alpha\beta,t})<2$, in which case the model is in the lazy phase, we can apply small $\zeta$-perturbation theory, and the loss converges.
The other possibility is that $\eta \lambda_{\text{max}}(H_{\alpha\beta,t})>2$, in which case $z_{\alpha,t}$ grows exponentially fast again.
Following the arguments used in Appendix \ref{ssec:relu_net} to study positive labels in the two-layer ReLU MLP, we can argue that in this case there is a second catapult mechanism where $\bs{\theta}_t^2$ again receives $O(n)$, negative updates.
The arguments below will be identical in form to the ones used in Appendix \ref{ssec:relu_net}, so the reader can safely skip the remainder of this section.

Following the conventions of Appendix \ref{ssec:relu_net}, we will denote the time at which the first catapult mechanism ends by $t_1$, at which time $|\!|z_{\alpha,t_1}|\!|=O(1)$.
Since $\bs{\theta}^2_t$ received negative $O(\zeta^{-2})$ updates during the first catapult mechanism we have:
\begin{align}
|\bs{\theta}_{t_{*}}^2-\bs{\theta}_{t_1}^2|&=O(\zeta^{-2}),\label{eq:change_theta_1}
\\
\bs{\theta}_{t_1}^2&<\bs{\theta}_{t_{*}}^2.\label{eq:change_theta_2}
\end{align}
Next, because the learning rate is still super-critical at time $t=t_1$, the output $z_{\alpha,t}$ grows exponentially fast and after an $O(1)$ number of steps becomes exponentially large again.
We will denote the time at which $z_{\alpha,t}$ becomes exponentially large by $t_2$ and take $1\ll t_2-t_1\ll \log(\zeta^{-1})$.
If we take $t_2-t_1$ to be large, but $O(1)$ (i.e. it does not scale with $\zeta^{-1}$), then we have:
\begin{align}
|\bs{\theta}_{t_1}^2-\bs{\theta}_{t_2}^2|=O(1).\label{eq:change_theta_3}
\end{align}
Our assumption that $1\ll t_2-t_1\ll \log(\zeta^{-1})$ is crucial, the lower bound allows us to make the approximation $\epsilon_{\alpha,t_2}\approx z_{\alpha,t_2}$ while the upper bound allows us to use small $\zeta$-perturbation theory to compute $\bs{\theta}^2_{t_2}$ and argue it can change at most by an $O(1)$ amount when we evolve for an $O(1)$ number of steps. 
Together, \eqref{eq:change_theta_1}-\eqref{eq:change_theta_3} yield:
\begin{align}
\bs{\theta}_{t_{*}}^2-\bs{\theta}_{t_2}^2<0. \label{eq:change_theta_4}
\end{align}
Finally, \eqref{eq:change_theta_4} implies that the inequality \eqref{eq:eta_bound_upper_mult_data} continues to hold at time $t=t_2$:
\begin{align}
\eta < \frac{4}{\lambda_{\text{max}}\left(\Omega_{\alpha\mu,\beta\rho}\right)\bs{\theta}_{t_2}^2}, \quad \text{for} \ \ \ 1\ll t_2-t_1\ll \log(\zeta^{-1}).
\label{eq:eta_bound_upper_mult_data_version2}
\end{align}
The bound \eqref{eq:eta_bound_upper_mult_data_version2} implies that $\bs{\theta}_t^2$ will again receive $O(\zeta^{-2})$ negative corrections when $z_{\alpha,t}=O(\zeta^{-1})$ for the second time.
Given the fact the NTK is a positive semi-definite quantity which is bounded from above by $\bs{\theta}_t^2$, see \eqref{eq:H_bound_Theta}, eventually these large decreases in $\bs{\theta}^2_t$ will drive the model to the lazy phase, $\eta \lambda_{\text{max}}(H_{\alpha\beta,t})<2$.

We can now restate the condition \eqref{eq:eta_bound_upper_mult_data} in terms of the weights at initialization.
Because we assumed $t_{*}\ll \log(\zeta^{-1})$ we have $\bs{\theta}_{t_{*}}^2=\bs{\theta}_{0}^2+O(1)$.
Therefore, we can rewrite \eqref{eq:eta_bound_upper_mult_data} as:
\begin{align}
\eta < \frac{4}{\lambda_{\text{max}}\left(\Omega_{\alpha\mu,\beta\rho}\right)\bs{\theta}_{t_{*}}^2}=\frac{4}{\lambda_{\text{max}}\left(\Omega_{\alpha\mu,\beta\rho}\right)\bs{\theta}_{0}^2}+O(\zeta^2), \quad \text{for} \ \ \ 1\ll t_{*}\ll \log(\zeta^{-1}),
\label{eq:etabound_multi_pure_final}
\end{align}
where we have $O(\zeta^2)$ because of the implicit $\zeta^2$ in the definition of $\Omega$, see \eqref{eq:omega_definition}.
The $O(\zeta^2)$ correction is computable using small $\zeta$-perturbation theory, but we will not give it here.
Instead \eqref{eq:etabound_multi_pure_final} is our final result and the order $\zeta^2$ term in \eqref{eq:etabound_multi_pure_final} is the source of the ``$\lesssim$" in \eqref{eq:bound_H_mult_data_claim1}.

\subsubsection*{\textbf{Method 2}}
While the previous method works, the window \eqref{eq:bound_H_mult_data_claim1} can be very small or simply not exist.
For this reason we will consider a different argument for the existence of the catapult phase in the pure quadratic model.
We conjecture that if the learning rate is in the range:
\begin{align}
\frac{2}{\lambda_{\text{max}}(H_{\alpha\beta,0})}<\eta \lesssim \frac{4}{\zeta^2\bs{\theta}^{2}_0 \lambda_{\text{max}}(\bs{\psi}_{\text{eff}}^{2})},\label{eq:window_eta_claim2}
\end{align}
and $\lambda_{\text{max}}(\bs{\psi}_{\text{eff}}^{2})$ is $O(1)$, then the model is in the catapult phase.
The effective meta-feature function is defined by:
\begin{align}
\bs{\psi}_{\text{eff}}=\frac{1}{\sqrt{D}}\sum\limits_{\alpha=1}^{D}e^{\alpha}_{\text{max},0}\bs{\psi}_{\alpha},
\end{align}
where $e^{\alpha}_{\text{max},0}$ is the top eigenvector of the NTK at initialization.

To argue that the model is in the catapult phase, we follow the set-up of the previous section.
Assuming $\eta \lambda_{\text{max}}(H_{\alpha\beta,0})>2$, the output $z_{\alpha,t}$ grows exponentially fast at early times. At a time $t_*$ such that $1\ll t_*\ll \log(\zeta^{-1})$, the output $z_{\alpha,t_*}$ is exponentially large but we can still use small $\zeta$-perturbation theory.
The leading order solutions take the same form as before:
\begin{align}
z_{\alpha,t_*}&= c_{t_*}^{\text{max}}e^{\text{max}}_{\alpha,0} + O(\zeta),
\label{eq:z_small_zeta_pert}
\\
c^{\text{max}}_{t_*}&=e^{(1-\eta \lambda_{\text{max}}(H_{\alpha\beta,0}))t_*}c^{\text{max}}_{0} + O(\zeta),
\\
H_{\alpha\beta,t_*}&= H_{\alpha\beta,0} + O(\zeta),
\\
\bs{\theta}^{2}_{t_*}&= \bs{\theta}^{2}_{0}+ O(\zeta).\label{eq:approx_theta_claim2}
\end{align}

In contrast to the previous method, at this point we will assume that the eigenvectors of the NTK remain approximately static until $t=O(\log(\zeta^{-1}))$.
If we make this approximation, then the update equation \eqref{eq:update_equation_thetasq_pure} becomes:
\begin{align}
\bs{\theta}^{2}_{t_{*}+1}&=\bs{\theta}^{2}_{t_{*}}+\frac{\eta}{D} (c_{t_*}^{\text{max}})^2 \left(\eta\sum\limits_{\alpha,\beta=1}^{D}e^{\text{max}}_{\alpha,0}e^{\text{max}}_{\beta,0}H_{\alpha\beta,t_{*}}-4\right).
\end{align}
The above update equation for $\bs{\theta}^2_t$ is essentially the same as the update equation in the pure quadratic model with a single datapoint, see \eqref{eq:weight_sq_evolution}.
The important difference is that when we had a single data-point the NTK was simply a number while here we are studying a particular matrix element of the NTK. 

Following our analysis of the model with a single datapoint given in Appendix \ref{app:proof_single_data_quad}, we can bound the NTK in terms of the meta-feature functions:
\begin{align}
\sum\limits_{\alpha,\beta=1}^{D}e^{\text{max}}_{\alpha,0}e^{\text{max}}_{\beta,0}H_{\alpha\beta,t_{*}}\leq \zeta^2\bs{\theta}_{t_{*}}^2 \lambda_{\text{max}}(\bs{\psi}_{\text{eff}}^{2}).\label{eq:bound_effectivepsi}
\end{align}
Using the bound \eqref{eq:bound_effectivepsi}, we can show that if the learning rate $\eta$ obeys the following bound:
\begin{align}
\eta < \frac{4 }{\zeta^2\bs{\theta}^{2}_{t_{*}}\lambda_{\text{max}}(\bs{\psi}_{\text{eff}}^{2})}=\frac{4}{\zeta^2\bs{\theta}^{2}_0 \lambda_{\text{max}}(\bs{\psi}_{\text{eff}}^{2})}+O(\zeta^2) \quad \text{for} \ \ \ 1\ll t_{*}\ll \log(\zeta), \label{eq:bound_eta_method2}
\end{align}
then the weight norm $\bs{\theta}_t^2$ receives negative corrections while $z_{\alpha,t}$ is large \textit{and} aligned with the eigenvector $e_{\alpha,0}^{\text{max}}$.
The second assumption is crucial and underlies the difference between this method and the previous method.

Ignoring this subtlety for a moment, we know from Appendix \ref{app:early_time} that perturbation theory breaks down when $t=O(\log(\zeta^{-1}))$ and $|\!|z_{\alpha,t}|\!|=O(\zeta^{-1})$.
If the learning rate $\eta$ obeys the bound \eqref{eq:bound_eta_method2} and $z_{\alpha,t}$ is aligned with the top eigenvector of the NTK, then $\bs{\theta}_t^2$ decreases by an order $O(\zeta^{-2})$ amount when $|\!|z_{\alpha,t}|\!|=O(\zeta^{-1})$.
We conjecture that this large decrease in $\bs{\theta}_t^2$ is sufficient to ensure that the model does not diverge.

The above argument is not rigorous because we do not have analytic control over the evolution of the NTK eigenvectors when perturbation theory breaks down near $t=O(\log(\zeta^{-1}))$. 
Specifically, the issue is that the bound \eqref{eq:bound_eta_method2} depends on the top eigenvector of the NTK at initialization through our definition of $\bs{\psi}_{\text{eff}}$.
If the top eigenvector evolves significantly during training, and $z_{\alpha,t}$ becomes primarily aligned with this new direction in sample space, there is not a guarantee that \eqref{eq:bound_eta_method2} continues to hold with the new top eigenvector.
That being said, we find numerically that the bound \eqref{eq:bound_eta_method2} is always sufficient to ensure convergence and in fact is generically stronger than necessary. 
See Appendix \ref{app:more_experiments} for more details.

\subsection{Quadratic Model with Bias}
\label{app:derivation_quad_bias_mult}
Here we will extend the above analysis to include quadratic models with bias.
The arguments in this section will largely be the same as ``method 2" above for the pure quadratic model.
We will argue that if the learning rate lies in the range:
\begin{align}
\frac{2}{\lambda_{\text{max}}(H_{\alpha\beta,0})}<\eta\lesssim\frac{4}{2\bs{\phi}^{2}_{\text{eff}}+\zeta^2\lambda_{\text{max}}(\bs{\psi}^2_{\text{eff}})\left(\bs{\theta}^{2}_{0}+\frac{(\bs{\phi}^T_{\text{eff}}\bs{\theta}_{0})^2}{\bs{\phi}^2_{\text{eff}}}\right)},
\label{eq:inequality_eta_bias_mult_init}
\end{align}
then the model converges.
Here the effective (meta-)feature functions are defined by:
\begin{align}
\bs{\phi}_{\text{eff}}&=\frac{1}{\sqrt{D}}\sum\limits_{\alpha=1}^De^{\text{max}}_{\alpha,0}\bs{\phi}_{\alpha},
\\
\bs{\psi}_{\text{eff}}&=\frac{1}{\sqrt{D}}\sum\limits_{\alpha=1}^De^{\text{max}}_{\alpha,0}\bs{\psi}_{\alpha},
\end{align}
where $e^{\text{max}}_{\alpha,0}$ is the top eigenvector of the NTK at initialization.

We will study the quadratic model with bias at time $t_{*}$ such that $1\ll t_{*}\ll \log(\zeta^{-1})$.
In this regime we can use the results from small $\zeta$-perturbation theory given in \eqref{eq:z_small_zeta_pert}-\eqref{eq:approx_theta_claim2}.
If we assume that the eigenvectors of the NTK remain approximately static until $t=O(\log(\zeta^{-1}))$, then we can approximate the update equation for $\bs{\theta}^2_t$ \eqref{eq:weight_norm_early_time} as:
\begin{align}
\bs{\theta}_{t_*+1}^2&=\bs{\theta}^2_{t_*}
+\frac{\eta}{D}(c_{t_*}^{\text{max}})^{2}\left(
\eta\sum\limits_{\alpha,\beta=1}^{D}e_{\alpha,0}^{\text{max}}e_{\beta,0}^{\text{max}}H_{\alpha\beta,t_{*}}-4
\right)
+\frac{2\eta}{D}c_{t}^{\text{max}}\bs{\phi}_{\text{eff}}^T\bs{\theta}_{t_*}.
\label{eq:theta_sq_approx_multi_bias}
\end{align}
Following our analysis for the same model with a single data-point, see Appendix \ref{app:proof_single_data_quad_bias}, we will study the evolution of the quantity:
\begin{align}
\bs{\theta}_{t}^2+\frac{\left(\sum\limits_{\alpha=1}^{D}\hat{z}_{\alpha,t}\bs{\phi}_{\alpha}^T\bs{\theta}_t\right)^2}{\left(\sum\limits_{\alpha=1}^{D}\hat{z}_{\alpha,t}\bs{\phi}_{\alpha}\right)^2}.
\label{eq:monotonic_weight_bias_mult}
\end{align}
Here $\hat{z}_{\alpha,t}$ is the unit vector in sample space parallel to $z_{\alpha,t}$.
We will prove that under certain conditions on $\eta$, the quantity \eqref{eq:monotonic_weight_bias_mult} decreases whenever the loss is large.

In the range $1\ll t_{*}\ll \log(\zeta^{-1})$, the output $z_{\alpha,t_*}$ becomes aligned with the top eigenvector of the NTK and we have:
\begin{align}
\bs{\theta}_{t_{*}}^2+\frac{\left(\sum\limits_{\alpha=1}^{D}\hat{z}_{\alpha,t_{*}}\bs{\phi}_{\alpha}\bs{\theta}_{t_{*}}\right)^2}{\left(\sum\limits_{\alpha=1}^{D}\hat{z}_{\alpha,t_{*}}\bs{\phi}_{\alpha}^T\right)^2}=\bs{\theta}_{t_{*}}^2+\frac{\left(\bs{\phi}_{\text{eff}}^T\bs{\theta}_{t_{*}}\right)^2}{\bs{\phi}_{\text{eff}}^2}+O(1), \quad \text{for} \ \ \ 1\ll t_{*} \ll\log(\zeta^{-1}).
\end{align}
The update equation for $\bs{\phi}_{\text{eff}}^T\bs{\theta}_{t}$ in the range $1\ll t_{*} \ll\log(\zeta^{-1})$ is:
\begin{align}
\bs{\phi}_{\text{eff}}^T\bs{\theta}_{t_{*}+1}=\bs{\phi}_{\text{eff}}^T\bs{\theta}_{t_{*}}-\frac{\eta}{\sqrt{D}}\sum\limits_{\beta=1}^{D}\bs{\phi}^{T}_{\text{eff}}\bs{\phi}_{\beta}\epsilon^{\beta}_{t_{*}}
= 
\bs{\phi}_{\text{eff}}^T\bs{\theta}_{t_{*}}-\eta c_{t_{*}}^{\text{max}}\bs{\phi}_{\text{eff}}^2+O(\zeta),
\end{align}
and the update equation for $\frac{(\bs{\phi}_{\text{eff}}^T\bs{\theta}_{t})^2}{\bs{\phi}_{\text{eff}}^2}$ is:
\begin{align}
\frac{(\bs{\phi}_{\text{eff}}^T\bs{\theta}_{t_{*}+1})^2}{\bs{\phi}_{\text{eff}}^2}\approx
\frac{(\bs{\phi}_{\text{eff}}^T\bs{\theta}_{t_{*}})^2}{\bs{\phi}_{\text{eff}}^2}-2\eta\bs{\phi}^{T}_{\text{eff}}\bs{\theta}_{t_{*}}+\eta^2(c_{t_{*}}^{\text{max}})^2\bs{\phi}_{\text{eff}}^2. \label{eq:update_multi_bias_pt2}
\end{align}
Together \eqref{eq:theta_sq_approx_multi_bias} and \eqref{eq:update_multi_bias_pt2} give the update equation:
\begin{align}
\bs{\theta}_{t_{*}+1}^2+\frac{\left(\bs{\phi}_{\text{eff}}^T\bs{\theta}_{t_{*}+1}\right)^2}{\bs{\phi}_{\text{eff}}^2}\approx\bs{\theta}_{t_{*}}^2+\frac{\left(\bs{\phi}_{\text{eff}}^T\bs{\theta}_{t_{*}}\right)^2}{\bs{\phi}_{\text{eff}}^2}+\eta(c_{t_*}^{\text{max}})^2\left(\eta\left(\frac{1}{D}\sum\limits_{\alpha,\beta=1}^{D}e^{\text{max}}_{\alpha,0}e^{\text{max}}_{\beta,0}H_{\alpha\beta,t_{*}}+\bs{\phi}_{\text{eff}}^2\right)-4
\right).
\label{eq:update_bias_multiple_data}
\end{align}

Finally, we can bound the size of the $\eta^2$ term in \eqref{eq:update_bias_multiple_data} in terms of the effective meta-feature function:
\begin{align}
\frac{1}{D}\sum\limits_{\alpha,\beta=1}^{D}e^{\text{max}}_{\alpha,0}e^{\text{max}}_{\beta,0}H_{\alpha\beta,t}+\bs{\phi}_{\text{eff}}^2&\leq 2\bs{\phi}_{\text{eff}}^{2}+\zeta^2\lambda_{\text{max}}(\bs{\psi}_{\text{eff}}^2)\bs{\theta}_{t}^2 
\nonumber
\\
&\leq 2\bs{\phi}_{\text{eff}}^{2}+\zeta^2\lambda_{\text{max}}(\bs{\psi}_{\text{eff}}^2)\left(\bs{\theta}_{t}^2 +\frac{\left(\bs{\phi}_{\text{eff}}^T\bs{\theta}_{t}\right)^2}{\bs{\phi}_{\text{eff}}^2}\right).
\label{eq:bound_NTK_bias_multiple}
\end{align}
To get the first inequality we used that $\bs{\psi}_{\text{eff}}^2$ is a positive semi-definite matrix. 
The second inequality is trivial since we simply added a positive term.
Finally, we find that if $\eta$ obeys the following bound for $1\ll t_{*}\ll \log(\zeta^{-1})$:
\begin{align}
\eta&\leq\frac{4}{2\bs{\phi}^{2}_{\text{eff}}+\zeta^2\lambda_{\text{max}}(\bs{\psi}^2_{\text{eff}})\left(\bs{\theta}^{2}_{t_{*}}+\frac{(\bs{\phi}^T_{\text{eff}}\bs{\theta}_{t_{*}})^2}{\bs{\phi}^2_{\text{eff}}}\right)}
\nonumber
\\
&
=
\frac{4}{2\bs{\phi}^{2}_{\text{eff}}+\zeta^2\lambda_{\text{max}}(\bs{\psi}^2_{\text{eff}})\left(\bs{\theta}^{2}_{0}+\frac{(\bs{\phi}^T_{\text{eff}}\bs{\theta}_{0})^2}{\bs{\phi}^2_{\text{eff}}}\right)} +O(\zeta),
\label{eq:inequality_eta_bias_mult_init_t0}
\end{align}
then the quantity $\bs{\theta}_{t}^2+\frac{\left(\bs{\phi}_{\text{eff}}^T\bs{\theta}_{t}\right)^2}{\bs{\phi}_{\text{eff}}^2}$ receives large $O(\zeta^{-2})$ negative corrections during the catapult phase.
The argument is essentially the same as ``method 2" of Appendix \ref{app_multi_data_pure}.
If the inequality \eqref{eq:inequality_eta_bias_mult_init_t0} holds at $t=t_{*}$, then from the approximate update equation \eqref{eq:update_bias_multiple_data} we see that $\bs{\theta}_{t}^2+\frac{\left(\bs{\phi}_{\text{eff}}^T\bs{\theta}_{t}\right)^2}{\bs{\phi}_{\text{eff}}^2}$ decreases after one step of gradient descent. This then implies the inequality \eqref{eq:inequality_eta_bias_mult_init_t0} continues to hold at $t=t_{*}+1$.
This procedure continues until the small $\zeta$ expansion breaks down when $z_t=O(\zeta^{-1})$, at which point $\bs{\theta}_{t}^2+\frac{\left(\bs{\phi}_{\text{eff}}^T\bs{\theta}_{t}\right)^2}{\bs{\phi}_{\text{eff}}^2}$ decreases by an $O(\zeta^{-2})$ amount.
Given that the NTK is bounded from above in terms of this quantity by an order-one amount, see \eqref{eq:bound_NTK_bias_multiple}, we expect the large decrease in $\bs{\theta}_{t}^2+\frac{\left(\bs{\phi}_{\text{eff}}^T\bs{\theta}_{t}\right)^2}{\bs{\phi}_{\text{eff}}^2}$ will drive the model back to the lazy phase, $\eta \lambda_{\text{max}}(H_{\alpha\beta,t})<2$.

As with our previous version of the above argument, method 2 of Appendix \ref{app_multi_data_pure}, this argument is not completely rigorous because we do not have analytic control over the evolution of the NTK's eigenvectors during training.
To make the argument rigorous we need to understand the evolution of $\bs{\phi}_{\text{eff}}$ and $\bs{\psi}_{\text{eff}}$ when perturbation theory breaks down near $t=O(\log(\zeta^{-1}))$.
That said, we empirically observe in Appendix \ref{app:more_experiments} that the bound \eqref{eq:inequality_eta_bias_mult_init_t0} is stronger than necessary to ensure convergence at super-critical learning rates.

\subsection{Homogenous, Two-Layer, MLPs}
\label{app:mult_data_two_MLPs}
As our final example, we will study two-layer MLPs with a scale-invariant activation function and multiple datapoints, see \eqref{eq:z_homogenous_two_layer_MLP} and
\eqref{eq:H_homogenous_two_layer_MLP} for $z_{\alpha,t}$ and $H_{\alpha\beta,t}$ in this model.
In practice, we will find our bounds for this MLP will only be non-trivial for uni-dimensional data, $d=1$, although we believe there should be a way to extend our type of arguments to general $d$.

The weight norm in this model is given by:
\begin{align}
\bs{\theta}_{t}^2=\Tr(\bs{U}^T_t\bs{U}_t)+\bs{v}_t^2.
\end{align}
Under gradient descent it evolves as:
\begin{align}
\bs{\theta}_{t+1}^2=\bs{\theta}_{t}^2-\frac{4\eta}{D}\sum\limits_{\alpha=1}^{D}\epsilon_{\alpha,t}z_{\alpha,t}+\frac{\eta^2}{D}\sum\limits_{\alpha,\beta=1}^{D}\epsilon_{\alpha,t}\epsilon_{\beta,t}H_{\alpha\beta,t}.
\label{eq:update_thetasq_MLP_generic}
\end{align}
If we assume $\eta\lambda_{\text{max}}(H_{\alpha\beta,0})>2$ then the output $z_{\alpha,t}$ becomes exponentially large and in an order $O(n^0)$ number of steps we can make the approximation $\epsilon_{\alpha,t}=z_{\alpha,t}-y_{\alpha}\approx z_{\alpha,t}$.
Then for $1\ll t_{*}\ll \log(n)$, we again have $|\!|y_\alpha|\!|\ll |\!|z_{\alpha,t_*}|\!|\ll 1/\sqrt{n}$ and the $\bs{\theta}^2$ update equation can be approximated as:
\begin{align}
\bs{\theta}_{t_{*}+1}^2\approx \bs{\theta}_{t_{*}}^2
+\frac{\eta}{D}|\!|z_{\alpha,t_{*}}|\!|^2\left(
\eta\sum\limits_{\alpha,\beta=1}^{D}\hat{z}_{\alpha,t_{*}}\hat{z}_{\beta,t_{*}}H_{\alpha\beta,t_{*}}-4
\right),
\end{align}
where $\hat{z}_{\alpha,t}$ is a unit vector in sample space and $|\!|z_{\alpha,t_{*}}|\!|$ is the $L_2$ norm in sample space.

To derive a bound on $\eta$ we need to bound the NTK in terms of the weight norm:
\begin{align}
\sum\limits_{\alpha,\beta=1}^D\hat{z}_{\alpha,t}\hat{z}_{\beta,t}H_{\alpha\beta,t}&\leq \frac{a_+^2}{nD} \sum\limits_{\alpha\beta}\hat{z}_{\alpha,t}\hat{z}_{\beta,t}\left(\bs{x}_{\alpha}^T\bs{U}_t^T\bs{U}_t\bs{x}_{\beta}+\bs{x}_{\alpha}^T\bs{x}_{\beta}\bs{v}_t^2
\right)
\nonumber
\\
& \leq \frac{a_+^2}{nD} \sum\limits_{\alpha,\beta=1}^D\hat{z}_{\alpha,t}\hat{z}_{\beta,t}\bs{x}_{\alpha}^{T}\bs{x}_{\beta}\left(\Tr(\bs{U}^T_t\bs{U}_t)+\bs{v}_t^2\right)
\nonumber
\\
& \leq \frac{a_+^2}{nD} \sum\limits_{\alpha,\beta=1}^D \lambda_{\text{max}}(\bs{x}_{\alpha}^{T}\bs{x}_{\beta})\left(\Tr(\bs{U}_t^T\bs{U}_t)+\bs{v}_t^2\right)
\nonumber
\\
&=\frac{a_+^2}{nD}\lambda_{\text{max}}(\bs{x}_{\alpha}^{T}\bs{x}_{\beta})\bs{\theta}_{t}^{2}.
\label{eq:bound_norm_NTK_MLP_generic}
\end{align}
To obtain the first inequality we used our assumption $0<a_0\leq a_+$. The second inequality then follows from a Cauchy-Schwarz inequality.
The final inequality follows from the fact $\bs{x}_{\alpha}^{T}\bs{x}_{\beta}$ is a positive semi-definite matrix and $\hat{z}_{\alpha}$ is a unit vector in sample space.\footnote{Note that the second inequality is generically very weak because we used an inequality of the form:
\begin{align}
\bs{e}^{T}\bs{U}^T\bs{U}\bs{e}\leq \bs{e}^{T}\bs{e}\Tr(\bs{U}^T\bs{U}),
\label{eq:weak_bound_UU}
\end{align}
when we could have used the stronger inequality:
\begin{align}
\bs{e}^{T}\bs{U}^T\bs{U}\bs{e}\leq \bs{e}^{T}\bs{e}\lambda_{\text{max}}(\bs{U}^T\bs{U}),
\end{align}
since $\bs{U}^T\bs{U}$ is a positive semi-definite matrix. 
The fact the bound \eqref{eq:weak_bound_UU} is weak implies our final upper bounds on $\eta$ may be stronger then necessary to ensure convergence. However, using the bound \eqref{eq:weak_bound_UU} is a necessary evil if we want to bound the NTK in terms of the weight norm $\bs{\theta}_t^2$.
To derive weaker bounds on $\eta$ it may be necessary to bound the NTK in a different manner.}

We can then show that if at time $t_{*}$, such that $1\ll t_{*}\ll \log(n)$, the learning rate $\eta$ obeys the bound:
\begin{align}
\frac{2}{\lambda_{\text{max}}(H_{\alpha\beta,0})}<\eta<\frac{4nD}{a_+^2 \lambda_{\text{max}}(\bs{x}_{\alpha}^T\bs{x}_{\beta})\bs{\theta}_{t_{*}}^2}=\frac{4nD}{a_+^2 \lambda_{\text{max}}(\bs{x}_{\alpha}^T\bs{x}_{\beta})\bs{\theta}_{0}^2}+O(1/n),
\label{eq:final_bound_generic_scale_invariant}
\end{align}
then the model is in the catapult phase.
The argument for \eqref{eq:final_bound_generic_scale_invariant} is identical in form to the argument given in Appendix \ref{app_multi_data_pure}, so we will not repeat it here.
Instead we will use that the update equation for $\bs{\theta}_t^2$ in the two-layer, homogenous MLP \eqref{eq:update_thetasq_MLP_generic} has the same form as the $\bs{\theta}^2_t$ update equation in the pure quadratic model \eqref{eq:update_equation_thetasq_pure} and also that the upper bound for the NTK \eqref{eq:bound_norm_NTK_MLP_generic} has the same form as the same bound in the pure quadratic model \eqref{eq:H_bound_Theta}
 with the replacement:
\begin{align}
\lambda_{\text{max}}(\Omega_{\alpha\mu,\beta\rho}) \ \ \Longrightarrow \ \ \frac{a_+^2}{nD}\lambda_{\text{max}}(x^T_{\alpha}x_{\beta}).
\label{eq:replacement_generic_MLP}
\end{align}
Then if we make the replacement \eqref{eq:replacement_generic_MLP} in our result for the pure quadratic model, \eqref{eq:etabound_multi_pure_final}, we find the result \eqref{eq:final_bound_generic_scale_invariant}.

\subsection{\textcolor{black}{Justifications for the Simplifications}}

\textcolor{black}{The above analysis of the training dynamics under the regime of small-$\zeta$ employs several controlled approximations to simplify the proofs. Although some of these approximations are heuristic, they remain justified through a combination of perturbation theory and assumptions consistently applied throughout our work, specifically the condition $\zeta^{-1}\gg |y_{\alpha}|$ for all $\alpha$. We summarize the main approximations below:}

\textcolor{black}{1) To derive equation (\ref{eq:approx_thetasq_update_claim1}), we made the approximation $\epsilon_{\alpha,t_*} \approx z_{\alpha,t_*}$ when $1 \ll t_* \ll \log(\zeta^{-1})$. As shown in Appendix A, $z_{\alpha,t}$ grows exponentially in this early phase, ensuring that the approximation $\epsilon_{\alpha,t_*}= z_{\alpha,t_*}-y_{\alpha} \approx z_{\alpha,t_*}$ becomes increasingly accurate as $t_*$ and $\zeta^{-1}$ grow, provided the relationship $1 \ll t_* \ll \log(\zeta^{-1})$ is maintained.}

\textcolor{black}{2) In establishing the sharp inequality (\ref{eq:eta_bound_upper_mult_data}) within Appendix \ref{app_multi_data_pure}, we initially bounded the expression in terms of $\theta_{t_*}^2$ and subsequently converted it into a bound on $\theta_{0}^2$. This conversion relies on small-$\zeta$ perturbation theory to argue that the difference $\theta_{t_*}^2-\theta_{0}^2$ is $O(1)$, as detailed in equation (47). The outcome is captured clearly in equation (\ref{eq:etabound_multi_pure_final}). Such approximations closely follow analogous procedures used in prior work (\cite{Belkin_quadratic}, \cite{lewkowycz2020large}).}

\textcolor{black}{3) Lastly, we considered the case where gradient descent induces $z_{\alpha,t}$ to decrease from $\Theta(\zeta^{-1})$ to $o(\zeta^{-1})$, while maintaining a super-critical learning rate. Under these conditions, the model effectively reverts to its initial dynamic regime, validating the reuse of approximation (1) above. However, the key distinction here is a significant reduction in $\theta_t^2$ by an amount of $O(\zeta^{-2})$, allowing the inequality analogous to equation (\ref{eq:eta_bound_upper_mult_data}) to remain applicable in analyzing the second "catapult mechanism."}

\textcolor{black}{Furthermore, We also realize the time-scale $t_1$ can be defined more clearly as the first time $t$ after the catapult mechanism has started such that $\theta_t^2$ receives a positive, additive correction. This is almost equivalent to the definition we gave since $\theta^2_t$ can only receive a positive correction when $|\!|z_{\alpha,t}|\!|=O(1)$ but the converse is not necessarily true.}

\textcolor{black}{These approximations align with the spirit of previous theoretical analyses (\cite{Belkin_quadratic}, \cite{lewkowycz2020large}), particularly concerning NTK approximations during early and peak phases. Besides, our application of the small-$\zeta$ expansion in a newly introduced time-scale ($1 \ll t_* \ll \log(\zeta^{-1})$), coupled with the identification and analysis of double-peaked loss curves (e.g., Figures 2a and 9a), are our contributions beyond existing studies.}

\section{More Experiments}
\label{app:more_experiments}
In this appendix we will explain our experimental set-up and present new numerical results.
All experiments were carried out in PyTorch \cite{NEURIPS2019_9015} and the models were trained using full-batch gradient descent.
In figures \ref{fig:random_quad_linear}-\ref{fig:CIFAR10_01_ReLU_two_hidden_layer} the different colored lines in plots (a)-(c), which give the time evolution of the training loss $L_t$, the weight norm $\bs{\theta}_t^2$, and $\eta \lambda_{\text{max}}(H_{\alpha\beta,t})$, correspond to different initial choices of $\eta \lambda_{\text{max}}(H_{\alpha\beta,0})$.
In addition, the vertical dashed lines (blue or red) in our plots of $|\!|\eta H_{\alpha\beta,\infty}|\!|$, the weight norm $\bs{\theta}_{\infty}^2$, and the generalization loss correspond to the theoretical predictions of Appendix \ref{app:single_datapoint} and \ref{app:multiple_datapoints}.
That is, the methods of Appendices \ref{app:single_datapoint} and \ref{app:multiple_datapoints} imply the models should converge for super-critical learning rates to the left of those vertical lines.
For each experiment we fix a random seed in order to compare different learning rates with a fixed weight initialization.
\subsection{Quadratic Model}
\label{app:exp_quad}
\subsubsection{Definitions}
Here we will give more details about how we set up the quadratic model,
\begin{align}
z_\alpha=\bs{\phi}_{\alpha}^{T}\bs{\theta}+\frac{\zeta}{2}\bs{\theta}^T\bs{\psi}_{\alpha}\bs{\theta}.
\end{align}
We focus on the quadratic model with bias, but the pure quadratic model can be recovered by setting $\bs{\phi}=0$.
In the quadratic model with bias we need to impose:
\begin{align}
\bs{\psi}_{\alpha}\bs{\phi}_{\beta}=0, \quad \forall \alpha,\beta.
\end{align}
We will impose this by assuming that the feature and meta-feature functions span different subspaces of $\mathbb{R}^n$.
That is, we split the indices $\mu=\{1,\ldots,n\}$ into two sets $A=\{1\ldots,m\}$ and $B= \{m+1,\ldots,n\}$ so that $\psi_{\alpha,\mu_i\mu_j}\neq 0$ only if $i,j\in A$ and $\phi_{\alpha,\mu_i}\neq 0$ only if $i\in B$.
We will define the meta-feature hidden dimension to be $n_{\bs{\psi}}=m$ and the feature hidden dimension to be $n_{\bs{\phi}}=n-m$. 
In addition, we will assume $\bs{\phi}_{\alpha}^2=O(1)$ and $\zeta^2=1/m$ so that $z_{\alpha}$ is $O(1)$ at initialization when the eigenvalues of $\bs{\psi}_\alpha$ are $O(1)$.
At initialization the weights $\theta_{\mu,t=0}$ are all drawn from $\mathcal{N}(0,1)$.

We will consider the following feature and meta-feature functions:
\begin{align}
\bs{\phi}(\bs{x}_{\alpha})&=\bs{U}\bs{x}_{\alpha}, \label{eq:app_feat_func}
\\
\bs{\psi}(\bs{x}_{\alpha})&=g\left(\sum\limits_{i=1}^{d}\bs{W}^ix_{\alpha,i}\right) ,\label{eq:app_meta_feat_func}
\end{align}
where $\bs{x}_{\alpha}\in\mathbb{R}^{d}$, $\bs{U}\in\mathbb{R}^{n\times d}$, and $\bs{W}^i\in\mathbb{R}^{n\times n}$ for each $i$.
Here $g$ is a generic activation function. 
The matrices $\bs{U}$ and $\bs{W}^{i}$ are fixed throughout training.
We will draw the components of $\bs{U}$ from $\mathcal{N}(0,1)$.
For $\bs{W}^{i}$ we will parameterize it so we can tune its eigenvalues by hand.
Specifically, as in \eqref{eq:W_definition}, we take:
\begin{align}
W^{i}_{\mu\nu}=\sum\limits_{\sigma=1}^n\lambda^{i}_{\sigma}q_{\sigma \mu}^{i}q_{\sigma \nu}^{i},
\end{align}
where $\lambda^i_{\sigma}$ are the eigenvalues of $W^i_{\mu\nu}$ for a fixed $i$ and $q_{\sigma\mu}^i$ are the corresponding eigenvectors.
We will define $\bs{q}^i\in\mathbb{R}^{n\times n}$ as:
\begin{align}
\bs{q}^i=e^{\bs{B}^i},
\label{eq:q_definition}
\end{align}
where $\bs{B}^i$ is an anti-symmetric matrix whose independent components are drawn from $\mathcal{N}(0,1)$. 
In \eqref{eq:q_definition} we are taking the matrix exponential, and not the component-wise exponential, and using that the exponential of an anti-symmetric matrix gives an orthogonal matrix (more precisely it gives a matrix in SO$(N)$).

We will train the quadratic model on three types of datasets:
\begin{enumerate}
\item The toy dataset $(x,y)=(1,0)$.
\item Random datasets where we draw $\bs{x}_{\alpha}$ and $y_\alpha$ from cubes of the appropriate dimension, $\bs{x}_{\alpha}\in\mathcal{U}([-k,k]^d)$ and $y_{\alpha}\in\mathcal{U}([-k,k])$.
\item Teacher-student set-ups where we train a student quadratic model to recover the predictions of a teacher quadratic model. Here the datapoints are drawn uniformly from a $d$-dimensional cube and the labels are produced from the teacher model.
\end{enumerate}

To explain the teacher-student set-up, we generate the labels using teacher (meta-)feature functions:
\begin{align}
y_{\alpha}=\bs{\phi}_{\text{teacher},\alpha}^{T}\bs{\theta}_{\text{teacher}}+\frac{\zeta_{\text{teacher}}}{2}\bs{\theta}^T_{\text{teacher}}\bs{\psi}_{\text{teacher},\alpha}\bs{\theta}_{\text{teacher}},
\end{align}
and the predictions are generated by student (meta-)feature functions:
\begin{align}
z_{\alpha}=\bs{\phi}_{\text{student},\alpha}^{T}\bs{\theta}_{\text{student}}+\frac{\zeta_{\text{student}}}{2}\bs{\theta}_{\text{student}}^T\bs{\psi}_{\text{student},\alpha}\bs{\theta}_\text{student}.
\end{align}
The student (meta-)feature functions are defined by projecting the corresponding teacher (meta-)feature functions:
\begin{align}
\bs{\phi}_{\text{student}}(\bs{x}_{\alpha})&=\bs{Q}\bs{\phi}_{\text{teacher}}(\bs{x}_{\alpha}),
\\
\bs{\psi}_\text{student}(\bs{x}_{\alpha})&=\bs{Q}\bs{\psi}_{\text{teacher}}(\bs{x}_{\alpha})\bs{Q}^T,
\end{align}
where $\bs{Q}$ is a projector which maps from the teacher to the student hidden dimension, $\mathbb{R}^{n_{\text{teacher}}}\rightarrow \mathbb{R}^{n_\text{student}}$.
The two $\zeta$'s are defined independently and chosen such that $z_{\alpha,0}, y_{\alpha}=O(1)$.

\subsubsection{Pure Quadratic Model}

\begin{figure*}[!ht]
\centering
\begin{subfigure}[b]{0.3\textwidth}
\centering
\includegraphics[scale=.27]{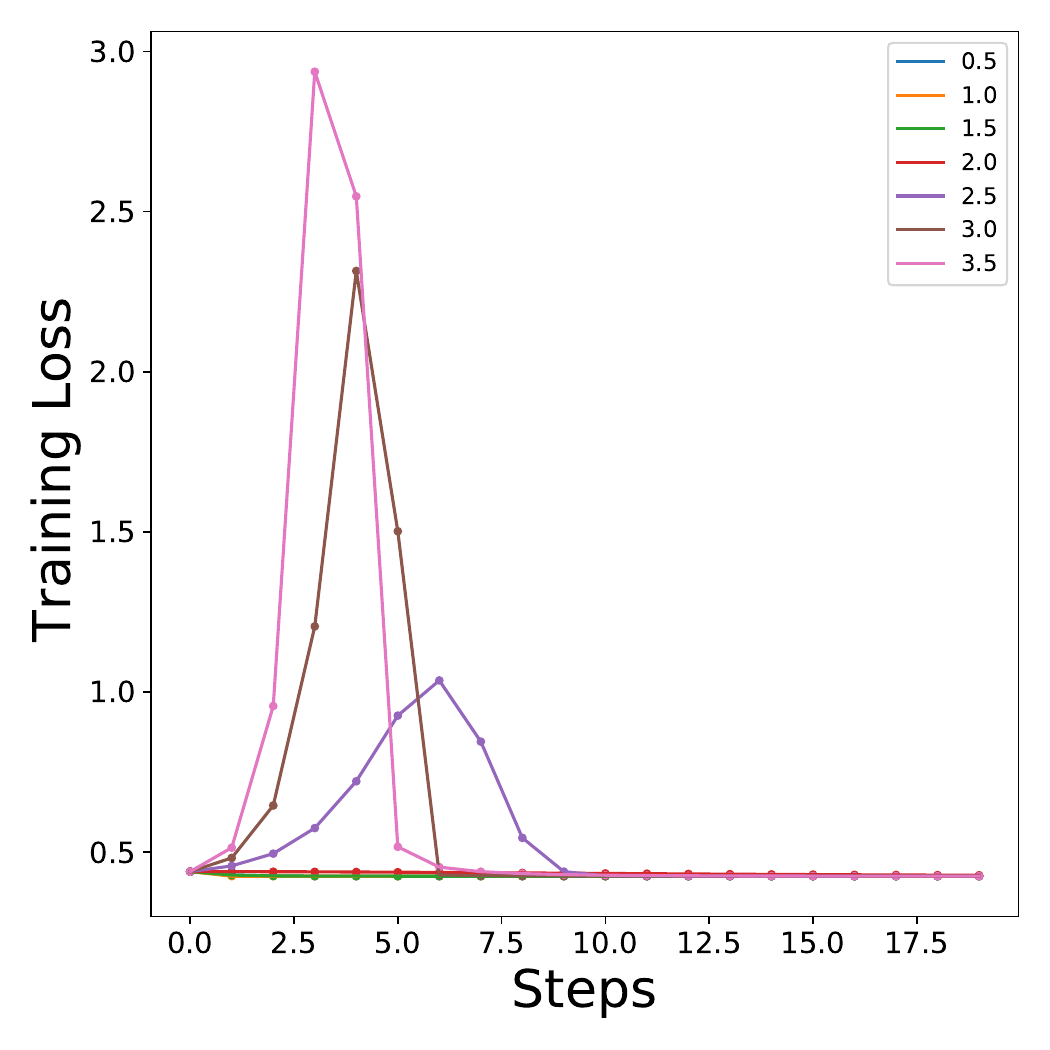}
\caption{}
\end{subfigure}
\hfill
\begin{subfigure}[b]{0.3\textwidth}
\centering
\includegraphics[scale=.27]{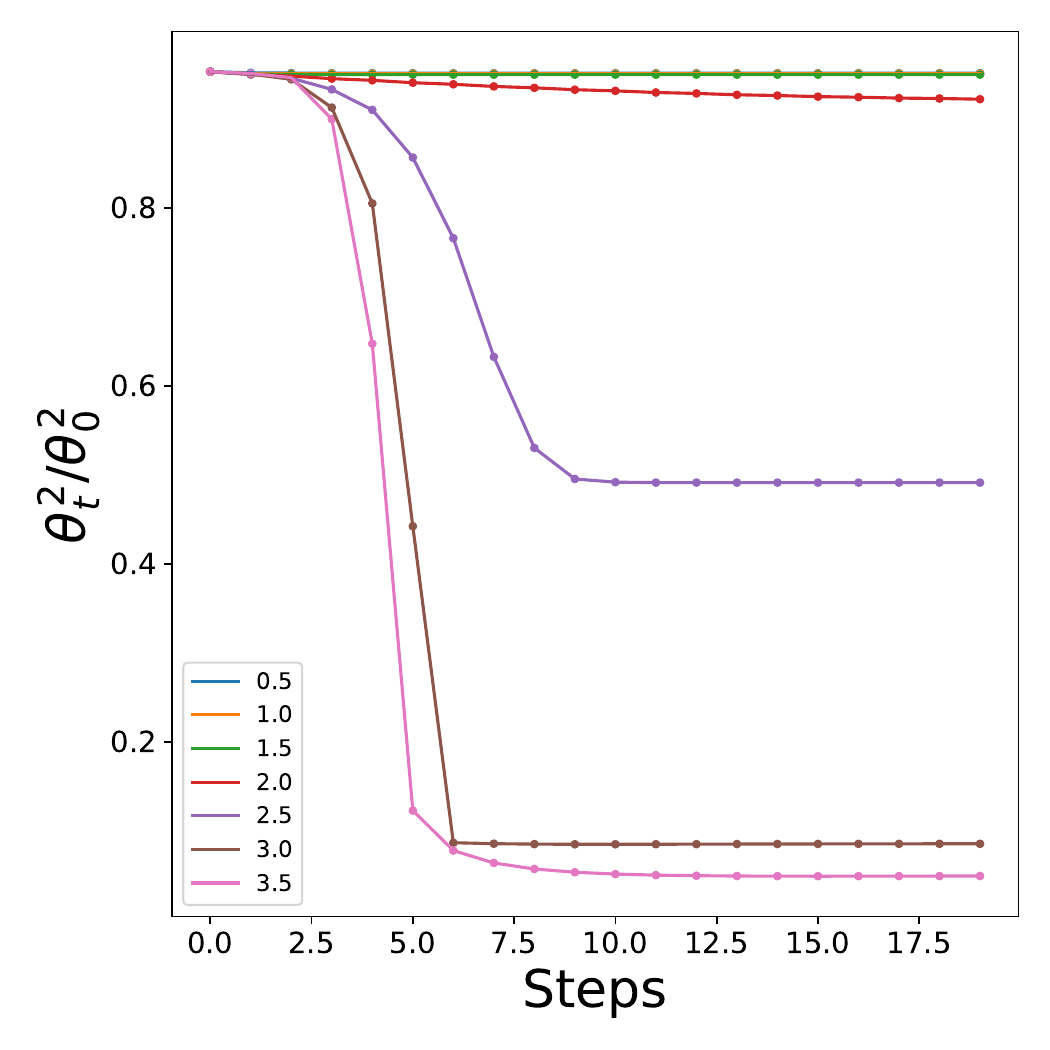}
\caption{}

\end{subfigure}
\hfill
\begin{subfigure}[b]{0.3\textwidth}
\centering
\includegraphics[scale=.27]{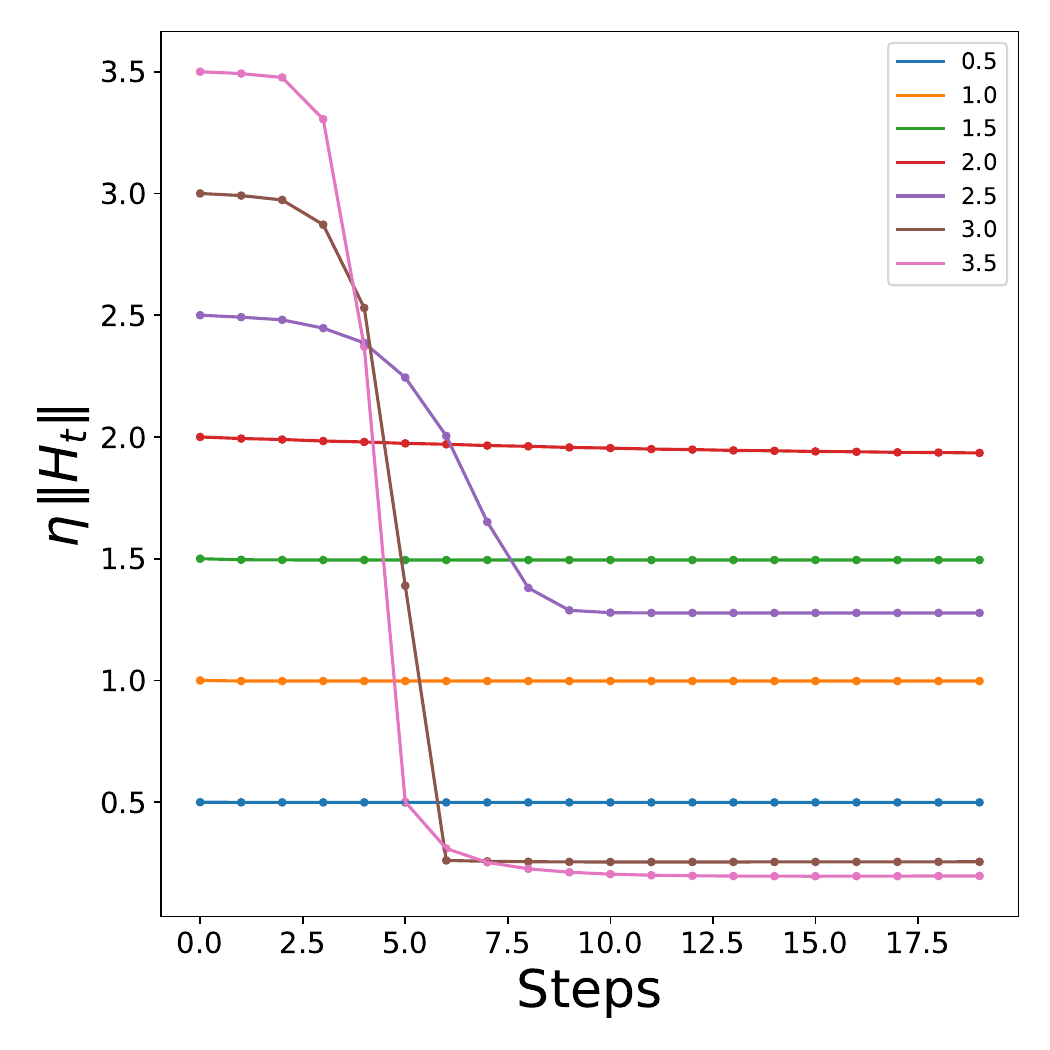}
\caption{}

\end{subfigure}
\hfill
\centering
\begin{subfigure}[t]{0.3\textwidth}
\centering
\includegraphics[scale=.27]{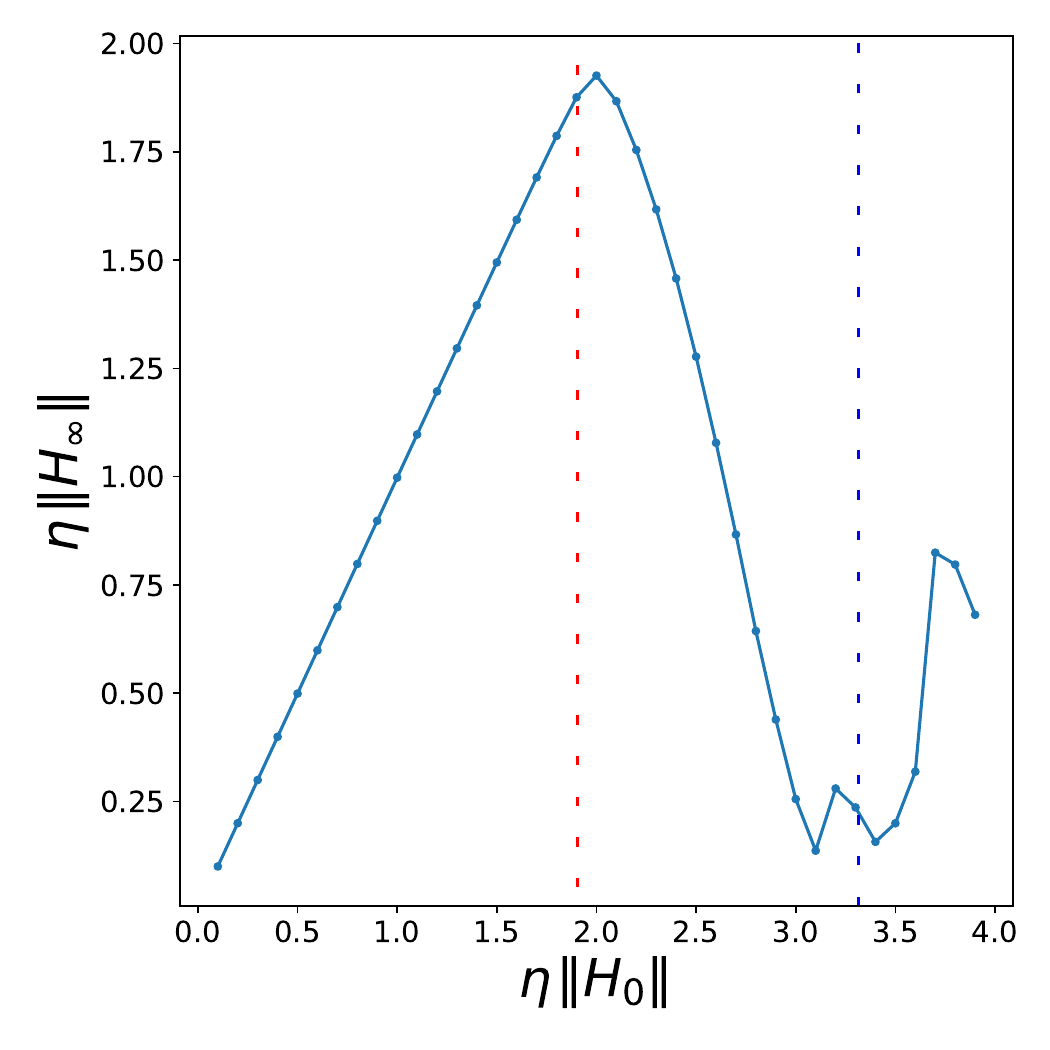}
\caption{}
\label{fig:random_quad_linear_lrNTK}
\end{subfigure}
\begin{subfigure}[t]{0.3\textwidth}
\centering
\includegraphics[scale=.27]{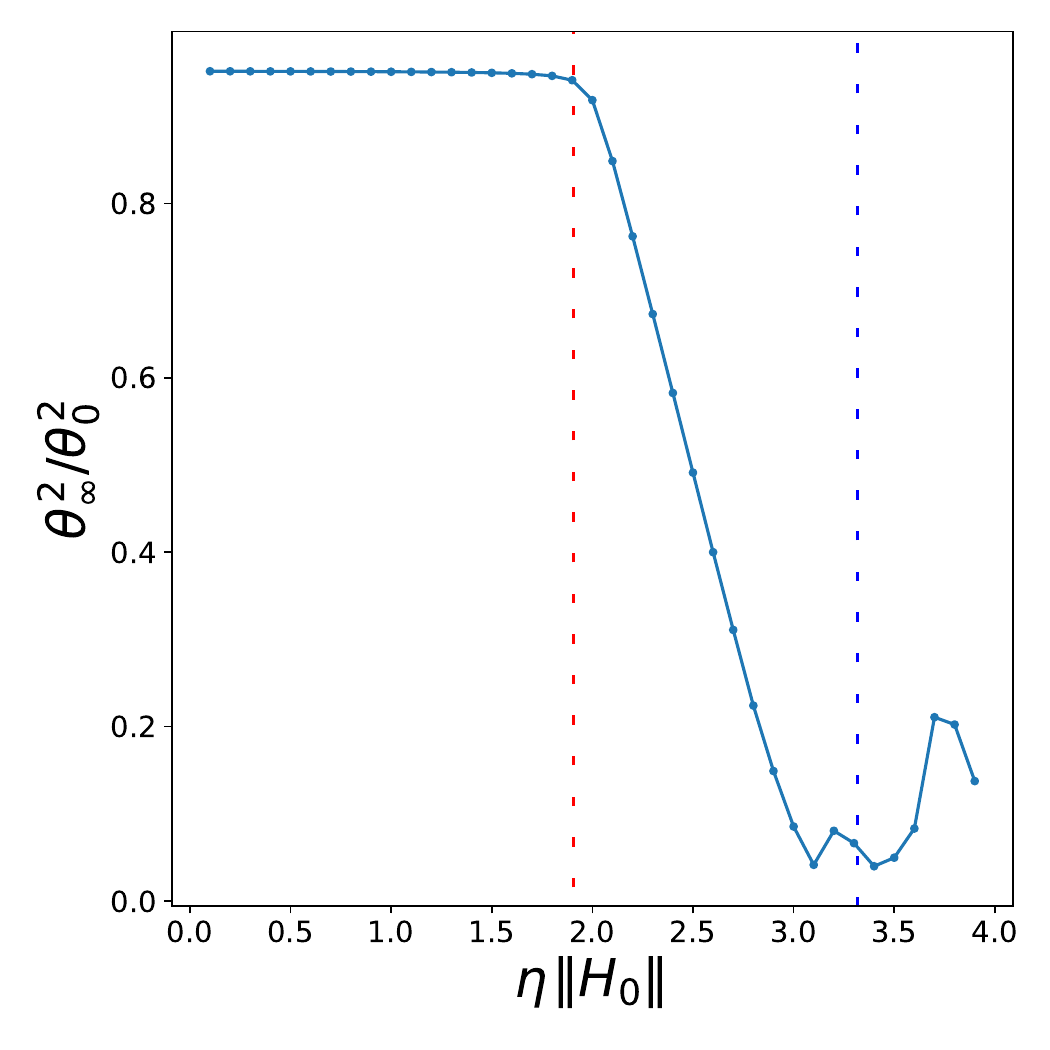}
\caption{}
\label{fig:random_quad_linear_final_weight}
\end{subfigure}
\caption{Results for the pure quadratic model trained on random two-dimensional data and one-dimensional labels. The plots are the same as in figure \ref{fig:linear_meta_toy_dataset_ev1_2}: (a)-(c) give the time evolution of the loss, the weight norm, and the NTK while (d)-(e) give the final value of the NTK and weight norm as a function of the (normalized) learning rate. The only difference is we have one dashed red line and one dashed blue line corresponding to predictions from method 1 and method 2 of Appendix \ref{app_multi_data_pure}, respectively.}
\label{fig:random_quad_linear}
\end{figure*}

In Figure \ref{fig:random_quad_linear} we plot the results for a pure quadratic model trained with a random dataset.
We take $\bs{x}_{\alpha}\sim \mathcal{U}([1/2,1/2]^2)$ (i.e. the interior of a unit, two-dimensional cube) and the labels are $y_{\alpha}\sim\mathcal{U}([-1/2,1/2])$.
We take the activation function $g$ in \eqref{eq:app_meta_feat_func} to be the identity function.
We also take the weights $\bs{\theta}\in\mathbb{R}^{500}$.
We draw the positive eigenvalues of $\bs{W}^i$ from $\mathcal{U}([.9,1.1])$ and take the negative eigenvalues to be exactly $-1$ times the positive eigenvalues.
Finally, we train the model for 100 epochs.

We observe this model has all the expected features of the catapult phase, the loss has the characteristic spike at early times and both the top eigenvalue of the NTK and the weight norm decrease significantly.
The blue and red dashed lines in figures \ref{fig:random_quad_linear_lrNTK} and \ref{fig:random_quad_linear_final_weight} correspond to the predictions from method 1 and 2 of Appendix \ref{app_multi_data_pure}, respectively.
Note that here the prediction from method 1 is trivial. 
In order to get a non-trivial result we need the red line to sit to the right of $\eta\lambda_{\text{max}}(H_{\alpha\beta,0})=2$, see the bound in \eqref{eq:bound_H_mult_data_claim1}. 
Therefore, we cannot use method 1 here to argue that the model has a catapult phase.
On the other hand, the bound from method 2 is non-trivial because the blue dashed line sits to the right of $\eta\lambda_{\text{max}}(H_{\alpha\beta,0})=2$, see \eqref{eq:window_eta_claim2}.
The results of Figure \ref{fig:random_quad_linear} illustrate how our bounds are sufficient, but not necessary, for the existence of the catapult phase.

\begin{figure*}[!ht]
\centering
\begin{subfigure}[b]{0.3\textwidth}
\centering
\includegraphics[scale=.27]{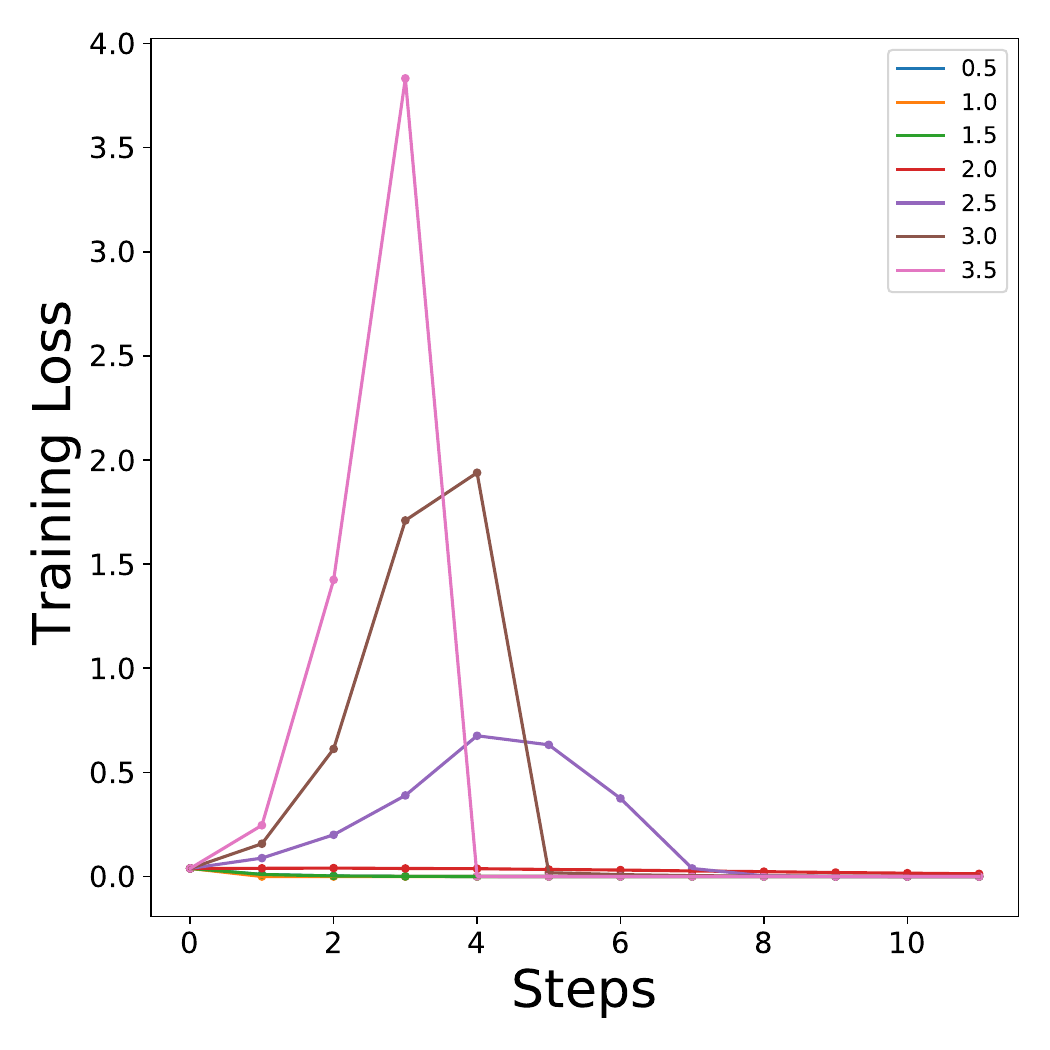}
\caption{}
\end{subfigure}
\hfill
\begin{subfigure}[b]{0.3\textwidth}
\centering
\includegraphics[scale=.27]{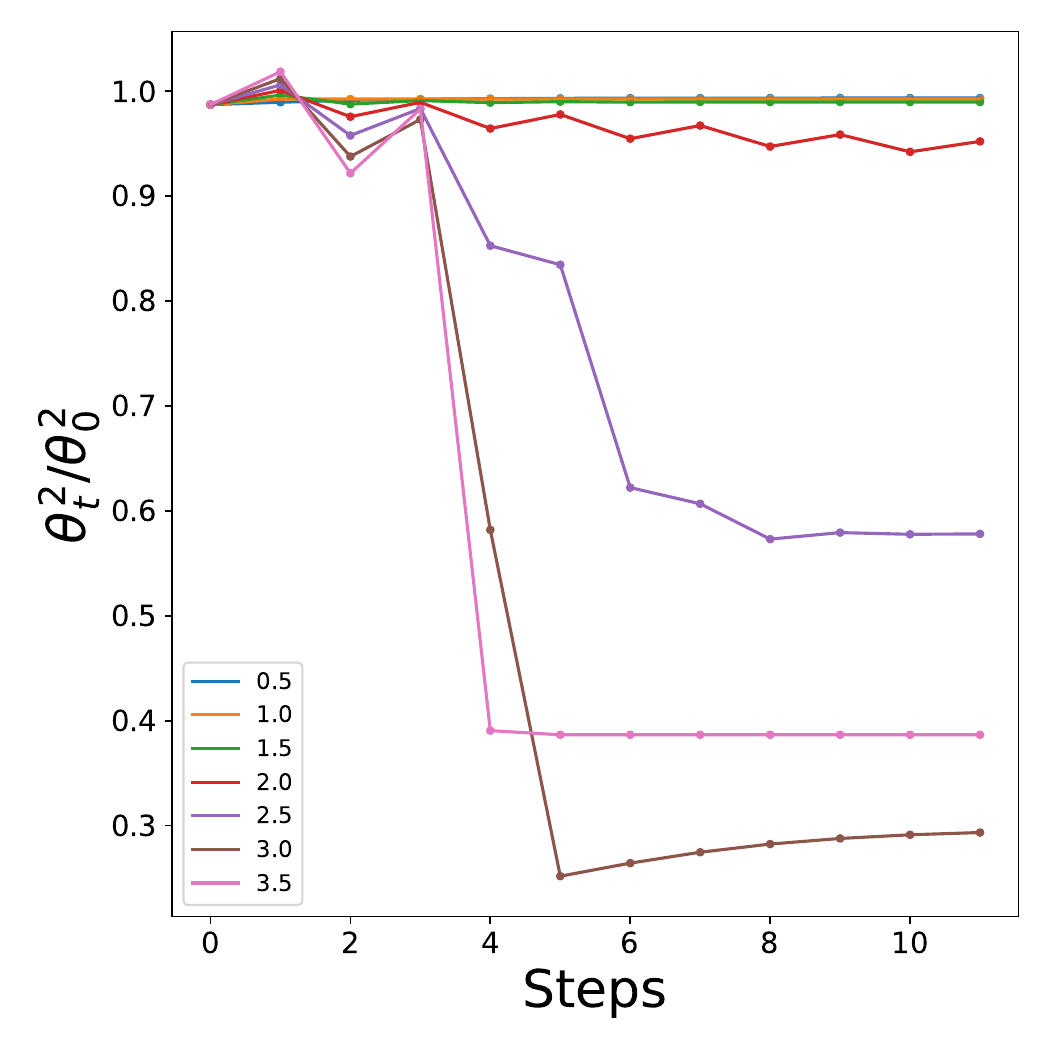}
\caption{}
\end{subfigure}
\hfill
\begin{subfigure}[b]{0.3\textwidth}
\centering
\includegraphics[scale=.27]{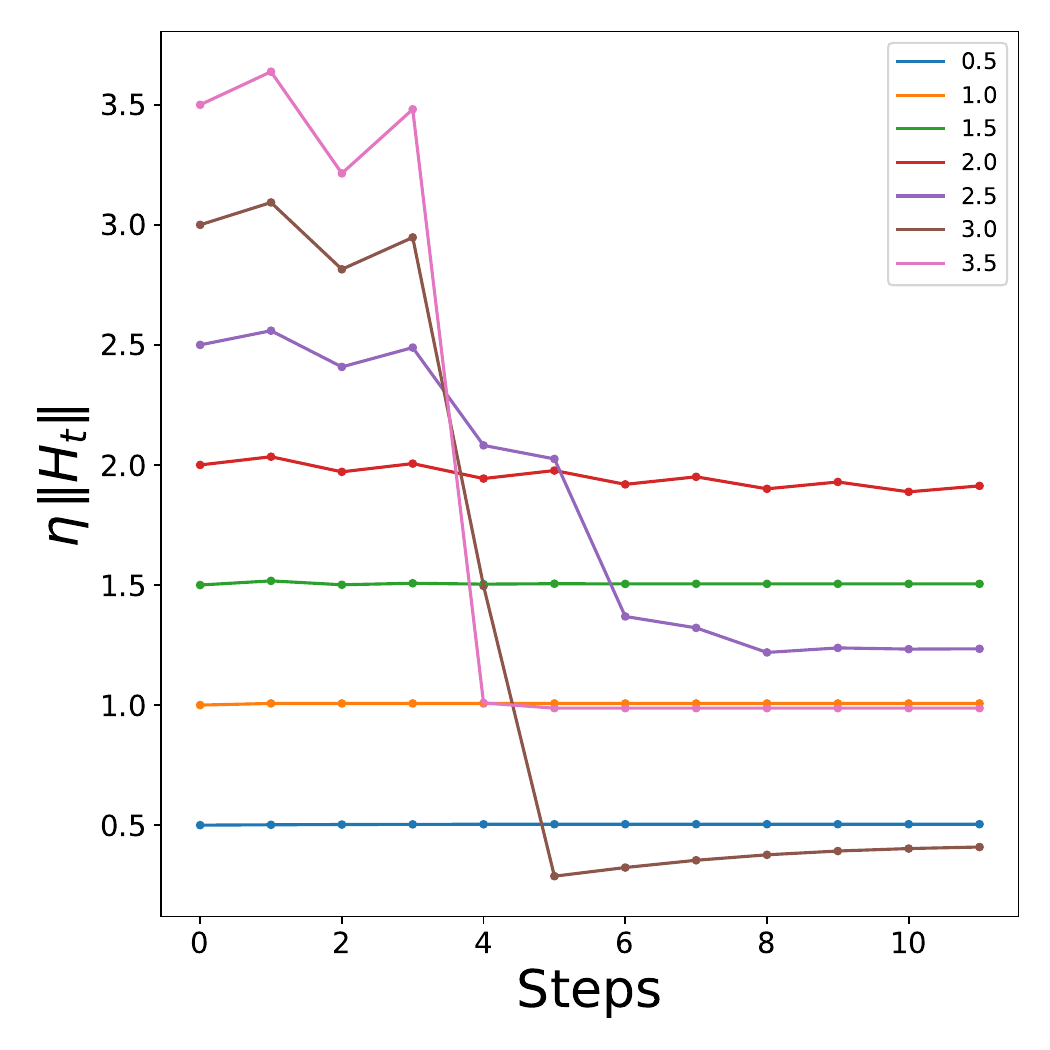}
\caption{}
\end{subfigure}
\hfill
\centering
\begin{subfigure}[t]{0.3\textwidth}
\centering
\includegraphics[scale=.27]{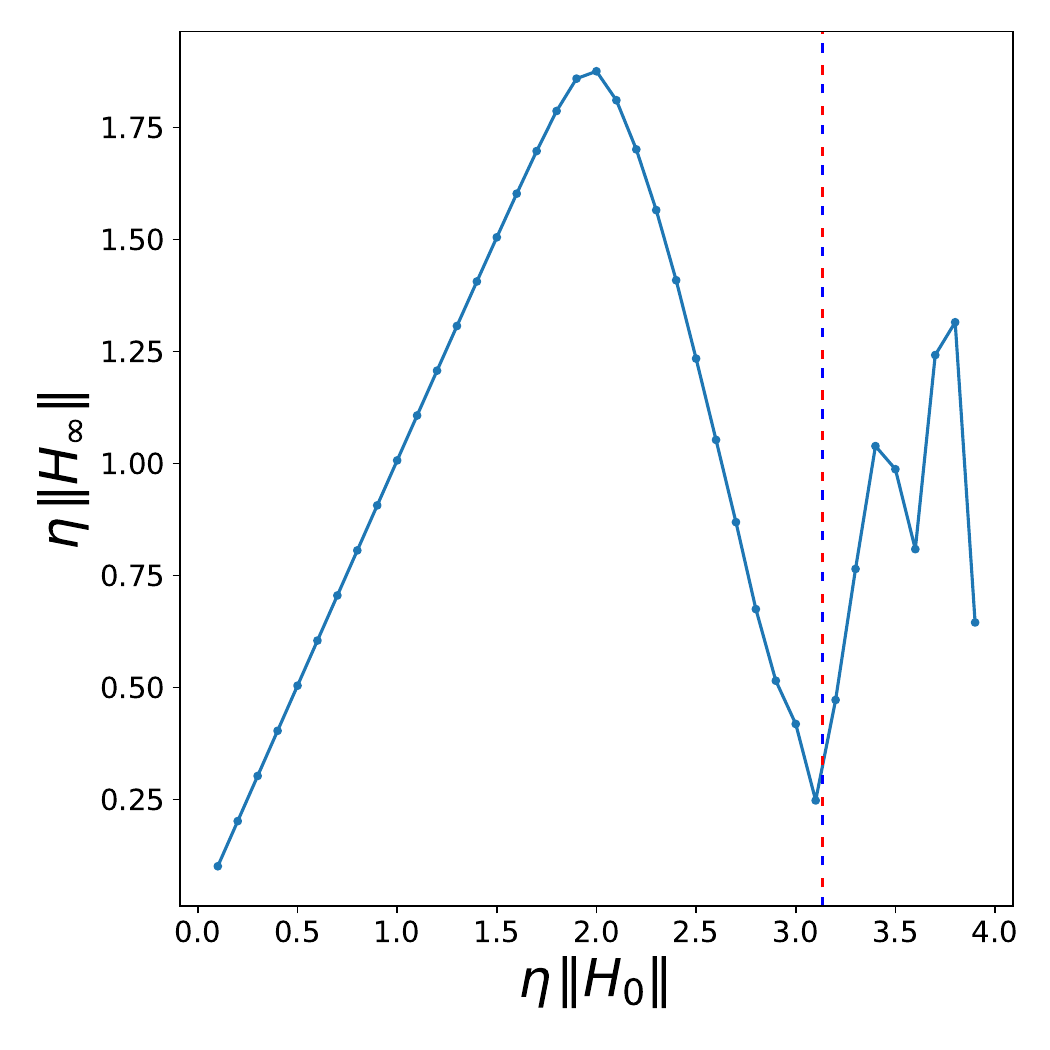}
\caption{}
\label{fig:teacher_quad_tanh_final_NTK}
\end{subfigure}
\hfill
\begin{subfigure}[t]{0.3\textwidth}
\centering
\includegraphics[scale=.27]{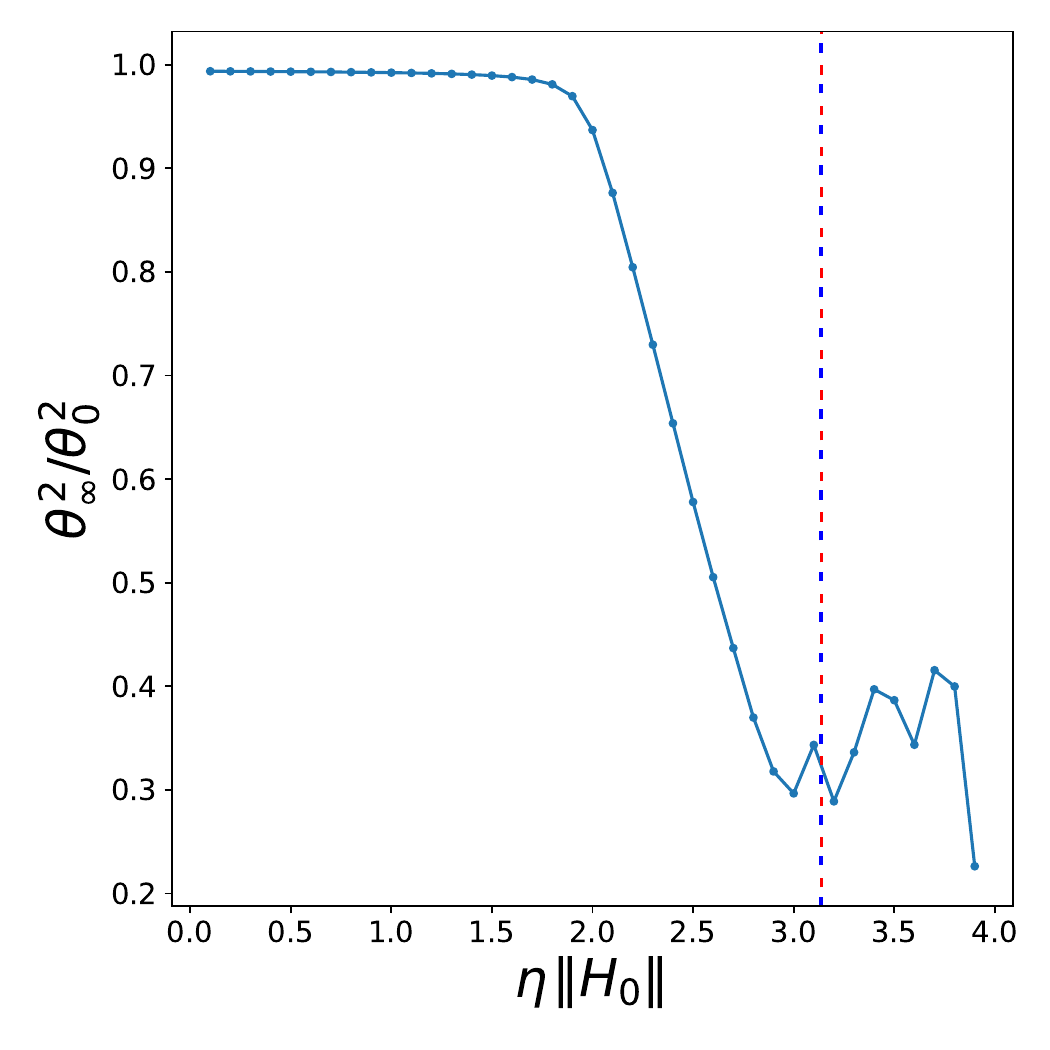}
\caption{}
\label{fig:teacher_quad_tanh_final_weight}
\end{subfigure}
\hfill
\begin{subfigure}[t]{.3\textwidth}
\centering
\includegraphics[scale=.27]{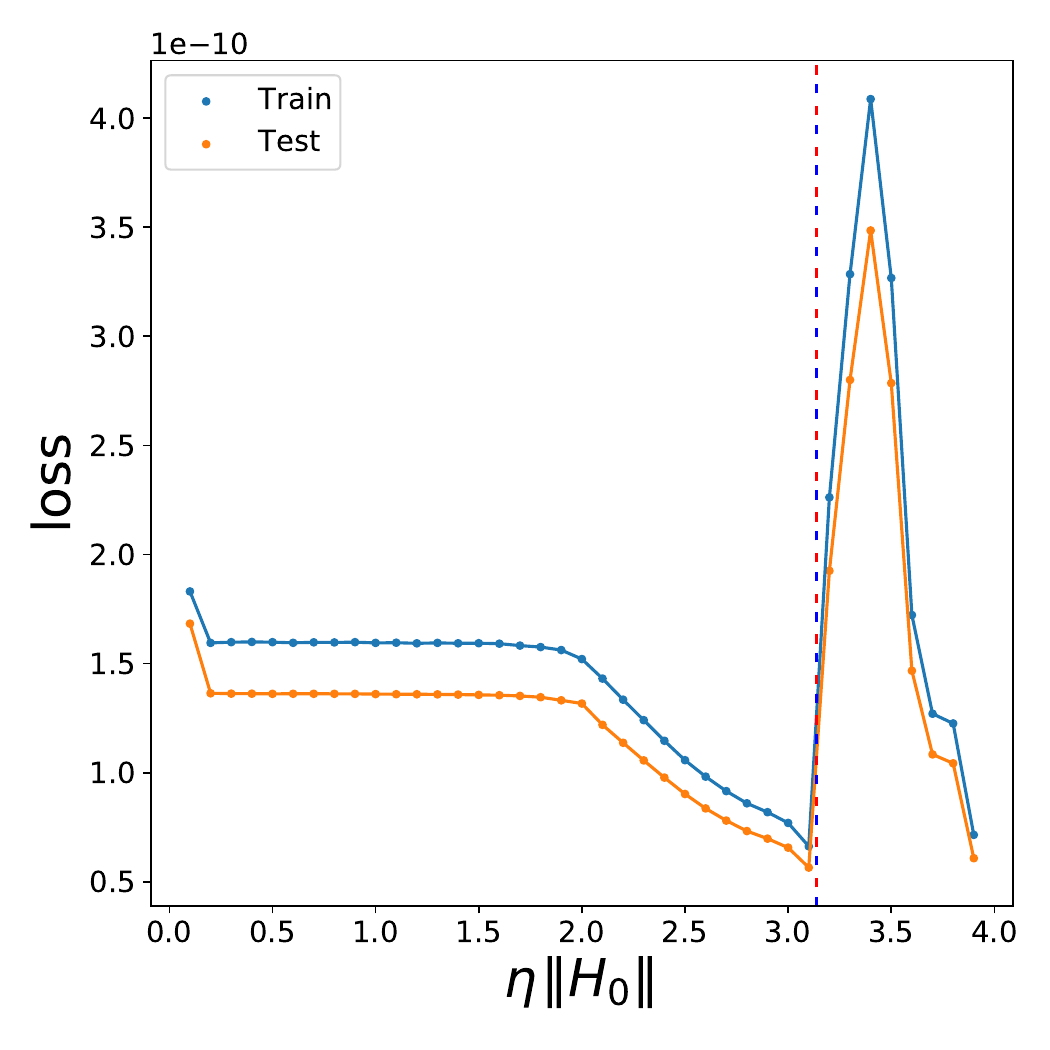}
\caption{}
\label{fig:teacher_quad_tanh_gen_loss}
\end{subfigure}

\caption{Results for the pure quadratic model trained using a teacher-student set-up. }
\label{fig:teacher_quad_tanh}
\end{figure*}

In Figure \ref{fig:teacher_quad_tanh} we plot the results for a pure quadratic model trained using the teacher student set-up.
We take the teacher meta-feature function to have rank 500 and the student meta-feature function to have rank 400.
In order to produce a non-linear function, in \eqref{eq:app_meta_feat_func} we take the activation function $g$ to be the tanh function.
We also take the eigenvalues of $\bs{W}^{i}$ to be $\pm 1$ for all $i$.
We draw the datapoints $x_{\alpha}\sim \mathcal{U}([-1/2,1/2])$ and take the training and test set to have size $32$ and $1000$, respectively.
 
Here we observe that both method 1 (the red dashed line) and method 2 (the blue dashed line) of Appendix \ref{app_multi_data_pure} give non-trivial predictions. 
Moreover, our experimental results match our theoretical predictions since the model converges for super-critical learning rates to the left of both dashed lines in figures \ref{fig:teacher_quad_tanh_final_NTK}-\ref{fig:teacher_quad_tanh_gen_loss}.
For this experiment the predictions from both methods agree to a high degree of precision and are virtually indistinguishable on the plots.

\subsubsection{Quadratic Model with Bias}
\begin{figure*}[!ht]
\centering
\begin{subfigure}[t]{0.3\textwidth}
\centering
\includegraphics[scale=.27]{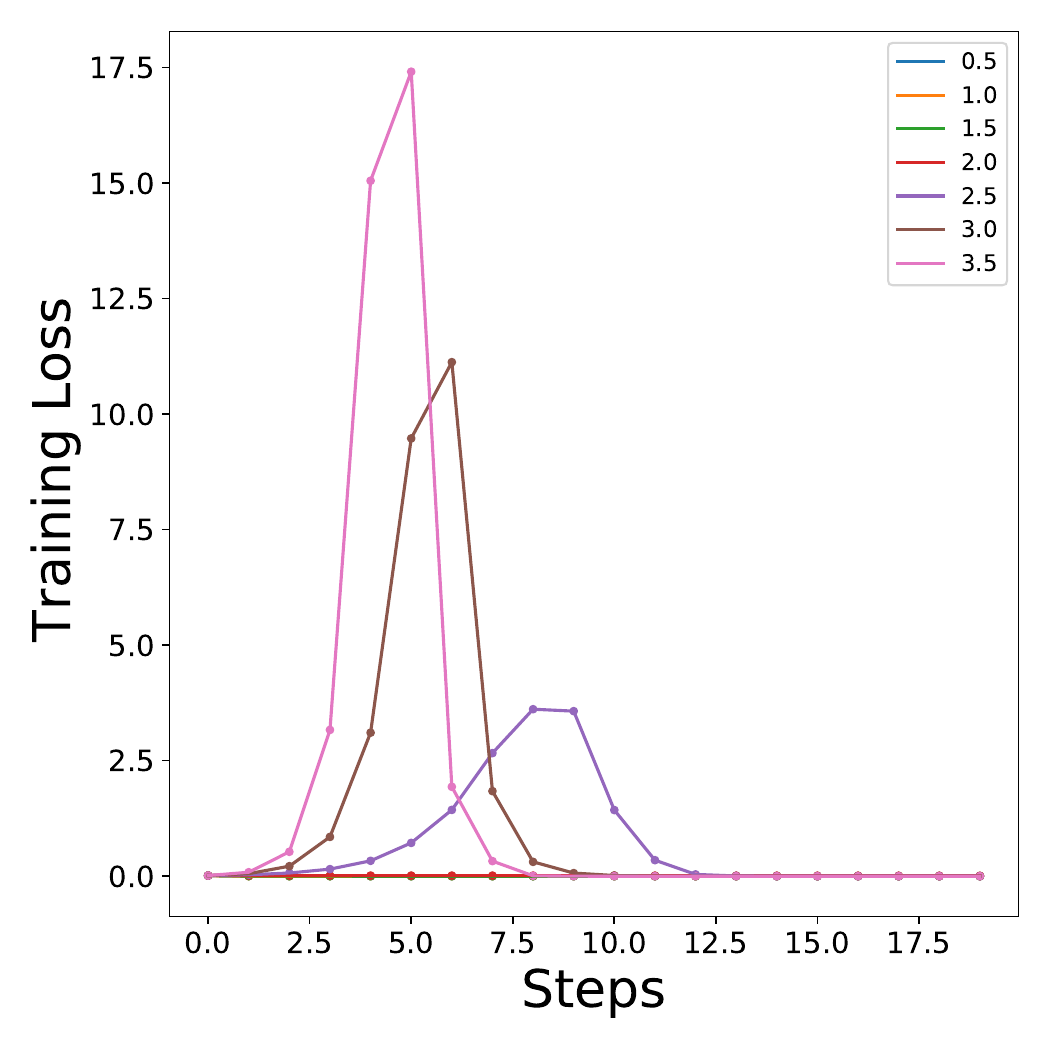}
\caption{}
\end{subfigure}
\begin{subfigure}[t]{0.3\textwidth}
\centering
\includegraphics[scale=.27]{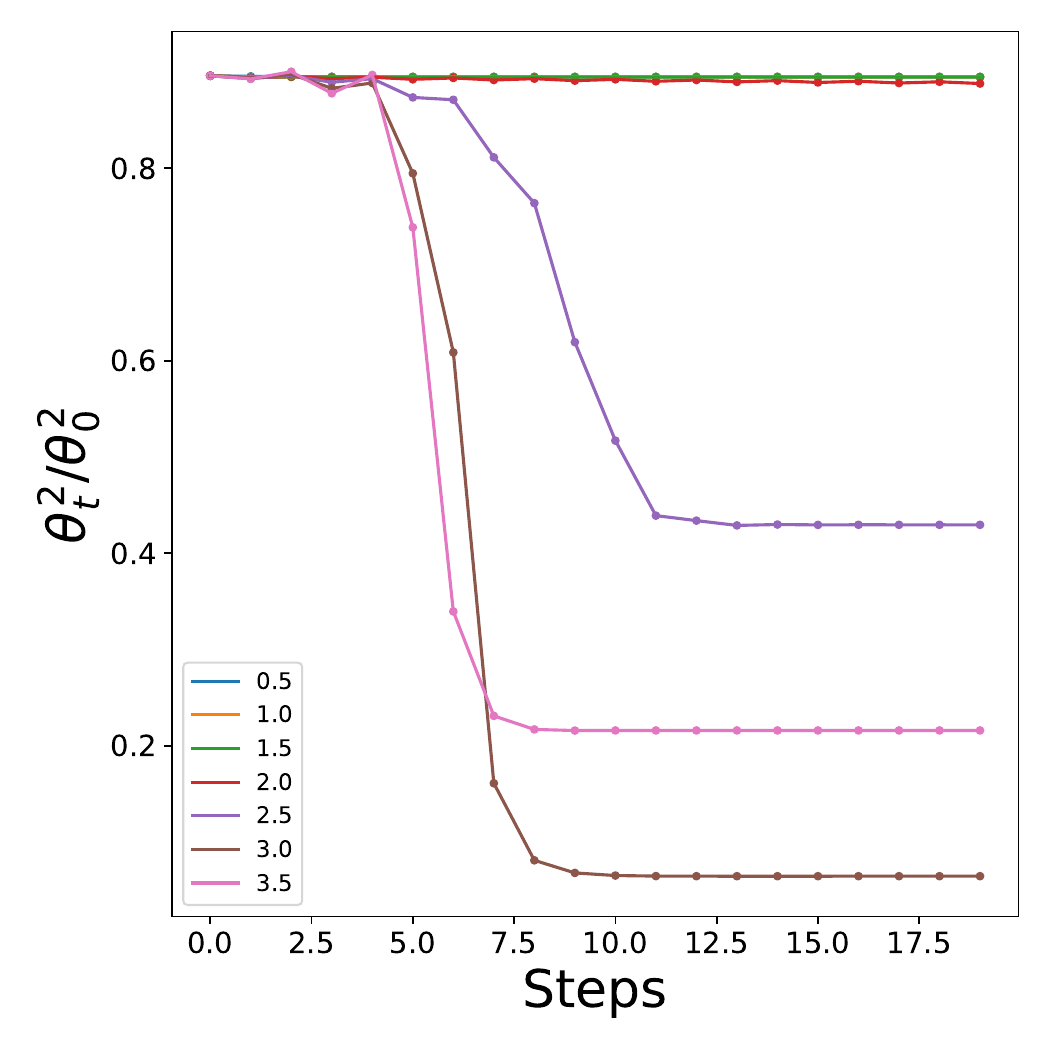}
\caption{}
\end{subfigure}
\begin{subfigure}[t]{0.3\textwidth}
\centering
\includegraphics[scale=.27]{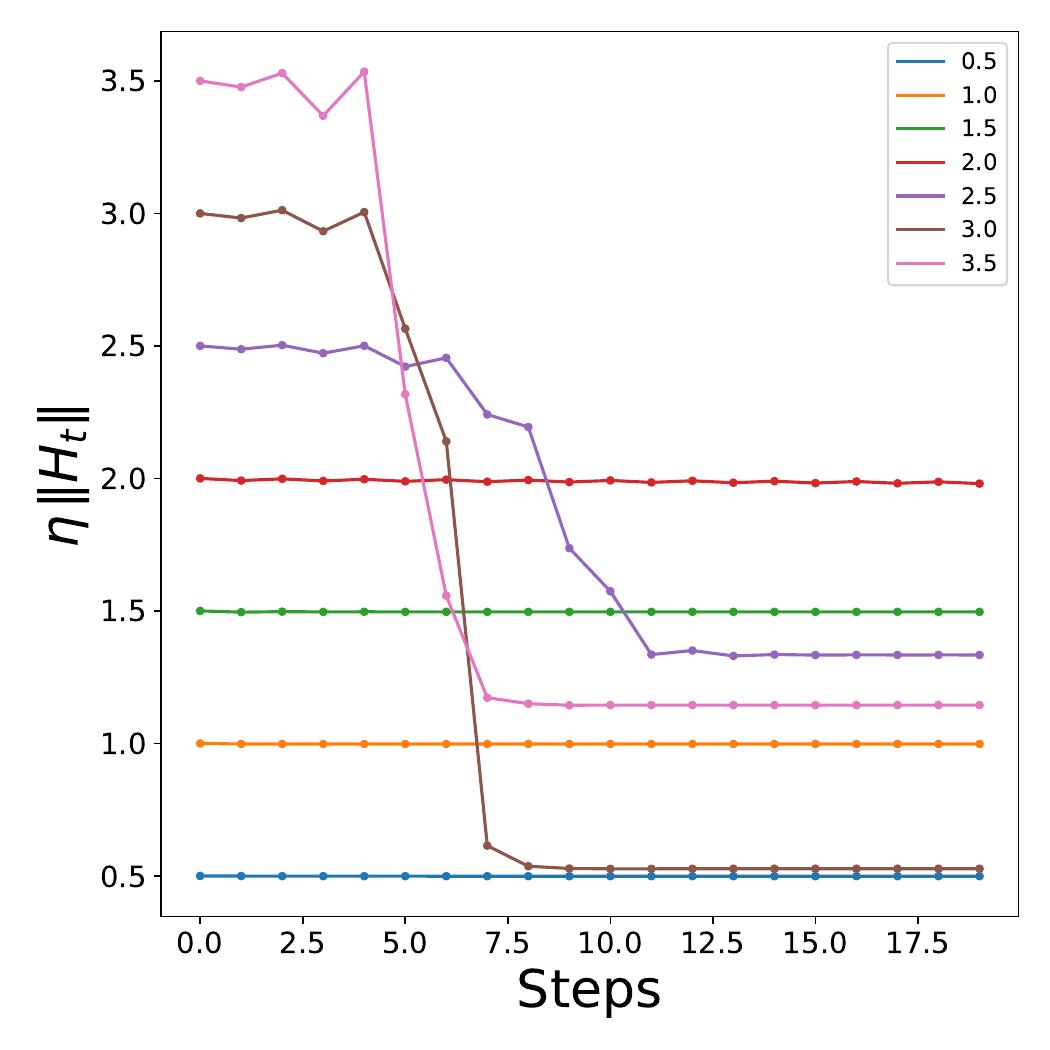}
\caption{}
\end{subfigure}
\centering
\begin{subfigure}[b]{0.4\textwidth}
\centering
\includegraphics[scale=.27]{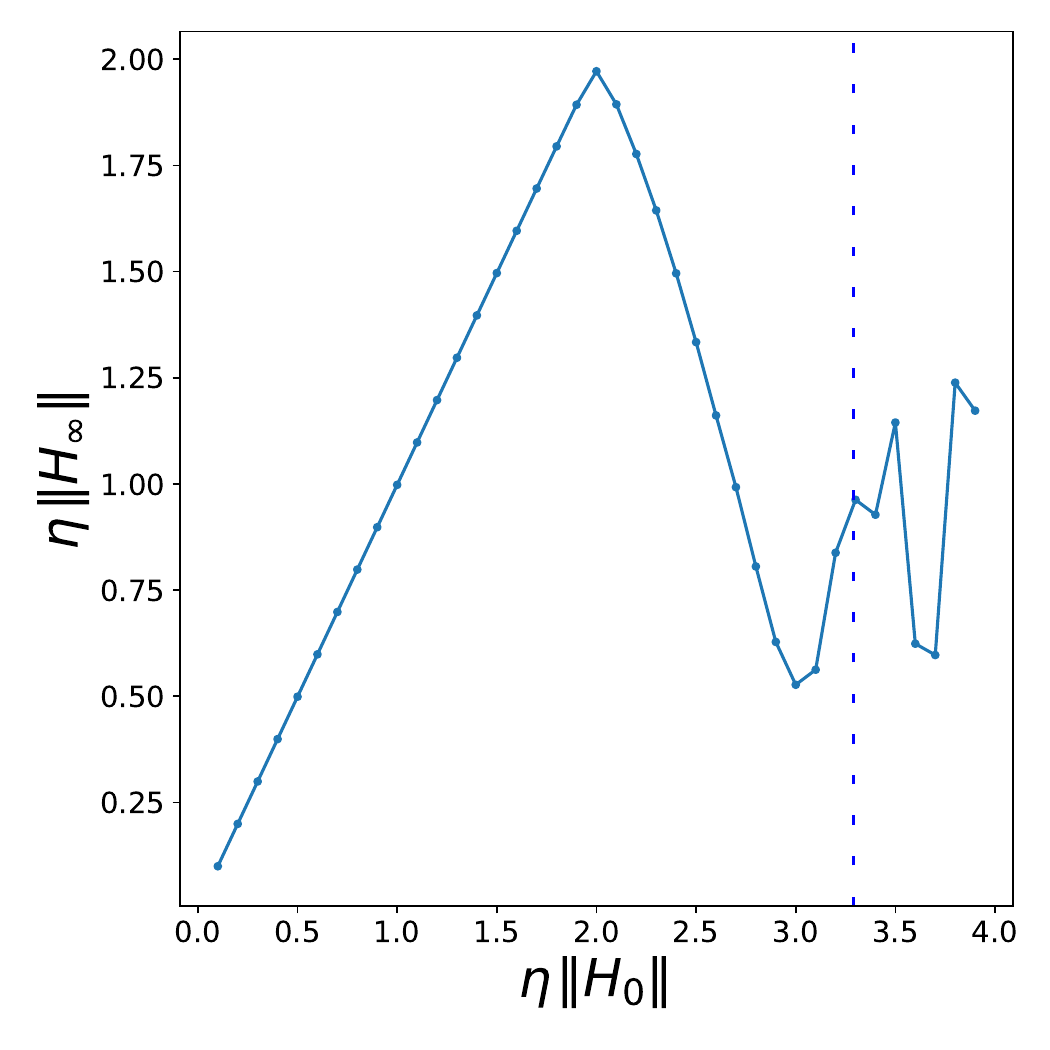}
\caption{}
\label{fig:linear_meta_toy_dataset_bias_final_lrNTK}
\end{subfigure}
\hspace{.25in}
\begin{subfigure}[b]{0.4\textwidth}
\centering
\includegraphics[scale=.27]{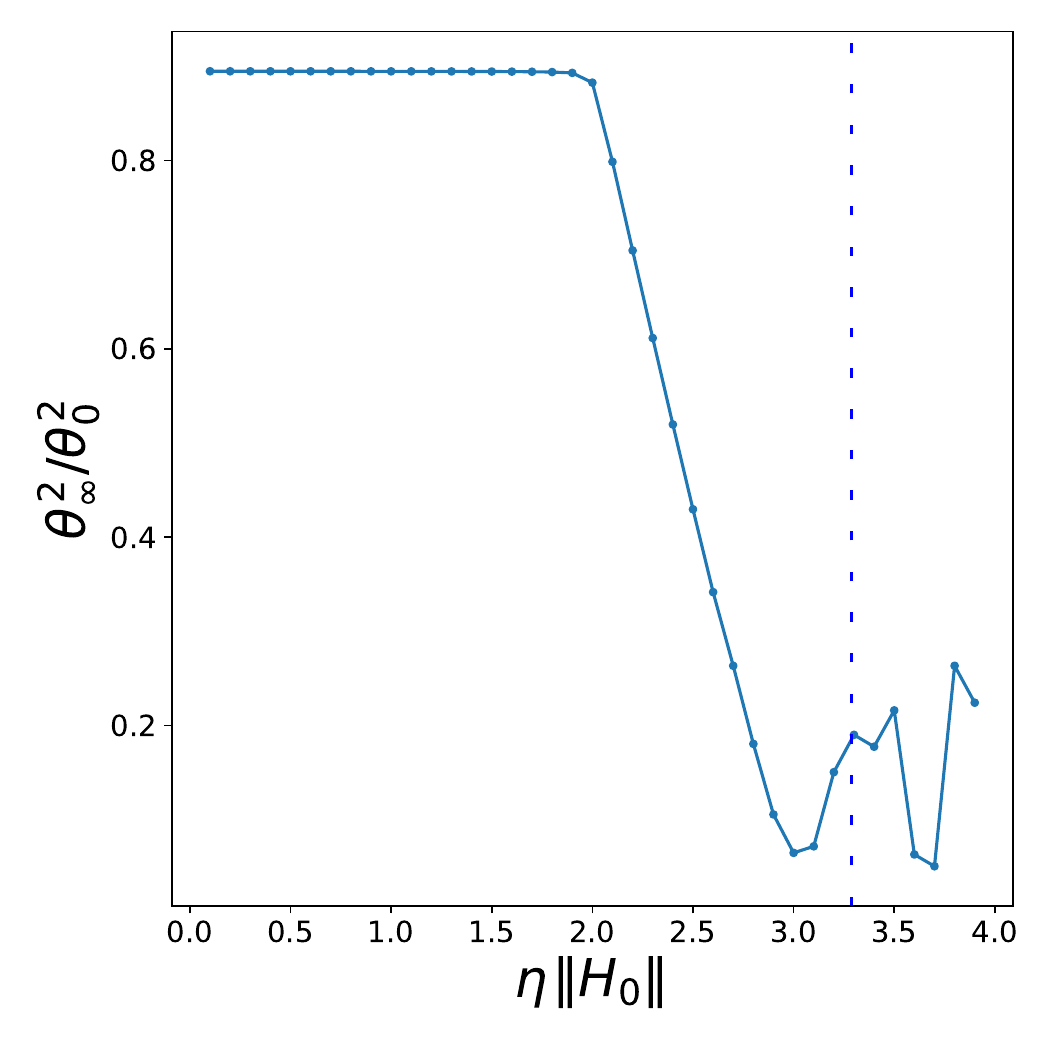}
\caption{}
\label{fig:linear_meta_toy_dataset_bias_final_weight_norm}
\end{subfigure}
\hfill
\caption{Results for the quadratic model with bias trained on the toy dataset $(x,y)=(1,0)$.}.
\label{fig:linear_meta_toy_dataset_qbias}
\end{figure*}

In Figure \ref{fig:linear_meta_toy_dataset_qbias} we summarize the results for a quadratic model with bias trained on the toy dataset $(x,y)=(1,0)$. Specifically, we take the meta-feature hidden dimension to be $n_{\bs{\psi}}=100$ and the feature hidden dimension to be $n_{\bs{\phi}}=10$.
For this model we take the activation function in \eqref{eq:app_meta_feat_func} to be the identity function, $g(x)=x$.
Finally, we take the eigenvalues of the meta-feature function to be $\lambda_{i}(\bs{\psi})=\pm 1$.

In this model we observe the expected behavior from our theoretical analysis.
The blue, vertical dashed line in figures \ref{fig:linear_meta_toy_dataset_bias_final_lrNTK} and \ref{fig:linear_meta_toy_dataset_bias_final_weight_norm}
correspond to the theoretical prediction given in \eqref{eq:inequality_eta_bias_V2_init}.
We observe that the model does indeed converge for super-critical learning rates to the left of this line.
The model also converges for a small interval to the right of this line, which means our sufficiency condition can likely be weakened.

\begin{figure*}[!ht]
\centering
\begin{subfigure}[b]{0.3\textwidth}
\centering
\includegraphics[scale=.27]{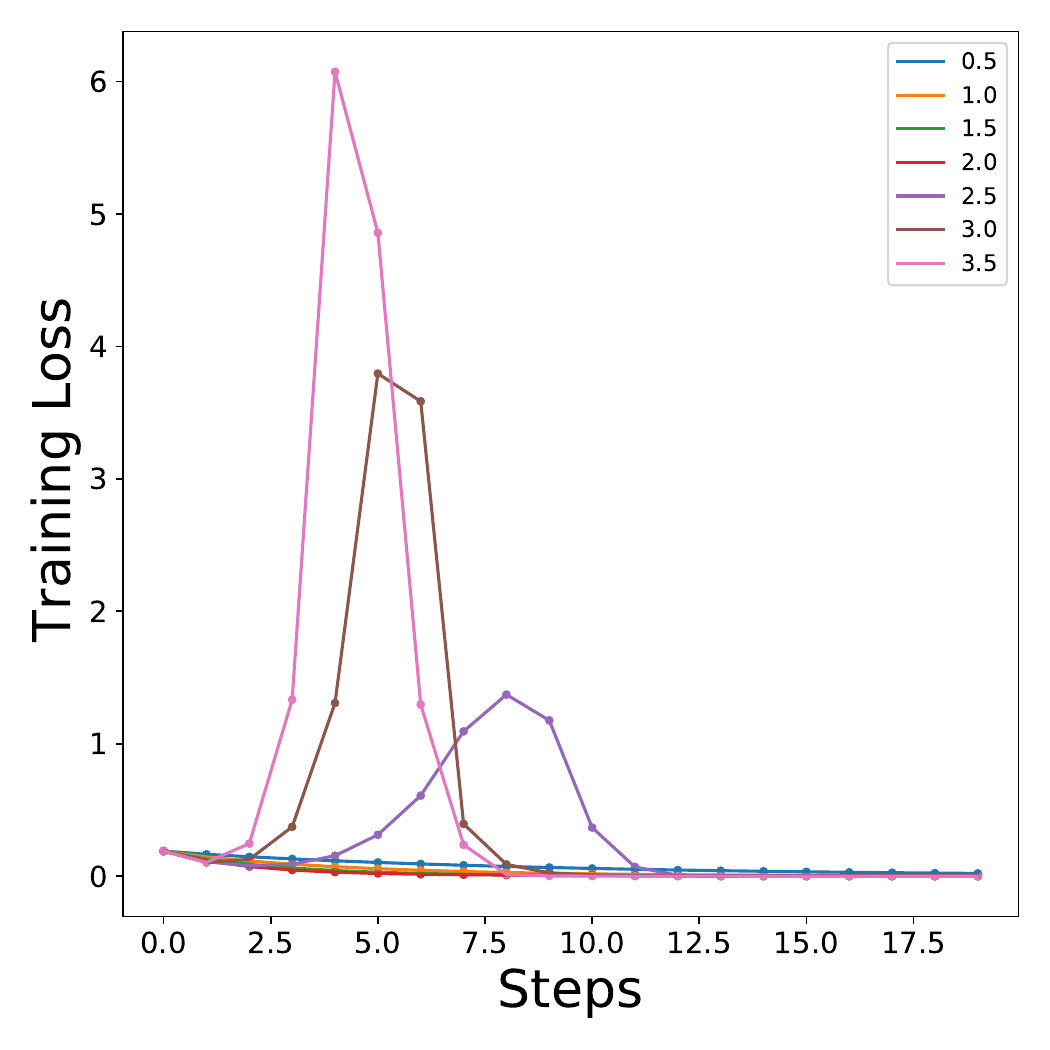}
\caption{}
\end{subfigure}
\hfill
\begin{subfigure}[b]{0.3\textwidth}
\centering
\includegraphics[scale=.27]{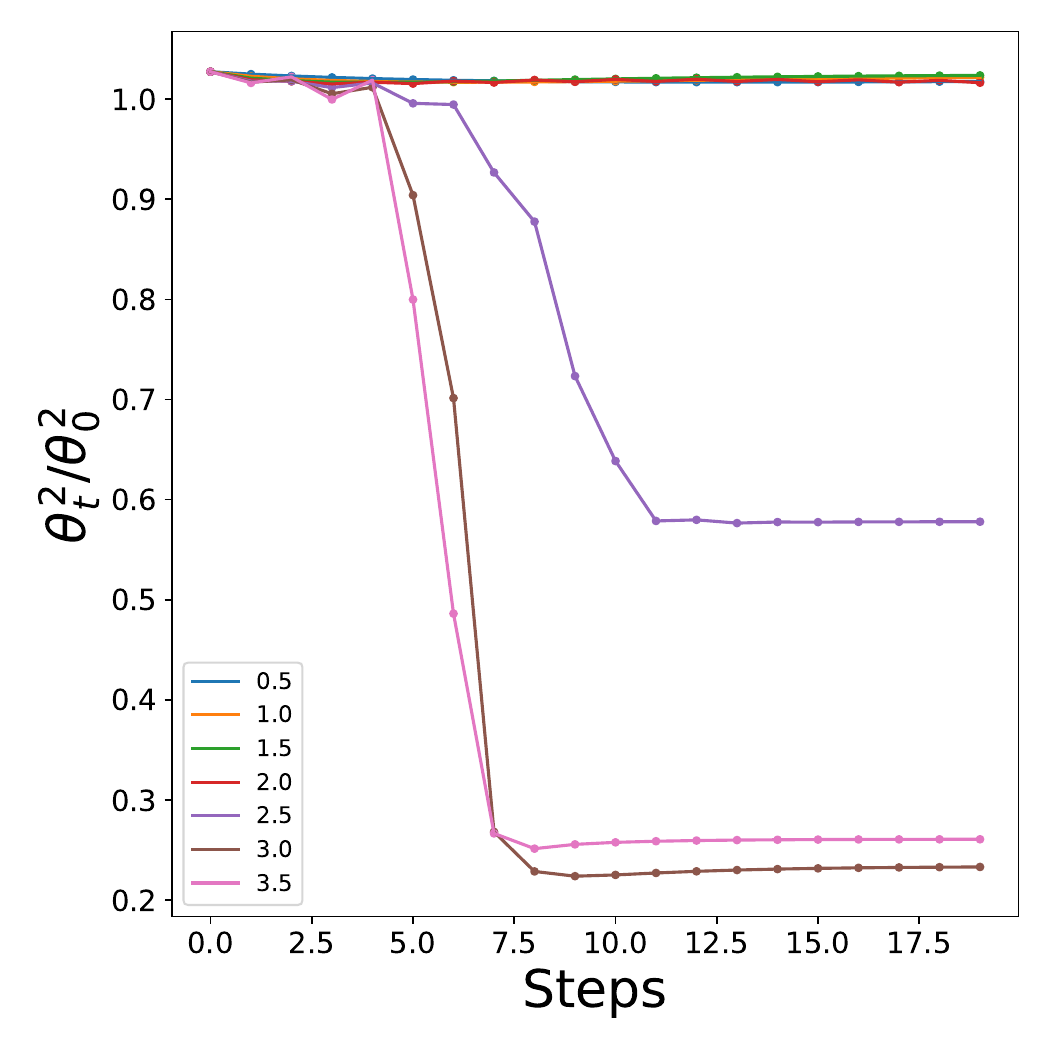}
\caption{}
\end{subfigure}
\hfill
\begin{subfigure}[b]{0.3\textwidth}
\centering
\includegraphics[scale=.27]{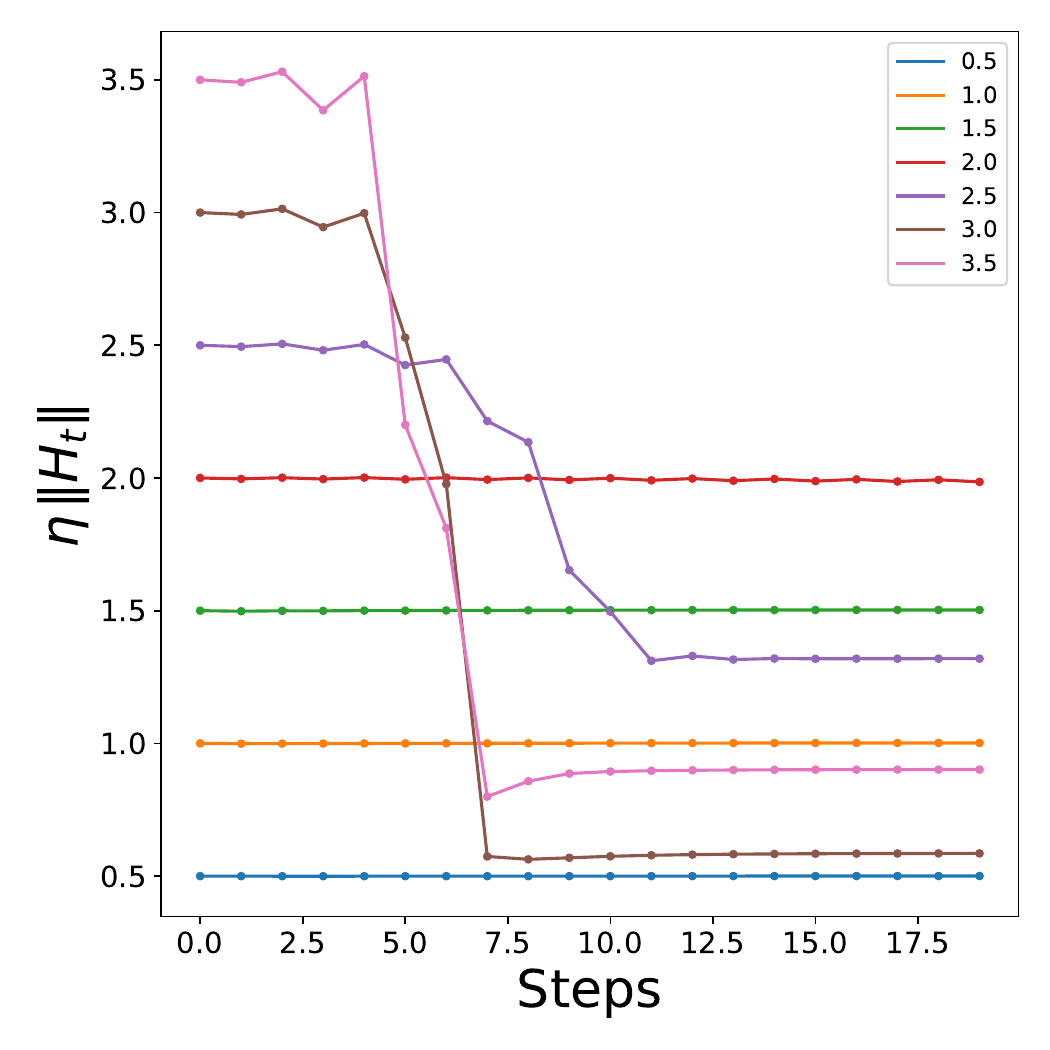}
\caption{}
\end{subfigure}
\hfill
\centering
\begin{subfigure}[t]{0.3\textwidth}
\centering
\includegraphics[scale=.27]{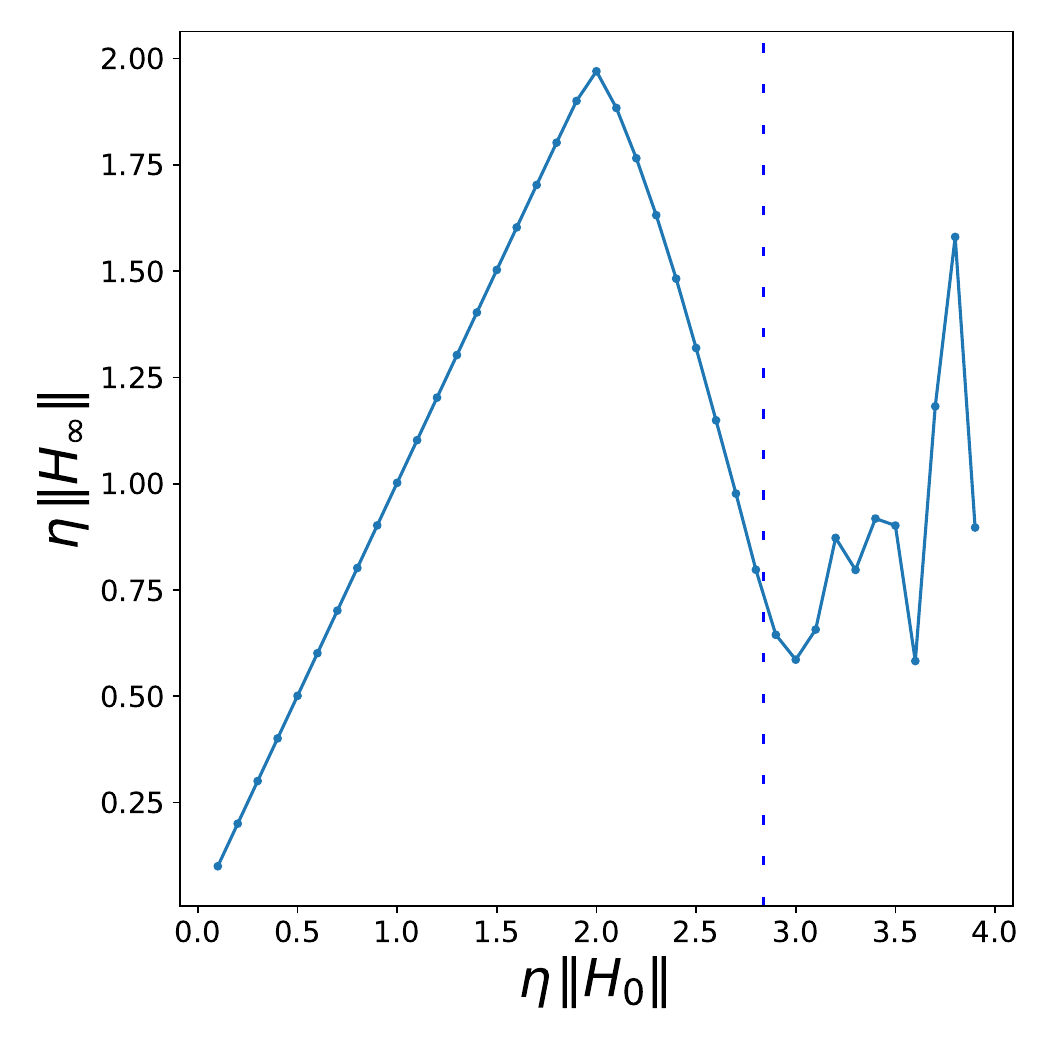}
\caption{}
\end{subfigure}
\hfill
\begin{subfigure}[t]{0.3\textwidth}
\centering
\includegraphics[scale=.27]{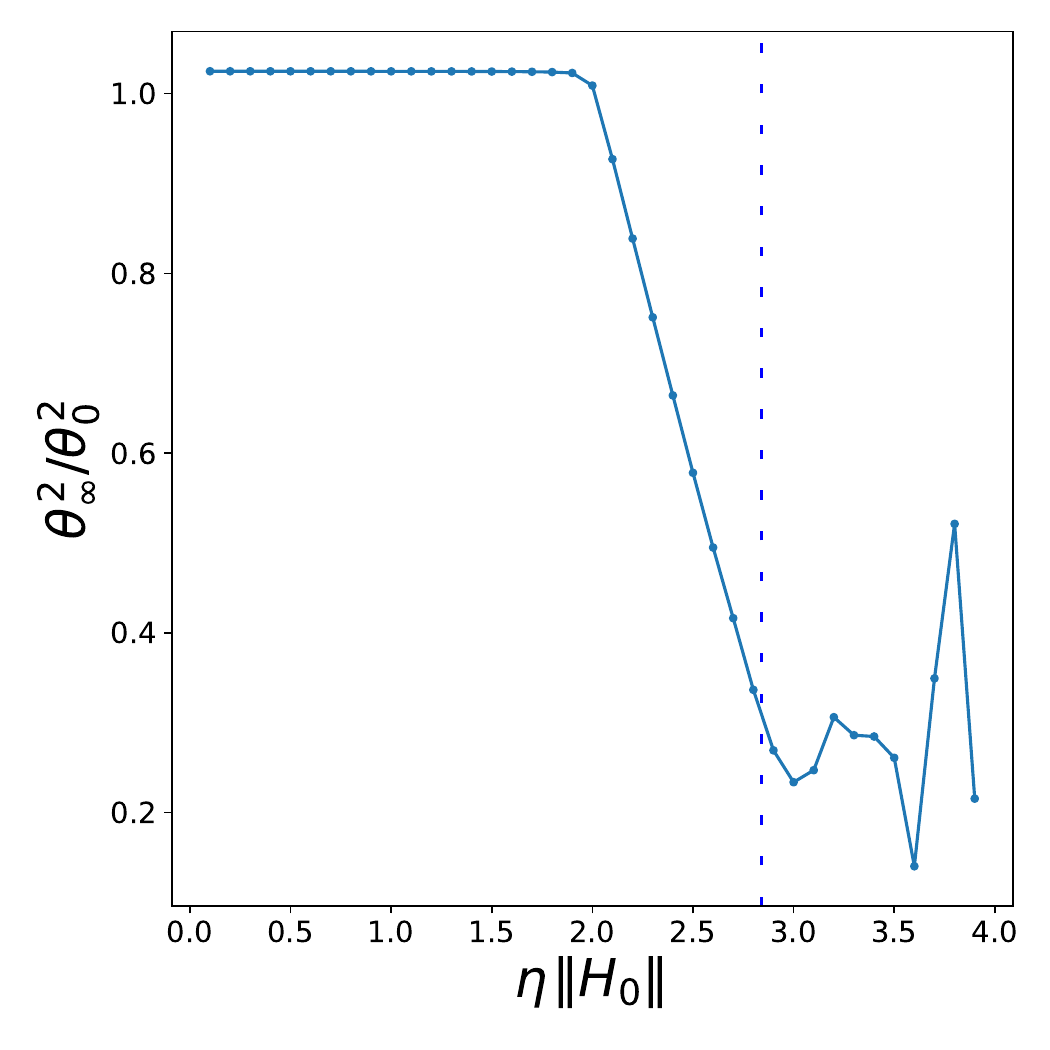}
\caption{}
\end{subfigure}
\hfill
\begin{subfigure}[t]{.3\textwidth}
\centering
\includegraphics[scale=.27]{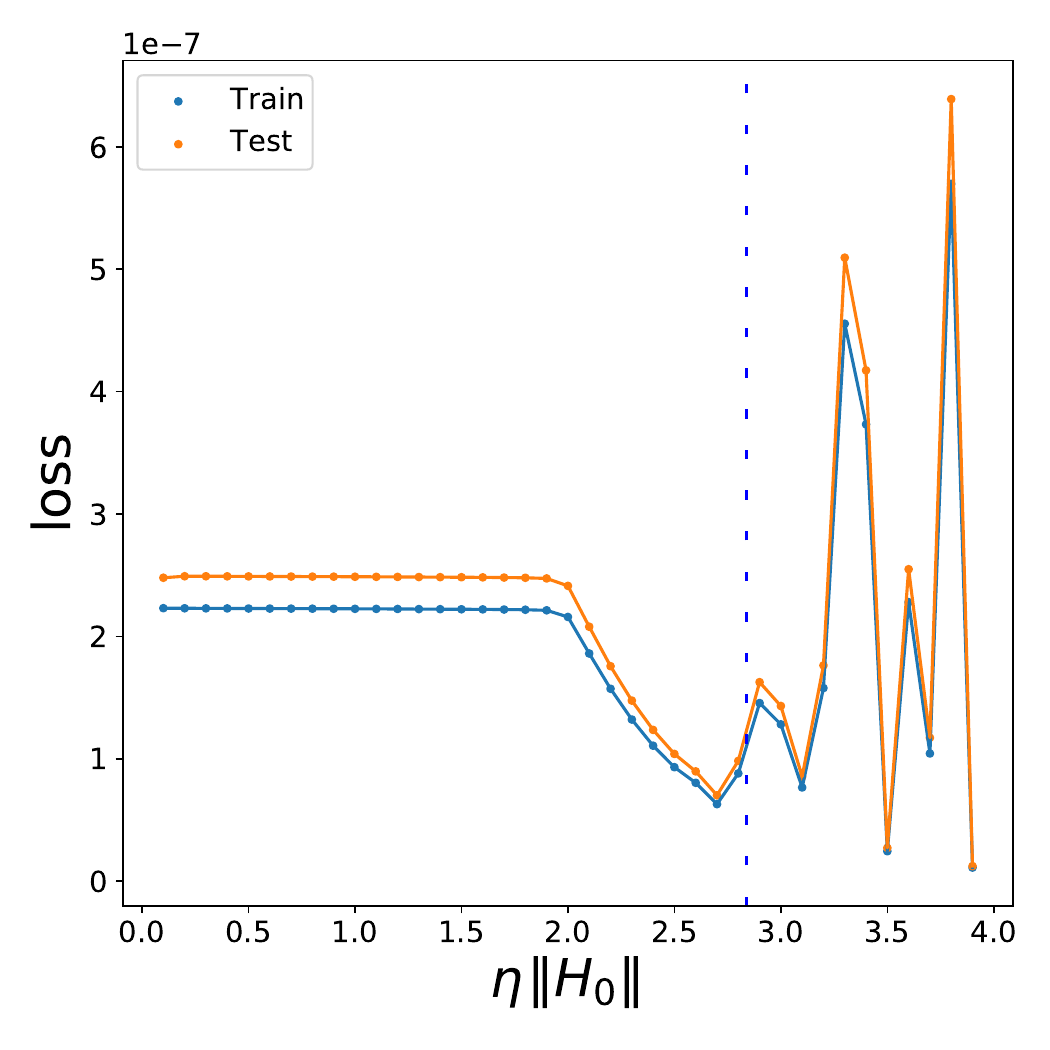}
\caption{}
\end{subfigure}

\caption{Results for the quadratic model with bias trained using a teacher-student set-up. }
\label{fig:teacher_quad_bias_tanh}
\end{figure*}

In Figure \ref{fig:teacher_quad_bias_tanh} we perform a similar experiment, but for the quadratic model with bias trained using a teacher-student set-up. We take the inputs to be $1d$ and draw them from $x_{\alpha}\sim\mathcal{U}([-1/2,1/2])$.
We take the training set and test set to have size 32 and 1000, respectively. 
The teacher meta-feature function has rank $n_{\bs{\psi}}= 200$ and the teacher feature functions have dimension $n_{\bs{\phi}}=20$.
We take the student meta-feature function to have rank $n_{\bs{\psi}}=150$ and the student feature functions to have dimension $n_{\bs{\phi}}=10$.
In \eqref{eq:app_meta_feat_func} we take $g(x)=\tanh(x)$.
Finally in \eqref{eq:app_meta_feat_func} we take $\lambda_{a}(\bs{W}^i)=\pm1$ for all $a$ and $i$.

In Figure \ref{fig:teacher_quad_bias_tanh} the vertical, dashed lines corresponds to the prediction from \eqref{eq:inequality_eta_bias_mult_init}. We see agreement between the theoretical and experimental results since the model converges for super-critical learning rates to the left of the vertical lines.
As with our other experiments, we observe that the bounds we derive are sufficient, but not necessary, to ensure convergence.

\subsection{Generic Two-Layer Homogenous MLPs}
\label{app:scale_inv_MLP_exp}
Here we will study the two-layer homogenous net trained on the toy dataset $(x,y)=(1,0)$ and on random datasets.
These examples will be simple enough to illustrate the learning dynamics of these models in the catapult phase.
In the next section we will study the evolution of the weight norm in ReLU MLPs when trained on more realistic datasets.

\begin{figure*}[!ht]
\centering
\begin{subfigure}[t]{0.3\textwidth}
\centering
\includegraphics[scale=.27]{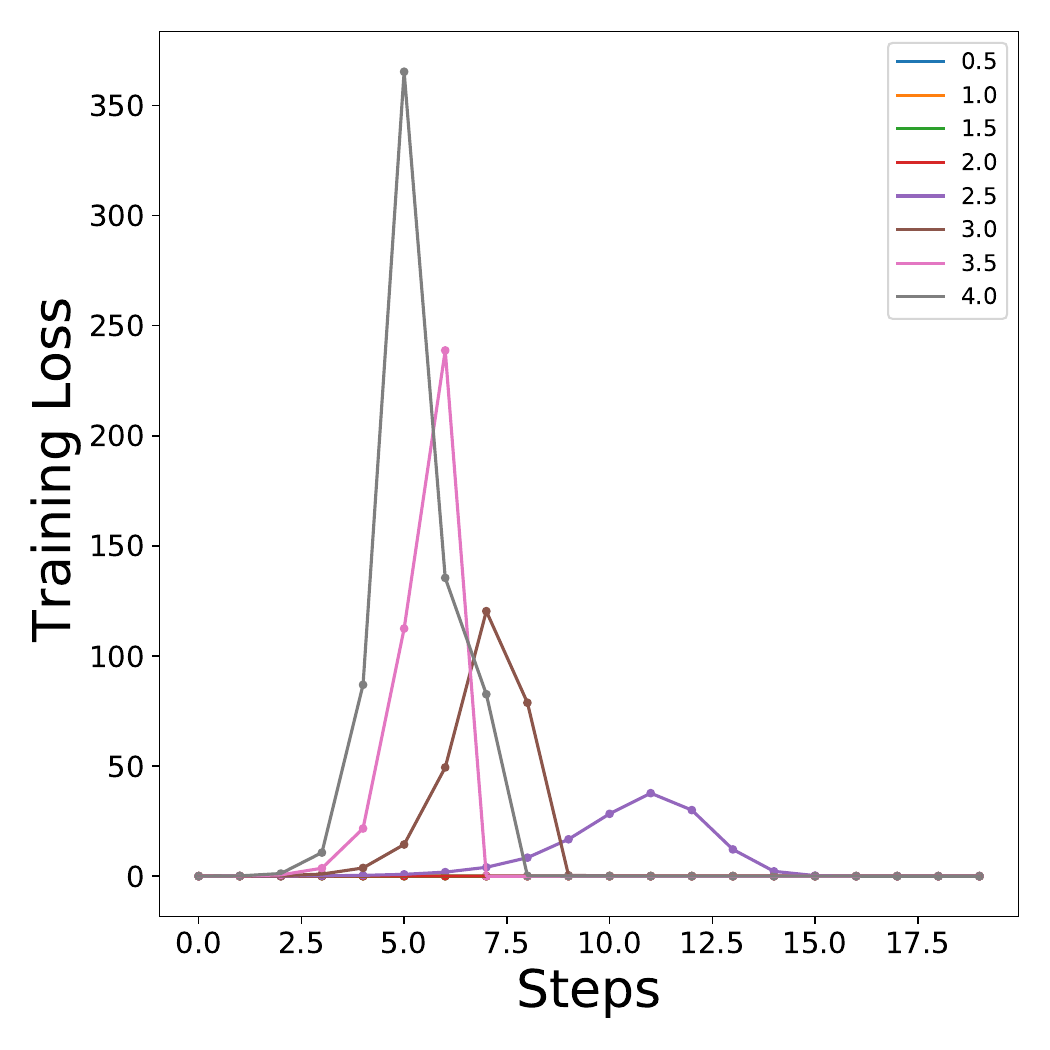}
\caption{}
\end{subfigure}
\hfill
\begin{subfigure}[t]{0.3\textwidth}
\centering
\includegraphics[scale=.27]{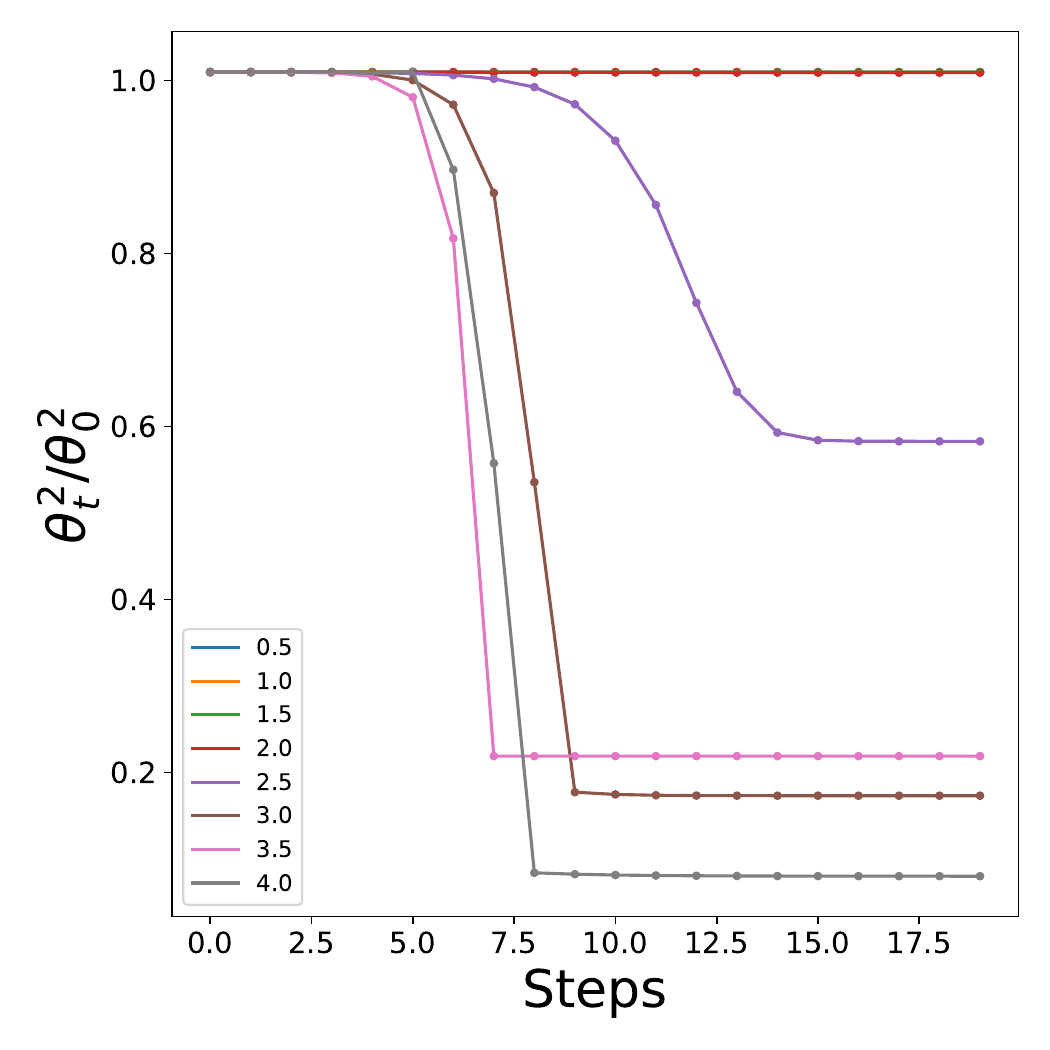}
\caption{}
\end{subfigure}
\hfill
\begin{subfigure}[t]{0.3\textwidth}
\centering
\includegraphics[scale=.27]{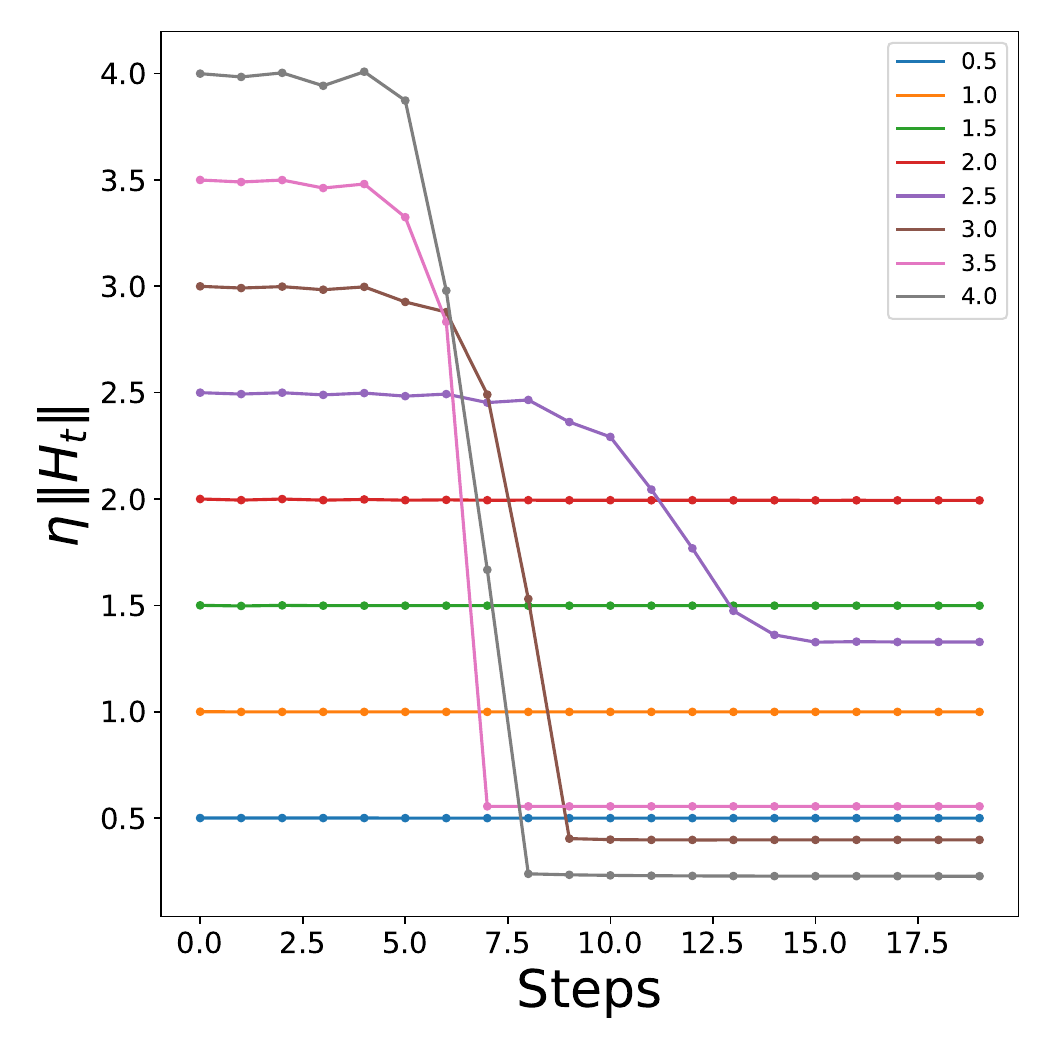}
\caption{}
\end{subfigure}
\centering
\begin{subfigure}[b]{0.4\textwidth}
\centering
\includegraphics[scale=.27]{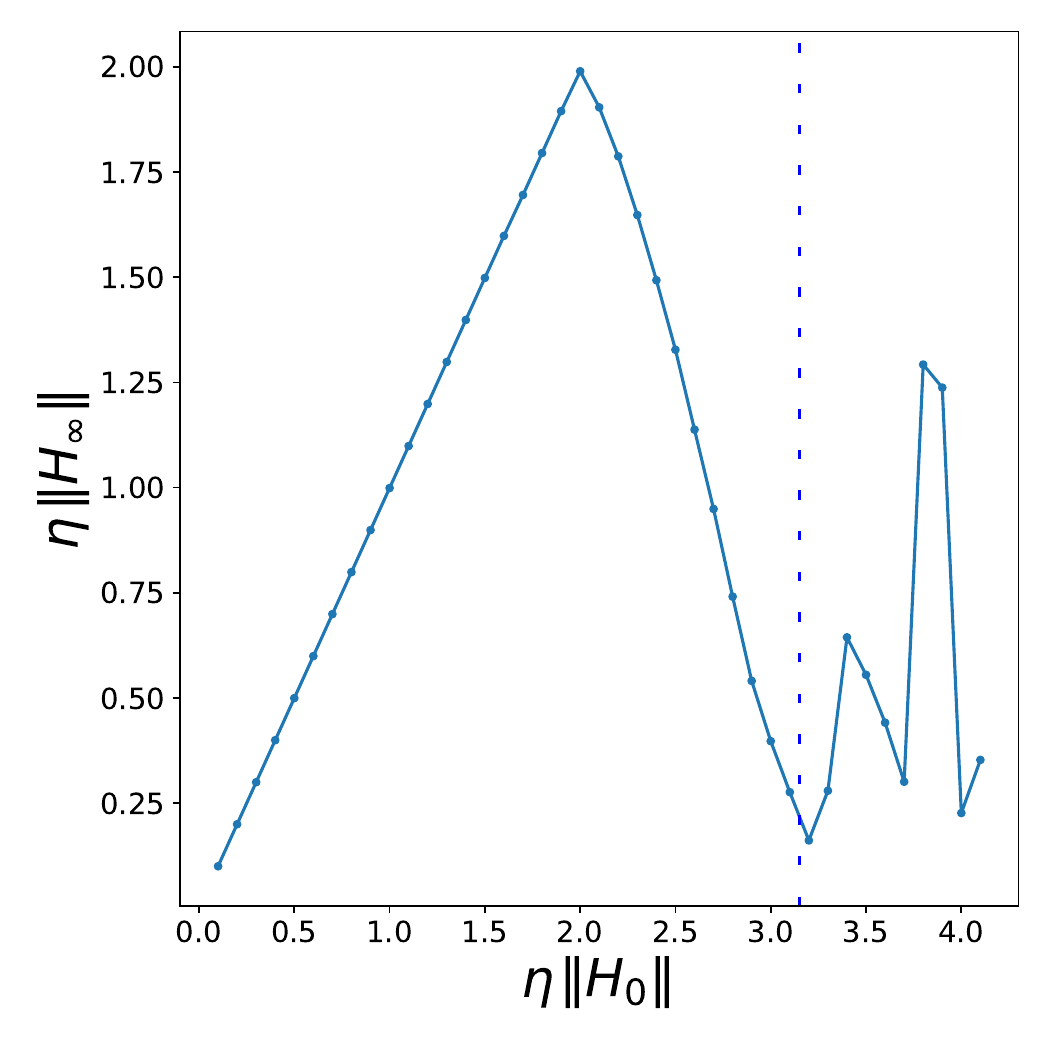}
\caption{}
\label{fig:linear_scale_invariant_toy_dataset_34_final_lrNTK}
\end{subfigure}
\hspace{.25in}
\begin{subfigure}[b]{0.4\textwidth}
\centering
\includegraphics[scale=.27]{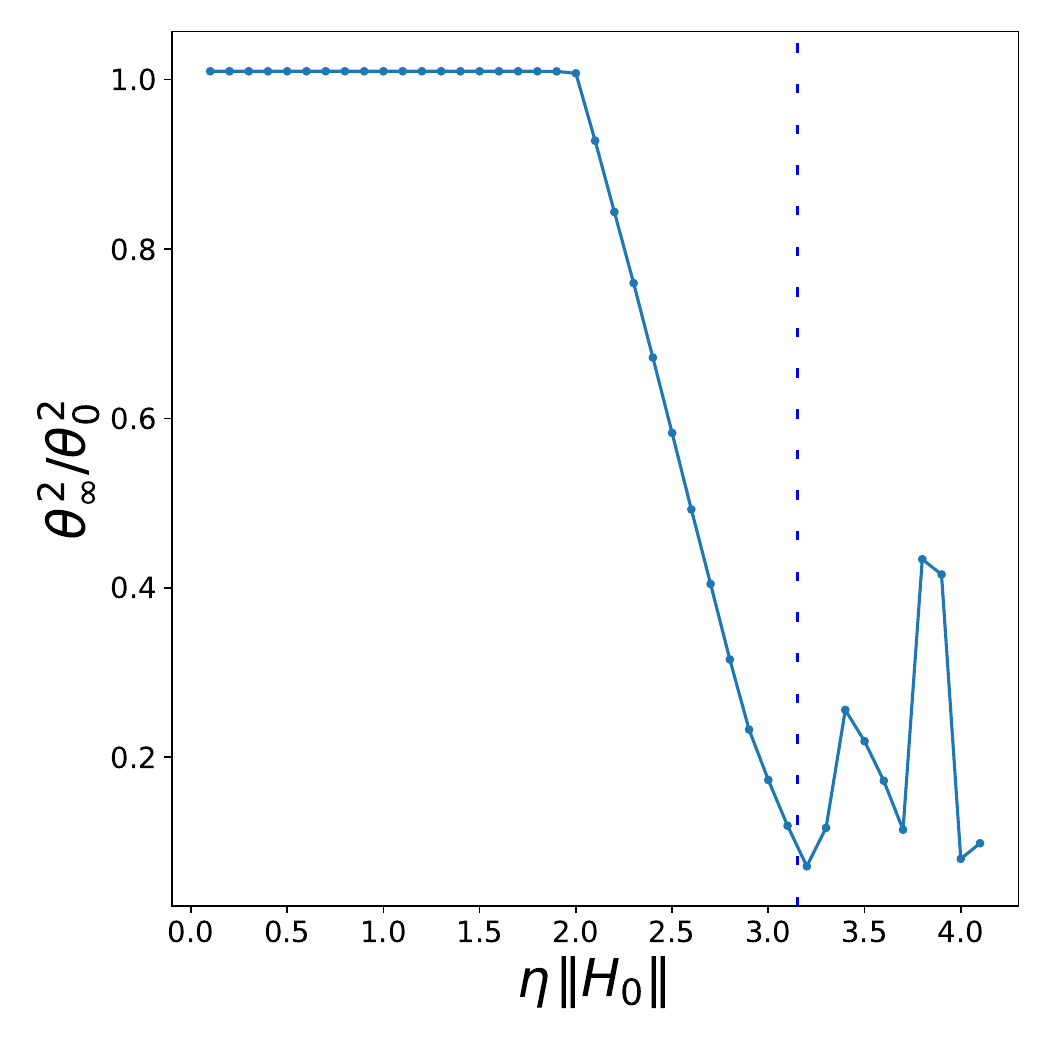}
\caption{}
\label{fig:linear_scale_invariant_toy_dataset_34_final_weight}
\end{subfigure}
\hfill
\caption{Results for the homogenous net with $a_+=1$ and $a_-=3/4$ trained on the toy dataset.}
\label{fig:linear_scale_invariant_toy_dataset_34}
\end{figure*}

In Figure \ref{fig:linear_scale_invariant_toy_dataset_34} we show the results for the two-layer MLP trained on the toy dataset $(x,y)=(1,0)$. For the activation function we take $a_+=1$ and $a_-=3/4$. We also take the width of the hidden layer to be 1024 and train the model until it converges.
In comparison to Figure \ref{fig:linear_scale_invariant_toy_dataset_half}, we can note that our theoretical prediction \eqref{eq:upper_bound_generic_scale}, which corresponds to the dashed blue lines in Figures \ref{fig:linear_scale_invariant_toy_dataset_34_final_lrNTK} and \ref{fig:linear_scale_invariant_toy_dataset_34_final_weight}, has moved to the right.
This result is not surprising since we derived the bound on the learning rate \eqref{eq:upper_bound_generic_scale} using the upper bound on the NTK given in \eqref{eq:bound_NTK_single_data_scale_invariant}, which we reproduce below:
\begin{align}
H_{t}\leq \frac{a_+^2}{n}\bs{\theta}_t^2. \label{eq:bound_NTK_single_data_scale_invariant_V2}
\end{align}
In general, the closer this upper bound is to being saturated, the weaker the upper bound \eqref{eq:upper_bound_generic_scale} on $\eta$ will be.
The upper bound \eqref{eq:bound_NTK_single_data_scale_invariant_V2} is saturated when $a_+=a_-$, in which case the bound on $\eta$ \eqref{eq:upper_bound_generic_scale} becomes optimal and agrees with the bound derived in \cite{lewkowycz2020large} for the two-layer linear net.
On the other hand, as we decrease $a_-$ the NTK decreases, so the bound \eqref{eq:bound_NTK_single_data_scale_invariant_V2} becomes less tight, which in turn causes the upper bound on $\eta$ to decrease.

\begin{figure*}[!ht]
\centering
\begin{subfigure}[t]{0.3\textwidth}
\centering
\includegraphics[scale=.27]{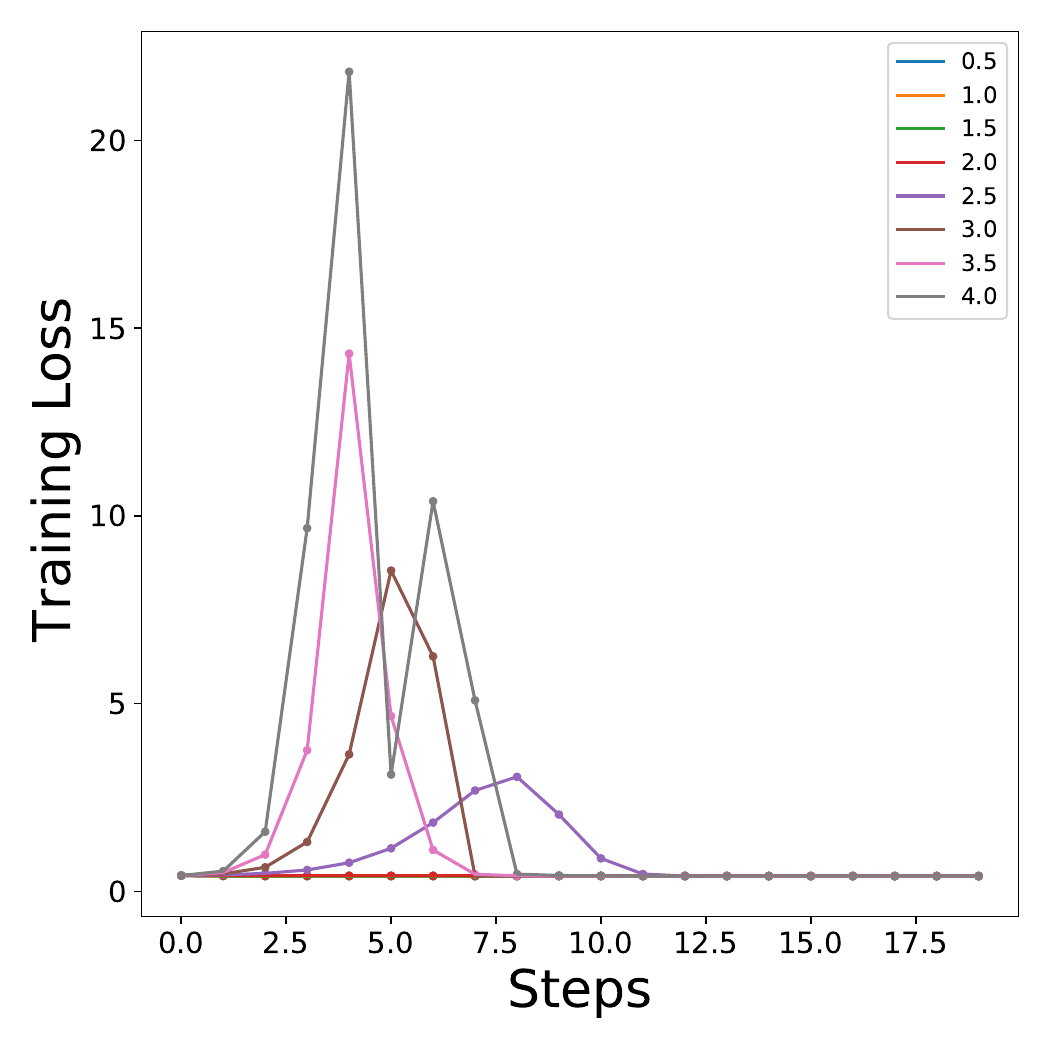}
\caption{}
\end{subfigure}
\hfill
\begin{subfigure}[t]{0.3\textwidth}
\centering
\includegraphics[scale=.27]{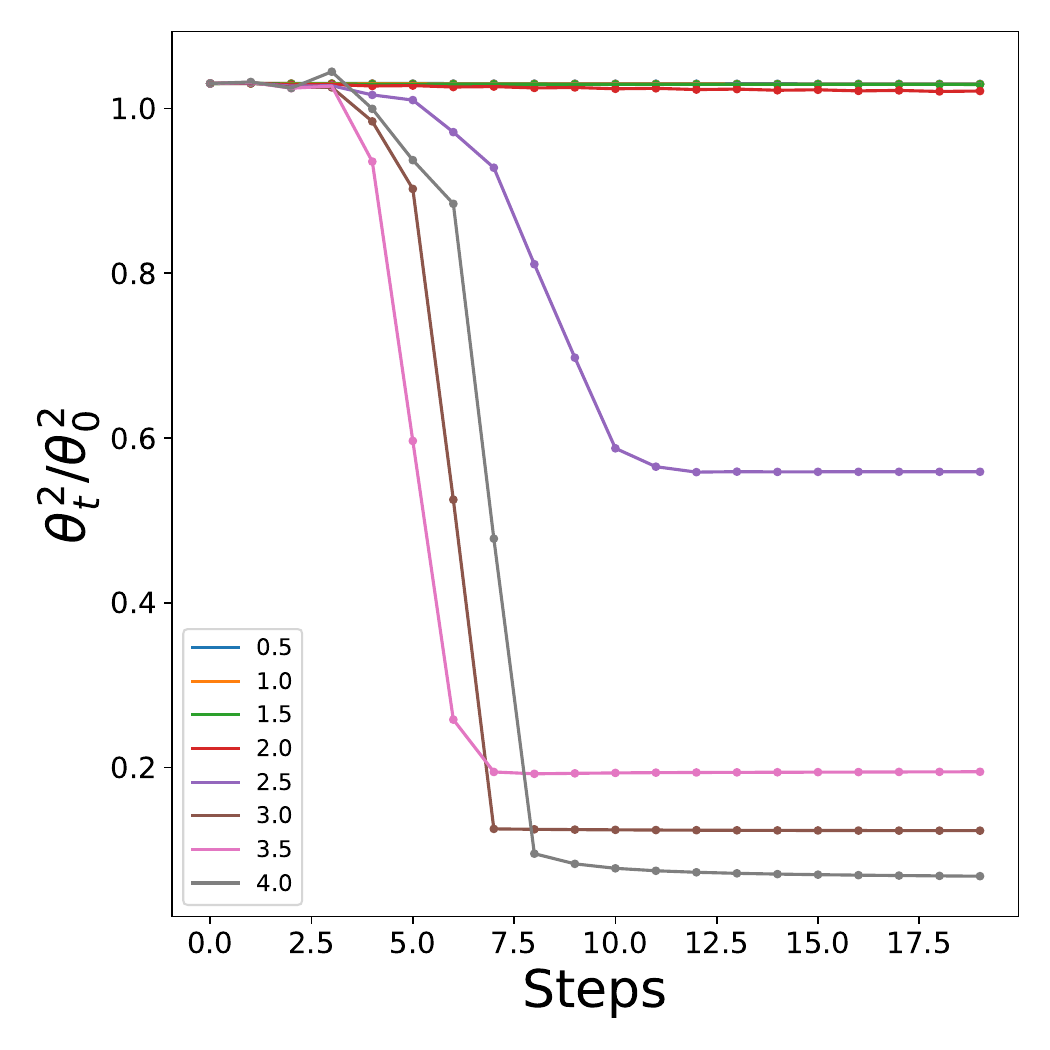}
\caption{}
\end{subfigure}
\hfill
\begin{subfigure}[t]{0.3\textwidth}
\centering
\includegraphics[scale=.27]{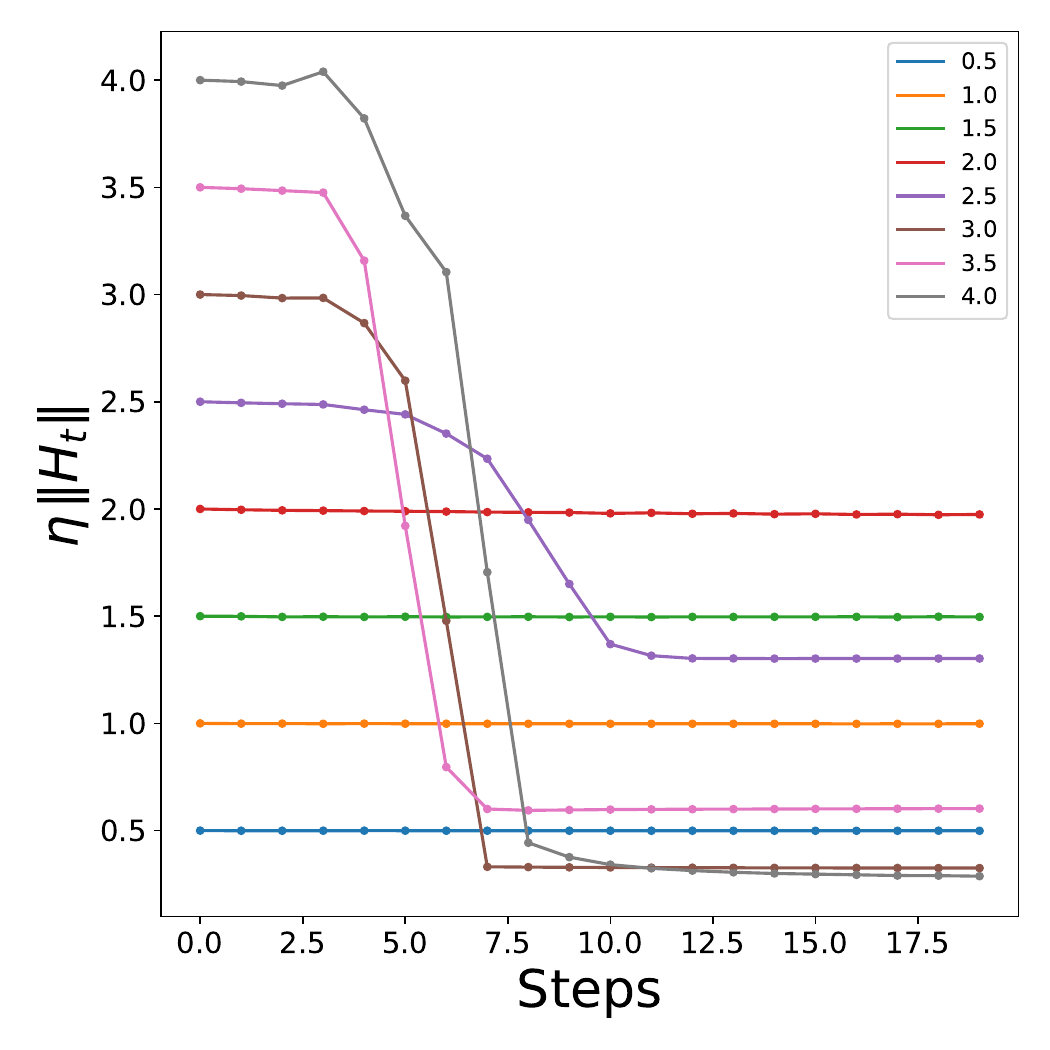}
\caption{}
\end{subfigure}
\centering
\begin{subfigure}[b]{0.4\textwidth}
\centering
\includegraphics[scale=.27]{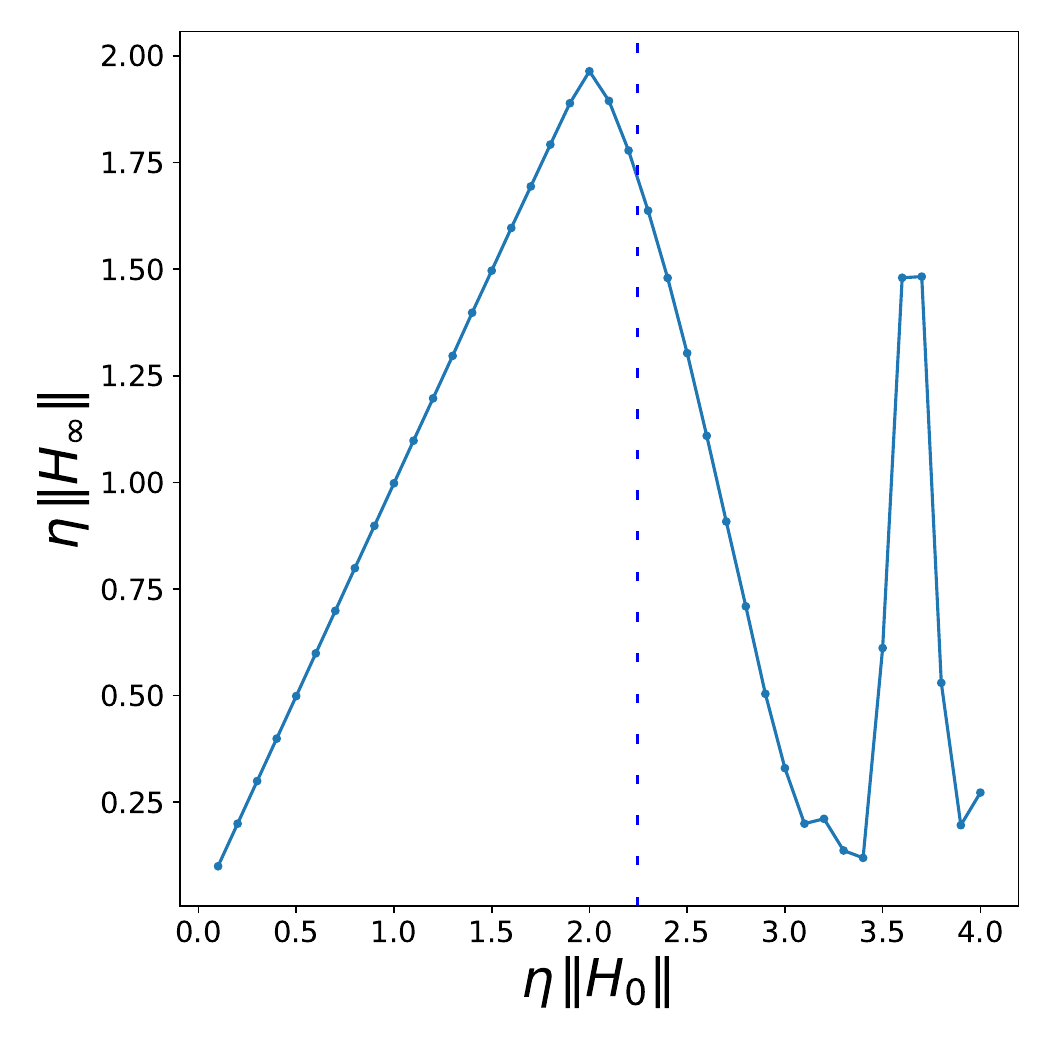}
\caption{}
\end{subfigure}
\hspace{.25in}
\begin{subfigure}[b]{0.4\textwidth}
\centering
\includegraphics[scale=.27]{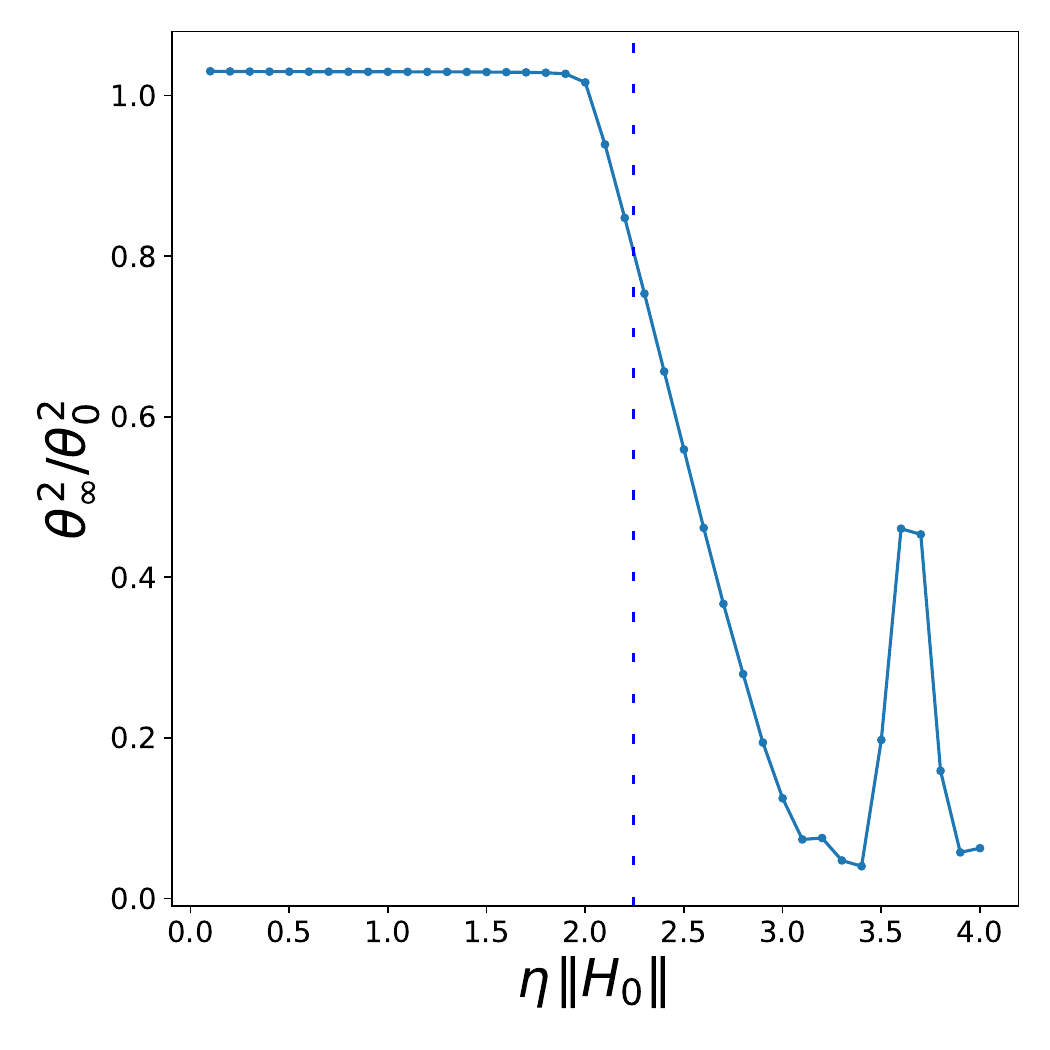}
\caption{}
\end{subfigure}
\hfill
\caption{Results for the homogenous net with $a_+=1$ and $a_-=1/2$ trained on random data.}
\label{fig:random_data_mutl_data_leaky}
\end{figure*}

Next, we test the prediction of Appendix \ref{app:mult_data_two_MLPs}, see \eqref{eq:final_bound_generic_scale_invariant}, by studying homogenous MLPs on random data.
Here we will set $a_+=1$ and $a_-=1/2$ and set the width to be $n=1024$.
We will draw both the datapoints $x_\alpha$ and the labels $y_\alpha$ from $\mathcal{U}([-1/2,1/2])$.
We take the training set to have size 32 and train the model for 1000 epochs.
The results are summarized in Figure \ref{fig:random_data_mutl_data_leaky}.
Here we observe that our upper bound is close to 2.25. 
Although our result is still non-trivial (it indicates that the model can converge for super-critical learning rates) the fact the model empirically converges up to $\eta\lambda_{\text{max}}(H_{\alpha\beta,0})=4$ indicates there is room for improvement.

We can also note that most of the results in Figure \ref{fig:random_data_mutl_data_leaky} are qualitatively similar to the results in Figure \ref{fig:linear_scale_invariant_toy_dataset_half}, indicating that adding multiple datapoints does not significantly alter the training dynamics, at least for random data.
One difference however is that in Figure \ref{fig:linear_scale_invariant_toy_dataset_half_final_weight} the model converged for learning rates up to $\eta \lambda_{\text{max}}(H_{\alpha\beta,0})=4.5$, while here the model stops converging around $\eta \lambda_{\text{max}}(H_{\alpha\beta,0})=4$.
This indicates that where the catapult phase ends and the divergent phase begins must be data-dependent.

\begin{figure*}[!ht]
\centering
\begin{subfigure}[t]{0.3\textwidth}
\centering
\includegraphics[scale=.27]{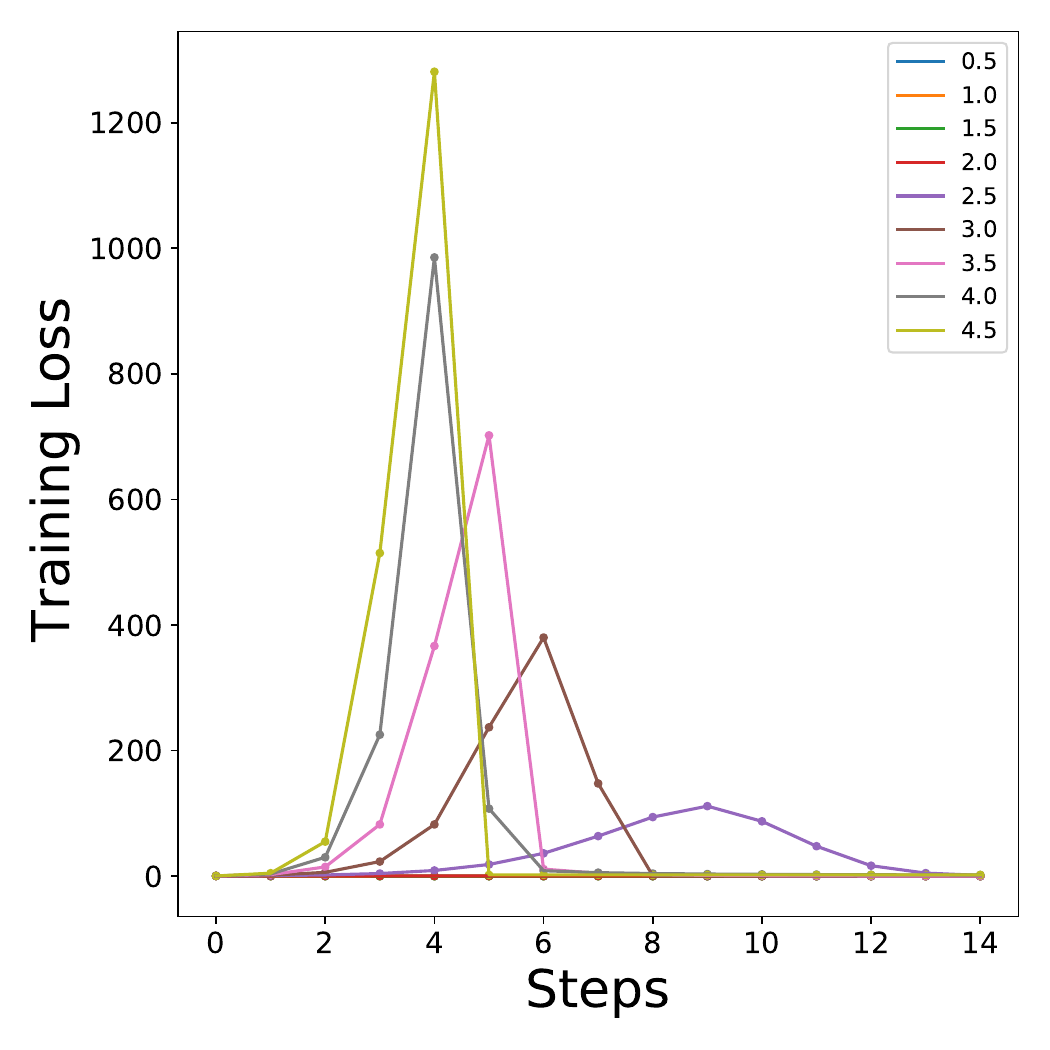}
\caption{}
\end{subfigure}
\hfill
\begin{subfigure}[t]{0.3\textwidth}
\centering
\includegraphics[scale=.27]{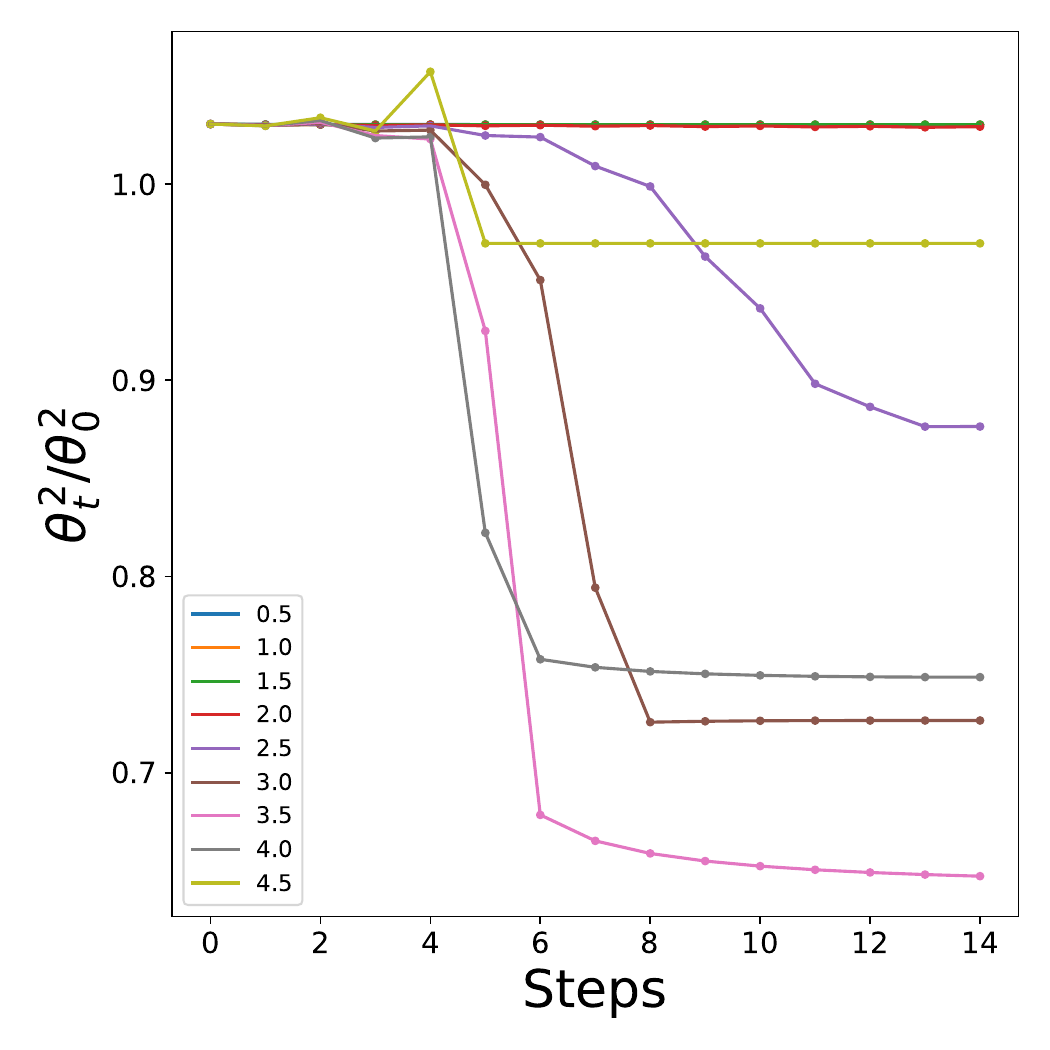}
\caption{}
\label{fig:relu_single_datapoint_weight}
\end{subfigure}
\hfill
\begin{subfigure}[t]{0.3\textwidth}
\centering
\includegraphics[scale=.27]{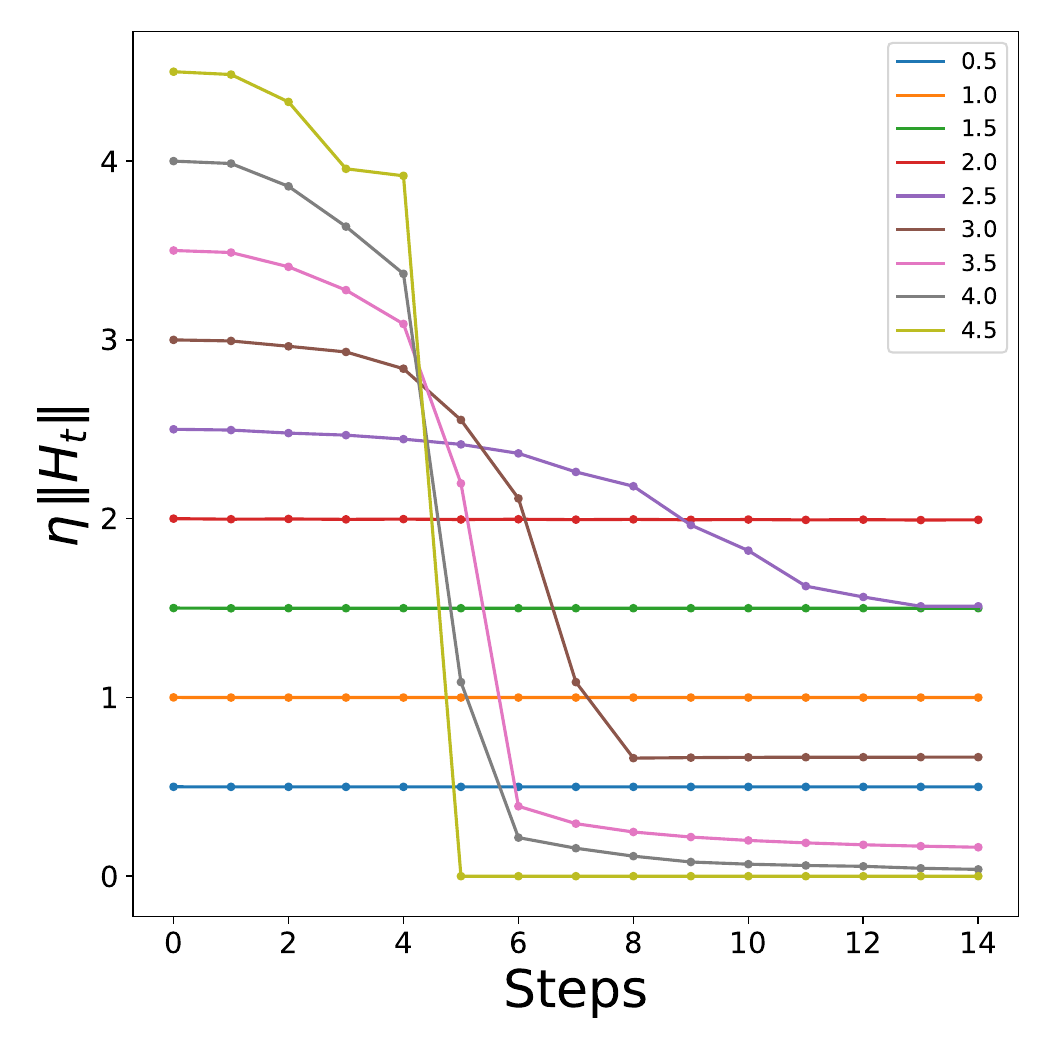}
\caption{}
\end{subfigure}
\centering
\begin{subfigure}[b]{0.4\textwidth}
\centering
\includegraphics[scale=.27]{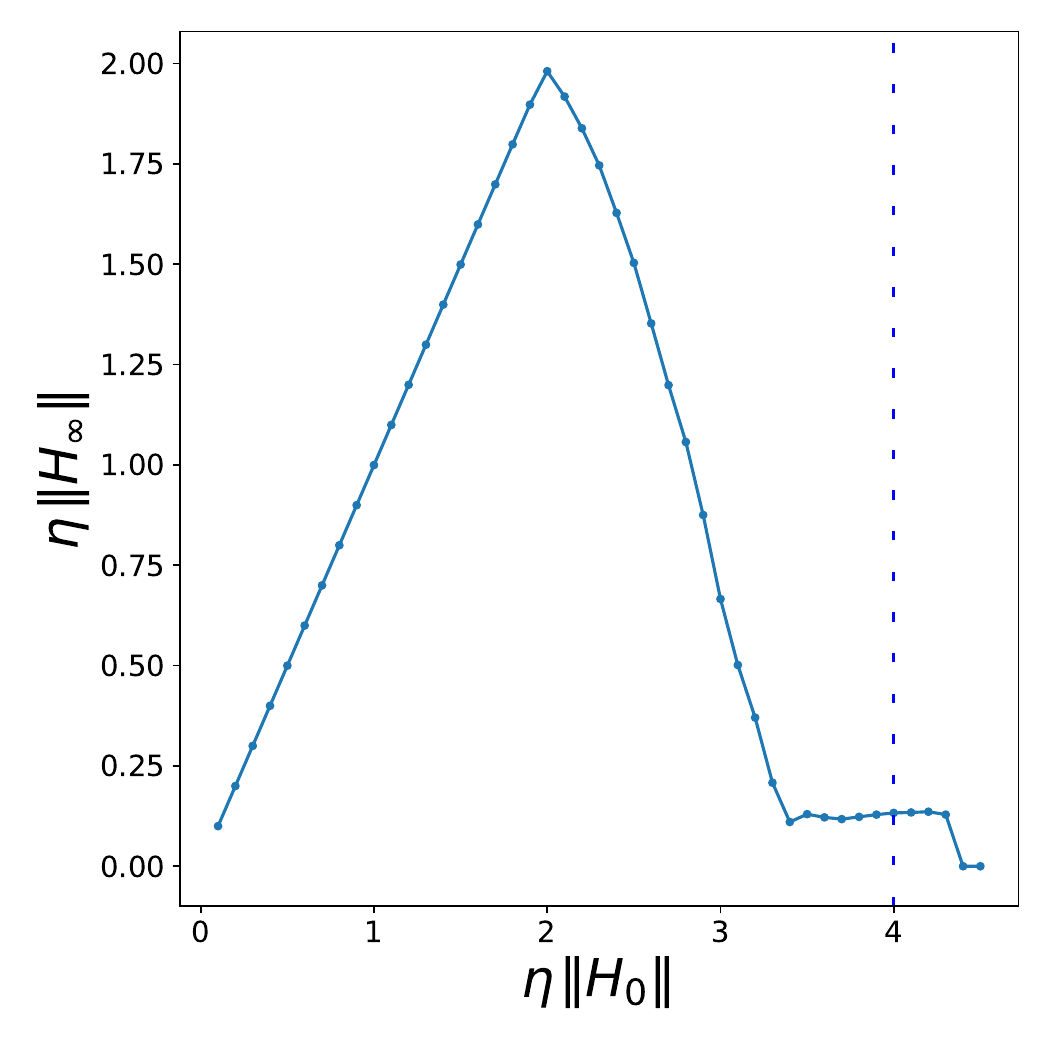}
\caption{}
\end{subfigure}
\hspace{.25in}
\begin{subfigure}[b]{0.4\textwidth}
\centering
\includegraphics[scale=.27]{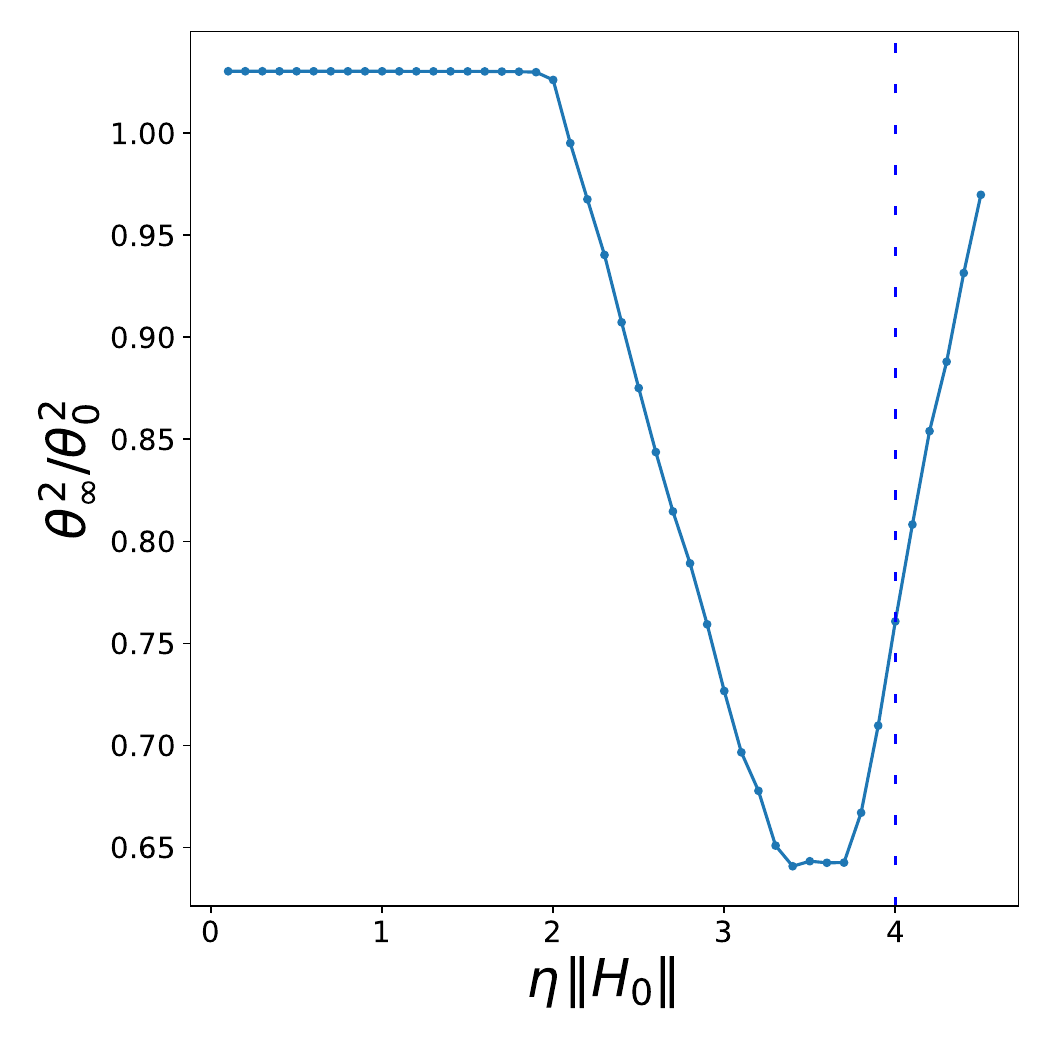}
\caption{}
\end{subfigure}
\hfill
\caption{Results for a two-layer ReLU net trained on the dataset $(x,y)=(4,2)$.}
\label{fig:relu_single_datapoint}
\end{figure*}

Finally, we will study the convergence of a two-layer ReLU net trained on a single datapoint in order to test the predictions of Appendix \ref{ssec:relu_net}.
We take the width of the model to be $n=1024$ and the dataset to be $(x,y)=(4,2)$. The results are summarized in Figure \ref{fig:relu_single_datapoint}.
We observe that our theoretical predictions, the dashed vertical lines, again agree with the data since the model converges to the left of the line $\eta\lambda_{\text{max}}(H_{\alpha\beta,0})= 4$.
In particular, $\bs{\theta}_t^2$ receives large negative updates when $\eta\lambda_{\text{max}}(H_{\alpha\beta,0})\leq 4$, but can receive positive corrections when $\eta\lambda_{\text{max}}(H_{\alpha\beta,0})\approx 4.5$, see Figure \ref{fig:relu_single_datapoint_weight}. We can also note that around this value, $\eta\lambda_{\text{max}}(H_{\alpha\beta,0})\approx 4.5$, the model becomes trivial (the NTK vanishes) and the neurons die.
A similar phenomenon was also observed in \cite{lewkowycz2020large} for two-layer ReLU nets trained on one data-point, although they found the model became trivial around $\eta\lambda_{\text{max}}(H_{\alpha\beta,0})\approx 12$.

\subsection{ReLU MLPs for Images}
\subsubsection{Architectures and Datasets}
In this appendix we present results for ReLU MLPs without bias. The networks have the form:
\begin{align}
z_{\alpha}=\frac{1}{n^{\frac{k+1}{2}}}\bs{v}^T\sigma_{\text{ReLU}}(\bs{W}_1\sigma_{\text{ReLU}}( \ldots \bs{W}_k\sigma_{\text{ReLU}}(\bs{U} \bs{x}_{\alpha}))),
\end{align}
where $\bs{U}\in \mathbb{R}^{n\times d}$, $\bs{W}_{i}\in\mathbb{R}^{n\times n}$, $\bs{v}\in\mathbb{R}^{n}$, and $\bs{x}_{\alpha}\in\mathbb{R}^{d}$. In practice we will set $k=0$ or $k=1$, i.e. we consider two- and three-layer networks.
The components of all the weights are drawn from $\mathcal{N}(0,1)$. 
We set the width to be $n=1024$ for both $k=0$ and $k=1$.

We will train these MLPs on three datasets, which are two-class versions of MNIST, FMNIST, and CIFAR-10. 
For MNIST the model is trained to distinguish images of ``0" and ``1", for FMNIST it is trained to distinguish T-shirts and trousers, and on CIFAR-10 it is trained to distinguish airplanes and automobiles.
For all cases we give the first image a label of $-1$, the second image a label of $1$, and train the model to minimize the MSE.
We take the training set to consist of 128 images from the original training set and take the test set to consist of all of the corresponding images in the original test set.
Finally, we train the models until the training loss changes by less than $10^{-8}$.

\subsubsection{Two Layers}
\begin{figure*}[!ht]
\centering
\begin{subfigure}[b]{0.3\textwidth}
\centering
\includegraphics[scale=.27]{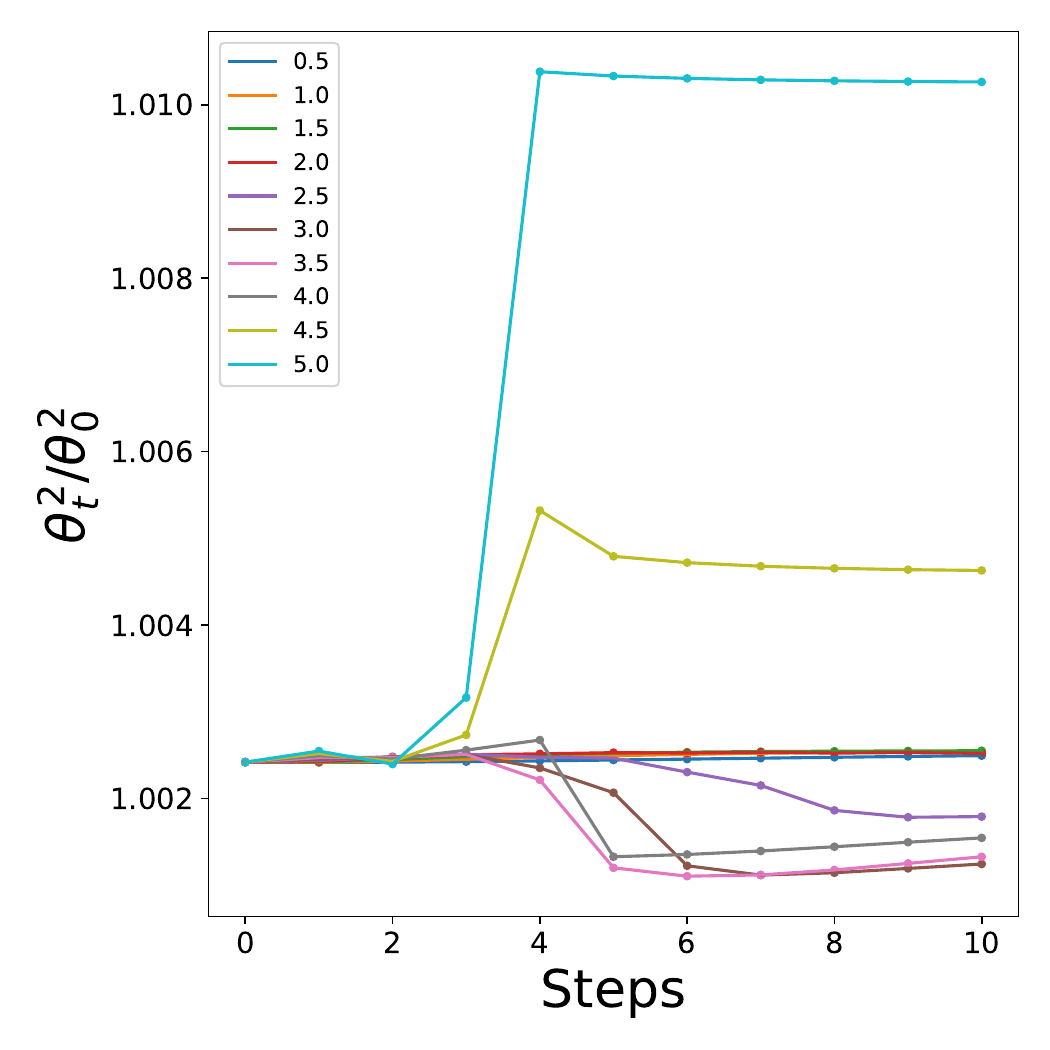}
\caption{}
\end{subfigure}
\begin{subfigure}[b]{0.3\textwidth}
\centering
\includegraphics[scale=.27]{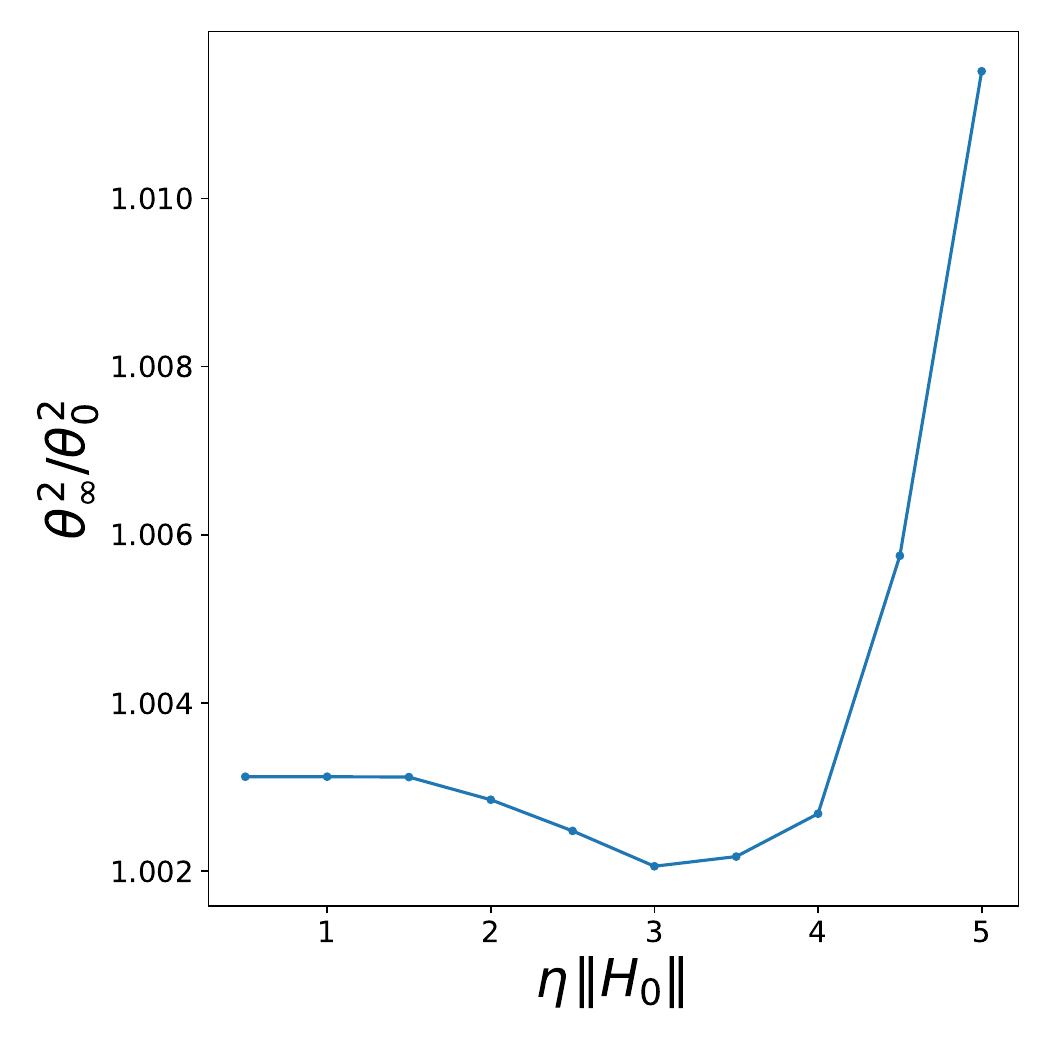}
\caption{}
\label{fig:FMNIST_011_ReLU_one_hidden_layer_final_weight_norm}
\end{subfigure}
\caption{Evolution of $\bs{\theta}_t^2$ in a two-layer ReLU MLP trained on the two-class version of MNIST. Here we have zoomed into the region $\eta\lambda_{\text{max}}(H_{\alpha\beta,0})\leq 5$ to see how the evolution in $\bs{\theta}_t^2$ changes qualitatively as we increase the learning rate. For the evolution of these quantities for larger learning rates see Figure \ref{fig:MNIST_011_ReLU_one_hidden_layer}.}
\label{fig:MNIST_011_ReLU_one_hidden_layer_zoomed_in}
\end{figure*}

Here we will study two-layer MLPs trained on the two-class versions of MNIST, FMNIST and CIFAR-10. The primary results for the two-class version of MNIST are given in Figure \ref{fig:MNIST_011_ReLU_one_hidden_layer}.
However, in Figure \ref{fig:MNIST_011_ReLU_one_hidden_layer} it is difficult to see how the weight norm behaves for $\eta\lambda_{\text{max}}(H_{\alpha\beta,0})\lesssim 4$, so in Figure \ref{fig:MNIST_011_ReLU_one_hidden_layer_zoomed_in} we plot the evolution of $\bs{\theta}_t^2$ zoomed in to this region.
We observe the expected behavior, for $\eta\lambda_{\text{max}}(H_{\alpha\beta,0})\lesssim 4$ the weight norm decreases during training, but for larger learning rates it receives large, positive corrections.

\begin{figure*}[!ht]
\centering
\begin{subfigure}[b]{0.3\textwidth}
\centering
\includegraphics[scale=.27]{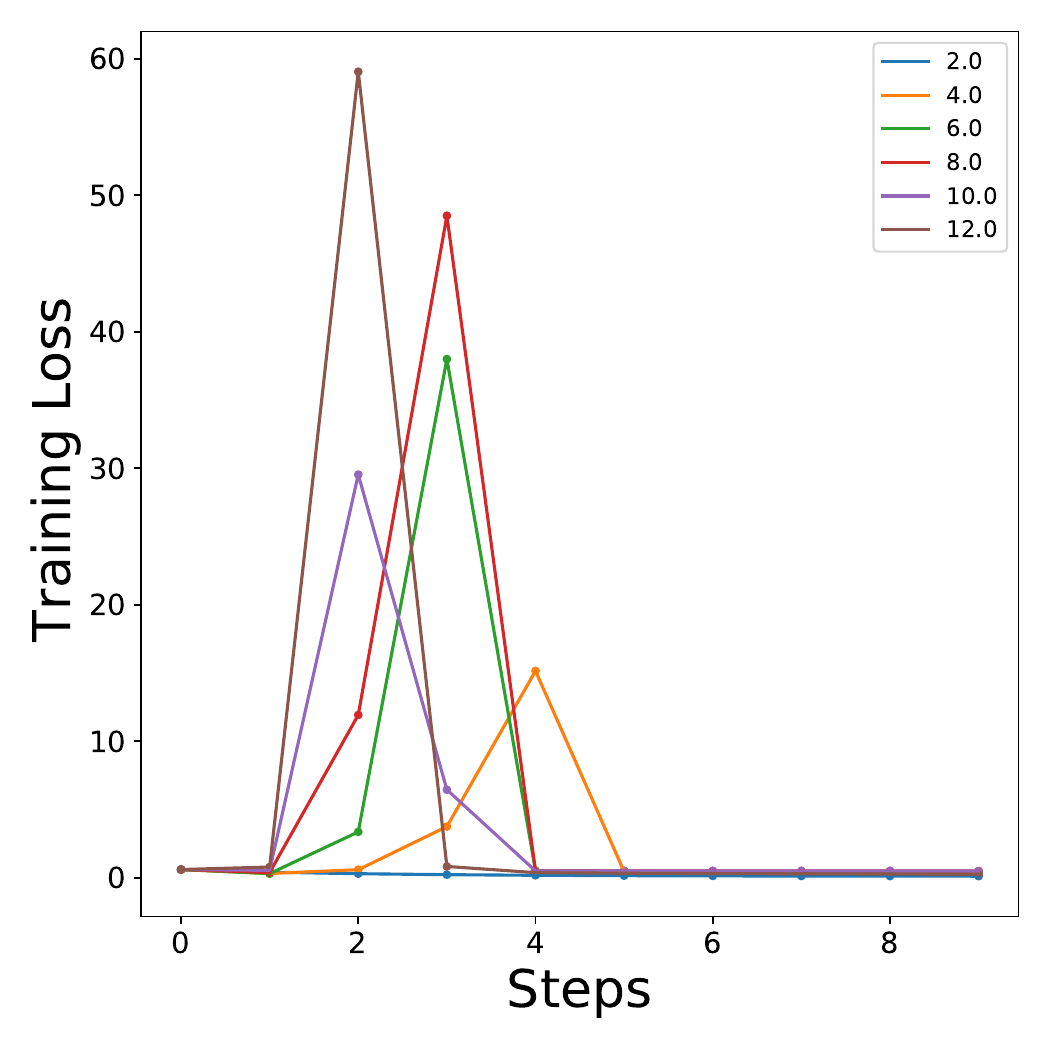}
\caption{}
\end{subfigure}
\hfill
\begin{subfigure}[b]{0.3\textwidth}
\centering
\includegraphics[scale=.27]{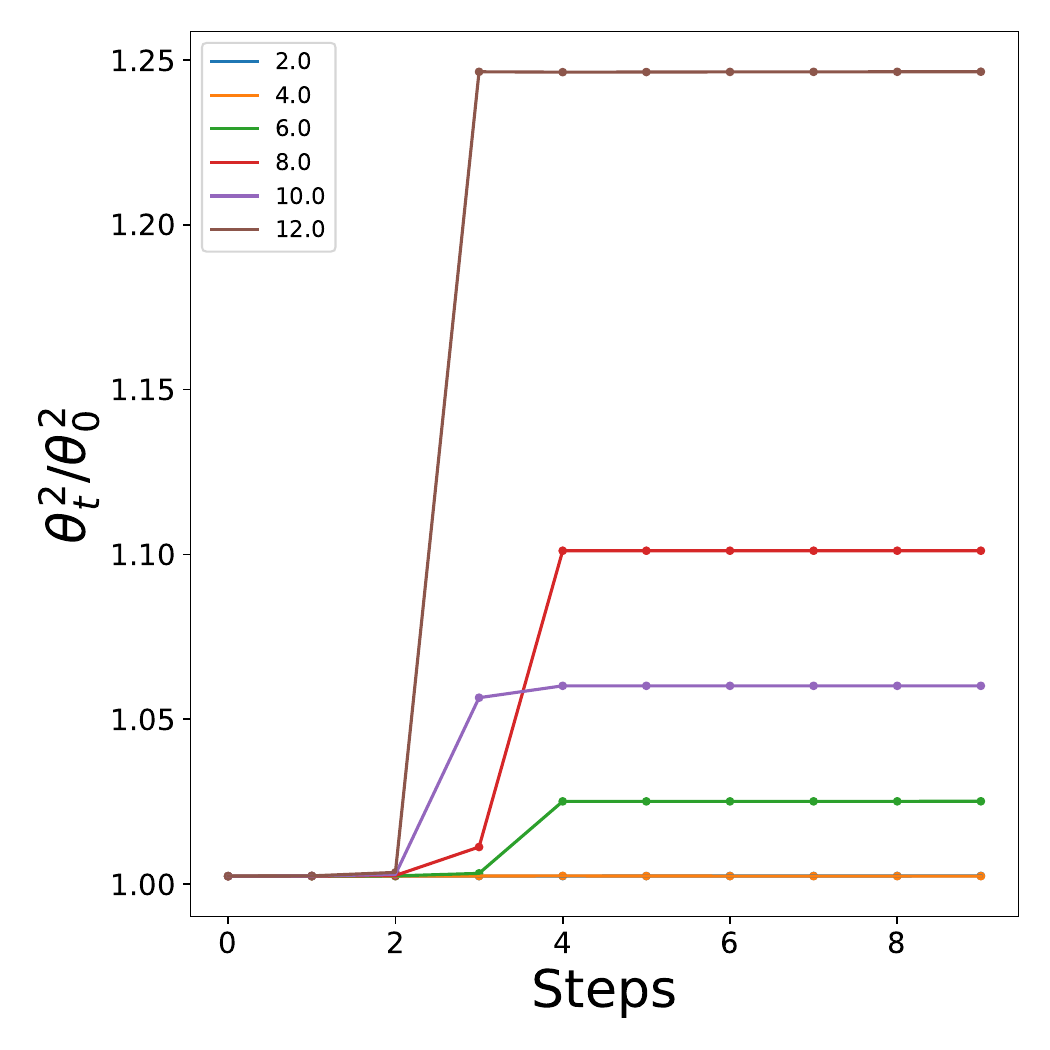}
\caption{}
\label{fig:FMNIST_011_ReLU_one_hidden_layer_weight_norm}

\end{subfigure}
\hfill
\begin{subfigure}[b]{0.3\textwidth}
\centering
\includegraphics[scale=.27]{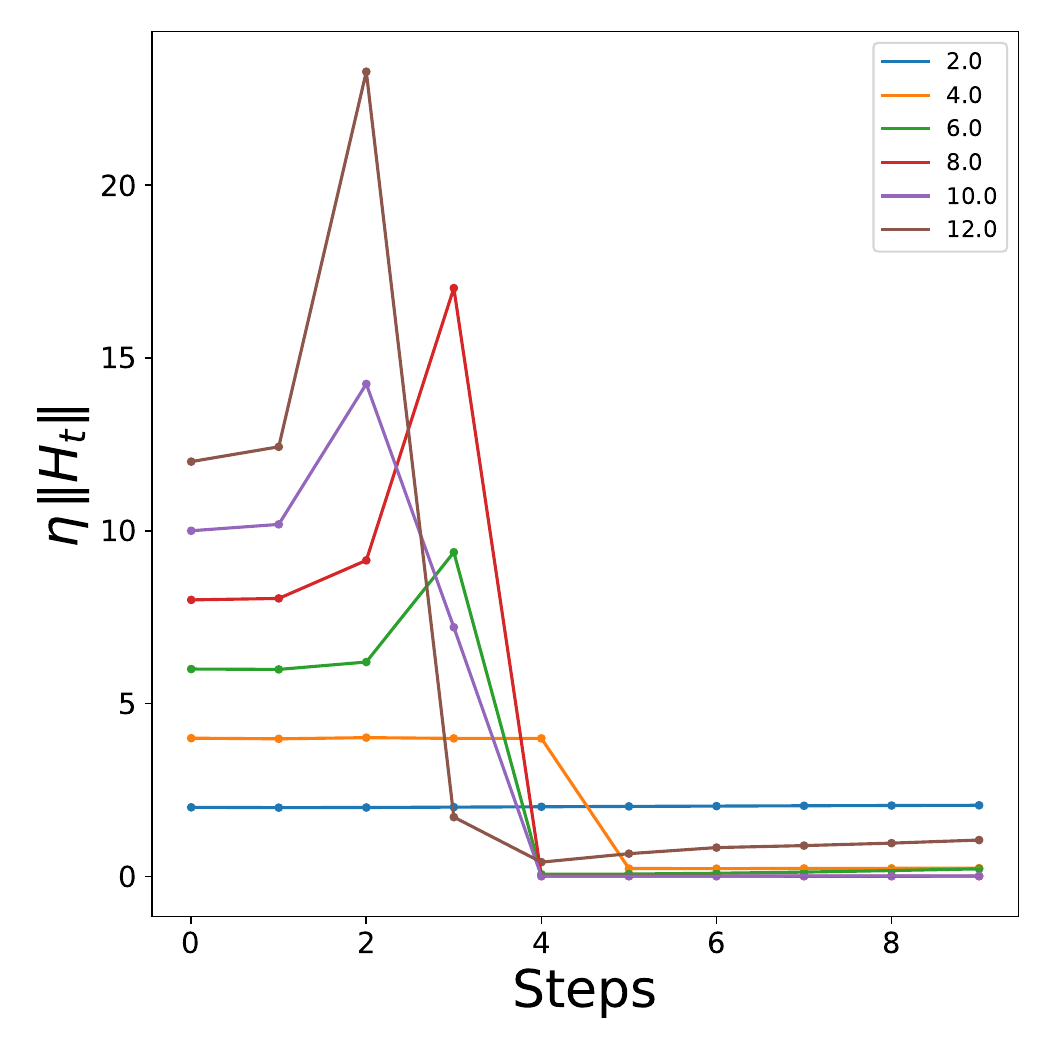}
\caption{}
\label{fig:FMNIST_011_ReLU_one_hidden_layer_NTK}

\end{subfigure}
\centering
\begin{subfigure}[t]{0.3\textwidth}
\centering
\includegraphics[scale=.27]{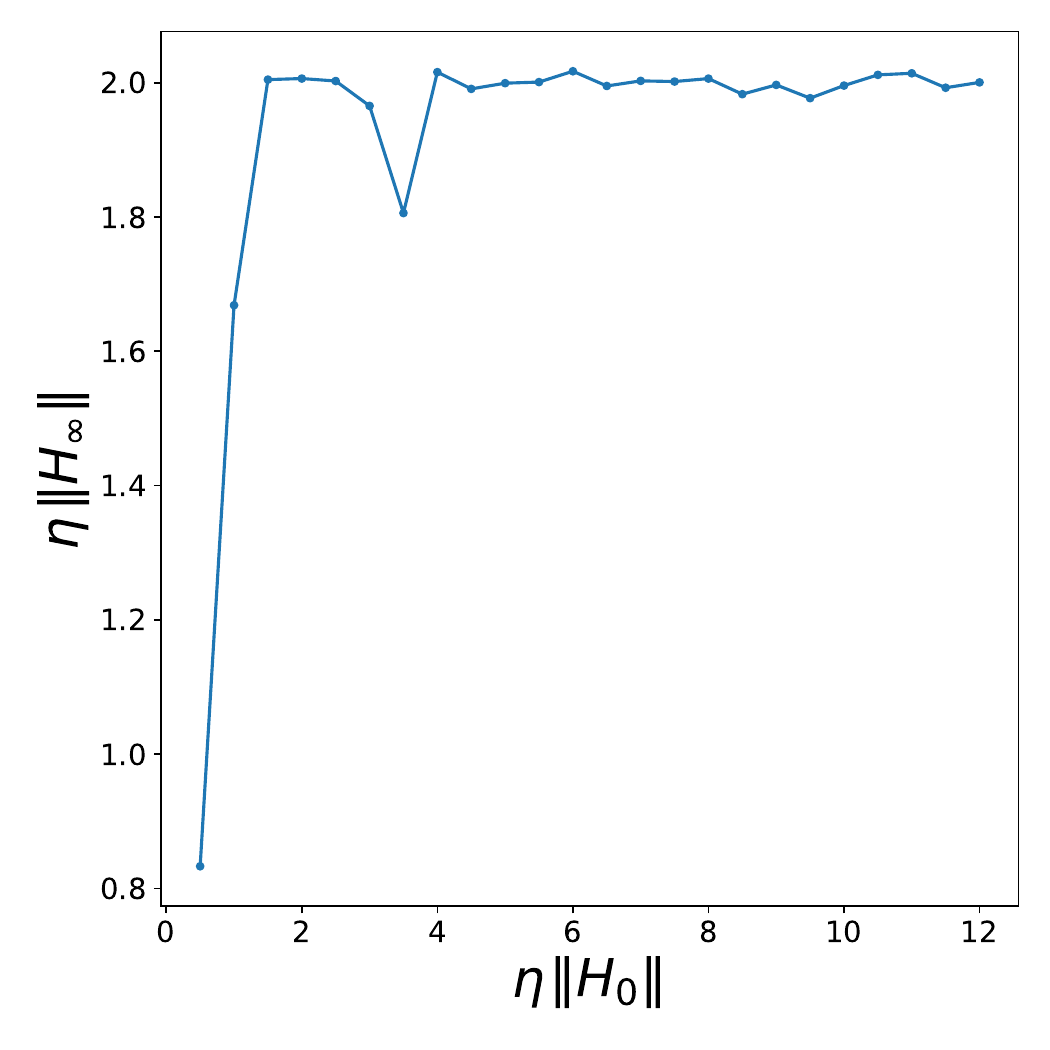}
\caption{}
\label{fig:FMNIST_011_ReLU_one_hidden_layer_final_NTK}

\end{subfigure}
\hfill
\begin{subfigure}[t]{0.3\textwidth}
\centering
\includegraphics[scale=.27]{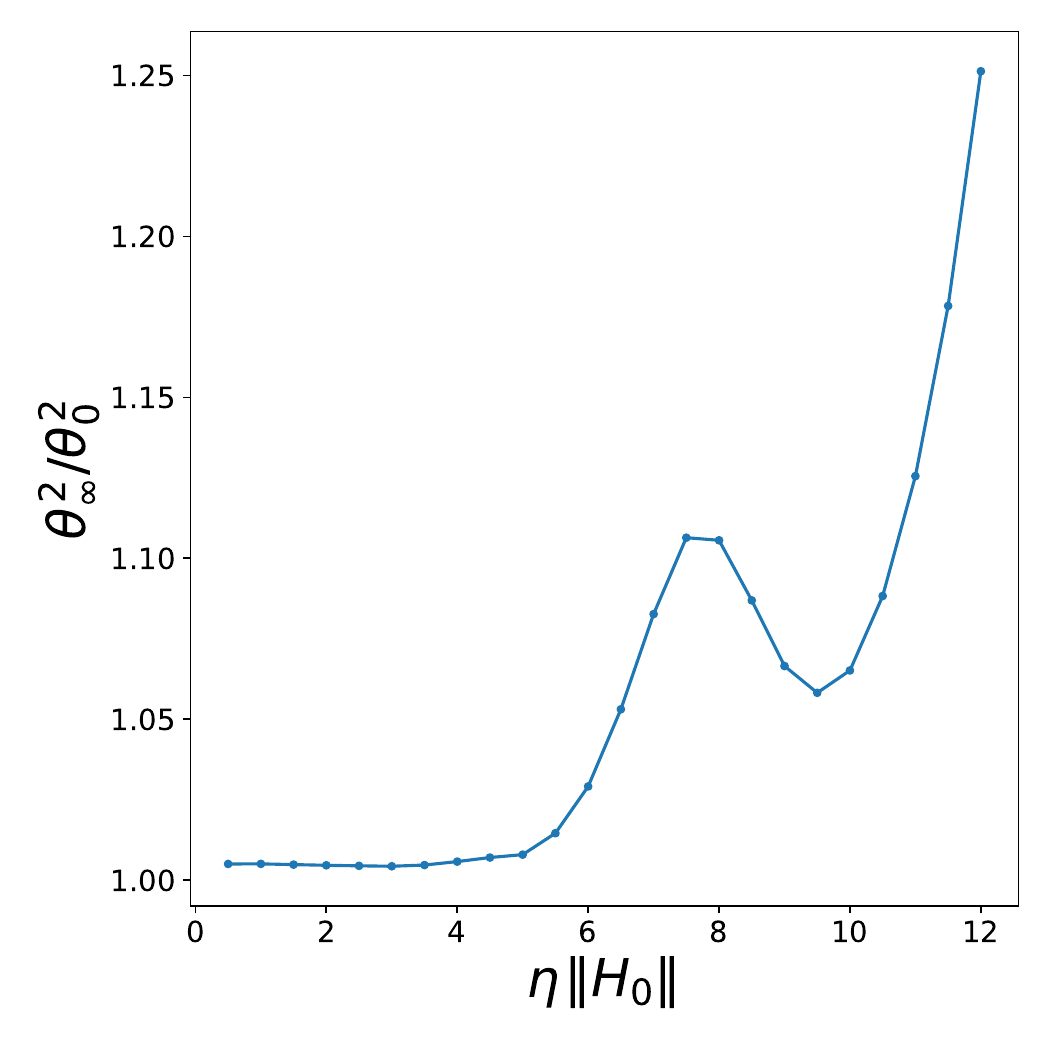}
\caption{}
\label{fig:FMNIST_011_ReLU_one_hidden_layer_final_weight}

\end{subfigure}
\hfill
\begin{subfigure}[t]{0.3\textwidth}
\centering
\includegraphics[scale=.27]{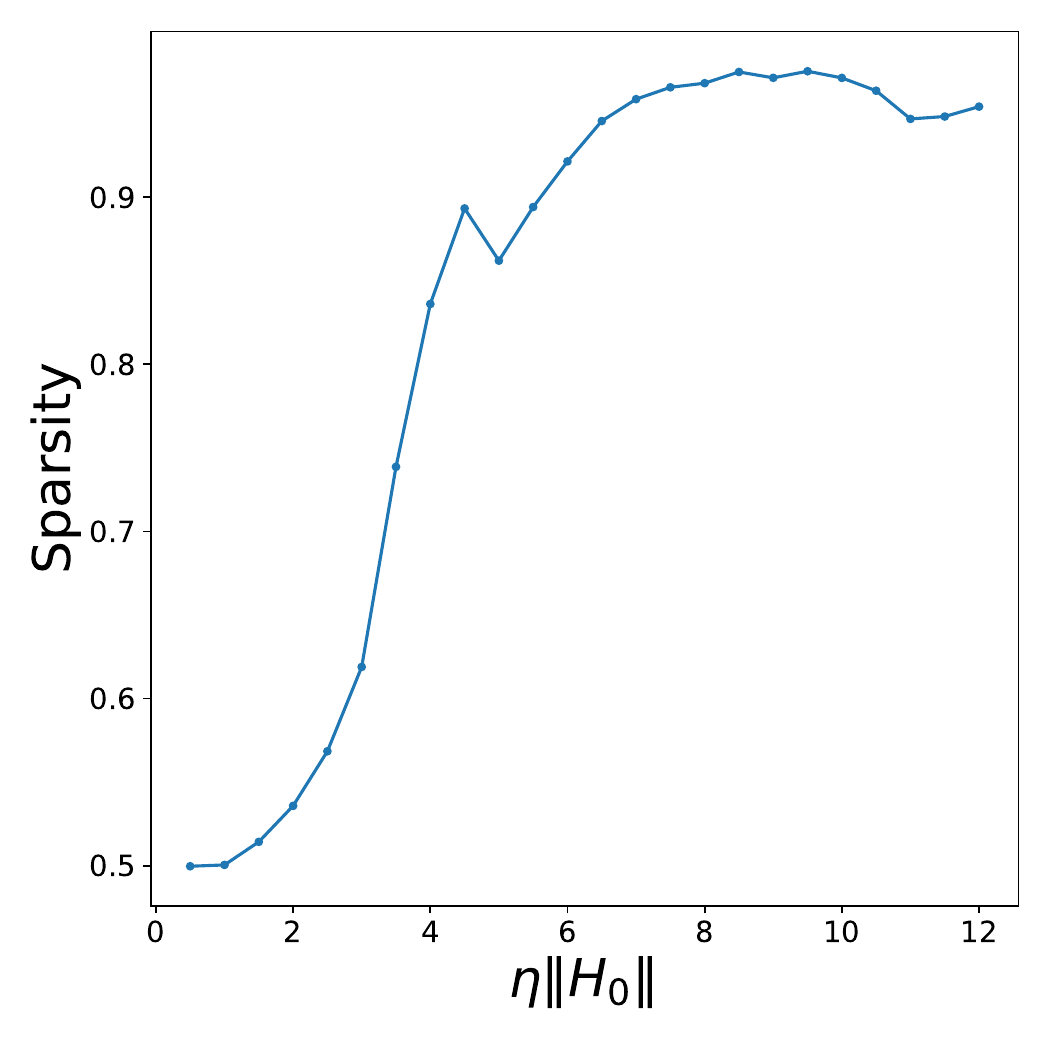}
\caption{}
\label{fig:FMNIST_011_ReLU_one_hidden_layer_sparsity}
\end{subfigure}
\\
\hfill
\begin{subfigure}[t]{\textwidth}
\centering
\includegraphics[scale=.27]{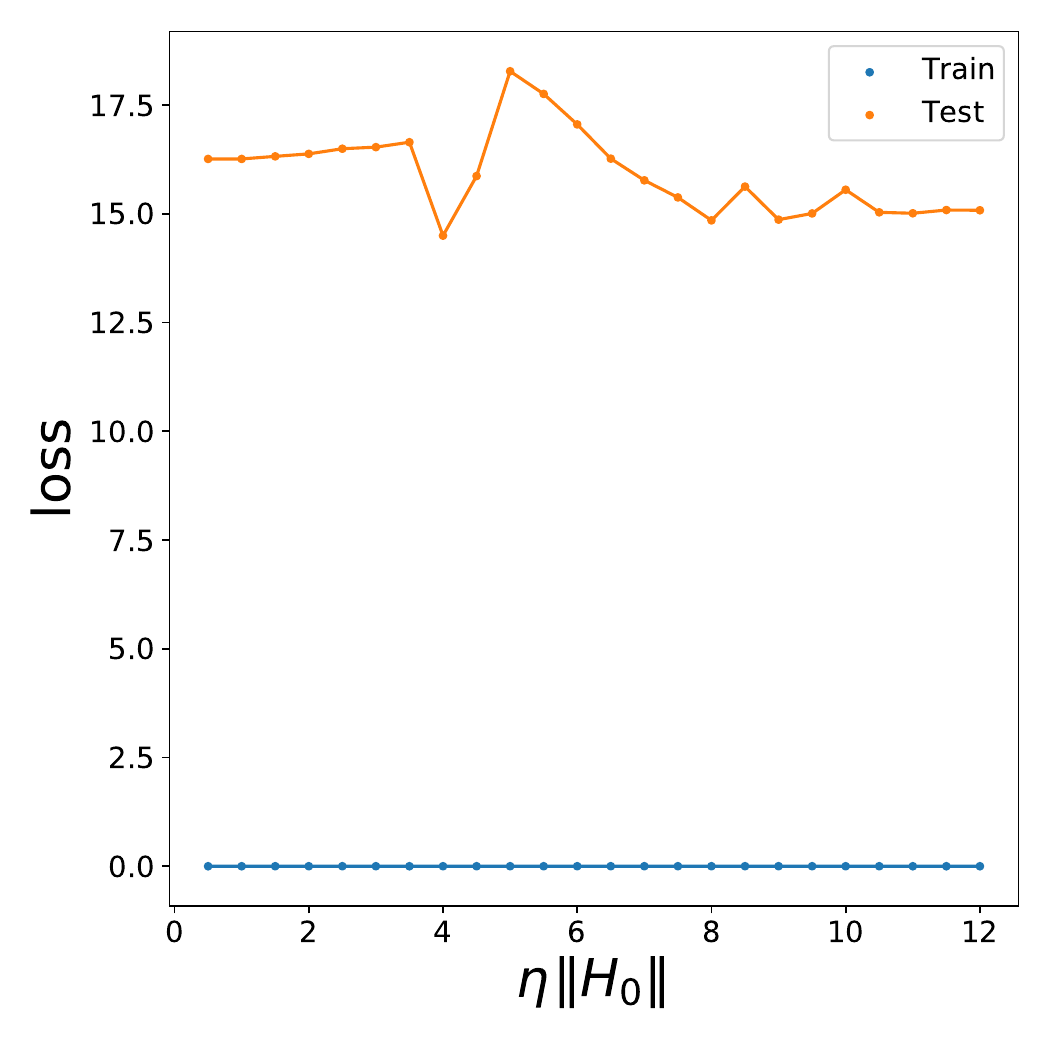}
\caption{}
\label{fig:FMNIST_011_ReLU_one_hidden_layer_gen_loss}
\end{subfigure}
\caption{Results for a two-layer ReLU net trained on the two-class version of FMNIST.
The plots (a)-(g) are the same as in Figure \ref{fig:MNIST_011_ReLU_one_hidden_layer}.
}
\label{fig:FMNIST_011_ReLU_one_hidden_layer}
\end{figure*}

\begin{figure*}[!ht]
\centering

\begin{subfigure}[b]{0.3\textwidth}
\centering
\includegraphics[scale=.27]{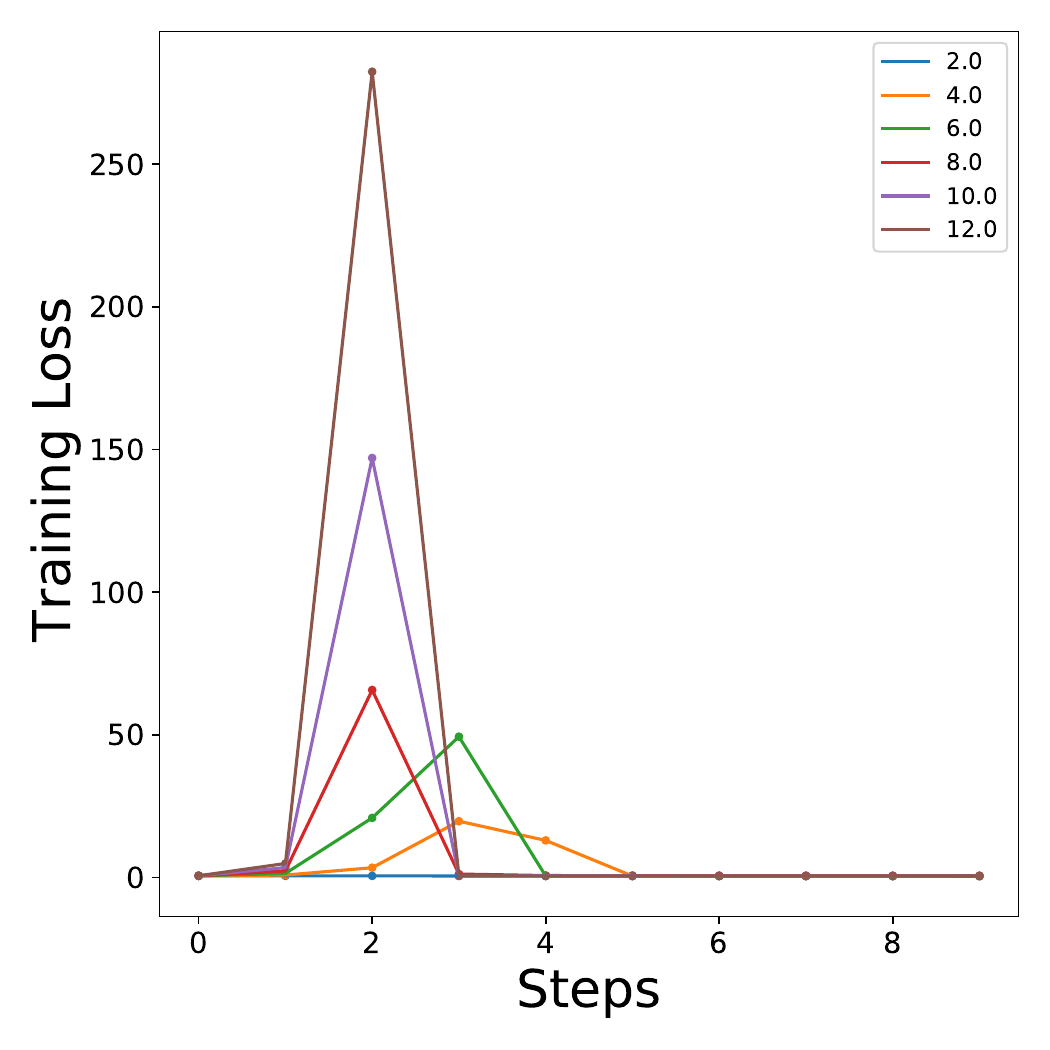}
\caption{}
\end{subfigure}
\hfill
\begin{subfigure}[b]{0.3\textwidth}
\centering
\includegraphics[scale=.27]{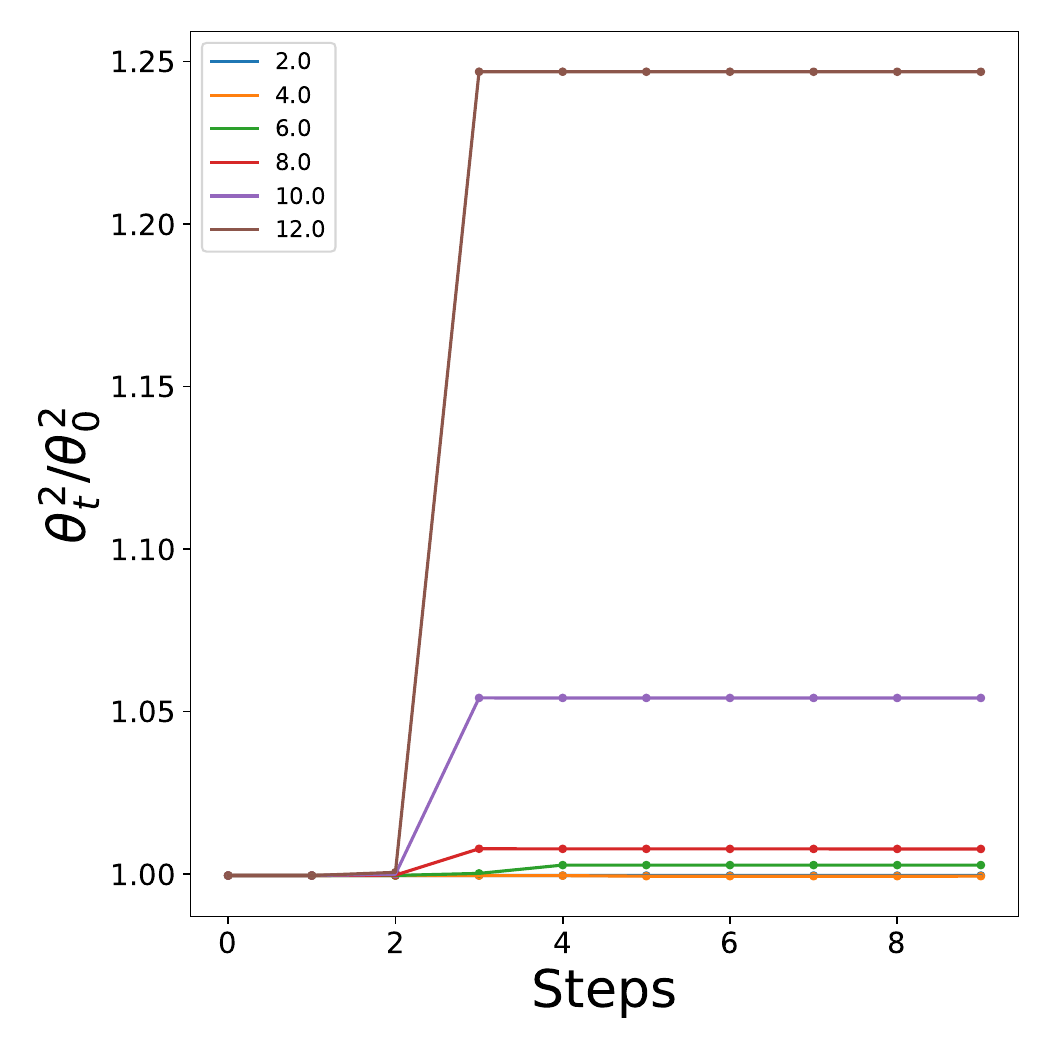}
\caption{}
\label{fig:CIFAR10_01_ReLU_one_hidden_layer_weight_norm}

\end{subfigure}
\hfill
\begin{subfigure}[b]{0.3\textwidth}
\centering
\includegraphics[scale=.27]{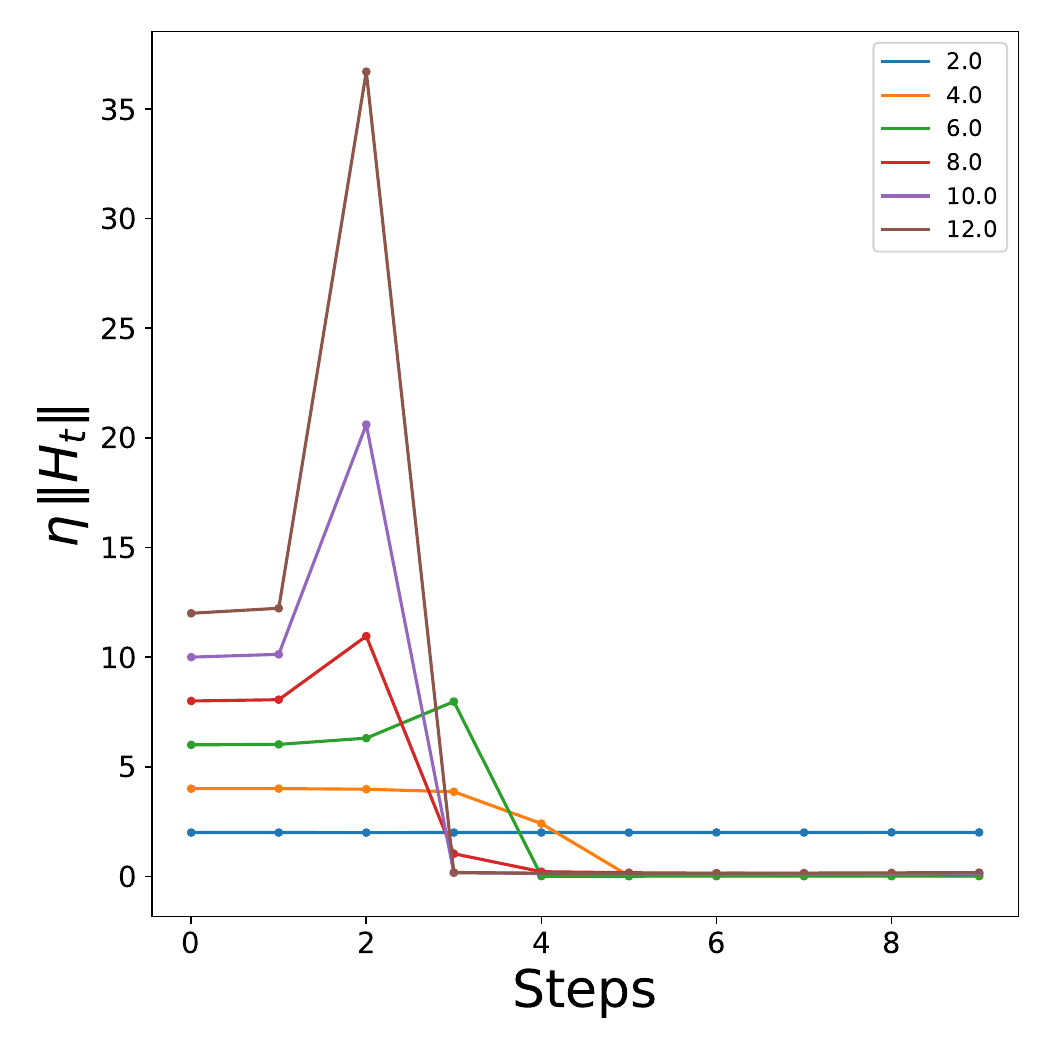}
\caption{}
\label{fig:CIFAR10_01_ReLU_one_hidden_layer_NTK}

\end{subfigure}
\centering
\begin{subfigure}[t]{0.3\textwidth}
\centering
\includegraphics[scale=.27]{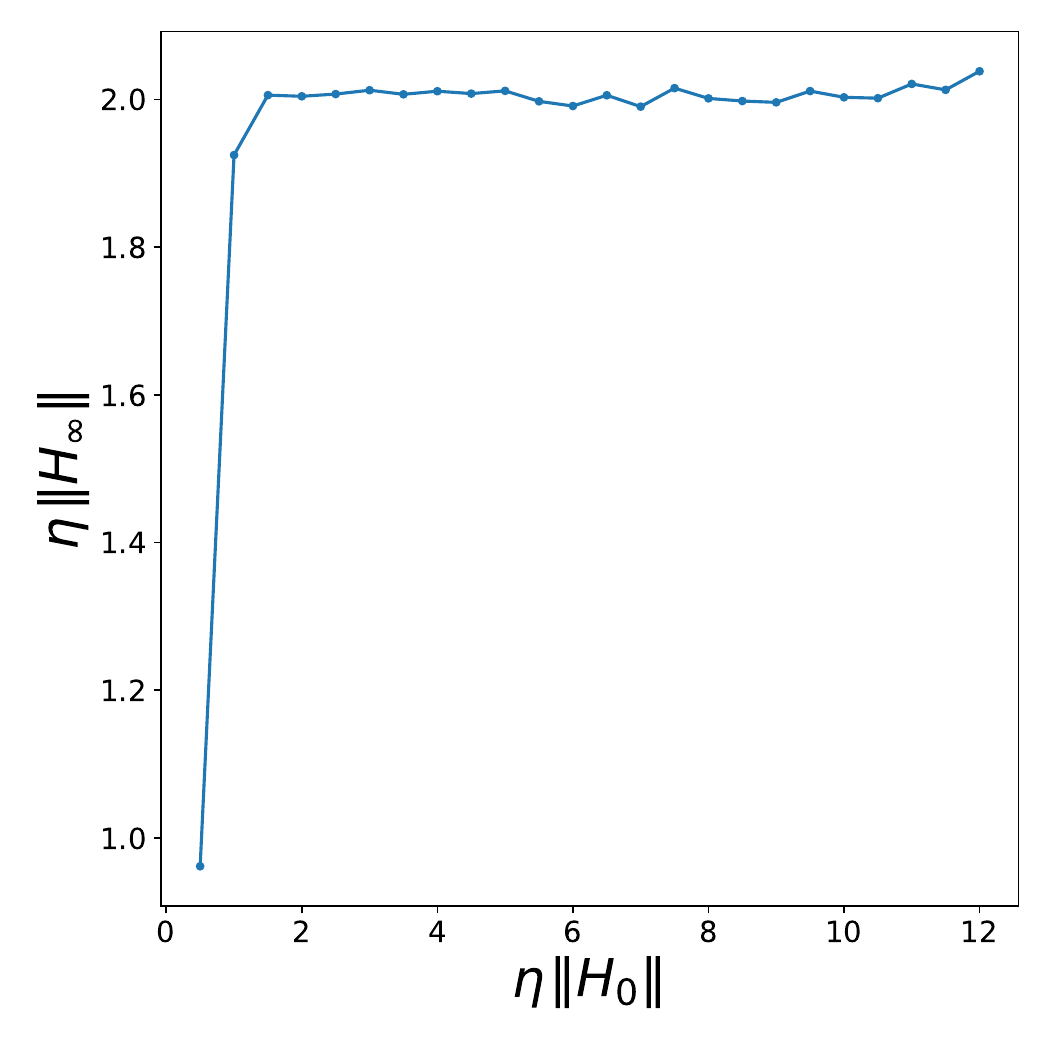}
\caption{}
\label{fig:CIFAR10_01_ReLU_one_hidden_layer_final_NTK}

\end{subfigure}
\hfill
\begin{subfigure}[t]{0.3\textwidth}
\centering
\includegraphics[scale=.27]{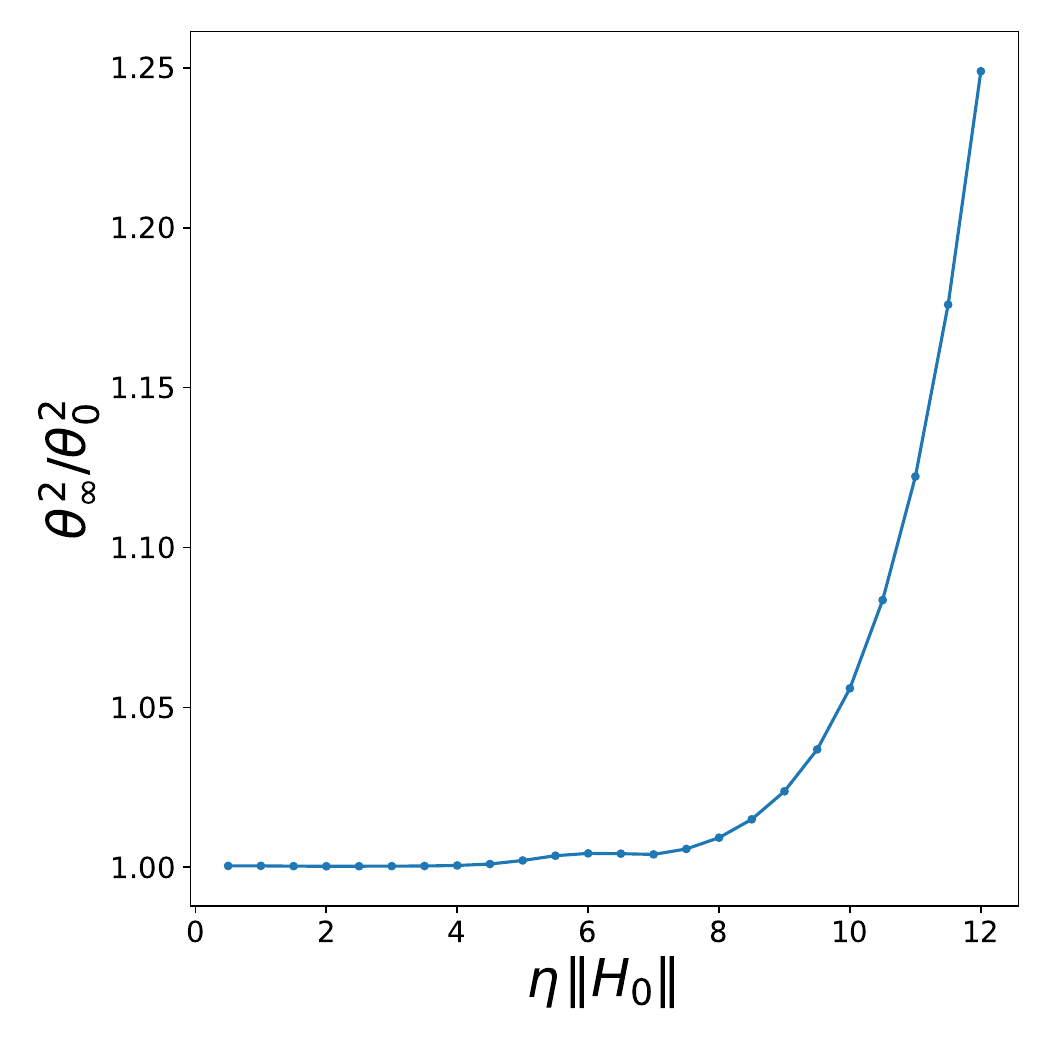}
\caption{}
\label{fig:CIFAR10_01_ReLU_one_hidden_layer_final_weight}

\end{subfigure}
\hfill
\begin{subfigure}[t]{0.3\textwidth}
\centering
\includegraphics[scale=.27]{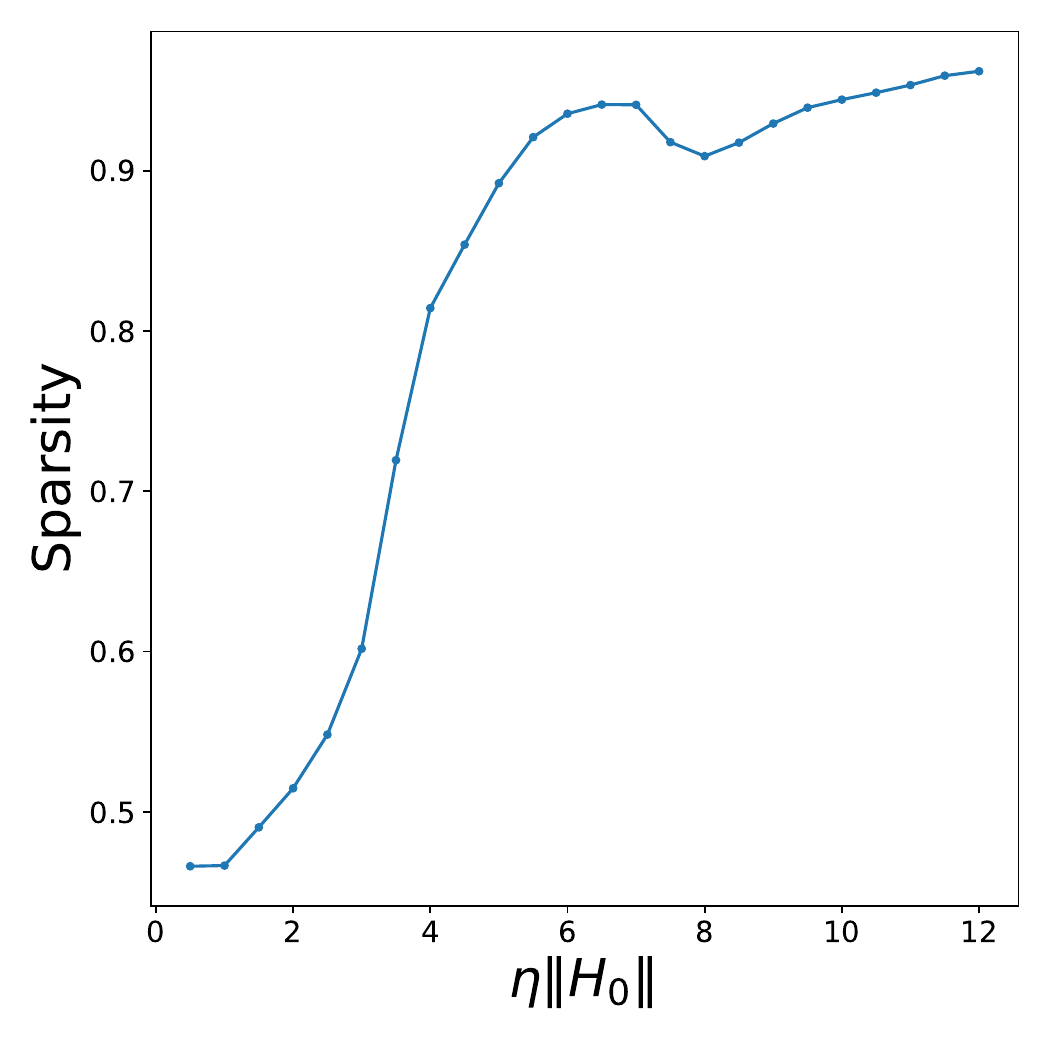}
\caption{}
\label{fig:CIFAR10_01_ReLU_one_hidden_layer_sparsity}
\end{subfigure}
\\
\hfill
\begin{subfigure}[t]{\textwidth}
\centering
\includegraphics[scale=.27]{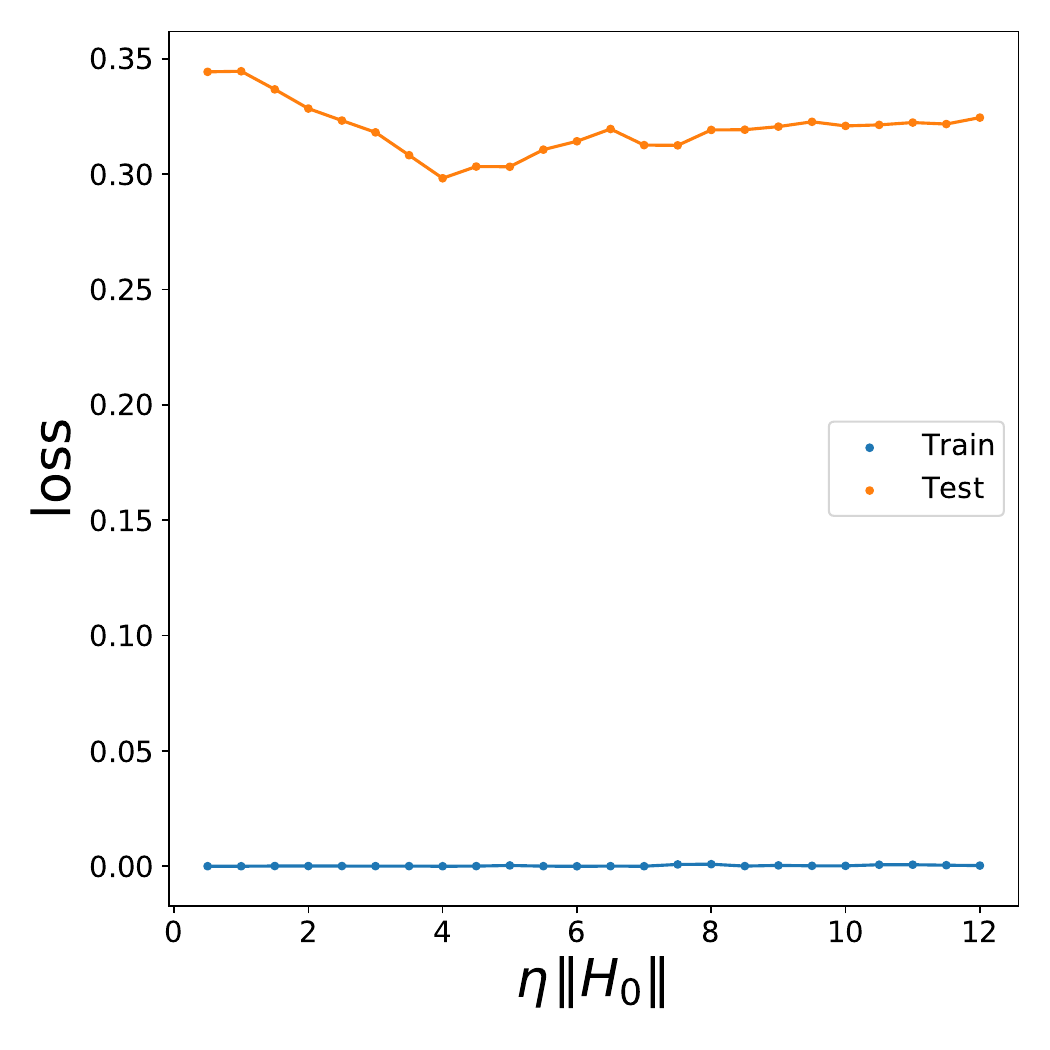}
\caption{}
\label{fig:CIFAR10_01_ReLU_one_hidden_layer_gen_loss}
\end{subfigure}
\caption{Results for a two-layer ReLU net trained on the two-class version of CIFAR-10.
The plots (a)-(g) are the same as in Figure \ref{fig:MNIST_011_ReLU_one_hidden_layer}.}
\label{fig:CIFAR10_01_ReLU_one_hidden_layer}
\end{figure*}

The results for the two-layer MLP trained on the two-class version of FMNIST are given in Figure \ref{fig:FMNIST_011_ReLU_one_hidden_layer}.
For the most part, the results are qualitatively the same as for MNIST. 
One qualitative difference is that in Figure \ref{fig:FMNIST_011_ReLU_one_hidden_layer_final_NTK} we observe the final value of $\eta\lambda_{\text{max}}(H_{\alpha\beta,t})$ converges to $2$ for almost all learning rates in the catapult phase.
This is a pronounced example of how training a model with full-batch gradient descent causes the top eigenvalue of the NTK to hover around the edge of stability \cite{https://doi.org/10.48550/arxiv.2103.00065,Pennington_second_order}.
We also observe, as before, that the weight norm $\bs{\theta}_t^2$ increases for $\eta\lambda_{\text{max}}(H_{\alpha\beta,0})\gtrsim 4$ and the increase becomes particularly pronounced when $\eta\lambda_{\text{max}}(H_{\alpha\beta,0})\gtrsim 6$. Finally, as with MNIST, we observe the activation map becomes increasingly sparse as we increase the learning rate.
The analogous results for CIFAR-10 are shown in Figure \ref{fig:CIFAR10_01_ReLU_one_hidden_layer}.
The results are almost identical in form to what we observed for FMNIST.
In particular, we observe that the trained model has a sparse activation map and that the top eigenvalue of the trained NTK hovers around the edge of stability.

\subsubsection{Three Layers}
\begin{figure*}[!ht]
\centering

\begin{subfigure}[b]{0.3\textwidth}
\centering
\includegraphics[scale=.27]{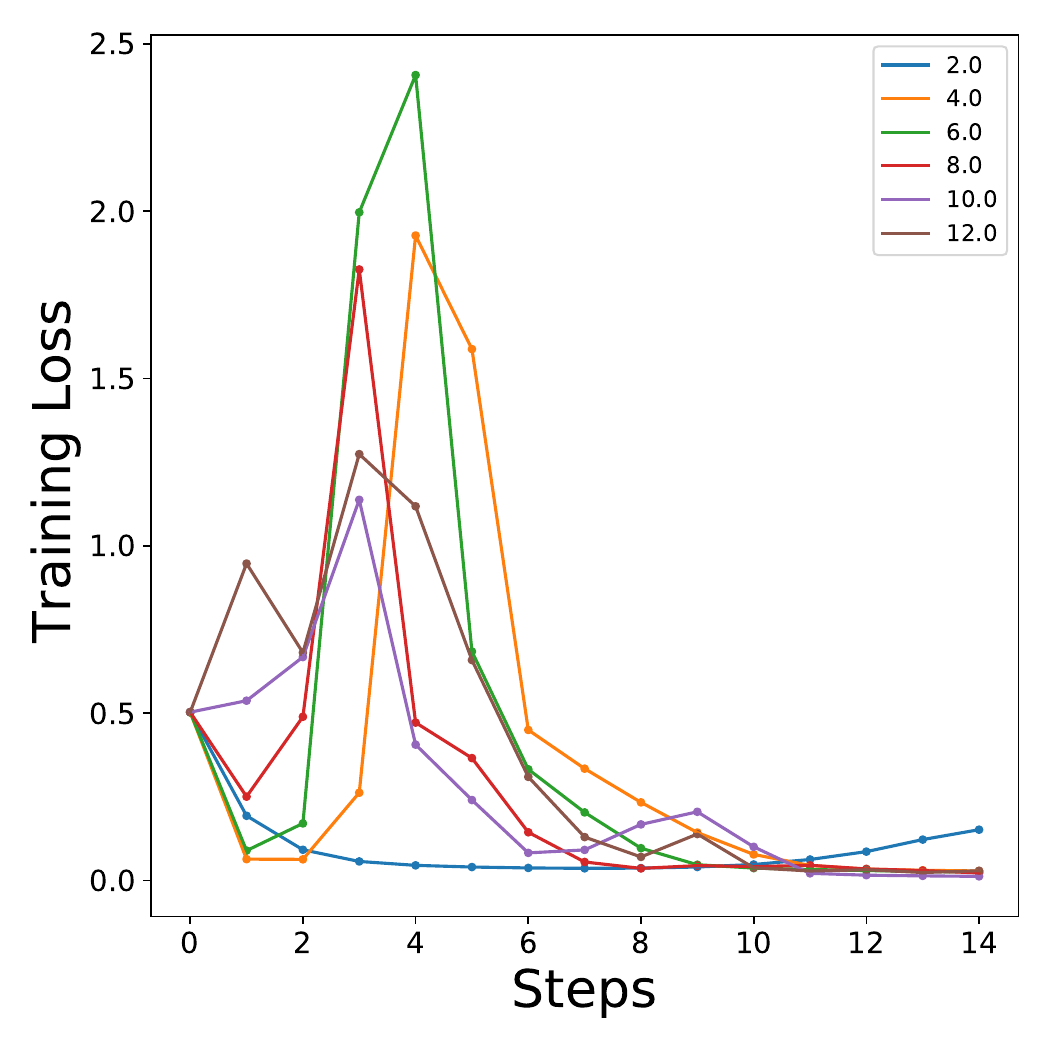}
\caption{}
\end{subfigure}
\hfill
\begin{subfigure}[b]{0.3\textwidth}
\centering
\includegraphics[scale=.27]{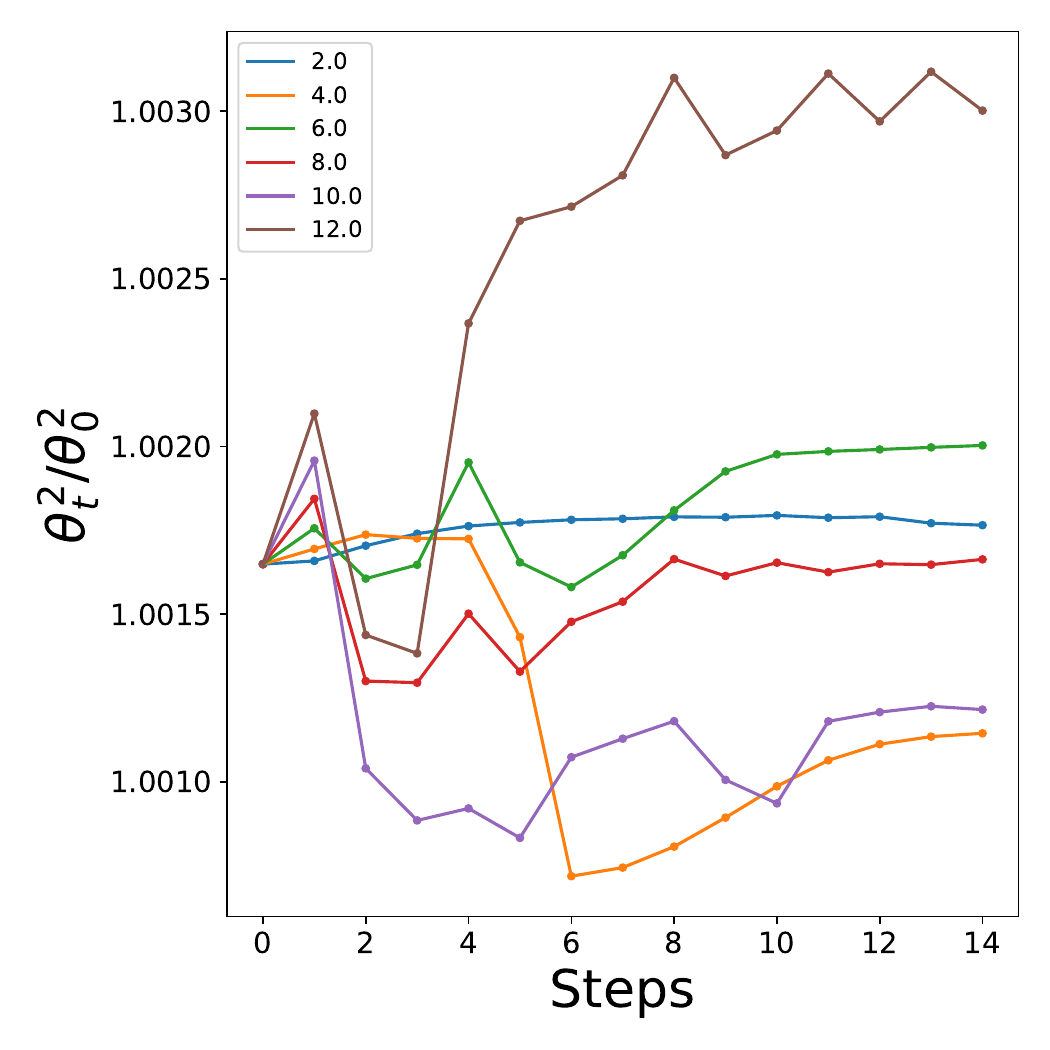}
\caption{}

\end{subfigure}
\hfill
\begin{subfigure}[b]{0.3\textwidth}
\centering
\includegraphics[scale=.27]{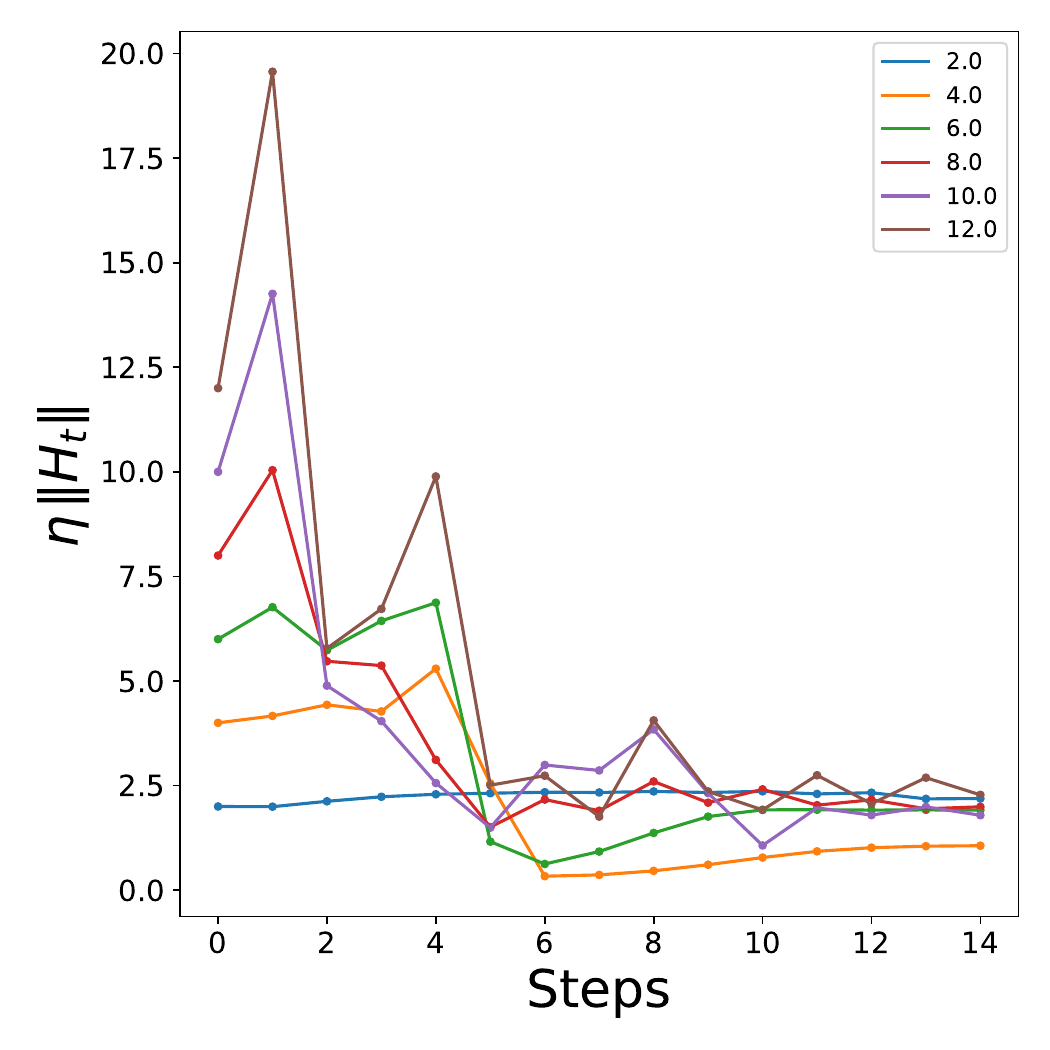}
\caption{}

\end{subfigure}
\centering
\begin{subfigure}[t]{0.3\textwidth}
\centering
\includegraphics[scale=.27]{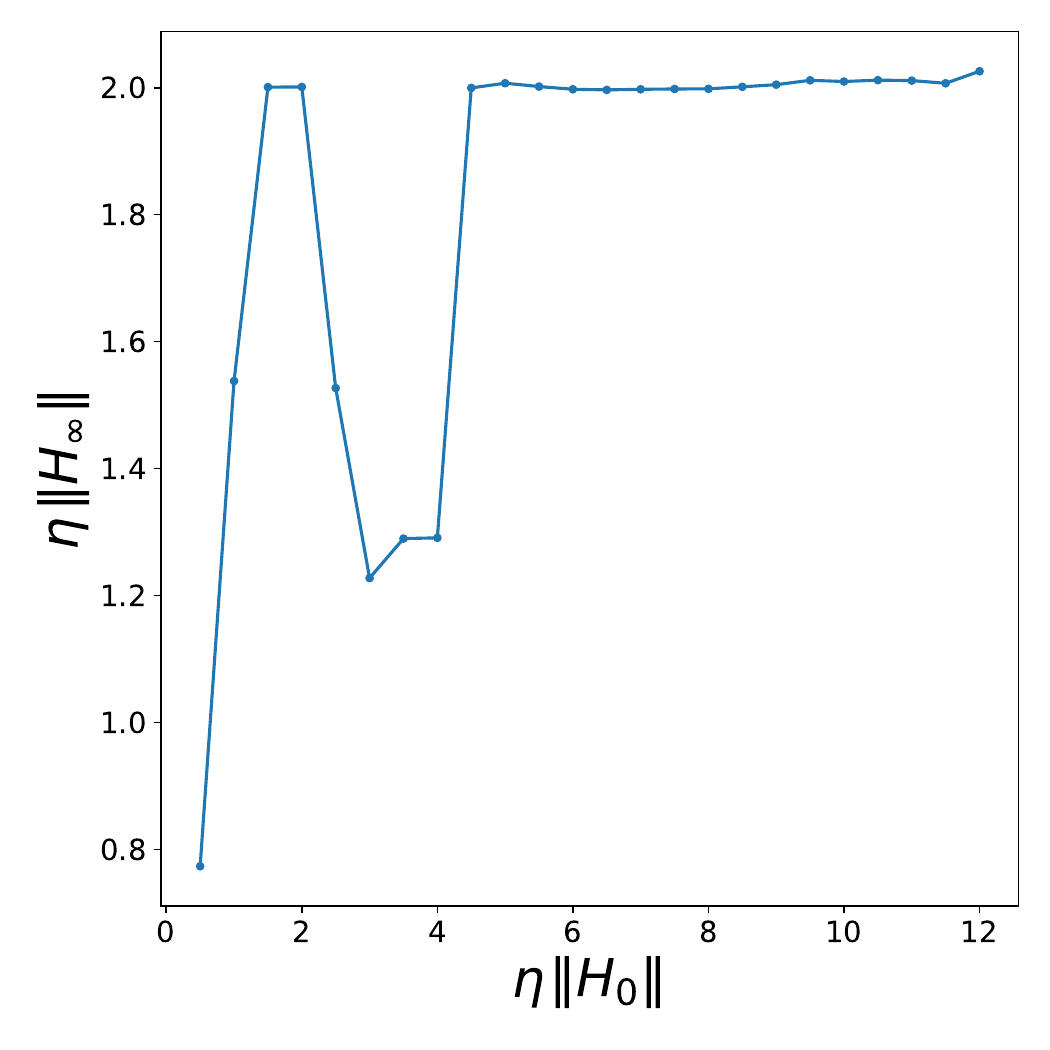}
\caption{}

\end{subfigure}
\hfill
\begin{subfigure}[t]{0.3\textwidth}
\centering
\includegraphics[scale=.27]{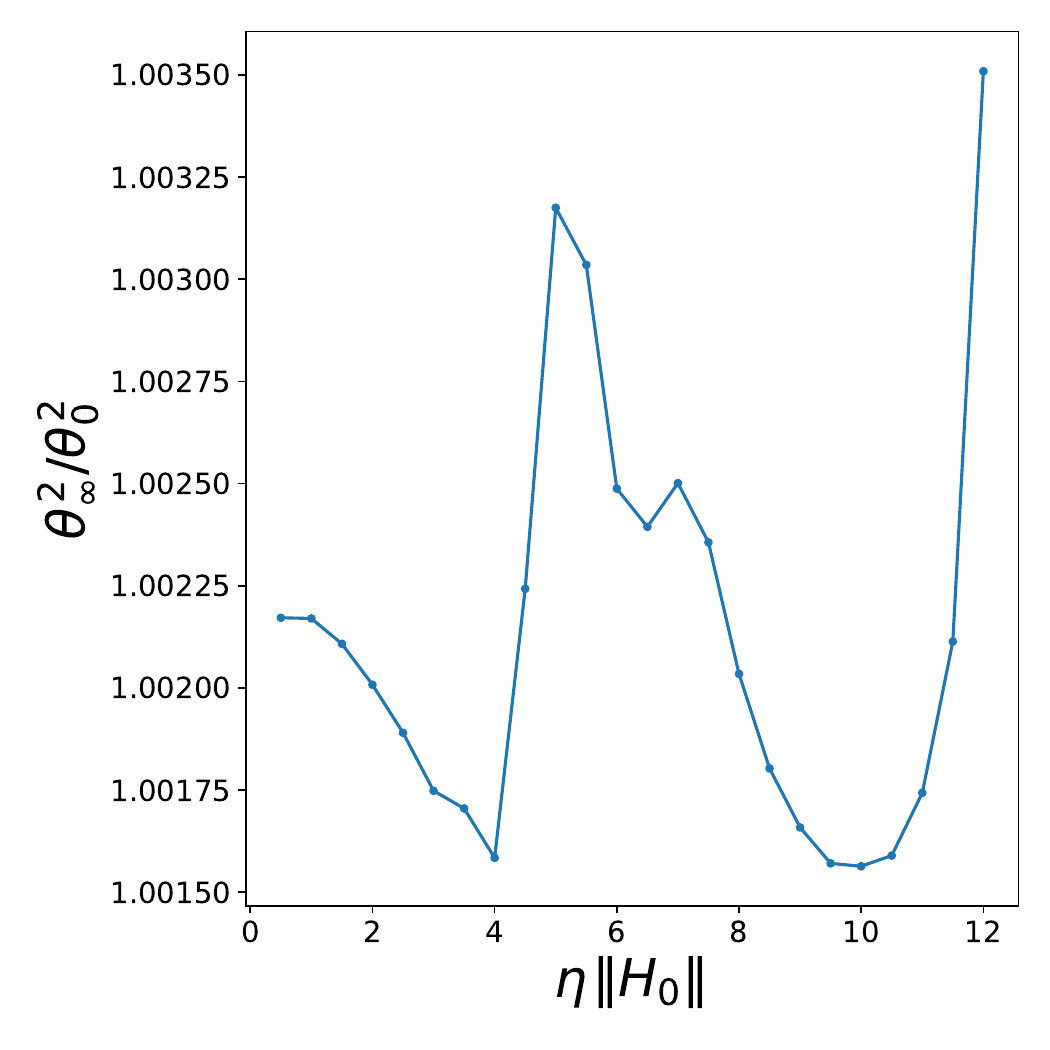}
\caption{}

\end{subfigure}
\hfill
\begin{subfigure}[t]{0.3\textwidth}
\centering
\includegraphics[scale=.27]{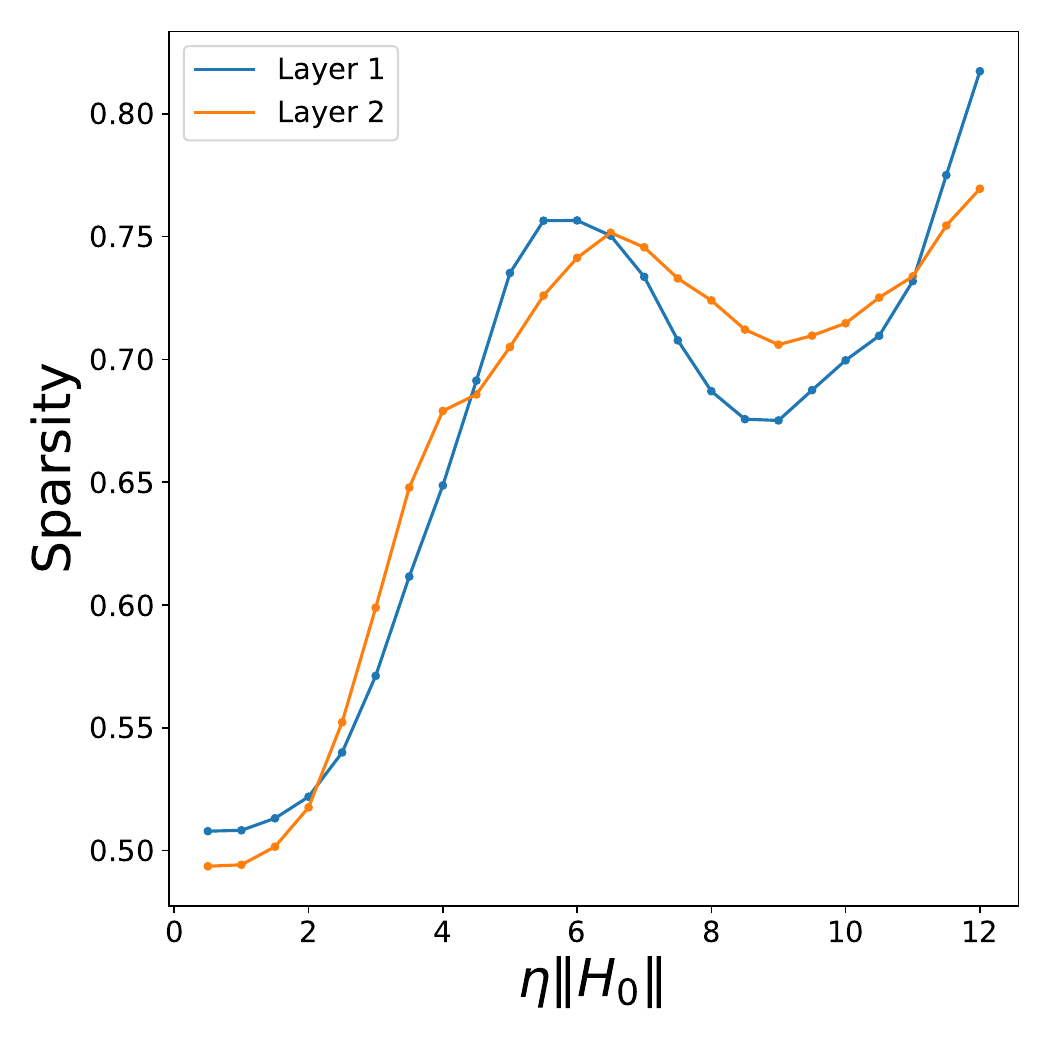}
\caption{}
\end{subfigure}
\\
\hfill
\begin{subfigure}[t]{\textwidth}
\centering
\includegraphics[scale=.27]{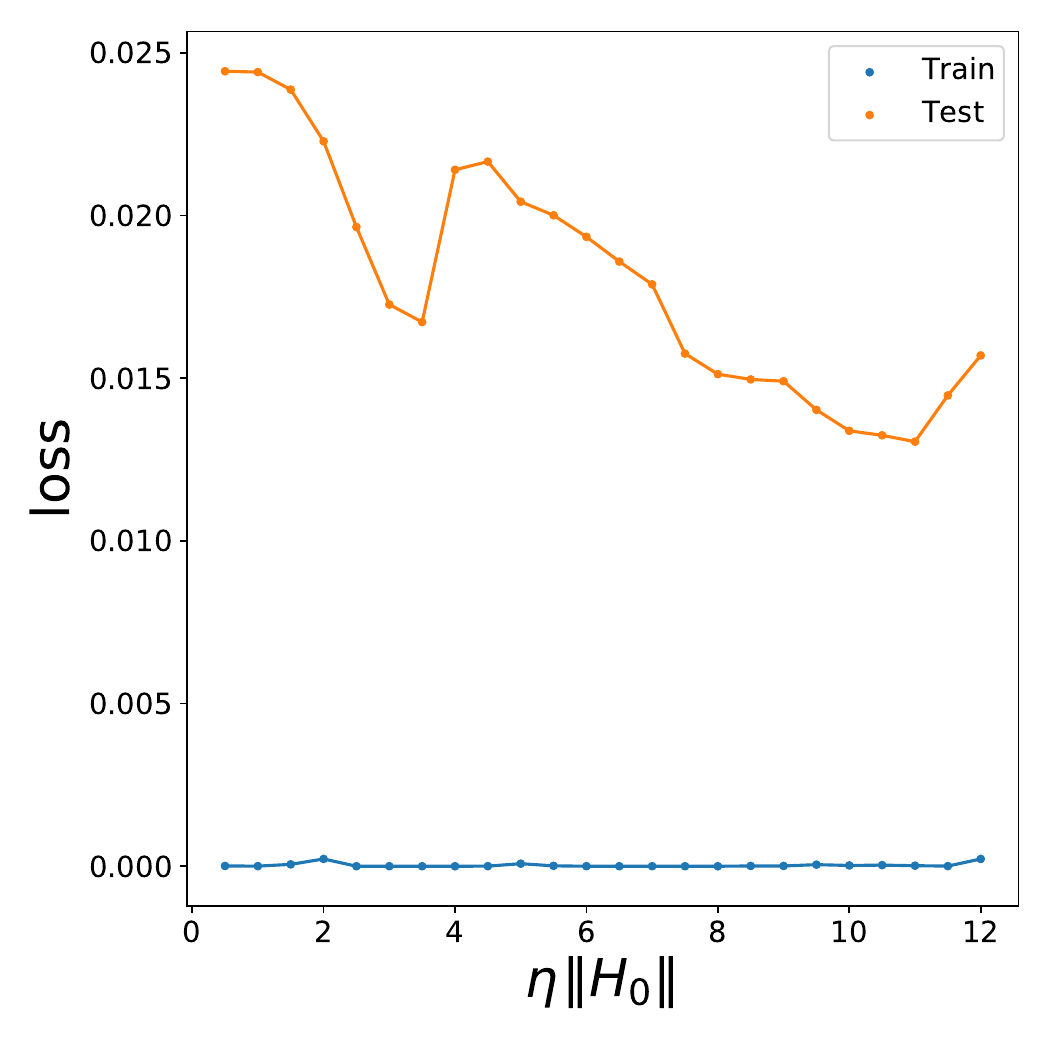}
\caption{}
\end{subfigure}
\caption{Results for the three-layer ReLU net trained on the two-class version of MNIST.}
\label{fig:MNIST_011_ReLU_two_hidden_layer}
\end{figure*}

\begin{figure*}[!ht]
\centering

\begin{subfigure}[b]{0.3\textwidth}
\centering
\includegraphics[scale=.27]{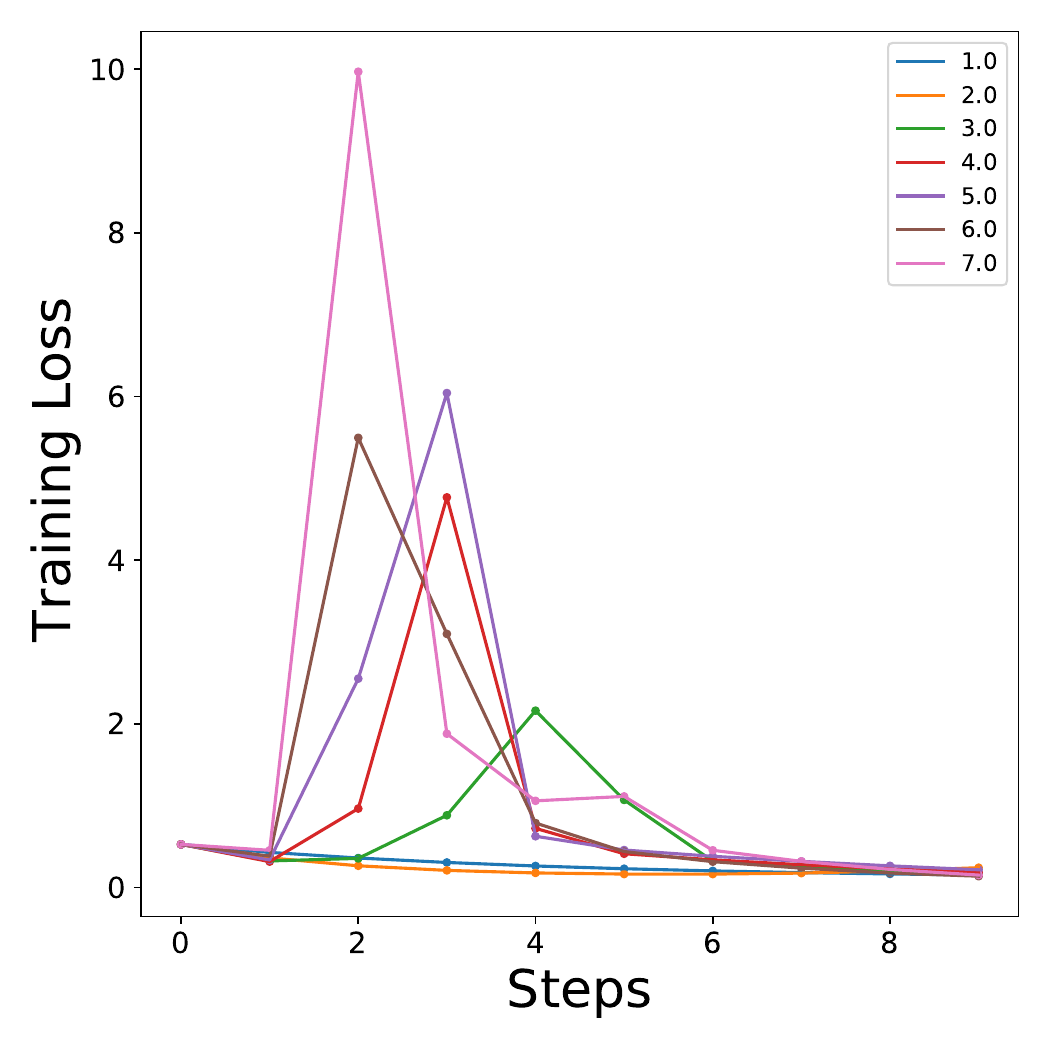}
\caption{}
\end{subfigure}
\hfill
\begin{subfigure}[b]{0.3\textwidth}
\centering
\includegraphics[scale=.27]{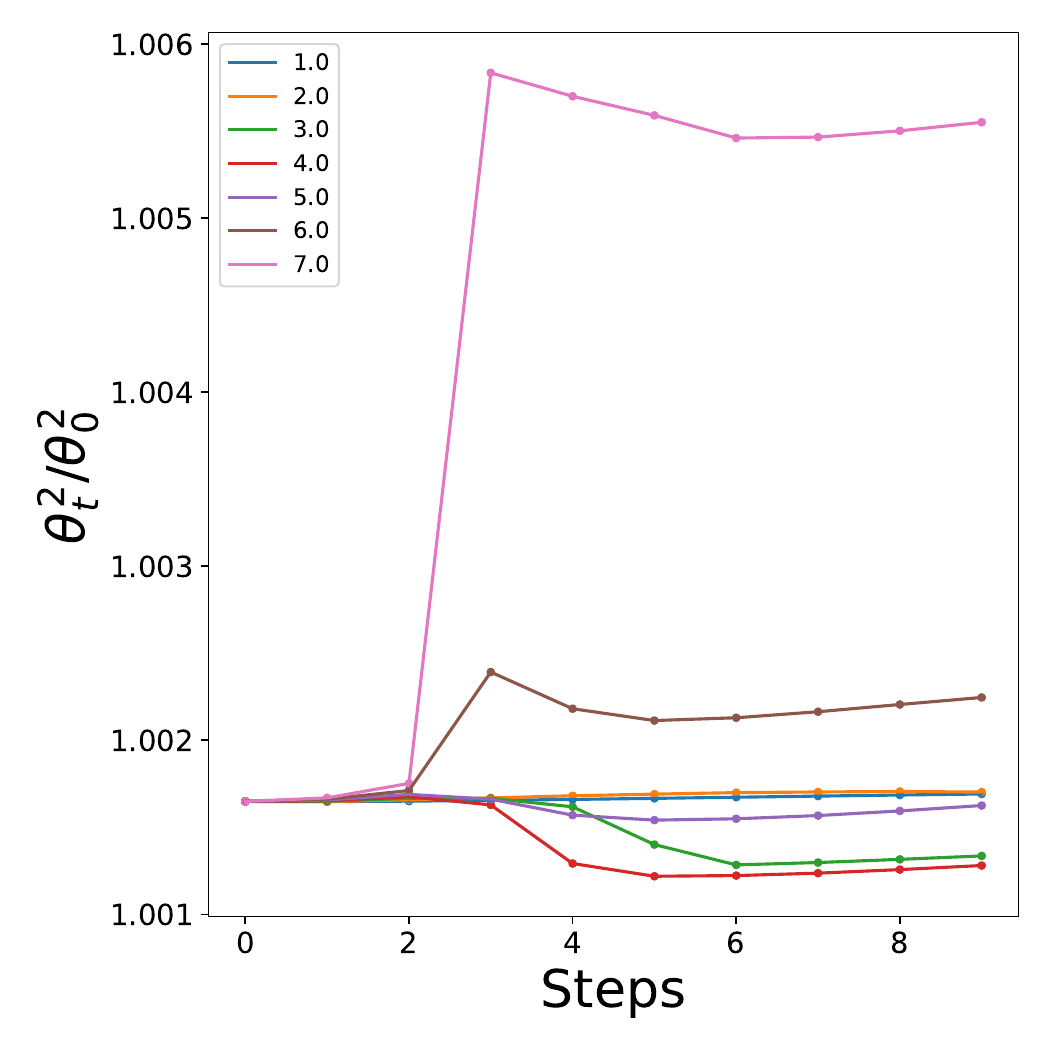}
\caption{}

\end{subfigure}
\hfill
\begin{subfigure}[b]{0.3\textwidth}
\centering
\includegraphics[scale=.27]{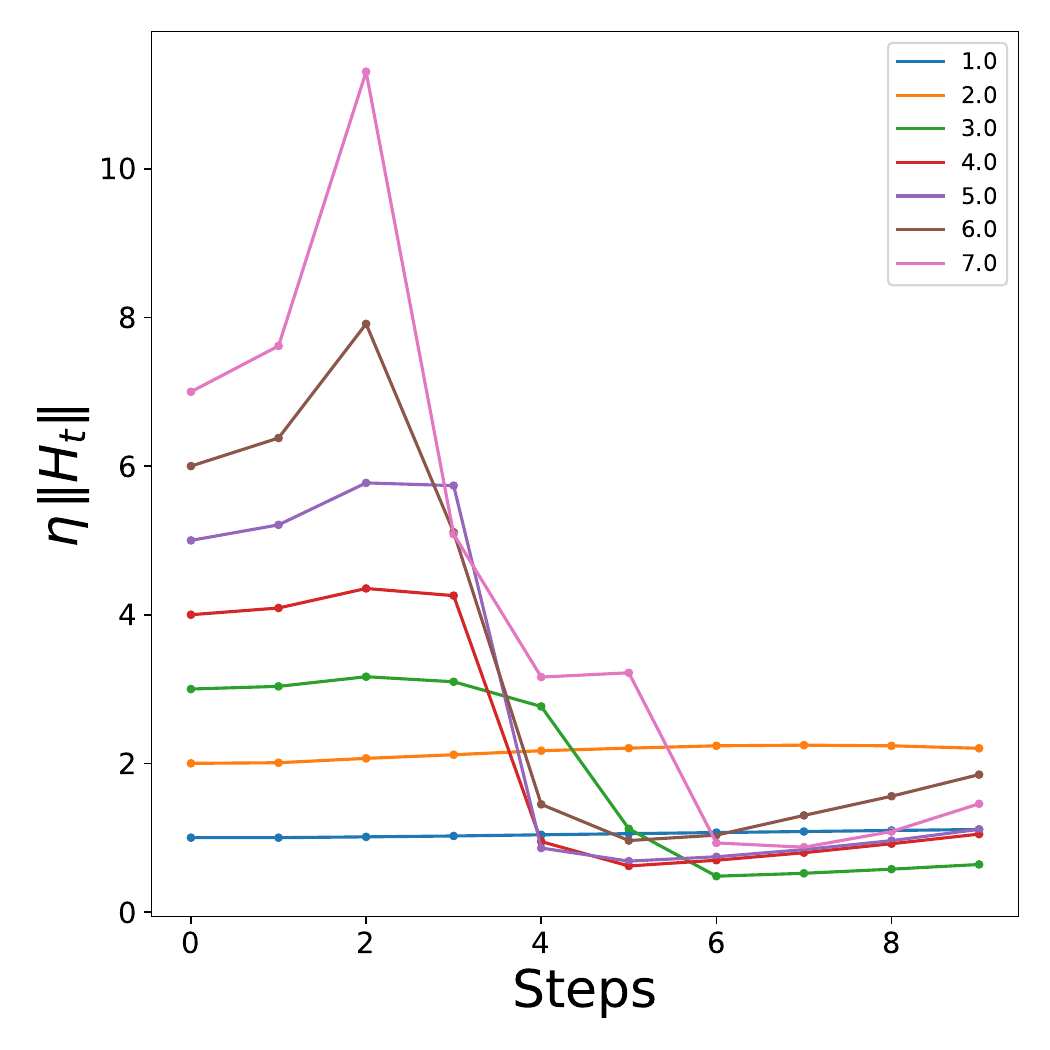}
\caption{}

\end{subfigure}
\centering
\begin{subfigure}[t]{0.3\textwidth}
\centering
\includegraphics[scale=.27]{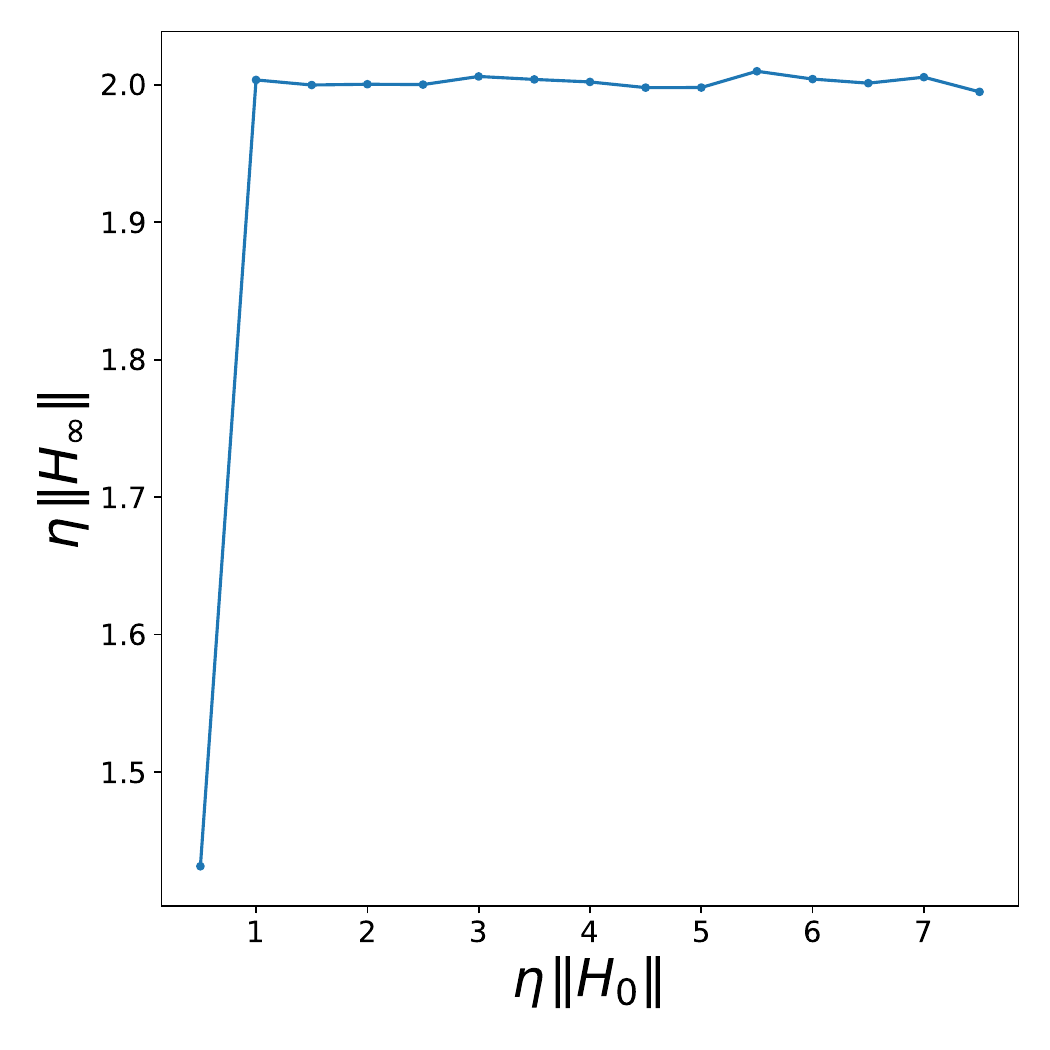}
\caption{}

\end{subfigure}
\hfill
\begin{subfigure}[t]{0.3\textwidth}
\centering
\includegraphics[scale=.27]{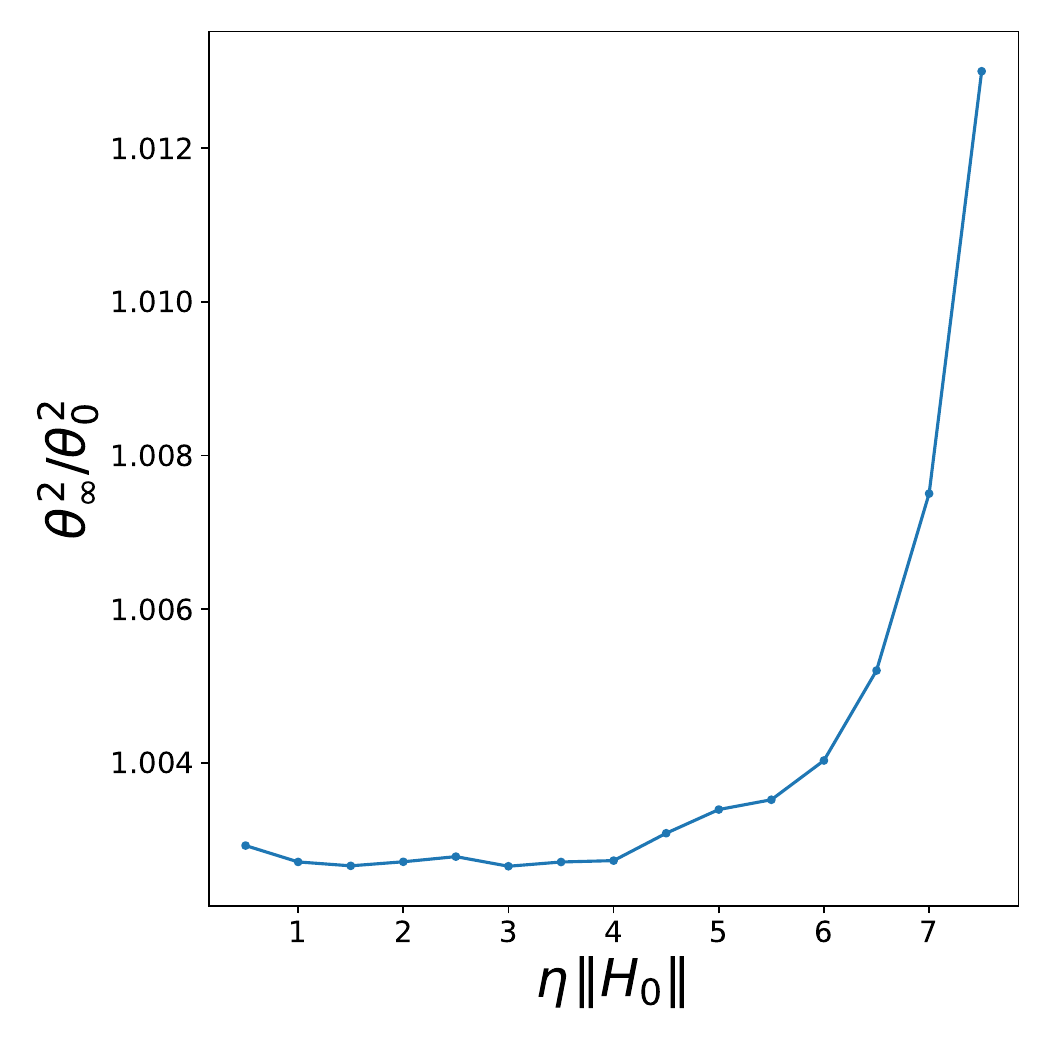}
\caption{}

\end{subfigure}
\hfill
\begin{subfigure}[t]{0.3\textwidth}
\centering
\includegraphics[scale=.27]{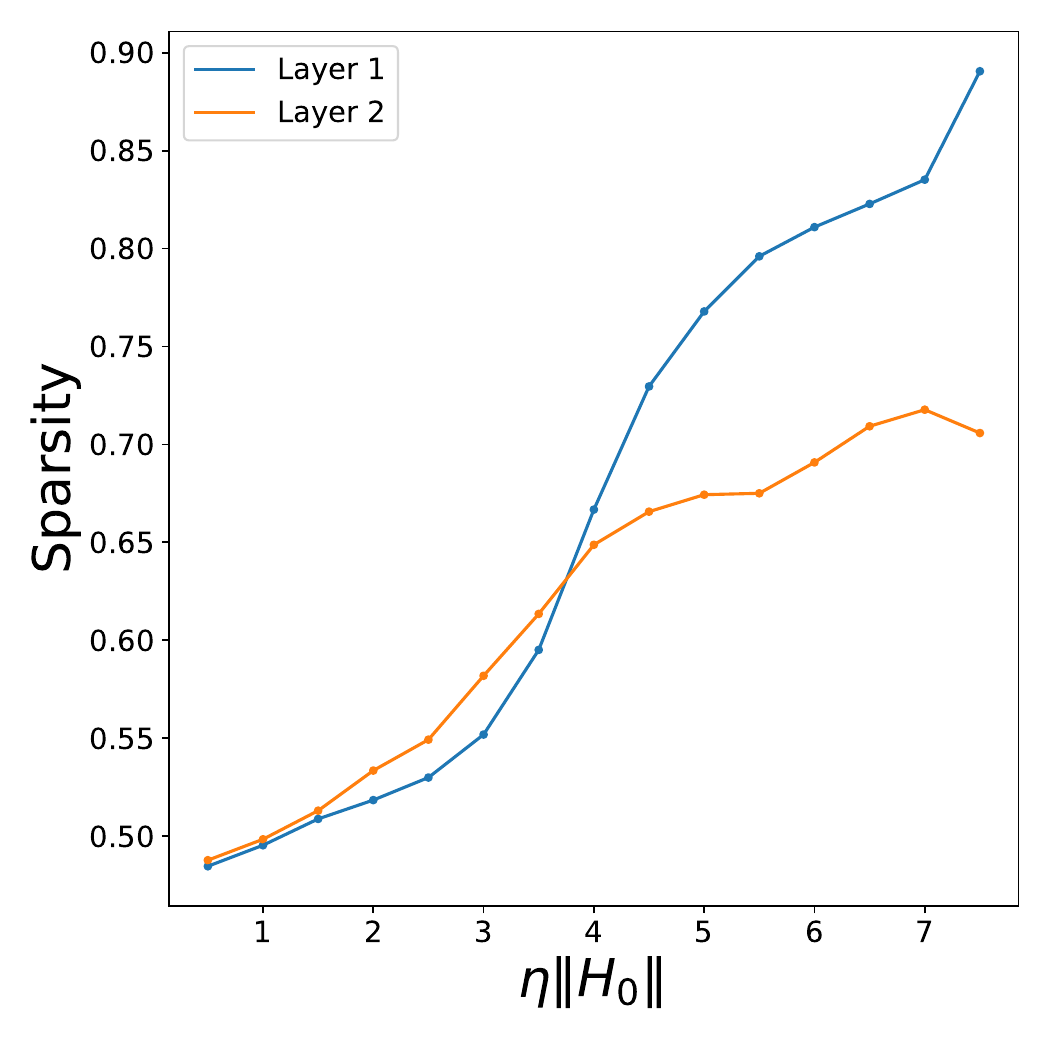}
\caption{}
\end{subfigure}
\\
\hfill
\begin{subfigure}[t]{\textwidth}
\centering
\includegraphics[scale=.27]{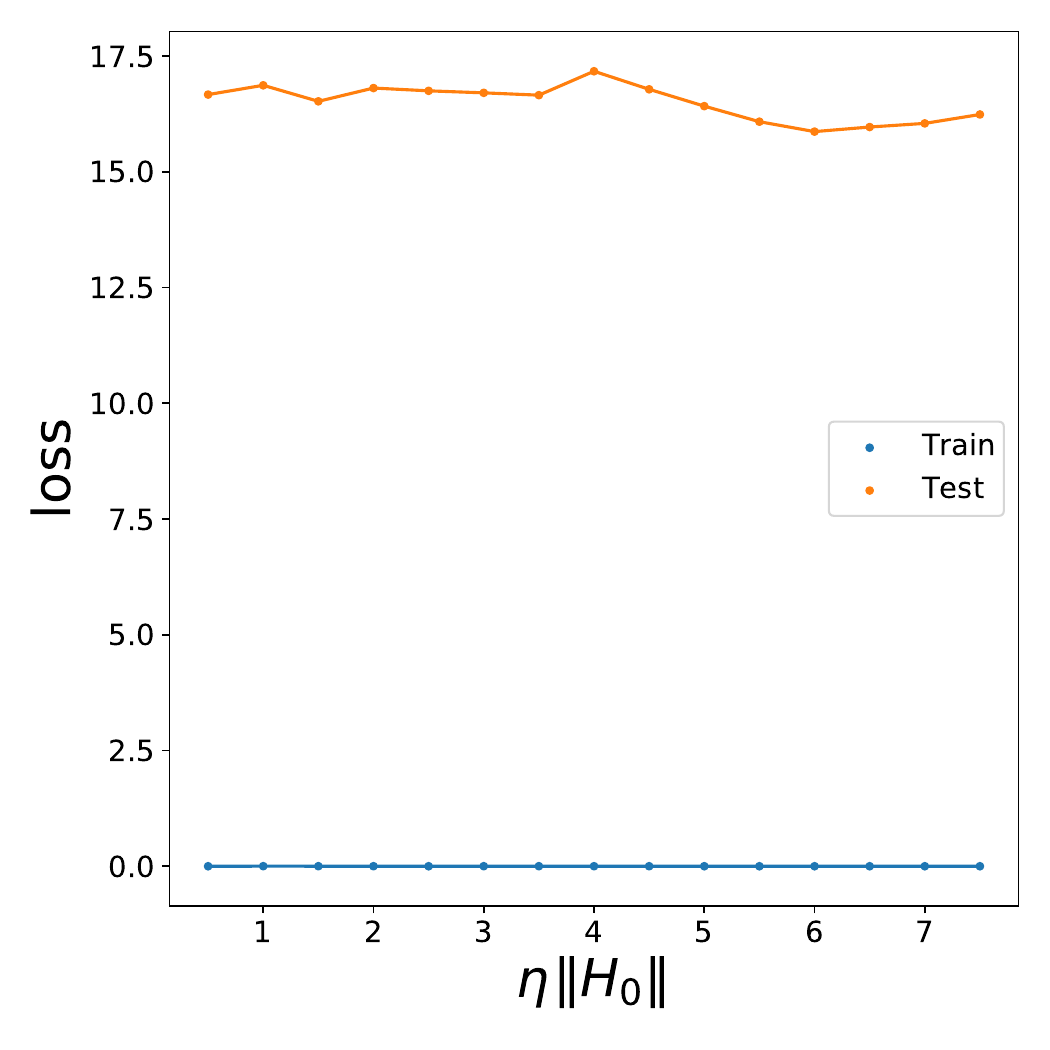}
\caption{}
\end{subfigure}
\caption{Results for the three-layer ReLU net trained on the two-class version of FMNIST.}
\label{fig:FMNIST_011_ReLU_two_hidden_layer}
\end{figure*}

\begin{figure*}[!ht]
\centering

\begin{subfigure}[b]{0.3\textwidth}
\centering
\includegraphics[scale=.27]{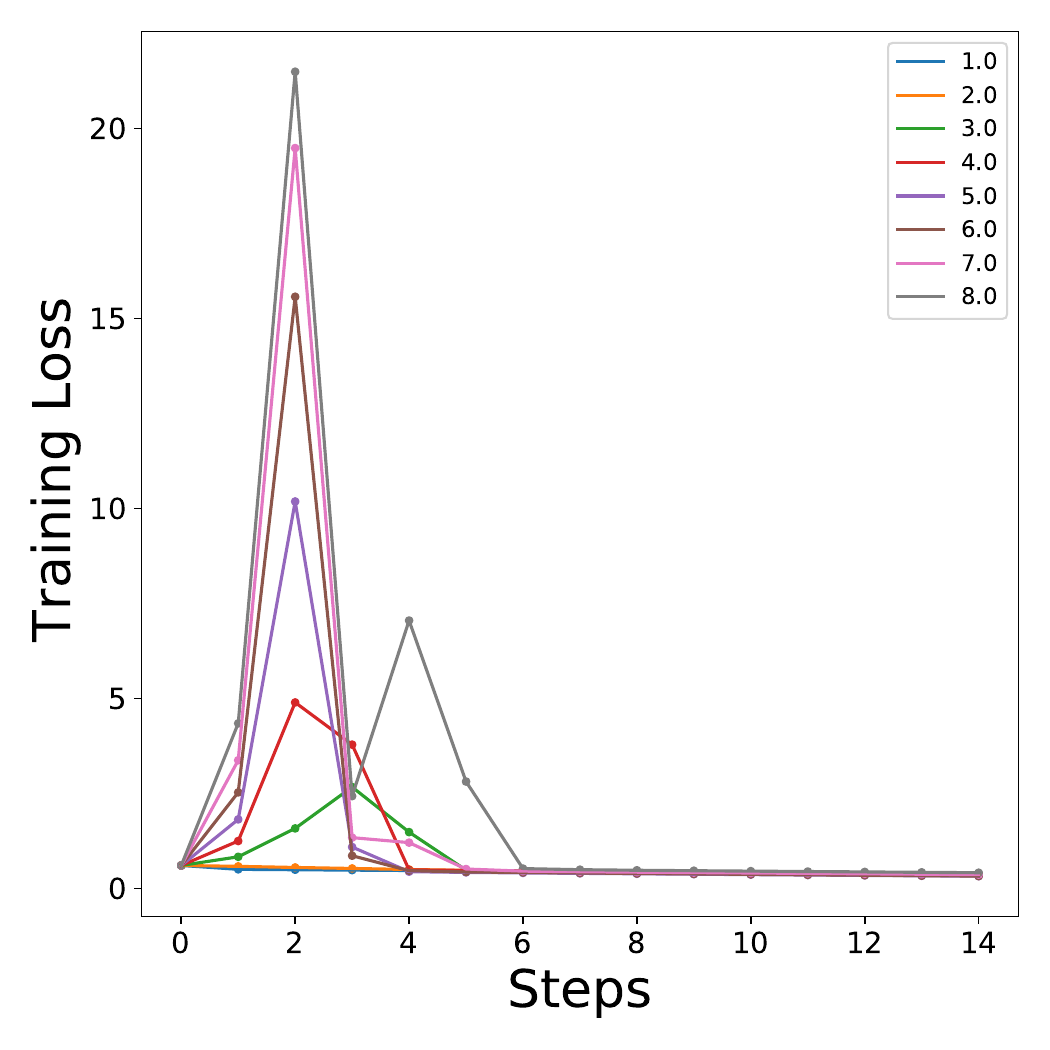}
\caption{}
\end{subfigure}
\hfill
\begin{subfigure}[b]{0.3\textwidth}
\centering
\includegraphics[scale=.27]{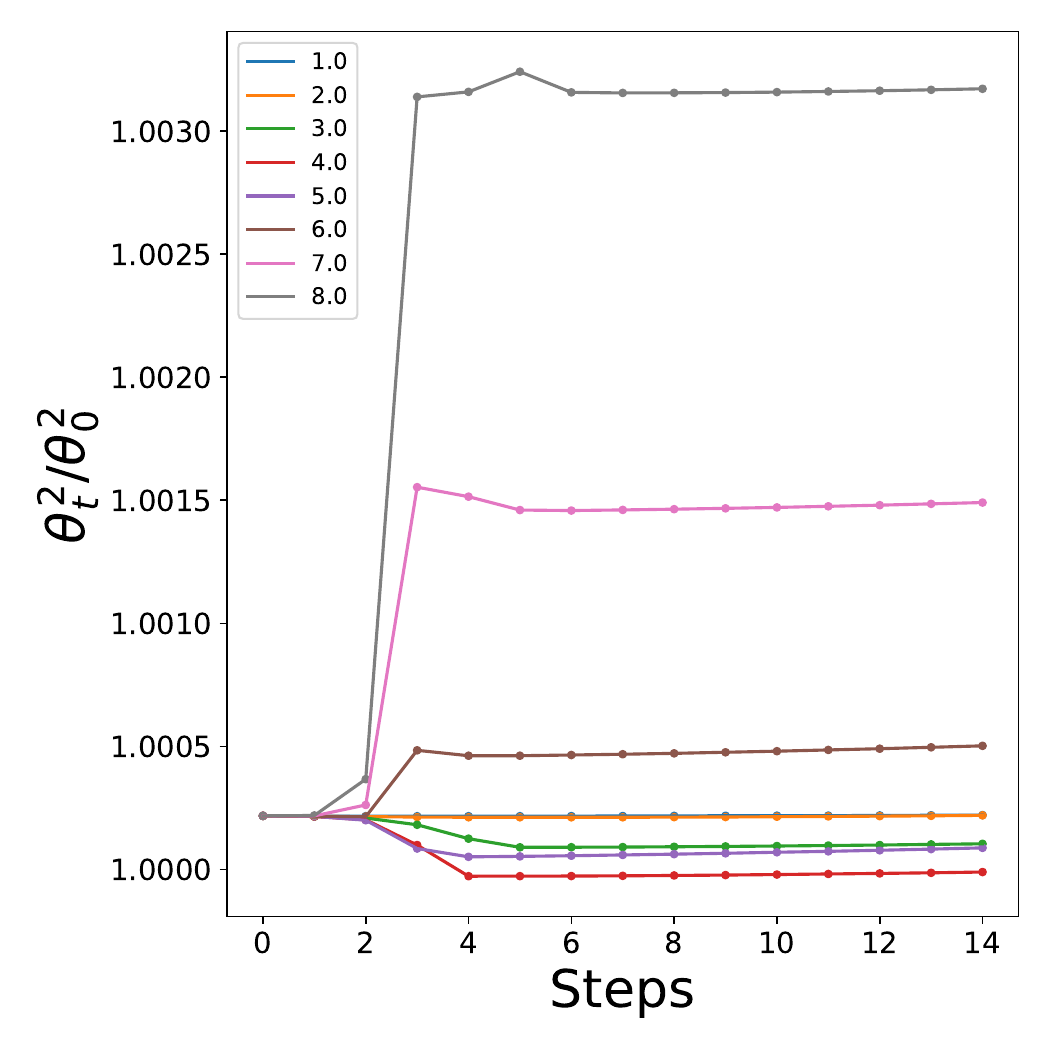}
\caption{}

\end{subfigure}
\hfill
\begin{subfigure}[b]{0.3\textwidth}
\centering
\includegraphics[scale=.27]{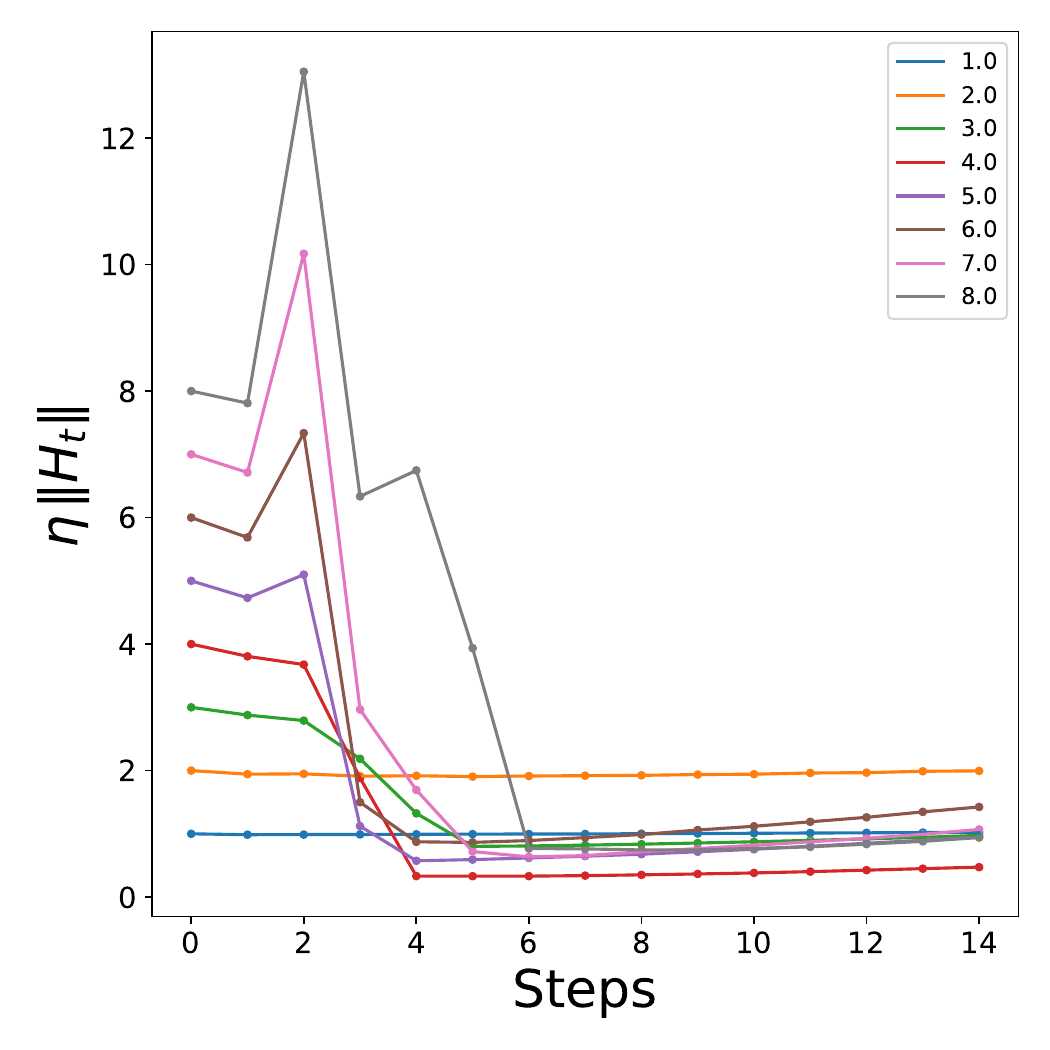}
\caption{}

\end{subfigure}
\centering
\begin{subfigure}[t]{0.3\textwidth}
\centering
\includegraphics[scale=.27]{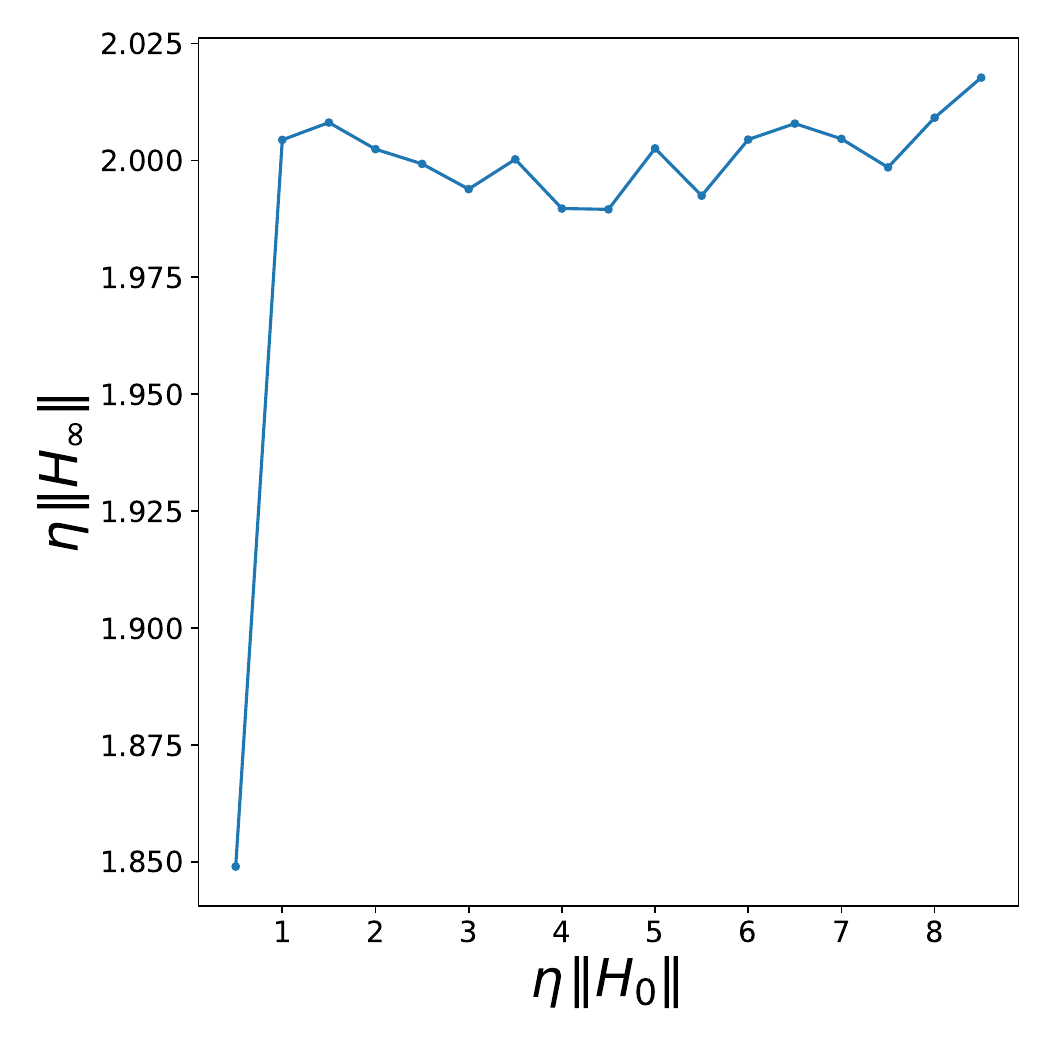}
\caption{}

\end{subfigure}
\hfill
\begin{subfigure}[t]{0.3\textwidth}
\centering
\includegraphics[scale=.27]{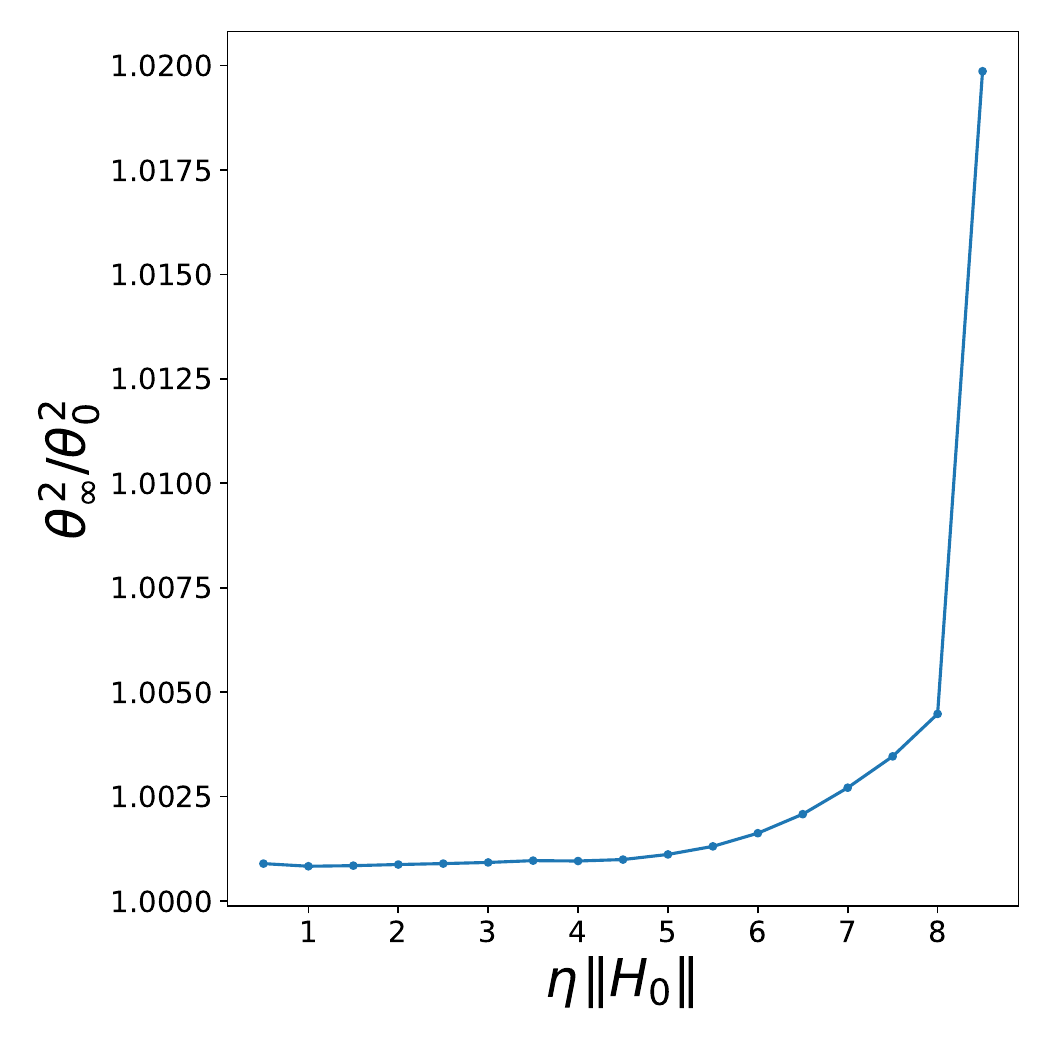}
\caption{}

\end{subfigure}
\hfill
\begin{subfigure}[t]{0.3\textwidth}
\centering
\includegraphics[scale=.27]{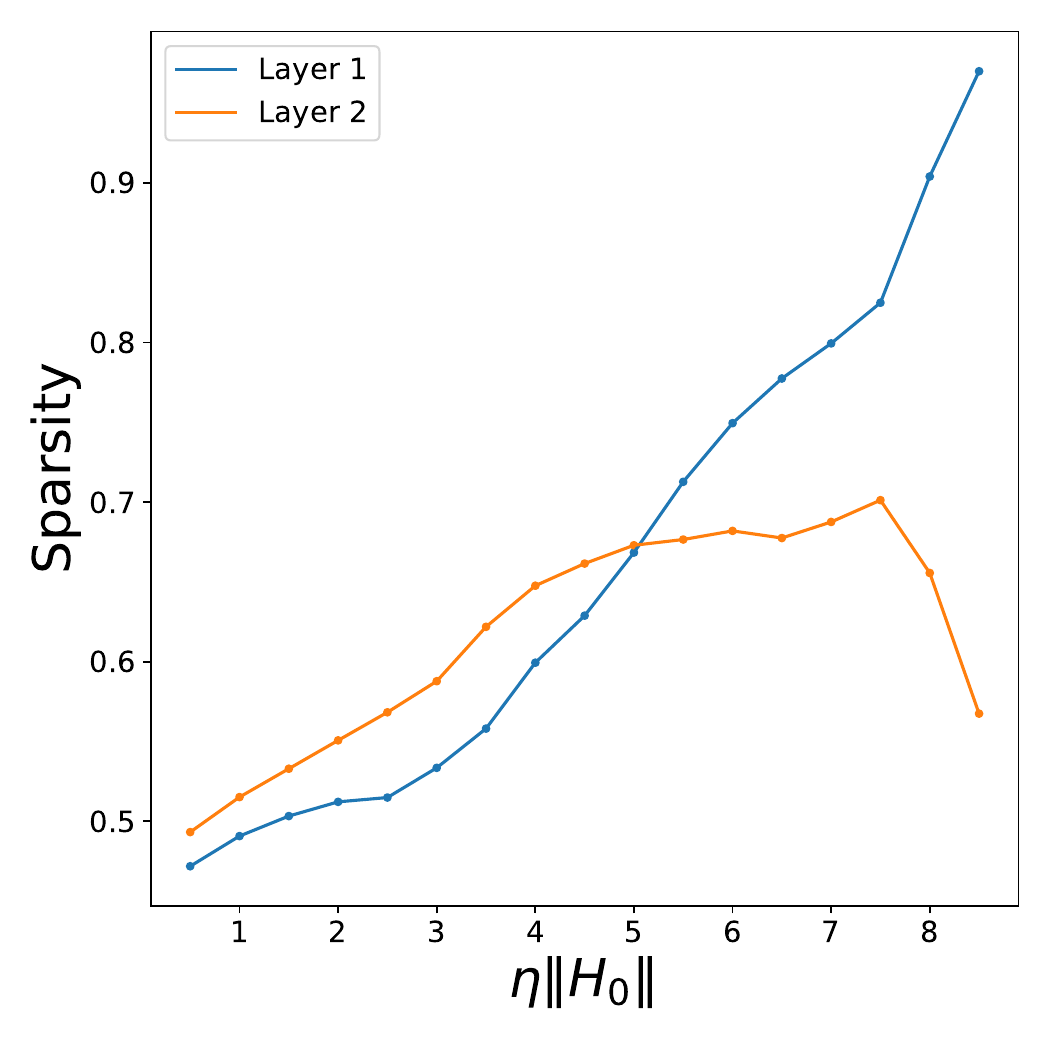}
\caption{}
\end{subfigure}
\\
\hfill
\begin{subfigure}[t]{\textwidth}
\centering
\includegraphics[scale=.27]{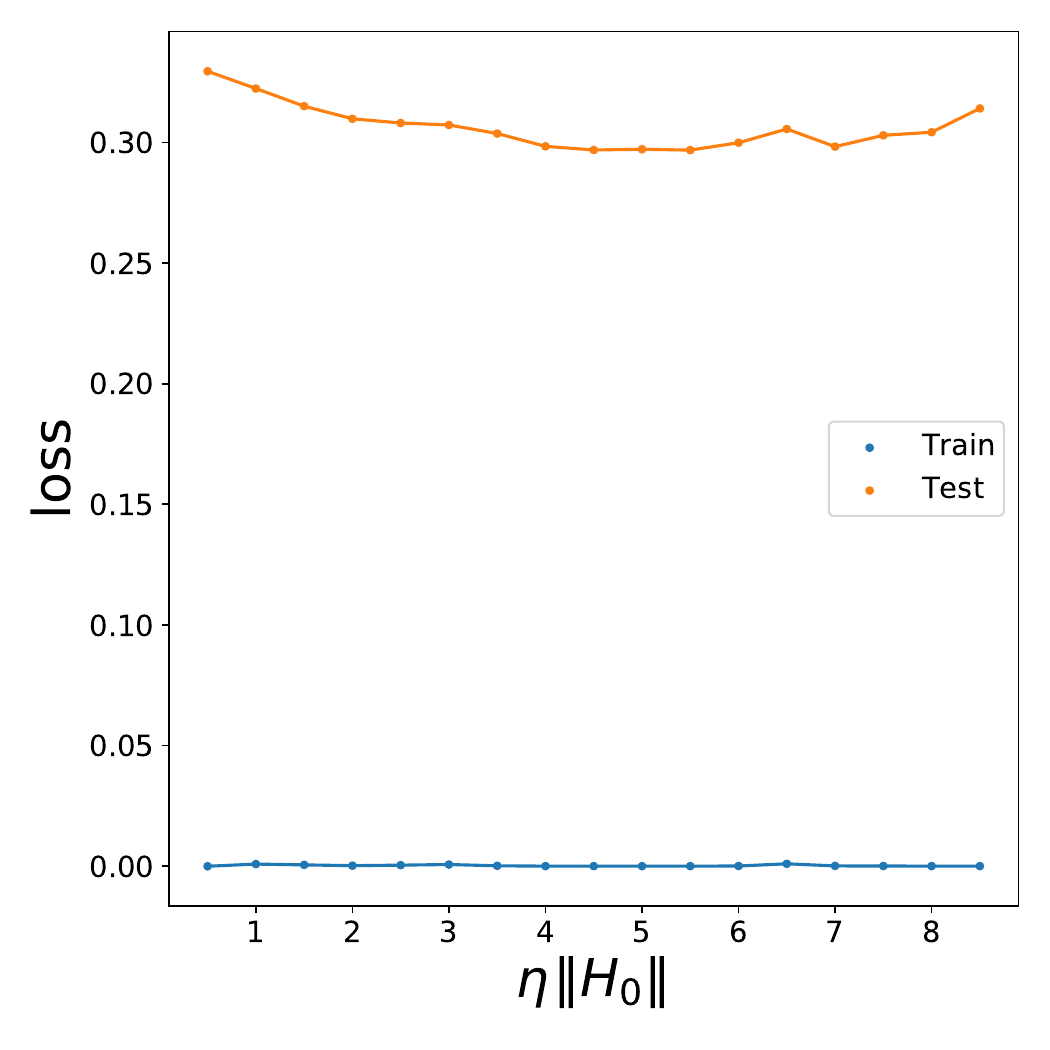}
\caption{}
\end{subfigure}
\caption{Results for the three-layer ReLU net trained on the two-class version of CIFAR-10.}
\label{fig:CIFAR10_01_ReLU_two_hidden_layer}
\end{figure*}

In this appendix we repeat the above experiments for ReLU nets with two hidden layers.
The results for the three-layer ReLU MLP trained on the two-class versions of MNIST, FMNIST, and CIFAR-10 are given in figures \ref{fig:MNIST_011_ReLU_two_hidden_layer}, \ref{fig:FMNIST_011_ReLU_two_hidden_layer}, and \ref{fig:CIFAR10_01_ReLU_two_hidden_layer}, respectively.

The results are largely the same as before: increasing the learning rate produces a sparser activation map and the top eigenvalue of the trained NTK hovers around the edge of stability.
We also observe that, above a critical learning rate, the weight norm $\bs{\theta}_t^2$ can increase over the course of training.
One difference, in comparison to the two-layer case, is that the models diverge for a smaller learning rate.
We also observe that the activation map for the first ReLU layer is sparser than the activation map for the second layer when the model is trained on the two-class versions of FMNIST and CIFAR-10.
The first and second layers are comparably sparse when the model is trained on MNIST.

\end{document}